\documentclass{article} 
\usepackage{iclr2027_conference,times}

\usepackage{amsmath,amsfonts,bm}

\def\eqref#1{equation~\ref{#1}}

\def\1{\bm{1}}

\DeclareMathAlphabet{\mathsfit}{\encodingdefault}{\sfdefault}{m}{sl}
\SetMathAlphabet{\mathsfit}{bold}{\encodingdefault}{\sfdefault}{bx}{n}

\usepackage{wrapfig}
\usepackage{hyperref}
\usepackage{url}
\usepackage{makecell}
\usepackage{pifont}
\usepackage{booktabs}
\usepackage{caption}
\usepackage[table]{xcolor}
\usepackage{booktabs}
\usepackage{makecell}  
\usepackage{fancyvrb}
\usepackage{float}
\usepackage{bbm}
\usepackage{simpleicons}

\definecolor{githubblue}{HTML}{0969DA}
\definecolor{hforange}{HTML}{B56A00}

\newcommand{\githublink}{%
  \href{https://github.com/SJTU-DeepDeltaLab/CausalCascade}{%
    \textcolor{githubblue}{%
      \raisebox{-0.2\height}{%
        \scalebox{1}{\simpleicon{github}}%
      }%
      \hspace{0.15em}GitHub%
    }%
  }%
}

\newcommand{\hflink}{%
  \href{https://huggingface.co/datasets/deepdeltalab/CausalCascade}{%
    \textcolor{hforange}{%
      \raisebox{-0.2\height}{%
        \includegraphics[height=1em]{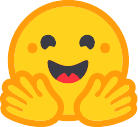}%
      }%
      \hspace{0.15em}Hugging Face%
    }%
  }%
}

\newenvironment{promptblock}
{%
    \par\medskip
    \noindent\rule{\linewidth}{0.6pt}
    \par\vspace{5pt}
    \begingroup
    \small
    \setlength{\parindent}{0pt}
    \setlength{\parskip}{2pt}
}
{%
    \par
    \endgroup
    \vspace{5pt}
    \noindent\rule{\linewidth}{0.6pt}
    \par
}

\newcommand{\promptheading}[1]{%
    \par\smallskip
    \noindent\textbf{#1}
    \par
}

\newcommand{\promptitem}[1]{%
    \par
    \noindent
    \hangindent=1.2em
    \hangafter=1
    - #1
}

\definecolor{heatblue}{RGB}{177,201,225}
\newcommand{\cmark}{\ding{51}}
\newcommand{\xmark}{\ding{55}}
\newcommand{\pmark}{$\circ$} 

\title{A Plausible Future Is Not a Causal Future: Can Video Models Generate How the World Unfolds?}

\author{
Jialiang Li$^{1,*}$ \quad
Zhengning Xu$^{1,*}$ \quad
Zongyang Ma$^{2}$ \quad
Shaobo Cui$^{1,\dagger}$ \\[0.4em]
{\small $^{1}$ DeepDelta Lab, School of Artificial Intelligence, Shanghai Jiao Tong University} \\ 
{\small $^{2}$ Xi'an Jiaotong-Liverpool University} \\[-0.1em]
{\small $^{*}$ Equal contribution \qquad
$^{\dagger}$ Corresponding author}
}

\usepackage{xspace}
\usepackage{graphicx}
\newcommand{\benchmarkname}[0]{CausalCascade\xspace}

\iclrfinalcopy 
\begin{document}

\maketitle

\vspace{-16pt}
\begin{abstract}
Large-scale video generation models are increasingly viewed as a promising path toward world models, yet their ability to capture causally coherent physical dynamics remains poorly understood.
Existing evaluations largely emphasize visual quality, prompt adherence, or broad physical plausibility, leaving open whether a model can faithfully realize a complete causal cascade, in which an initiating event propagates through causally dependent interactions to a terminal outcome.
We introduce \textbf{CausalCascade}, a simulator-grounded benchmark centered on 50 event-rich Blender scenes with structured event graphs, graph-grounded diagnostic VQA, three levels of future-event specification, and 2,832 controlled counterfactual variants, and use it to evaluate nine image-to-video models with complementary motion-, event-, and counterfactual-level diagnostics.
Across the evaluated models, generated motion is more reliably localized spatially than realized with correct spatiotemporal evolution, while richer future-event specification improves event recovery but leaves physical realization limited.
Counterfactual evaluation further reveals weak selective responsiveness to physical interventions, while most audited failures are foundational, arising at scene consistency or event occurrence.
Together, these results position causal cascade generation as a process-level test of world modeling and highlight the need to jointly recover the intended future, faithfully realize its physical evolution, selectively respond to physical interventions, and maintain a coherent physical world over time.
Code and dataset are available at
\githublink{} and \hflink{} respectively.

\end{abstract}

\section{Introduction}

Recent advances in large-scale video generation have made video generation models a promising route toward world models, as they show potential for predicting how physical scenes evolve under given conditions~\citep{sora, genie, nvidia2025cosmosworldfoundationmodel, worldsimbench}. Such predictive capability is central to applications such as robotic manipulation~\citep{pmlr-v235-zhou24f}, autonomous driving~\citep{NEURIPS2024_a6a066fb, wang2024drivedreamer}, and interactive world simulation~\citep{genie}, where effective decision-making depends on anticipating how the physical world will respond to actions and environmental changes. For such predictions to be useful, however, it is not enough for models to generate a single visually plausible event; they must produce a chain of events that is consistent with the physical consequences implied by the scene and the actions~\citep{physics-iq}, as physical consistency concerns not only whether individual events appear plausible, but also whether the interactions connecting them form a physically coherent evolution over time.

Many physical consequences unfold through a sequence of causally dependent events. A collision may set an object in motion, causing it to strike another body, which in turn triggers a downstream outcome. We refer to this multi-stage, causally connected process as a \emph{causal cascade}. Recent work has begun to recognize such cross-event causal progression as an important challenge for physically plausible video generation~\citep{wang2026chaineventcentriccausalthought, gillman2026goalforce}, but existing evaluations do not directly assess this capability. On the one hand, video-generation benchmarks assess perceptual quality, physical plausibility, adherence to physical principles, or agreement with reference dynamics~\citep{VBench_Huang_2024_CVPR, VBenchpp, EvalCrafter_Liu_2024_CVPR, videophy, videophy2, phygenbench, videosciencebench, physics-iq, worldmodelbench}, but they do not directly determine whether the causal dependencies between events are correctly realized. On the other hand, causal and counterfactual benchmarks probe dynamics prediction, causal relations, and responses to hypothetical changes~\citep{PHYRE, intphys, physion, physionpp, CLEVRER, CoPhy, craft, causalvqa}, but typically evaluate causal prediction or reasoning rather than whether causal relations are instantiated in generated content.

\begin{figure*}[t]
    \centering
    \vspace{-8pt}
    \includegraphics[width=\textwidth]{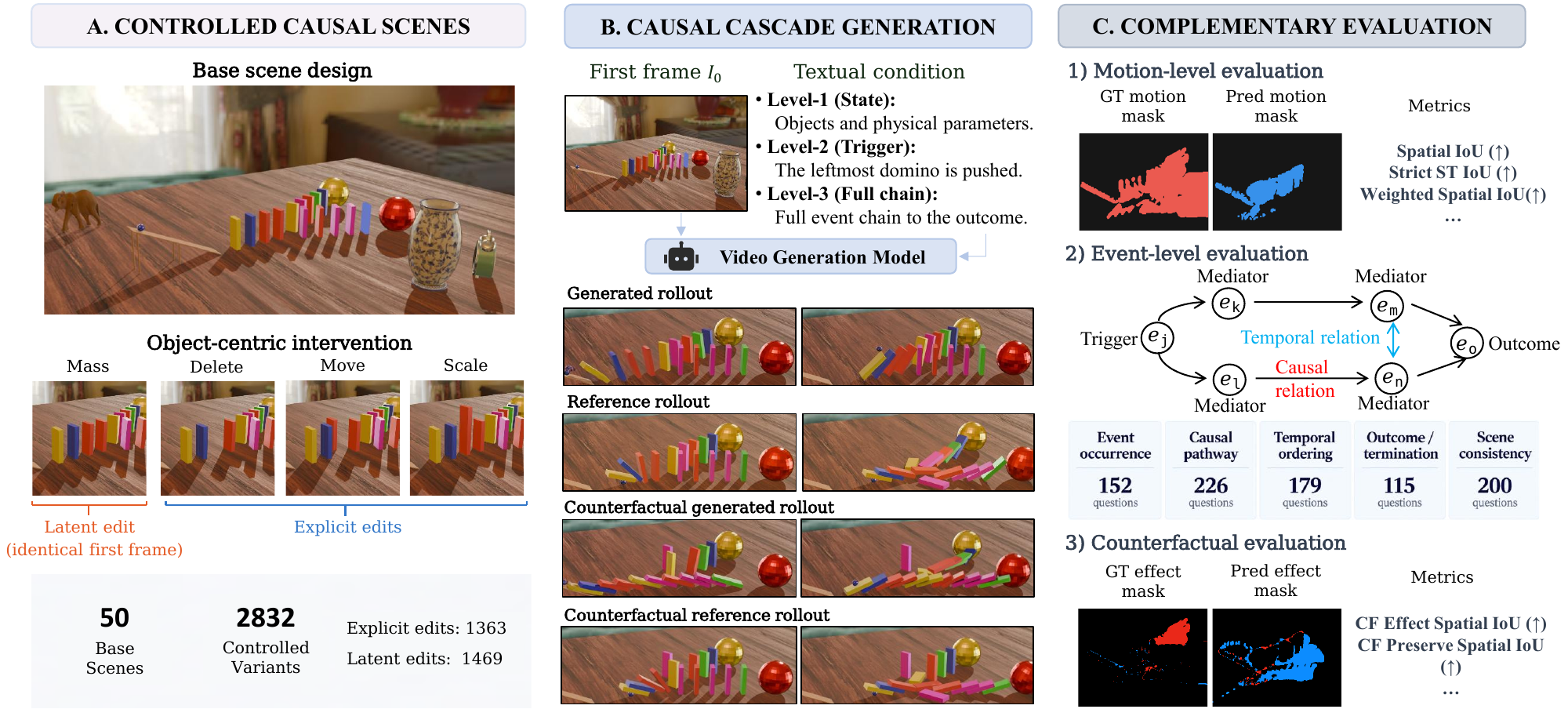}
    \vspace{-12pt}
\caption{
\textbf{Overview of \benchmarkname.}
\textbf{(A)} Base scene construction and controlled object-centric interventions.
\textbf{(B)} Simulator-grounded causal cascade generation with factual and counterfactual rollouts.
\textbf{(C)} Complementary motion-, event-, and counterfactual evaluations.
}
    \vspace{-12pt}
    \label{fig:main}
\end{figure*}

To address this gap, we introduce \benchmarkname, a simulator-grounded benchmark to evaluate whether image-to-video (I2V) models realize the complete causal process underlying a rollout (Figure~\ref{fig:main}). We construct 50 event-rich Blender scenes and represent reference rollouts with a structured event graph. This representation enables evaluation at two levels: motion-based metrics measure whether physical activity is realized in the correct space and time, while graph-grounded VQA diagnoses whether the events and dependencies are recovered. To separate knowing what should happen from realizing how it happens, we further consider three future-event specification levels: state-only, trigger-aware, and full event-chain prompts. We also construct 2,832 controlled variants through \emph{delete}, \emph{move}, \emph{scale}, and \emph{mass} interventions, enabling evaluation of whether intervention effects propagate to the dynamics that depend on them while unaffected dynamics remain preserved.

Across nine current I2V models, our evaluations reveal three gaps that limit their use as faithful world models.
First, more explicit information about \emph{what} should happen helps event recovery, but both event-level performance and the faithful realization of \emph{how} the causal process unfolds remain limited. 
Second, a world model must do more than produce a different future when the physical world changes.
It should selectively propagate the consequences of that change through affected dynamics while preserving those that are causally unaffected, yet current models struggle with both requirements.
Third, many rollouts fail at scene consistency or event occurrence, revealing a more foundational limitation: models often fail to establish a persistent physical world in which stable entities and events can support sustained causal propagation. 
These failures suggest that faithful causal cascade generation requires maintaining a persistent physical world with coherent entities and physical states, a foundational requirement for video models to function as causal world models.

Our main contributions are:

\begin{itemize}

\item We formulate \emph{causal cascade generation} as a process-level evaluation target for I2V models, asking whether causally dependent events are faithfully realized as a coherent process.

\item  We introduce \benchmarkname, a simulator-grounded benchmark that combines structured event graphs, controlled future-event specification, and physical interventions to evaluate causal cascade generation from complementary perspectives.

\item We evaluate nine I2V models, exposing a gap between event recovery and causal realization: future-event specification improves \emph{what} happens more than \emph{how} it unfolds, intervention responses remain weakly selective, and audited failures cluster at foundational stages.

\end{itemize} 

\section{Related Work}

\textbf{Physical Evaluation of Video Generation.}
General-purpose video generation benchmarks evaluate perceptual quality, temporal consistency, motion quality, and prompt alignment~\citep{VBench_Huang_2024_CVPR,VBenchpp,EvalCrafter_Liu_2024_CVPR}. Physics-oriented benchmarks further evaluate whether generated dynamics are physically plausible or agree with reference behavior~\citep{videophy,videophy2,phygenbench,videosciencebench,morpheus,worldmodelbench,phyworldbench,paibench}. For example, Physics-IQ~\citep{physics-iq} derives motion representations from generated and reference videos and evaluates their agreement using spatial, spatiotemporal, and motion-frequency metrics. WorldBench~\citep{worldbench} constructs tests that isolate physical concepts to attribute failures to specific aspects of physical understanding. These benchmarks provide increasingly detailed measures of physical fidelity, but place less emphasis on whether interactions are organized into the correct multi-stage causal process and satisfy the dependencies between successive events.

\textbf{Causal and Counterfactual Evaluation.}
Physical and causal reasoning benchmarks study dynamics prediction, temporal and causal reasoning, and counterfactual inference as reasoning or prediction tasks~\citep{PHYRE,intphys,physion,physionpp,CLEVRER,craft,CoPhy,clevrerhumans,comphy,contphy,causalvqa}. More recent work extends intervention-based evaluation to video and world models. CRONOS~\citep{cronos} evaluates whether predictions of physical events remain consistent under scene and object changes, while YoCausal~\citep{yocausal} probes temporal causal sensitivity using forward and time-reversed real-world videos. What-If World~\citep{whatifworld} is closely related to our setting: it evaluates paired generations in which one physical detail is changed, testing whether shared scene content is preserved while the expected physical consequence changes.

Taken together, these lines of work evaluate either the physical fidelity of generated dynamics or causal and counterfactual responses, but place less emphasis on whether a model realizes the complete causal process connecting successive events. Also, causal cascade generation further requires examining whether interventions alter causally dependent dynamics while preserving unaffected ones. To support this evaluation, \benchmarkname provides complete causal-process rollouts with controlled interventions and combines motion-, event-, and counterfactual-level diagnostics. Table~\ref{tab:benchmark_comparison} compares \benchmarkname with existing physical, causal, and counterfactual benchmarks.

\begin{table*}[t]
\centering
\small
\setlength{\tabcolsep}{4.2pt}
\renewcommand{\arraystretch}{1.12}

\caption{
\textbf{Comparison with representative physical, causal, and counterfactual benchmarks.}
\cmark, \pmark, and \xmark ~denote explicit, partial, and absent coverage, respectively.
}
\label{tab:benchmark_comparison}

\resizebox{\textwidth}{!}{%
\begin{tabular}{lcccccccc}
\toprule
\textbf{Benchmark}
& \makecell{\textbf{Video Generator}\\\textbf{Eval.}}
& \makecell{\textbf{Reference}\\\textbf{Rollout}}
& \makecell{\textbf{Controlled}\\\textbf{Intervention}}
& \makecell{\textbf{Paired}\\\textbf{CF Eval.}}
& \makecell{\textbf{Localized}\\\textbf{CF Effect}}
& \makecell{\textbf{Ref.-based}\\\textbf{Motion Metric}}
& \makecell{\textbf{Structured}\\\textbf{Event Graph}}
& \makecell{\textbf{Event-level}\\\textbf{Diagnostics}} \\
\midrule

WorldBench~\citep{worldbench}
& \cmark
& \cmark
& \xmark
& \xmark
& \xmark
& \cmark
& \xmark
& \xmark \\

CLEVRER-Humans~\citep{clevrerhumans}
& \xmark
& \xmark
& \xmark
& \xmark
& \xmark
& \xmark
& \cmark
& \cmark \\

CoPhy~\citep{CoPhy}
& \xmark
& \cmark
& \cmark
& \cmark
& \xmark
& \pmark
& \xmark
& \xmark \\

CRONOS~\citep{cronos}
& \cmark
& \cmark
& \cmark
& \pmark
& \pmark
& \cmark
& \xmark
& \pmark \\

What-If World~\citep{whatifworld}
& \cmark
& \xmark
& \cmark
& \cmark
& \pmark
& \xmark
& \xmark
& \pmark \\

\midrule

\textbf{\benchmarkname~(Ours)}
& \cmark
& \cmark
& \cmark
& \cmark
& \cmark
& \cmark
& \cmark
& \cmark \\

\bottomrule
\end{tabular}%
}
\end{table*}

\section{Benchmark Formulation}
\label{sec:benchmark_formulation}

\label{sec:research_questions}

We organize \benchmarkname around three research questions:
\begin{itemize}
    \item \textbf{RQ1:} How well can current I2V models faithfully generate complete causal cascades?
    \item \textbf{RQ2:} Can the I2V models selectively update causal futures under physical interventions?
    \item \textbf{RQ3:} What are the main failure modes in current causal cascade generation?
\end{itemize}

Together, these questions examine complementary aspects of causal cascade generation: faithful realization, selective responsiveness to physical interventions, and failure localization.

\subsection{Causal Cascade Generation Task}

A causal cascade is a multi-stage physical process in which downstream events depend on states produced by earlier interactions. We consider a physical scene $\mathcal{S}$ with a current physical state and its corresponding future evolution under the simulator.
The simulator renders an initial frame $I_0$ and a reference rollout $V^*$ that records the event-rich physical evolution of the scene from that state.
We formulate causal cascade generation as the task of generating the complete process from only the initial observation $I_0$ and a textual condition $u$:
\begin{equation}
    \hat{V} = F(I_0,u).
\end{equation}

The initial frame specifies the observable entities, geometry, and spatial relations of the scene, while the textual condition provides complementary physical information.
The generation target is not merely a plausible event or final outcome but a coherent process in which causal consequences are sustained across successive interactions.
Notably, the textual condition also provides a controlled interface for varying how explicitly the intended future process is specified.

\subsection{Causal Cascade Fidelity}
\label{sec:factual_eval}

Because a causal cascade unfolds through dependent stages, fidelity must be maintained throughout the process rather than judged from isolated events or the final outcome.
We therefore characterize causal fidelity through two complementary requirements.
First, causally relevant activity should occur in the appropriate regions and evolve consistently through time.
Second, the generated rollout should realize the expected events and causal dependencies that connect the initiating event, intermediate interactions, and downstream outcome.
To represent these event dependencies explicitly, we associate each reference rollout $V^*$ with a causal event graph $G^*$ to capture the organization of the intended cascade beyond pixel-level similarity.

Faithfully generating a causal cascade requires the model both to infer \emph{what} future process is implied by the current state and to physically realize \emph{how} that process unfolds.
These demands are coupled because the model must infer not only an immediate event but also the downstream event structure linked through successive interactions.
We therefore use the textual condition $u$ as an interface for varying how explicitly the future process is specified.
Increasing the explicitness of future-event specification reduces the need to infer the intended event structure while retaining the requirement to generate the corresponding physical evolution.
Comparisons across specification conditions thus help distinguish future-event inference from physical realization within the same generation task.

\subsection{Causal Responsiveness to Physical Interventions}
\label{sec:cf_formulation}

For a world model, causal fidelity for a given scene is not enough; the generated future should also respond selectively when the underlying physical state changes.
This requires capturing the causal dependencies through which the effects of that change propagate across the cascade.
In a multi-stage process, changing one physical factor may alter downstream interactions that depend on it while leaving other dynamics unaffected.
To test this selective propagation, we introduce controlled interventions on selected physical factors of the scene.

Given a \emph{factual} scene $\mathcal{S}^{f}$, an intervention $T_k$ modifies a controlled physical factor to produce a corresponding \emph{counterfactual} scene
\begin{equation}
    \mathcal{S}^{cf} = T_k(\mathcal{S}^{f}),
    \label{eq:counterfactual_state}
\end{equation}
yielding matched reference rollouts $(V^{f*},V^{cf*})$ and generated rollouts $(\hat V^{f},\hat V^{cf})$.

A causally responsive model should propagate the consequences of the intervention through the affected parts of the cascade without perturbing unaffected dynamics.
Dynamics that change between the reference rollouts $(V^{f*},V^{cf*})$ should change correspondingly between the generated rollouts $(\hat V^f,\hat V^{cf})$, while dynamics shared by the reference pair should remain shared in the generated pair.
Therefore, this paired formulation tests whether the generated causal process responds selectively to the modified physical state, rather than simply whether the paired generations differ.

\section{Benchmark Construction}
\label{sec:benchmark_construction}

\subsection{Base Scene Construction}
\label{sec:scene_construction}

We construct 50 event-rich Blender scenes containing 954 rigid bodies. Figure~\ref{fig:dataset_statistics} (left) summarizes their structural complexity in terms of objects, events, causal dependencies, and temporal depth.
The scenes are selected to satisfy four principles: \emph{multi-stage causality}, so success requires propagation through intermediate interactions; \emph{mechanism diversity}, including interaction-rich and instability-driven processes; \emph{intervention sensitivity}, so selected physical factors can causally alter downstream behavior; and \emph{observability}, so relevant interaction paths remain visible. Each scene is built by instantiating a physical interaction motif, varying its geometry or topology, extending it with upstream triggers or downstream consequences, and fixing the visual setup. We export initial frame $I_0$, the reference rollout $V^*$, and structured scene metadata. Detailed design principles, the full construction pipeline, and asset, physics, camera, and rendering settings are in Appendix~\ref{app:scene_creation_details}.

\subsection{Causal Annotation and Prompt Construction}
\label{sec:event_graph_qa}

For each reference rollout $V^*$, we annotate observable events and their direct causal dependencies to instantiate its causal event graph $G^*$. 
From each graph, we deterministically derive a set of closed-ended diagnostic questions $Q^*$ based on its events, dependencies, temporal stages, and participants.
Each question is therefore traceable to a specific constraint in the reference causal process and serves as an event-level evaluation instance.

We instantiate the controlled future-event specification introduced in Section~\ref{sec:factual_eval} using three prompt conditions.
Level~1 describes the current scene state and relevant physical conditions without specifying any future event.
Level~2 additionally specifies the initiating event, providing the model with the trigger of the upcoming causal process while leaving the subsequent evolution unspecified.
Level~3 provides a substantially more explicit description of the intended future by verbalizing the major event chain from initiation through intermediate interactions to the downstream outcome.
Across these conditions, the principal controlled factor is therefore the explicitness of future-event specification rather than literal textual inclusion between prompts. Graph schemas, QA templates, prompt templates, and examples are provided in Appendices~\ref{app:event_graph} and~\ref{app:prompt_templates} .

\subsection{Counterfactual Variant Construction}
\label{sec:variant_construction}

\begin{figure*}[t]
    \centering
    \vspace{-0.8em}
    \includegraphics[width=\linewidth]
    {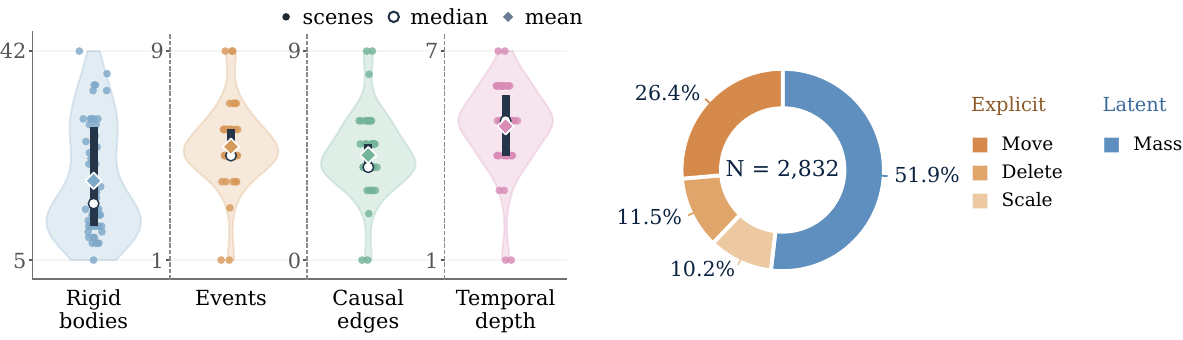}
    \caption{\textbf{Benchmark statistics.}
    \textbf{Left:} Distributions of structural statistics across 50 base scenes.
    \textbf{Right:} Distribution of intervention types across 2,832 counterfactual variants.}
    \label{fig:dataset_statistics}
    \vspace{-0.8em}
\end{figure*}

Starting from each base scene, we construct object-centric counterfactual variants by applying a single intervention to one selected object at a time.
We consider four intervention types.
Three are visually explicit and alter the initial RGB observation: \emph{delete} removes the selected object, \emph{move} changes its initial position, and \emph{scale} changes its size or proportions.
The fourth, \emph{mass}, is visually latent, changing only the object's mass while leaving the initial frame unchanged.
Each counterfactual scene is independently re-simulated to obtain its reference rollout, yielding factual-counterfactual pairs grounded in separate physical simulations rather than manual video edits.
We retain variants that alter downstream event structure as well as variants that affect only finer-grained motion or timing.
After automatic validity checks and screening, the benchmark contains 2,832 counterfactual variants, whose intervention distribution is shown in Figure~\ref{fig:dataset_statistics} (right).
Detailed intervention generation and validation procedures are provided in Appendix~\ref{app:edit_generation}.

\subsection{Evaluation Measures and Protocol}
\label{sec:evaluation_protocol}


\textbf{Motion-level evaluation.}
Following Physics-IQ~\citep{physics-iq}, we extract spatially normalized motion maps from each reference and generated rollout.
We measure their agreement using Spatial IoU, Strict ST-IoU, and Weighted Spatial IoU.
Spatial IoU measures motion-region overlap, Strict ST-IoU additionally enforces fixed frame-level temporal correspondence, and Weighted Spatial IoU accounts for motion magnitude.
Motion extraction and metric definitions are in Appendix~\ref{app:evaluation_details}.

\textbf{Event-level evaluation.}
We evaluate whether the generated rollout realizes the reference causal process using the diagnostic questions $Q^*$ derived from $G^*$.
A tool-augmented video-QA evaluator combines full-rollout inspection with temporally aligned views and object-centric evidence, producing dimension-level scores and an overall VQA score. Both motion- and event-level evaluations cover all 50 base scenes.
Complete evaluation protocol and human validation are in Appendix~\ref{app:hybrid_vqa_protocol}.

\textbf{Counterfactual evaluation.}
We operationalize causal responsiveness to physical interventions using two complementary motion-based criteria: \emph{Effect}, which measures whether motion is correctly suppressed or induced by the intervention, and \emph{Preserve}, which measures whether motion shared across the factual and counterfactual references remains preserved.
We compute both in spatial and weighted-spatial forms, yielding CF Spatial/Weighted Effect IoU and CF Preserve Spatial/Weighted IoU.
Our designated counterfactual evaluation subset contains 225 model-blind factual-counterfactual pairs: 50 delete, 50 move, 50 scale, and 75 mass interventions.
Each pair uses matched Level~1 prompts, with only intervention-affected attributes updated between the factual and counterfactual conditions, and scores are macro-averaged across represented base scenes.
Screening and subset-selection details are in Appendix~\ref{app:manual_screening}, and exact metric definitions are in Appendix~\ref{app:cf_metrics}.

\section{Experiment and Analysis}

\textbf{Evaluated models.}
We evaluate nine representative open-source I2V models spanning diverse model families and scales:
CogVideoX-Fun-V1.5~\citep{yang2024cogvideox,cogvideoxfun15},
Cosmos3-Nano~\citep{cosmos3},
HunyuanVideo-I2V~\citep{hunyuanvideo},
LTX-Video-13B~\citep{hacohen2024ltxvideo},
MAGI-1-4.5B~\citep{sandai2025magi1},
SkyReels-V2-I2V~\citep{chen2025skyreelsv2},
Wan2.1-I2V-14B,
Wan2.2-TI2V-5B, and Wan2.2-I2V-A14B~\citep{wan}.
We select models that are publicly available, support our common input interface, and provide broad coverage of recent model families and capacities.
Full inference configurations are provided in Appendix~\ref{app:implementation_details}.

\subsection{How well can current I2V models generate causal cascades? (RQ1)}

RQ1 examines how faithfully current I2V models realize causal cascades under different degrees of future-event specification.
We first compare all nine models under Level~1 and then examine how performance changes under Levels~2 and~3.
Finally, we apply post-hoc temporal alignment to assess how much of the remaining spatiotemporal discrepancy can be attributed to timing mismatch.

\begin{figure}[t]
    \centering

    \begin{minipage}[t]{0.51\textwidth}
        \vspace{0pt}
        \centering
        \scriptsize
        \setlength{\tabcolsep}{3.3pt}
        \renewcommand{\arraystretch}{1.02}
        \begin{tabular}{@{}lccc@{}}
            \toprule
            Model
            & Spatial IoU $\uparrow$
            & Strict ST-IoU $\uparrow$
            & VQA Overall $\uparrow$ \\
            \midrule
            CogVideoX-Fun-V1.5
            & 0.147 & 0.037 & 0.095 \\

            Cosmos3-Nano
            & \underline{0.252} & 0.043 & \textbf{0.357} \\

            HunyuanVideo-I2V
            & 0.178 & 0.027 & 0.174 \\

            LTX-Video-13B
            & 0.182 & 0.037 & 0.189 \\

            MAGI-1-4.5B
            & 0.128 & 0.031 & 0.153 \\

            SkyReels-V2-I2V
            & 0.193 & 0.047 & 0.177 \\

            Wan2.1-I2V-14B
            & 0.204 & \underline{0.055} & 0.196 \\

            Wan2.2-TI2V-5B
            & 0.149 & 0.027 & 0.141 \\

            Wan2.2-I2V-A14B
            & \textbf{0.312} & \textbf{0.080} & \underline{0.252} \\
            \bottomrule
        \end{tabular}
    \end{minipage}
    \hfill
    \begin{minipage}[t]{0.46\textwidth}
        \vspace{0pt}
        \centering
        \includegraphics[
            width=0.97\linewidth
        ]{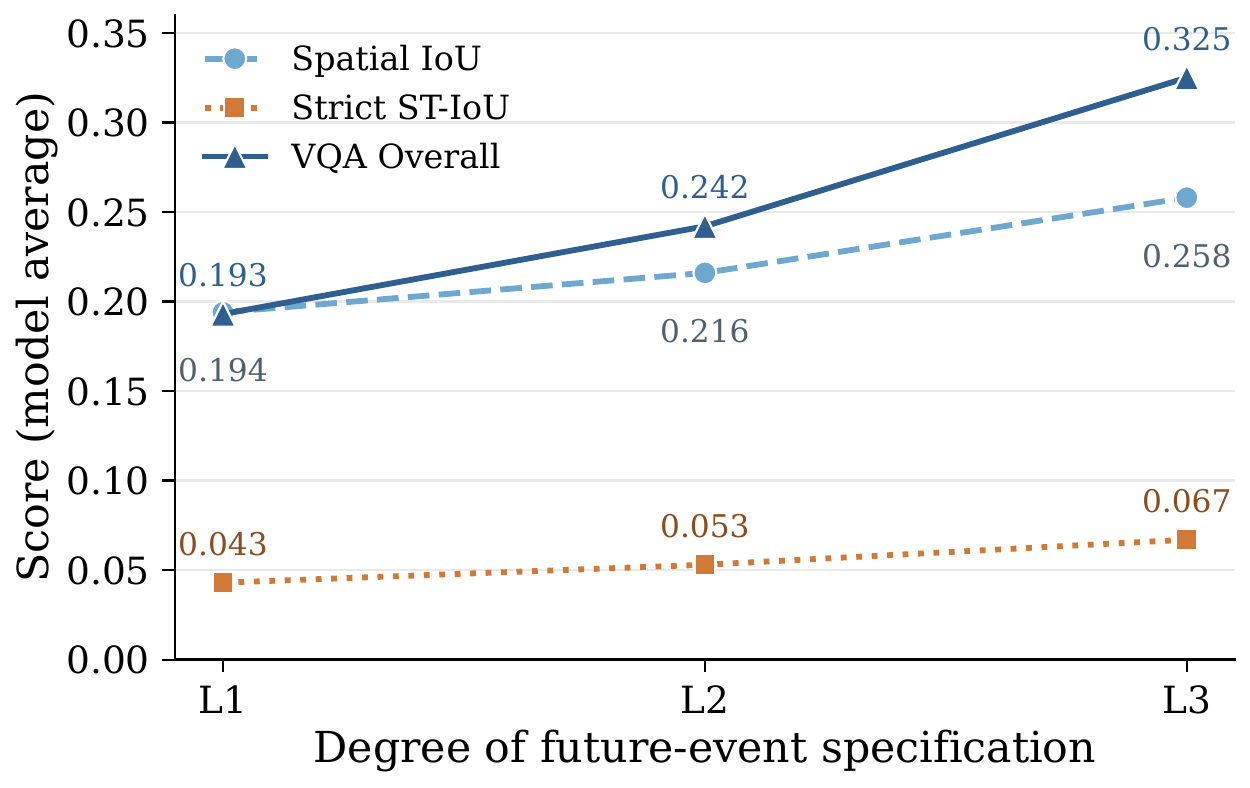}
    \end{minipage}

    \caption{
        \textbf{Main results for RQ1.}
        \textbf{Left:} Performance of current I2V models under the Level~1 prompt.
        \textbf{Right:} Model-average performance as future-event specification becomes more explicit.
    }
    \label{fig:main_results_summary}
    \vspace{-8pt}
\end{figure}

\textbf{Finding 1: Spatial localization exceeds spatiotemporal agreement.} 
Under the Level~1 setting, the model-average Spatial IoU reaches $0.194$, whereas Strict ST-IoU is only $0.043$ (Figure~\ref{fig:main_results_summary}, left). Spatial IoU only asks whether motion appears in the relevant spatial regions, while Strict ST-IoU further requires the motion to occur at the corresponding physical time. 
Even the strongest model on both metrics, Wan2.2-I2V-A14B, reaches only $0.080$ Strict ST-IoU despite a Spatial IoU of $0.312$.
These results suggest that models can roughly anticipate where physical change should take place, but cannot faithfully realize how the corresponding interaction evolves over time. 

\begin{figure}[t]
    \centering
    \includegraphics[width=\textwidth]{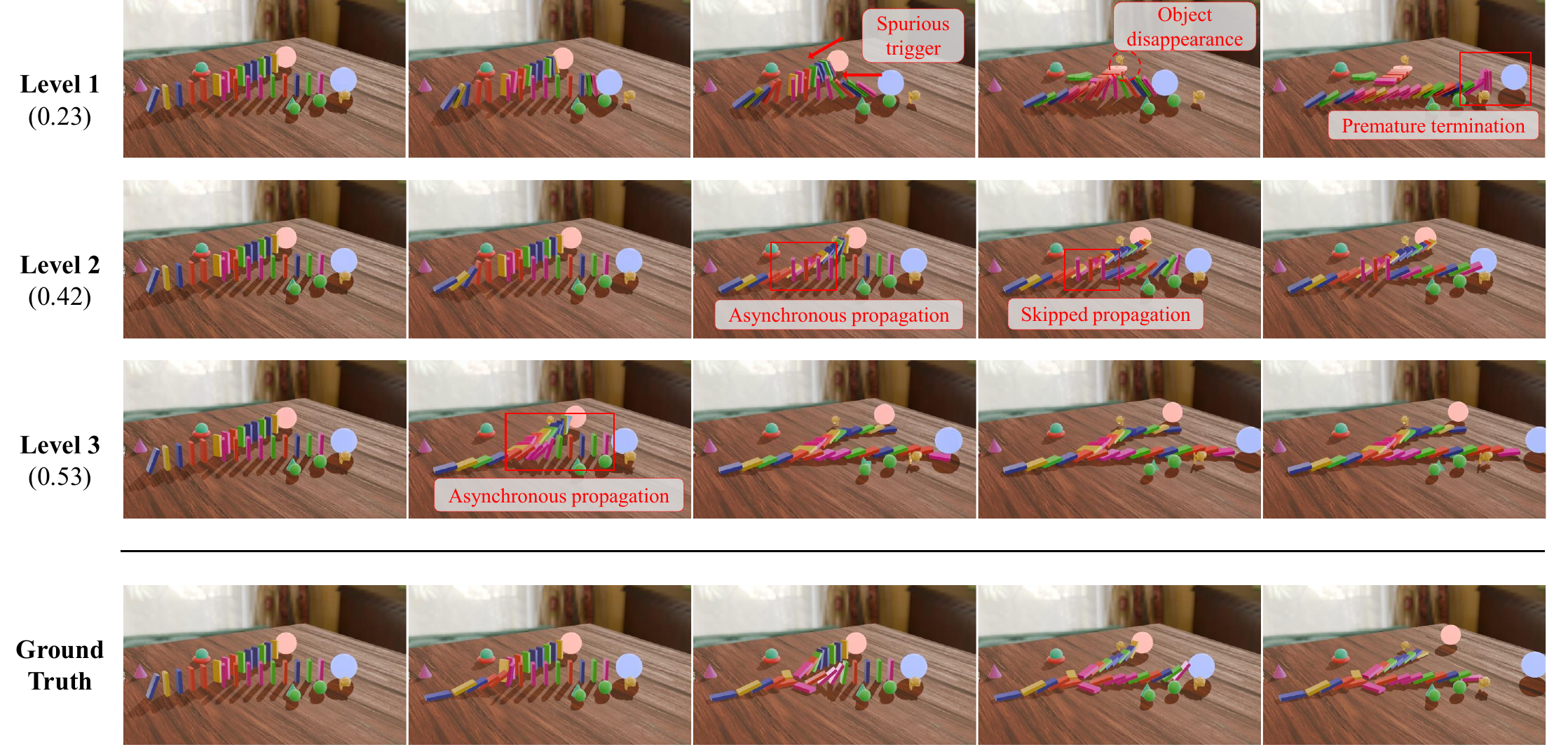}
    \caption{
\textbf{Wan2.2-I2V-A14B rollouts for one scene under three future-event specification levels.}
Numbers in parentheses denote VQA Overall scores.
Richer future-event specification improves event-structure recovery without comparable gains in faithful physical realization.
    }
    \label{fig:rq1_prompt_case}
    \vspace{-12pt}
\end{figure}

\textbf{Finding 2: Future-event specification helps event recovery, but physical realization remains limited.}
As future-event specification increases from Level~1 to Level~3, model-average VQA Overall rises from $0.193$ to $0.325$, while Strict ST-IoU increases from $0.043$ to $0.067$ (Figure~\ref{fig:main_results_summary}, right).
The same pattern appears for Wan2.2-I2V-A14B: VQA Overall rises from $0.252$ to $0.532$, while Strict ST-IoU increases only from $0.080$ to $0.131$.
Thus, more explicit future-event specification reduces the difficulty of recovering \emph{what} should happen, but event-level performance remains limited and faithfully realizing \emph{how} the causal process unfolds remains difficult.
The VQA dimensions further support the distinction that event occurrence is recovered more reliably than causal pathway, temporal ordering, and outcome or termination (Table~\ref{tab:vqa_dimension_levels}).
A representative case in Figure~\ref{fig:rq1_prompt_case} illustrates the same pattern, while complete per-model quantitative results are reported in Appendix~\ref{app:detailed_results}.

\begin{wrapfigure}[25]{r}{0.4\linewidth}
    \centering
    \scriptsize
    \setlength{\tabcolsep}{2.5pt}
    \renewcommand{\arraystretch}{0.95}
    \vspace{4pt}
    \begin{tabular}{lccc}
    \toprule
    Prompt
    & \makecell{Strict\\ST-IoU}
    & \makecell{Affine\\(+Global)}
    & \makecell{DTW\\(+Local)} \\
    \midrule
    Level~1 & 0.043 & 0.053 & 0.063 \\
    Level~2 & 0.053 & 0.069 & 0.082 \\
    Level~3 & 0.067 & 0.086 & 0.104 \\
    \bottomrule
    \end{tabular}

    \vspace{4pt}
    \includegraphics[
        width=0.98\linewidth
    ]{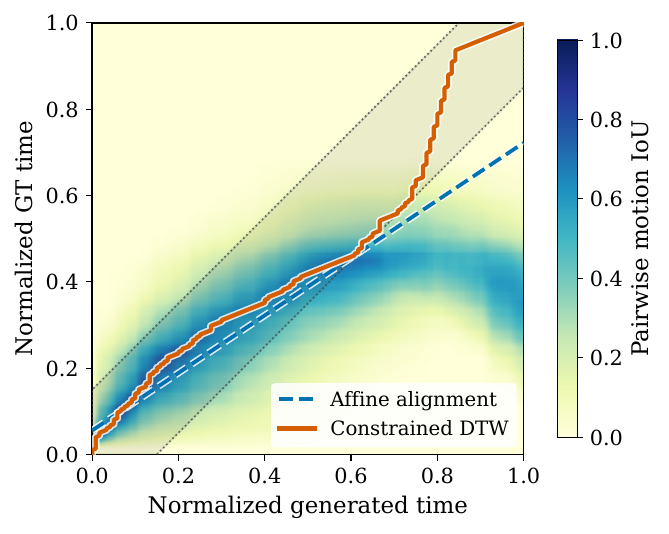}

    \caption{
    \textbf{Temporal alignment diagnostics.}
    \textbf{Top:} Model-average agreement under strict timing, global affine alignment, and constrained DTW.
    \textbf{Bottom:} Representative alignment paths over the pairwise motion-IoU matrix.
    }
    \label{fig:temporal_alignment}
    \vspace{-8pt}
\end{wrapfigure}

\textbf{Finding 3: Temporal alignment explains only part of the remaining gap.}
Even under Level~3, model-average Strict ST-IoU remains only $0.067$.
We therefore test whether the residual discrepancy can be explained by temporal misalignment.
Global affine alignment corrects rollout-wide temporal shift and rate change, whereas constrained DTW additionally allows local stretching or compression while preserving the order.
As shown in Figure~\ref{fig:temporal_alignment} (top), Level~3 agreement increases to $0.086$ after global alignment and $0.104$ after constrained DTW, with similar gains under Levels~1 and~2.
The representative case in Figure~\ref{fig:temporal_alignment} (bottom) comes from a Wan2.2-I2V-A14B rollout of a stepped-domino cascade.
The generated rollout exhibits both an overall timing mismatch and local replay of an earlier motion phase.
Affine alignment captures the broad temporal shift and rate mismatch, whereas constrained DTW can additionally follow local timing variations.
The remaining mismatch shows that flexible temporal alignment can recover correspondence between some motion phases without restoring the correct physical progression of the cascade.
Detailed formulations and alignment constraints are in Appendix~\ref{app:temporal_alignment}.

Overall, future-event specification improves recovery of \emph{what} should happen, while temporal alignment explains some mismatch in \emph{when} it happens. Yet neither closes the gap in faithfully realizing \emph{how} the causal cascade unfolds.

\subsection{Can models selectively update causal futures under interventions? (RQ2)}
\label{sec:rq2}


Using the counterfactual evaluation protocol defined in Section~\ref{sec:evaluation_protocol}, RQ2 examines whether current I2V models selectively update their generated futures under controlled physical interventions.

\begin{figure}[t]
    \centering
    \includegraphics[width=\linewidth]{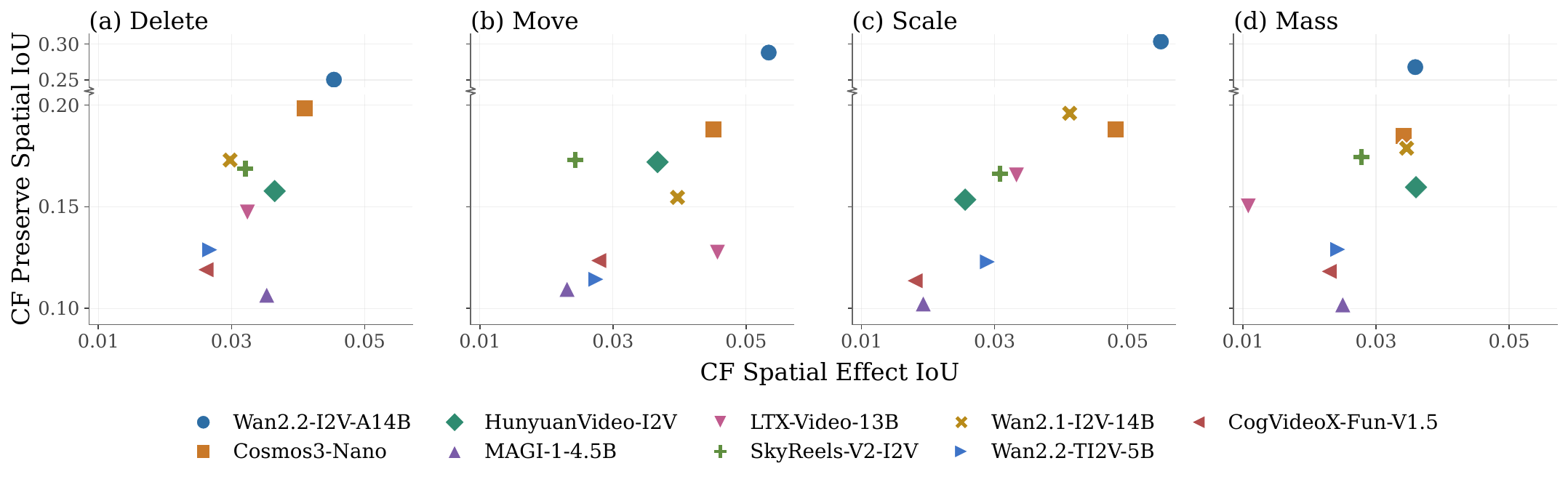}
    \caption{\textbf{Counterfactual motion response across intervention types.}
Each marker denotes one I2V model.
Higher values on both axes indicate stronger selective causal responsiveness.}
\vspace{-8pt}
    \label{fig:rq2_counterfactual}
\end{figure}

\textbf{Finding 4: Models struggle to realize the consequences of interventions.}
Figure~\ref{fig:rq2_counterfactual} plots CF Spatial Effect IoU against CF Preserve Spatial IoU for all nine models. Across the four types, models remain low in Effect performance, indicating that they rarely respond to changes in the physical state by producing the corresponding pattern of suppressed and induced motion. Inspection of complete factual-counterfactual rollouts reveals recurring manifestations of this behavior, including pairs that evolve nearly identically despite a consequential intervention and pairs that change but not according to the reference consequence.
Representative cases are analyzed in Appendix~\ref{app:cf_cases_effect}.

\textbf{Finding 5: Intervention responses are not sufficiently selective.}
Preserve performance is also limited across all four intervention types, with the highest CF Preserve Spatial IoUs remaining around $0.25$--$0.30$.
Thus, even when models respond to an intervention, they often fail to confine the response to the affected dynamics while preserving motion shared by the reference pair.
The case studies in Appendix~\ref{app:cf_cases_selectivity} further illustrate responses that disturb dynamics which should remain unchanged.

Weighted-spatial metrics exhibit the same pattern; complete results are reported in Appendix~\ref{app:cf_full_results}.
Collectively, current I2V models do not yet reliably update a generated future in accordance with interventions to the underlying physical state.

\subsection{Where does causal generation actually break? (RQ3)}

\begin{wrapfigure}[18]{l}{0.5\textwidth}
    \centering

    \includegraphics[width=\linewidth]
    {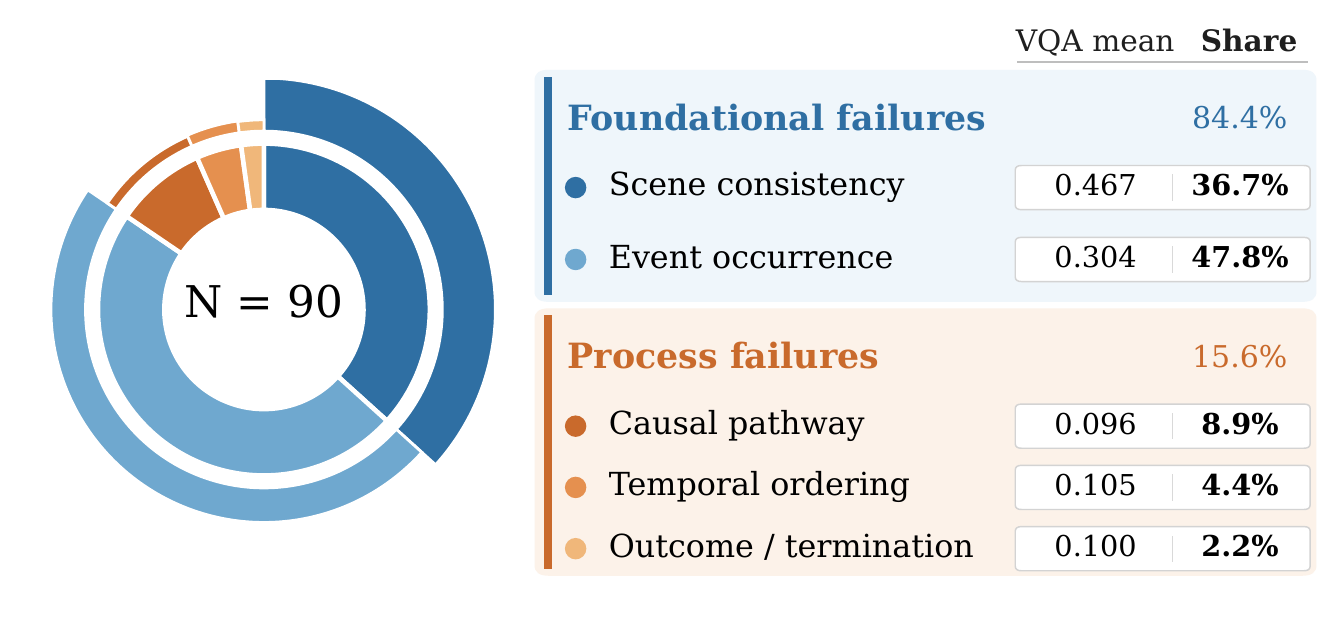}

    \caption{
    \textbf{Earliest primary failure distribution.}
    Angles show earliest-failure shares, while radial extents show mean dimension-level VQA scores across the audited subset.
Dimensions are grouped into foundational and process failures; the right panel reports exact values.
    }
    \label{fig:rq3_failure_distribution}

    \vspace{-6pt}
\end{wrapfigure}

Dimension-level VQA results already reveal uneven weaknesses across the causal process: scene consistency is relatively stronger, while event occurrence remains limited and causal pathway, temporal ordering, and outcome or termination are substantially lower (Table~\ref{tab:vqa_dimension_levels}).
However, these aggregate scores do not reveal where an individual rollout first departs from the reference process or how the resulting failures manifest.
We therefore combine structured failure localization on a fixed audit subset with broader qualitative case analysis.

\textbf{Finding 6: Most audited failures are foundational.}
For quantitative failure localization, we audit a fixed scene-level subset of 90 rollouts with matched scene-prompt coverage across all nine models, without selecting cases based on individual model success or failure.
We group scene consistency and event occurrence as \emph{foundational failures}, and causal pathway, temporal ordering, and outcome or termination as \emph{process failures}.
As shown in Figure~\ref{fig:rq3_failure_distribution}, 84.4\% of earliest failures are foundational, while only 15.6\% first occur at the later process dimensions.
Thus, the dominant bottleneck lies in establishing a persistent physical world and its event-level foundations before stable causal propagation can be sustained.
Detailed subset construction and annotation protocols are provided in Appendix~\ref{app:additional_qualitative_results}.

\begin{figure}[t]
    \centering
    \includegraphics[width=\textwidth]{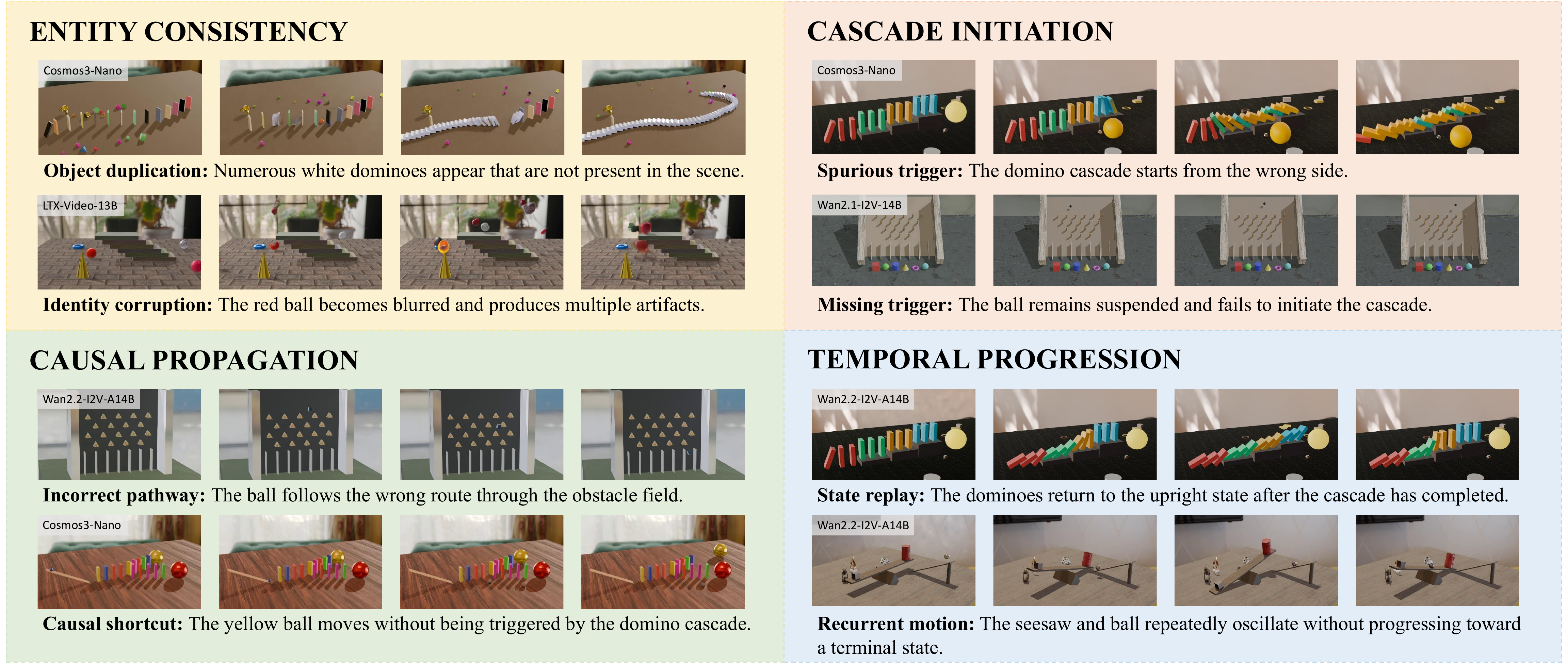}
    \caption{
    \textbf{Representative manifestations of foundational and process failures.} 
Examples of \emph{cascade initiation} and \emph{entity consistency} illustrate foundational failures, while examples of \emph{causal propagation} and \emph{temporal progression} illustrate process failures.
    }
    \label{fig:rq3_failure_cases}
\end{figure}
\textbf{Finding 7: Early deviations can reshape the downstream causal process.}
Broader qualitative inspection across the benchmark shows that once a rollout departs from the reference process, the resulting error need not remain local.
Figure~\ref{fig:rq3_failure_cases} provides representative examples of such departures.
Across these cases, an early deviation in the scene state or event process can change the conditions from which subsequent interactions unfold, causing later events to be omitted, misrouted, replayed, or otherwise deviate from the reference cascade.
These cases show that partial success does not guarantee sustained causal coherence: a local deviation can alter how later events unfold or whether the cascade remains physically consistent.
Detailed multilabel and downstream consequence analyses on the audited subset are provided in Appendix~\ref{app:recurring_failures}.

\section{Conclusion}

We introduced \benchmarkname, a simulator-grounded benchmark for evaluating causal cascade generation as a process-level capability of I2V models.
By combining structured event graphs, controlled future-event specification, physical interventions, and complementary motion-, event-, and counterfactual-level diagnostics, \benchmarkname examines three aspects of this capability: faithful realization of causal processes, selective responsiveness to changes in the physical world, and localization of where those processes break.

Across nine I2V models, we find that generated motion is more reliably localized spatially than realized with faithful spatiotemporal evolution, while richer future-event specification improves event recovery but leaves physical realization limited.
Counterfactual evaluation further shows that intervention responses are weakly selective: models struggle both to realize affected dynamics and to preserve those that should remain unchanged.
Finally, most audited earliest failures arise at foundational scene-consistency or event-occurrence stages, indicating that sustained causal propagation often breaks before later causal dependencies can be faithfully realized.
Together, these findings show that causal video generation requires more than producing plausible events: models must maintain a coherent physical world, realize how causal effects unfold over time, and selectively propagate the consequences of physical changes.
We hope that \benchmarkname provides a useful testbed for measuring progress toward video-based world models with these capabilities.

\bibliography{iclr2027_conference}
\bibliographystyle{iclr2027_conference}

\clearpage
\appendix
\section{Scene Creation Details}
\label{app:scene_creation_details}

We construct 50 event-rich base scenes in Blender~5.0, comprising 954 rigid-body instances in total. 
The scenes are designed around multi-stage physical processes rather than isolated interactions and also contain causally consequential scene factors, such that interventions on selected objects can alter downstream interactions and outcomes, enabling meaningful controlled counterfactual evaluation. 

To retain precise control over geometry and physical behavior, all dynamic objects and core mechanical components are constructed from simple geometric primitives in Blender, including cubes, spheres, cylinders, and meshes derived from these primitives. More complex mechanisms are assembled from combinations of these primitives; for example, a seesaw consists of a rectangular plank, a triangular support, and a hinge constraint. We assume uniform object density and set the center of mass of each rigid body to its geometric center to reduce ambiguities introduced by irregular asset geometry or uncertain mass distributions. External 3D assets are used only occasionally as decorative context to increase visual complexity.

Building on the primitives, we construct scene-specific instances of several recurring physical interaction motifs. We select these motifs because they pose challenges for causal dynamics generation.  Interaction-rich motifs, such as domino chains and Galton boards, involve dense sequences of contacts and collisions that are challenging to predict and track; instability-driven motifs, such as unbalanced seesaws and unstable structures, require reasoning about support relations or mass distribution from the initial configuration before predicting the subsequent evolution. These motifs serve as high-level organizational patterns rather than fixed scene templates. For each scene, we adapt their geometry, topology, scale, and object arrangement to produce a distinct instance. For example, a conventional linear domino chain may be transformed into an S-shaped, L-shaped, or Y-shaped branching  configuration, while a seesaw may use an off-center fulcrum or more complex object arrangements on its two sides.

\begin{wrapfigure}{R}{0.52\textwidth}
    \centering
    \includegraphics[width=\linewidth]{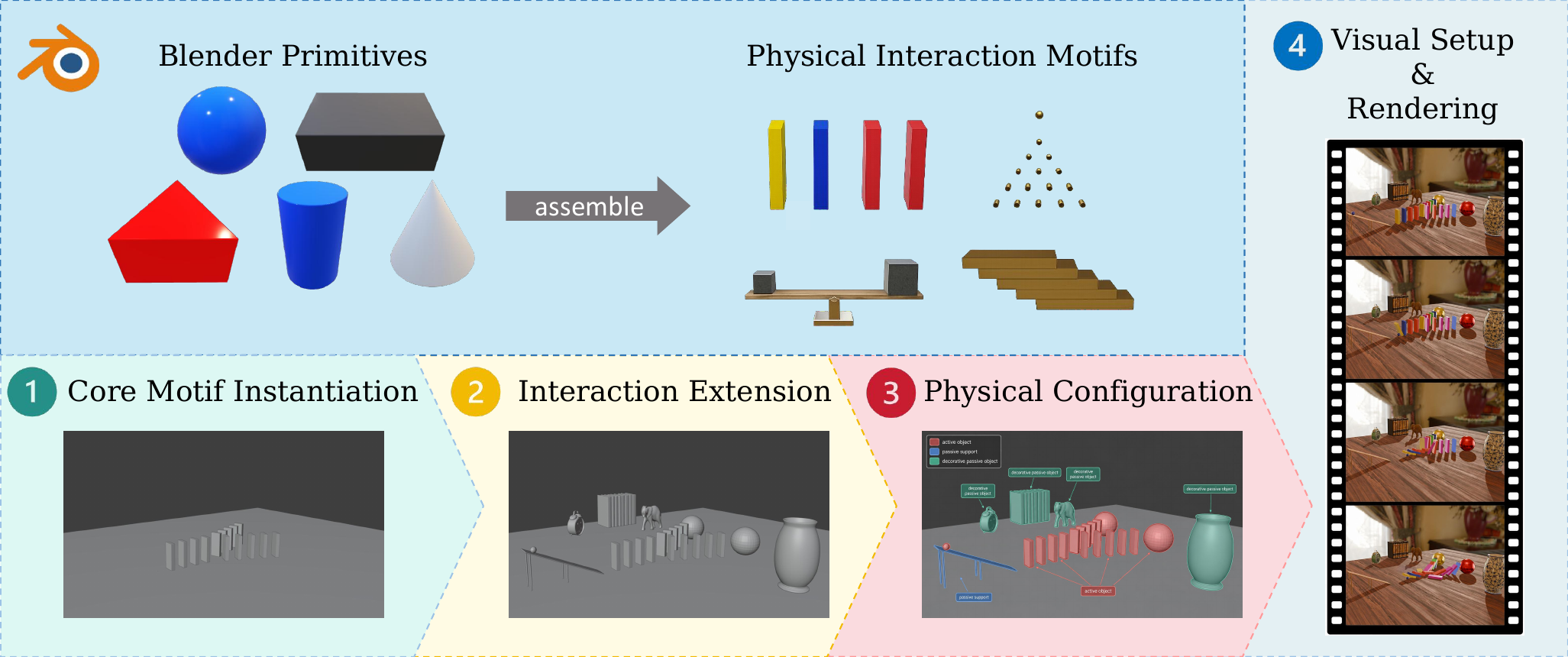}
    \caption{\textbf{Construction pipeline of each scene.}}
    \label{fig:dataset_curation}
\end{wrapfigure}
As shown in Figure~\ref{fig:dataset_curation}, the construction process comprises four stages. Each scene is constructed around such motif instances, which serve as the core interaction pattern of the scene. First, we establish a supporting environment, typically a tabletop or ground plane, and construct the selected motif instance from primitive-based components. Second, we optionally introduce upstream and downstream objects to extend the interaction beyond the core motif. Upstream objects provide the initiating condition, such as a ball rolling down a ramp and striking the first domino, whereas downstream objects transform the output of the motif into further collisions, falls, or other mechanical responses. We may also add a small number of stationary decorative objects to increase visual complexity. Third, we assign physical roles to all scene entities, distinguishing active rigid bodies from anchored passive supports and decorative objects. Finally, we assign PBR materials and HDRI environments selected from Poly Haven and position a fixed camera such that the causally relevant objects and their anticipated motion paths remain visible with minimal occlusion throughout the rollout.

After scene construction, we run the rigid-body simulation and render the resulting rollout at 24 frames per second. The rollout length is determined separately for each scene to ensure that all interactions have occurred and the objects have reached a stable state. We render each video at a resolution of $1920\times1080$ using Eevee. For every base scene, we export the initial RGB frame $I_0$, the complete ground-truth rollout $V^*$, and structured metadata describing the scene entities and their physical properties.  

\section{Event Graph and Diagnostic QA Details}
\label{app:event_graph}
\paragraph{Graph representation.}
Each scene is represented as a directed acyclic graph
\begin{equation}
    G^*=(V,E),
\end{equation}
where $V$ contains observable physical events and $E$ contains direct causal dependencies. Temporal order is represented by node attributes rather than additional temporal edges. Each event is represented as
\begin{equation}
    e_i=(d_i,P_i,\tau_i,r_i,s_i,o_i),
\end{equation}
where $d_i$ is a canonical natural-language description, $P_i$ is the set of participating entities, $\tau_i$ is the event type, $r_i$ is its role in the cascade, $s_i$ is a coarse temporal stage, and $o_i$ denotes observability. The corresponding serialized fields are \texttt{event\_id}, \texttt{canonical\_description}, \texttt{participants}, \texttt{event\_type}, \texttt{cascade\_role}, \texttt{temporal\_stage}, and \texttt{observability}.

Each graph additionally contains a participant registry that associates visually grounded entity descriptions with simulator metadata. A participant stores a stable graph-local identifier, canonical name, corresponding Blender metadata keys, participant type, and a visual referring expression. Repeated structures such as domino chains, peg arrays, and stacks are represented at the group level whenever their members share the same physical role. Individual members are separated only when they play a distinct causal role, such as an initiating domino, terminal target, or critical support. This keeps event granularity consistent across scenes with different numbers of repeated objects.

\paragraph{Event selection.}
An event is included only if it is visually observable from the RGB sequence, sufficiently atomic to be independently verified, relevant to the main causal cascade, diagnostically informative, and represented at a granularity consistent with other scenes. Event types include motion onset, contact/collision, support loss, tipping or rotation, falling, sliding, rolling, propagation, target-state change, and stabilization. Each event is assigned one of four cascade roles: \texttt{trigger}, \texttt{mediator}, \texttt{outcome}, or \texttt{terminal\_state}. Minor unrelated motion, repeated contacts, background variation, and changes that do not affect the principal cascade are excluded.

We use discrete temporal stages
\begin{equation}
    s(e_i)\in\{1,\ldots,K\},
\end{equation}
which encode reliable coarse ordering without requiring exact frame timestamps. Events assigned to the same stage may occur approximately simultaneously or overlap substantially in time.

\paragraph{Causal edges.}
Each directed edge
\begin{equation}
    e_i \xrightarrow{\texttt{leads\_to}} e_j
\end{equation}
indicates that $e_i$ is a direct physical cause or necessary enabling event for $e_j$. An edge is included only when the temporal stages are compatible, the rendered dynamics and scene mechanism support the dependency, and no already annotated mediator provides a more direct explanation. Pure temporal precedence is insufficient. We retain only direct dependencies rather than their transitive closure; for example, if $e_1\!\rightarrow\!e_2$ and $e_2\!\rightarrow\!e_3$ are present, $e_1\!\rightarrow\!e_3$ is normally omitted. We use the single relation type \texttt{leads\_to} throughout to avoid subjective distinctions between notions such as ``causes'' and ``enables.''

\paragraph{Annotation pipeline.}
For each base scene, we load the Blender scene metadata, initial rendered frame, and complete simulator-rendered frame sequence.
We sample every 10th frame and additionally include the final frame when it is not selected by the regular stride, arranging the sampled frames chronologically into review contact sheets.
A multimodal language-model \texttt{gpt-5.4-pro} API receives these visual summaries together with the annotation criteria defined above and produces an intermediate structured scene specification.
Each specification is then manually reviewed against the simulator-derived reference rollout and scene metadata, with event descriptions, participant assignments, temporal stages, cascade roles, and direct causal dependencies corrected where necessary.
Only the reviewed specification is deterministically converted into the final \texttt{event\_graph.json}.

\paragraph{Validation and statistics.}
Exported graphs are automatically checked for unique event identifiers, valid participant and edge references, required event fields, temporal consistency, duplicate edges, directed cycles, unsupported relation types, and redundant transitive edges. Across the 50 base scenes, the resulting graphs contain 267 event nodes and 226 direct causal edges. The nodes comprise 58 triggers, 94 mediators, 65 outcomes, and one terminal-state event for each of the 50 scenes.

\subsection{Graph-grounded Diagnostic QA}
\label{app:diagnostic_qa}

Diagnostic questions are deterministically generated from the annotated event nodes, causal edges, temporal stages, and participant registry using fixed templates. Each question retains provenance information linking it to its source events, edges, stages, and participants. All questions use closed answer spaces, enabling deterministic scoring and direct attribution of errors to specific aspects of the causal cascade.

\paragraph{Event occurrence.}
For every trigger and mediator event, we generate a binary question asking whether the event occurs in the evaluated video. This dimension tests whether the cascade is initiated correctly and whether intermediate propagation stages are realized. Outcomes are evaluated separately below.

\paragraph{Temporal ordering.}
Temporal questions are constructed from event pairs with meaningful direct causal relationships rather than arbitrary pairs. The answer space is
\begin{quote}
\small
A occurs before B; 

B occurs before A; 

A and B occur approximately simultaneously; 

one or both events do not occur.
\end{quote}
The final option prevents the evaluator from assigning a temporal relation when one of the required events is absent. For simple graphs we retain all eligible pairs; for more complex graphs we sample up to approximately six pairs distributed across early, middle, and late portions of the cascade.

\paragraph{Causal pathway.}
Each direct causal edge generates one four-way question:
\begin{quote}
\small
A directly leads to B; 

B directly leads to A; 

both events occur but no direct causal relation is visible; \

one or both events do not occur.
\end{quote}
This distinction separates correct propagation, reversed causal direction, missing direct mediation, and missing endpoint events. Consequently, reaching the correct final state without realizing an annotated intermediate causal pathway does not receive full pathway credit.

\paragraph{Outcome and termination.}
We evaluate up to three principal downstream outcomes per scene with binary occurrence questions. Each scene additionally contains one terminal-state question with three possible answers: the reference final state is reached and remains terminal; it is reached but subsequent motion violates terminality; or it is not reached. This distinguishes correct completion from both premature failure and continued dynamics after an otherwise correct outcome.

\paragraph{Scene consistency.}
Each scene contains four complementary consistency questions. \emph{Entity persistence} detects unexplained disappearance or duplication while allowing normal motion out of view. \emph{Trajectory continuity} detects teleportation or repeated spatial discontinuities while allowing temporary occlusion. \emph{Identity and geometry integrity} detects persistent identity changes, rigid-body deformation, fragmentation, or fusion while allowing ordinary viewpoint and illumination variation. \emph{Static-scene and camera stability} checks that anchored structures and the fixed camera remain stable, while permitting objects explicitly marked as dynamic to move.

\paragraph{Option-order balancing.}
For temporal-order and causal-pathway questions, the presentation order of events A and B is deterministically chosen from a SHA-256 hash of the scene identifier, edge identifier, and question type. This prevents the correct relation from being systematically associated with a fixed answer position while remaining fully reproducible. Among the temporal-order questions, 50.3\% have A-before-B as the ground-truth relation, 40.2\% have B-before-A, and 9.5\% describe approximately simultaneous events. And among the causal-pathway questions, the two causal directions account for 44.2\% and 55.8\%, respectively.

\paragraph{Question statistics.}
The final diagnostic set contains 872 questions across the 50 base scenes, distributed as shown in Table~\ref{tab:diagnostic_qa_stats}. The number of questions per scene varies according to graph complexity rather than being artificially equalized.

\begin{table}[t]
\centering
\caption{\textbf{Distribution of the diagnostic questions.}}
\label{tab:diagnostic_qa_stats}
\begin{tabular}{lr}
\toprule
Diagnostic dimension & \# Questions \\
\midrule
Event occurrence & 152 \\
Temporal ordering & 179 \\
Causal pathway & 226 \\
Outcome and termination & 115 \\
Scene consistency & 200 \\
\midrule
Total & 872 \\
\bottomrule
\end{tabular}
\end{table}

\subsection{Hybrid Video-QA Evaluation Protocol}
\label{app:hybrid_vqa_protocol}

For a diagnostic question $q$ with answer space $\mathcal{A}_q$, the evaluator receives the generated rollout $\hat V$, its simulator reference $V^*$, and the question and answer options, and returns
\begin{equation}
    \hat a_q = J(\hat V,V^*,q,\mathcal{A}_q).
\end{equation}

\paragraph{Judge model and execution harness.}
The semantic branch uses the vision-capable DeepSeek-V4.1-Flash. Each API turn uses temperature $0.2$, a maximum response budget of $32{,}768$ tokens, and enabled reasoning with a $24{,}576$-token budget. Each rollout-question pair is executed as an isolated session in a native tool-calling harness. 

Within each session, the judge can adaptively query the evidence tools summarized in Table~\ref{tab:hybrid_vqa_tools}. The interface separates direct visual inspection from computational diagnostics: \texttt{observe} provides global temporal context; \texttt{zoom}, \texttt{temporal\_pair}, and \texttt{paired\_zoom} resolve event boundaries and permit focused within- or across-rollout comparison; \texttt{flow\_info} and \texttt{segments} localize periods of activity; and \texttt{tracks}, \texttt{align}, and \texttt{traj\_diff} provide coarse object-motion and cross-rollout trajectory evidence. The image-based tools supply the primary semantic evidence, whereas the motion and tracking diagnostics guide attention and test candidate interpretations without determining the answer themselves. Every result is returned to the same session, allowing the judge to refine subsequent spatial crops or temporal windows in response to the evidence already observed.

\begin{table}[H]
\centering
\caption{Tools exposed to the semantic judge.}
\label{tab:hybrid_vqa_tools}
\small
\setlength{\tabcolsep}{3.5pt}
\renewcommand{\arraystretch}{1.5}
\begin{tabular}{@{}>{\raggedright\arraybackslash}p{0.18\textwidth}>{\raggedright\arraybackslash}p{0.30\textwidth}>{\raggedright\arraybackslash}p{0.47\textwidth}@{}}
\toprule
\textbf{Tool} & \textbf{Inputs and sampling} & \textbf{Evidence supplied to the judge} \\
\midrule
\texttt{observe} & Generated or reference video & A $5\!\times\!4$ contact sheet of 20 uniformly spaced frames, including the first and final frames. Cells are rendered at 336 pixels before the sheet is capped at 1680 pixels, and each cell carries its frame index and physical timestamp. \\
\texttt{zoom} & Video, interval, optional box & A high-resolution crop of up to 12 requested times; without explicit times, approximately ten samples cover the interval. The box accepts normalized or pixel coordinates.\\
\texttt{temporal\_pair} & Video, exact times, optional box & A two-column before/after comparison from one rollout, used for subtle changes in identity, support, orientation, and terminal state. \\
\texttt{paired\_zoom} & Interval or exact times & Generated and reference crops at the same nominal times, arranged in two rows; eight times are used by default. It identifies the intended object or endpoint but does not compute a verdict. \\
\texttt{flow\_info} & Generated or reference video & A global motion timeline from OpenCV DIS optical flow (medium preset) at $0.5\times$ spatial scale, including the 95th-percentile magnitude and candidate active windows.\\
\texttt{segments} & Generated or reference video & Active intervals whose flow exceeds the within-video 0.30 quantile for at least 0.25 seconds, used to select focused inspection windows. \\
\texttt{tracks} & Video side and optional overlay & Generic adaptive-background blob tracks at $0.5\times$ scale. Components occupy $0.2\%$--$25\%$ of the frame and are greedily associated with IoU $\geq0.10$ and centroid distance $\leq0.08$ of the image diagonal. Frames with foreground masks covering more than $35\%$ of the image are flagged as corrupted.\\
\texttt{align} & No additional input & Candidate generated/reference blob-track pairs, alignment coverage, mean/max spatial residual, and DTW cost. It indicates coarse structural correspondence and is unavailable when no track pair can be established. \\
\texttt{traj\_diff} & No additional input & For the best candidate pair, the aligned trajectory residual normalized by frame width; residuals $\leq0.035$, $0.035$--$0.10$, and $>0.10$ are reported as coarse consistent, partial, and mismatch diagnostics, respectively. \\
\bottomrule
\end{tabular}
\end{table}

\paragraph{Prompt construction and decision contract.}
The system prompt defines a literal event-verification policy. It specifies that every named participant, initial condition, direction, contact or support relation, repetition count, and endpoint is a required part of an event. The reference rollout may be used to identify the intended participant and relation, but cannot be copied as the verdict for the generated rollout. The prompt also distinguishes direct causation from temporal precedence, normal exit through an image boundary from disappearance, foreground motion from camera motion, and a briefly reached outcome from a terminal state that remains stable. 

The response contract is a JSON object containing exactly one supplied answer option, a \texttt{high}/\texttt{medium}/\texttt{low} self-assessed confidence, a two- to three-sentence evidence summary, and the tools supporting each claimed visual fact. For exact reproducibility, the complete fixed system instruction used by the semantic judge is provided below.
\newpage
\begin{promptblock}
\begin{Verbatim}
You are a careful comparative video evaluator for a physical causal-cascade benchmark. Judge
the GENERATED rollout. The GROUND-TRUTH rollout is a visual reference for the intended named
participants and relation only; it is not an answer key and must not be copied.

Read the supplied question literally. Treat every noun, modifier, direction, initial
condition, contact/support relation, count, repetition, and endpoint as required evidence.
An event is present only when its named participant(s) and the specified transition are
visibly recognizable in the generated video. A related motion, proximity, plausible later
pose, or motion of a different object is not a substitute. Conversely, minor blur or a brief
occlusion does not invalidate an otherwise recognizable transition.

Evidence protocol (use one tool call at a time):
1. Start with a whole-video contact sheet of GEN and then GT. Use the timestamps printed on
   the frames to locate the generated event window. Then inspect GEN at high-resolution zooms
   around the onset and endpoint of every named event. For persistence and terminal questions,
   inspect the first two seconds, the transition, and the last fifth; a short early anomaly can
   be decisive. Use `temporal_pair` to compare the same GEN object/support at exact before/after
   times when a shape, count, orientation, or terminal interval is subtle. Use paired GEN/GT
   zoom only to identify the intended object or endpoint, never to transfer the GT verdict to GEN.
2. Before selecting an answer, make a silent event checklist. For each named event record:
   present/absent, participant identity, required starting configuration, direction, required
   contact or support, all repeated/count endpoints, and completion. If any required component
   is absent or contradicted, the event is absent even if an adjacent or physically related
   action occurs.
3. For yes/no event questions, answer yes only when the complete literal event is visible.
   In particular, verify the starting orientation before motion: a plank that is not initially
   lowered on the stated side, an object that is already leaning but does not initiate the row,
   or two objects moving together without the stated lead/order does not establish the event.
   Check image-left/image-right endpoints directly rather than inferring direction from gravity
   or from the reference.
4. For temporal relations, first verify both complete events independently, including every
   named downstream object. Only after both checklists pass compare their onsets/completions.
   If either event is incomplete or absent, select the explicit absence option. Do not turn a
   partial domino chain, a ball entering a peg field, or a nearby later pose into a complete
   event. If both are present, use their visible temporal order, not expected physics.
5. For direct causal relations, event presence is a prerequisite. Require a visible contact,
   support change, or state/force transfer tied to the downstream onset. Mere precedence,
   co-directional motion, or a physically plausible explanation is insufficient. A brief
   recognizable contact may count if one frame is occluded, but a missing named endpoint or
   incomplete compound event forces the absence option. When one event is a necessary
   consequence of the other, still verify that the stated direction and endpoint actually occur.
6. For outcome and terminal-state questions, search the complete timeline for the first
   post-transition interval in which all stated conjuncts hold. Do not count the initial setup
   frame as a reached final outcome when the same condition is present before any relevant
   motion; evaluate it after the first process transition. If all conjuncts become true briefly
   and then a collapse, replay, black void, disappearance, or regression occurs, choose
   reached-but-does-not-remain-terminal. If no complete post-transition interval exists, choose
   not-reached. Inspect adjacent frames around the boundary: if the moving object has just
   exited and the support is unchanged for even a short recognizable interval before corruption
   begins, that interval counts as reached-but-does-not-remain-terminal. Do not judge from the
   last frame alone.
7. For anchored scene/camera questions, infer camera motion only from static background
   landmarks. Foreground seesaw or ball motion is not camera motion. Separately compare the
   named fixed support assembly (board, rail, plank, pivot, pegboard, or table) at the start,
   through the transition, and at the end. Any visible unexpected deformation, angle/placement
   change, appearance, disappearance, or identity change is an anchored-structure failure,
   even when the wall and camera remain fixed. If both camera and support change, use the
   combined option.
8. For identity/geometry, persistence, and trajectory questions, inspect all semantic objects
   rather than relying on one large object. Follow each small participant during the first two
   seconds as well as the final transition. A brief unsupported extra instance, reappearance,
   split/merge, or identity/appearance switch counts even if it later disappears. An object
   that visibly travels to and through the frame edge is not unexplained disappearance. For
   trajectory continuity, a visible object-level spatial jump, sudden reset to an earlier
   position, or multiple such jumps counts even when optical flow misses it; a tracker fragment
   or flow spike without a matching visual jump is not enough. If a scene collapse makes the
   named object disappear and reappear at a different position, treat that as a semantic
   discontinuity, not merely tracker noise.
9. Generic flow, tracks, alignment, and segments are diagnostic hints only. Use them after
   visual crops have established the candidate event, and never let their failures override a
   clear visual observation. In particular, color/texture changes, occlusion, camera view
   shifts, and background corruption can break pixel tracks without changing object identity.

When the answer options contain a combined or "multiple" category, use it whenever two or more
distinct violations are visibly established (for example, an appearance/identity switch plus
a separate rigid-body deformation, or both duplication and disappearance). Do not reduce a
clearly compound failure to whichever single symptom is most salient. For persistence and
identity questions, inspect the bottom/side object groups at close range; a central small
participant that changes shape, color, or position during the first two seconds is still an
identity/duplication event even if the large board remains unchanged.

Before answering, re-check every option against the generated video alone and resolve the most
specific literal condition first. Return exactly one of the supplied answer options in the
required JSON format:
{"answer":"<one option>","confidence":"high|medium|low","reasoning":"2-3 sentences",
 "evidence_tools":[{"tool":"<name>","claimed_fact":"<direct visual fact>"}]}
\end{Verbatim}
\end{promptblock}

\paragraph{Evidence loop and vote aggregation.}
The judge is instructed to begin with whole-video observations of the generated and reference rollouts, then request local evidence around the onset and endpoint of each named event. For persistence and termination it also inspects the first two seconds and final fifth of the generated rollout. Each tool call is executed by the harness, and its textual output and base64-encoded image are returned as a native tool result in the same session.

We run three independent sessions with the same model, prompt, parameters, and tool interface, and take a strict majority over normalized options. Three identical votes receive high vote consistency, a $2/3$ majority receives medium consistency, and an invalid vote or three-way disagreement receives low consistency.

\paragraph{Object-mask extraction for scene consistency.}
Scene-consistency judgments are particularly susceptible to misinterpretation by vision-language models because they require tracking subtle object-level changes over time. Small or repeated objects, temporary occlusion, appearance variation, and foreground motion can lead a semantic judge to report spurious duplication, disappearance, identity change, or camera movement. We therefore use a deterministic object-centric branch that grounds the four subdimensions in object lifecycle, mask geometry, camera-compensated trajectories, and background stability. This design makes each prediction temporally localized and auditable while applying consistent criteria across models and prompt levels.

The branch evaluates entity persistence and identity/geometry at nine uniformly spaced frames, trajectory continuity at 13 frames, and static-scene/camera stability over the complete rollout downsampled to $0.20\times$ resolution. Entity persistence retains at most 16 initial objects and enables sparse re-detection; identity/geometry retains at most 12; trajectory retains at most six objects selected by activity in the simulator reference. Reference activity is used only to prioritize initial participants and never supplies the generated-video answer. After establishing object identities, the pipeline records their temporal geometry and separates object motion from camera motion.

For object $i$ at sampled time $t$, the pipeline records normalized mask area $a_{it}=|M_{it}|/(HW)$, bounding-box fill and aspect ratio, connected-component count, HSV-histogram similarity to the initial mask, normalized centroid $c_{it}$, and visibility. A propagated object is visible when $a_{it}\geq0.18a_{i0}$. Pairwise camera motion is estimated from ORB features outside the union of all propagated object masks. Descriptor matches pass a 0.75 ratio test and at least eight matches are required; an affine transform is then fitted with RANSAC using a 3-pixel reprojection threshold. The transforms are accumulated and inverted to produce camera-compensated centroids $\tilde c_{it}$.

The step length is $d_{it}=\|\tilde c_{i,t+1}-\tilde c_{it}\|_2$. An object-specific jump threshold is
\begin{equation}
    \tau_i^{\mathrm{jump}}=\max\!\left(0.045,\;\operatorname{med}(d_i)+4\operatorname{MAD}(d_i),\;1.5\sqrt{a_{i0}}\right).
\end{equation}
The extractor also detects a state revisit when the compensated path length exceeds both 0.12 and three initial-object scales, the farthest displacement exceeds 2.5 scales, and the final displacement is less than 0.45 of the path length. An apparent death requires absence for at least two sampled frames; it is rejected as a normal boundary exit when the last visible centroid lies within 0.04 of an image boundary. Appearance change requires histogram similarity below 0.45 in at least two samples, while geometry change requires at least two samples whose area ratio is outside $[0.42,2.4]$ or aspect-ratio change is outside $[0.5,2.0]$. These measurements are retained for diagnosis, while the final answer is produced by the fixed option-projection rules described below.

\paragraph{Sparse re-detection and lifecycle evidence.}
Mask propagation alone cannot increase the number of fixed IDs initialized at $t=0$ and therefore cannot detect a genuinely new instance. For entity persistence, the system independently re-runs segmentation at an early sample, the midpoint, the penultimate sample, and the final sample. Initial instances are grouped by HSV appearance, area, and aspect ratio using descriptor similarity at least 0.78; re-detected masks are assigned to an initial instance at similarity at least 0.72. A repeated structure with at least four initial instances tolerates one extra segmentation component. A birth is retained when the same semantic cluster has an excess of at least two instances in at least two re-detection samples. Two terminal safeguards also retain a birth when multiple clusters grow and at least seven masks are unmatched, or when a positive count change co-occurs with at least ten unmatched terminal masks. A fusion/split candidate requires a tracked object of initial area at least 0.005 to shrink below 0.82 of its original area in at least two of its last three observations while an unmatched mask appears within normalized distance 0.12. These persistence requirements suppress one-frame mask fragmentation and color-only changes.

\paragraph{Static-scene and camera measurements.}
The static branch computes Farneb\"ack optical flow between consecutive frames at $0.20\times$ scale and records median, 90th-, and 99th-percentile magnitudes and spatial spread. It also computes absolute consecutive-frame differences over the full image, the outer 10\% border, the central 60\% region, and the four corner thirds, together with histogram and edge-density changes. Separately, ORB correspondences between the first frame and four points across the rollout are fitted with partial affine RANSAC; transform magnitude, inlier ratio, and post-transform residual are retained as audit features. The classifier uses two signals: median dense-flow magnitude above 0.10 denotes broad camera/background motion, and the 95th percentile of the mean absolute grayscale difference in the central region above 2.0 denotes an anchored-region change.

To convert these heterogeneous measurements into the benchmark's closed answer spaces, we apply a frozen rule layer. Each subdimension is first assigned one or more violation signals using only evidence from the generated rollout; these signals are then projected to the exact option wording, with combined outcomes retained whenever the answer set provides such an option. Table~\ref{tab:scene_consistency_rules} summarizes this deterministic decision layer.

\begin{table}[H]
\centering
\caption{Frozen decision rules and answer-option projection for scene consistency.}
\label{tab:scene_consistency_rules}
\small
\setlength{\tabcolsep}{3.5pt}
\renewcommand{\arraystretch}{1.5}
\begin{tabular}{@{}>{\raggedright\arraybackslash}p{0.20\textwidth}>{\raggedright\arraybackslash}p{0.36\textwidth}>{\raggedright\arraybackslash}p{0.40\textwidth}@{}}
\toprule
\textbf{Subdimension} & \textbf{Fixed violation signals} & \textbf{Exact-option projection} \\
\midrule
Entity persistence & Persistent multi-component mask, reliable sparse-re-detection birth, or terminal non-boundary death. & Neither signal --- consistent; death only --- unexplained disappearance; birth only --- unexplained duplication; both --- combined. \\
Identity and geometry integrity & A mask is split in at least two sampled frames, or a propagated identity is lost away from the boundary. & No signal maps to the first (preserved) option; any violation maps to the designated integrity-violation option.\\
Trajectory continuity & At least one compensated track satisfies the state-revisit test, or one semantic object has simultaneous multi-component paths. & No signal maps to spatial continuity; any violation maps to the multiple-discontinuity option. Jump and reversal statistics remain available for diagnosis. \\
Static scene/camera stability & Broad-motion signal $[\operatorname{median}(\mathrm{flow})>0.10]$ and anchored-region signal $[Q_{0.95}(\Delta_{\mathrm{centre}})>2.0]$. & Neither --- stable; anchored only --- structure change; broad motion only --- viewpoint change; both --- combined. \\
\bottomrule
\end{tabular}
\end{table}

\paragraph{Human validation.}
We validate the complete Hybrid evaluator on a question-definition-disjoint audit set of 100 generated-rollout questions, balanced to contain 20 questions from each diagnostic dimension and distributed across models and prompt levels. A human reviewer independently inspects the generated videos and selects an answer option, without being shown the evaluator prediction. We then compare this independent human answer with the evaluator prediction. Under this independent-annotation protocol, the Hybrid evaluator agrees with the human answer on 89 of 100 questions. The dimension-level results are reported in Table~\ref{tab:hybrid_vqa_validation}.

\begin{table}[H]
\centering
\caption{Agreement with independent human annotations on a 100-question audit set.}
\label{tab:hybrid_vqa_validation}
\small
\setlength{\tabcolsep}{3.5pt}
\renewcommand{\arraystretch}{1.5}
\begin{tabular}{lcc}
\toprule
Diagnostic dimension & Agreement & Accuracy \\
\midrule
Event occurrence & 17/20 & 85\% \\
Causal pathway & 19/20 & 95\% \\
Temporal ordering & 16/20 & 80\% \\
Outcome and termination & 20/20 & 100\% \\
Scene consistency & 17/20 & 85\% \\
\midrule
Overall & 89/100 & 89\% \\
\bottomrule
\end{tabular}
\end{table}

\subsection{Scoring Protocol}
\label{app:qa_scoring}

For each question $q$, we use exact-match accuracy,
\begin{equation}
    s_q=\mathbf{1}\!\left[\hat a_q=a_q^*\right],
\end{equation}
where $a_q^*$ is the reference answer and $\hat a_q$ is the predicted answer. To prevent scenes with larger event graphs from dominating the evaluation, scores are first averaged within each scene $c$ and diagnostic dimension $d$,
\begin{equation}
    S_{c,d}
    =
    \frac{1}{|Q_{c,d}|}
    \sum_{q\in Q_{c,d}} s_q,
\end{equation}
and then macro-averaged across scenes containing questions from that dimension,
\begin{equation}
    S_d
    =
    \frac{1}{|C_d|}
    \sum_{c\in C_d} S_{c,d},
\end{equation}
where $C_d$ denotes the set of applicable scenes.

For temporal ordering and causal pathway, we additionally distinguish failures caused by missing events from failures in the relation itself. \emph{Endpoint coverage} measures the fraction of relation questions for which both referenced events are present, while \emph{conditional relation accuracy} measures relation correctness only among those cases. Their product,
\begin{equation}
    S_d^{\mathrm{eff}}
    =
    \mathrm{Coverage}_d
    \times
    \mathrm{ConditionalAccuracy}_d,
\end{equation}
provides a complementary effective relation score that jointly reflects event realization and relation correctness.

\section{Prompt Templates and Examples}
\label{app:prompt_templates}

This section provides additional details of the prompt construction used for the three future-event specification conditions introduced in Section~\ref{sec:factual_eval}. 
Our prompt design aims to instantiate each future-event specification condition in a consistent and visually groundable form while keeping the prompts compact across models with different text encoders and context lengths.

\paragraph{Structured physical-state template.}
Levels~1 and~2 share the same structured representation of the current physical state. The first frame is treated as authoritative for visible geometry, object size, pose, contact relationships, spatial layout, and camera configuration. The textual prompt therefore supplements the image with physical roles and properties rather than redundantly describing the full scene geometry.

The structured prompt distinguishes \emph{dynamic active objects}, \emph{dynamic active object groups}, \emph{anchored static objects}, and \emph{decorative context}. Dynamic active objects are individually specified rigid bodies that may move, whereas anchored static objects are fixed to the world and decorative entities are instructed to remain visually unchanged.

To reduce prompt length in scenes containing many repeated objects, we introduce \emph{dynamic active object groups}. Objects sharing the same physical role and properties are represented collectively rather than enumerated individually. For example, a domino course can be specified by its visual organization, number of members, and per-object mass. This reduces redundant tokens while retaining the relevant physical information.

Each object or object group is assigned a human-readable reference name together with a short visual or spatial cue, such as color, shape, relative position, or relation to nearby scene components. These descriptions help associate textual physical properties with entities visible in the first frame without relying on simulator-specific object identifiers.

Shared physical assumptions are specified once under a \emph{Global rigid-body convention}. This includes common friction, restitution, linear damping, and angular damping, as well as the convention that the center of mass of each rigid body is set to its geometric center.

\paragraph{Construction of the three prompt levels.}
Level~1 contains only the structured physical-state prompt. Level~2 uses the identical prompt and adds a single \emph{Trigger event} field describing the initiating event. Neither level reveals the subsequent causal evolution or terminal outcome. Level~3 instead uses a standalone natural-language description of the complete reference event sequence, including the major future interactions and terminal state, without the structured physical-state specification used in Levels~1 and~2.

\paragraph{Level~3 future-event description construction with the aid of AI.}
Level~3 descriptions are generated from the simulator-rendered reference rollout using a multimodal language model.
For each scene, we read the rendered frame sequence and sample one frame every three frames, additionally including the final frame when it is not selected by the regular stride.
The sampled frames are arranged chronologically into contact sheets containing at most 20 frames each, with five columns and a tile width of 320 pixels.
The contact sheets are encoded as JPEG images with quality 90 and are provided to the model in temporal order.
Across the 50 base scenes, this procedure samples 3,174 frames from 9,380 rendered frames and produces 187 contact sheets, with two to six contact sheets per scene.
No scene metadata or complete video file is supplied to the model; the initial frame is naturally included as the first sampled frame.
We use \texttt{gpt-5.4-pro} API. 
The request consists of a single multimodal user message with no system message. The generated descriptions were manually inspected and proofread.
The textual instruction supplied before the contact-sheet images is:
\begin{quote}\small
The attached images are contact sheets sampled from one target physics video at regular intervals.
They are ordered chronologically by image, and within each image from left to right and top to bottom.
Write the final caption only; do not mention the contact sheet or sampling.

You are writing a description for a physics video.

Describe the full physical process in the video as a dense chronological event caption.
The caption will be used as the prompt for a video generation model, so it must explicitly reveal the causal event chain and final outcome.

Requirements:
\begin{itemize}
    \item Use no more than 512 tokens.
    \item Describe the scene layout, main objects, initial motion, key contacts/collisions, causal effects, temporal order, and final state.
    \item Focus on visible physical events and causal consequences.
    \item Do not mention uncertainty, frames, timestamps, or that you are analyzing a video.
    \item Write one fluent paragraph, not bullet points.
    \item In the end, add the sentence ``Keep the camera fixed.''
\end{itemize}
\end{quote}
\paragraph{Example.}
Figures~\ref{fig:prompt-level1}, \ref{fig:prompt-level2}, and~\ref{fig:prompt-level3} show the three prompt conditions for the same seesaw-mediated cascade.
Level~1 is shown in full, Level~2 displays only the additional initiating-event field because all remaining content is identical to Level~1, and Level~3 is shown in full as a standalone event-sequence description.

\begin{figure}[t]
\begin{promptblock}

Generate a physically realistic continuation video by predicting the subsequent physical evolution from the given fixed-camera first frame.

\promptheading{Important instructions:}
\promptitem{Preserve the fixed camera view.}
\promptitem{The first frame is authoritative for all visible geometry, object sizes, initial poses, contact relations, spatial layout, and camera view.}
\promptitem{Treat only the listed dynamic active objects and dynamic active object groups as freely moving rigid bodies.}
\promptitem{Treat anchored static colliders as fixed to the world.}

\promptheading{Global rigid-body convention:}
\promptitem{All rigid bodies use friction = 0.50, restitution = 0.10, linear damping = 0.04, angular damping = 0.10.}
\promptitem{For every rigid body, the geometric center is also its center of mass.}

\promptheading{Dynamic active objects:}
\promptitem{\textbf{small red seesaw ball:} the small red sphere resting near one end of the long lever on the left side of the tabletop scene. Mass = 2.0 kg.}

\promptitem{\textbf{blue seesaw ball:} the blue sphere on the long lever near the domino course. Mass = 50.0 kg.}

\promptitem{\textbf{long seesaw lever:} the horizontal plank-like lever carrying the two spheres. Mass = 20.0 kg.}

\promptitem{\textbf{large orange target sphere:} the large orange sphere positioned beyond the stepped domino row. Mass = 0.1 kg.}

\promptheading{Dynamic active object groups:}
\promptitem{\textbf{stepped domino row:} the six upright dominoes arranged across the lower step levels between the lever and the large target sphere. This group contains 6 dominoes. Each object has mass = 5.0 kg.}

\promptheading{Anchored static objects:}
\promptitem{\textbf{stepped platform:} the raised rectangular step supporting the domino row.}

\promptitem{\textbf{seesaw fulcrum:} the small fixed support underneath the long lever.}

\promptitem{\textbf{tabletop surface:} the broad surface supporting the whole scene.}

\promptheading{Decorative context:}
\promptitem{The rest of objects are anchored static decorators which should remain visually unchanged.}

\end{promptblock}

\caption{\textbf{Representative Level~1 prompt.} The prompt provides the structured physical state without specifying the initiating event or subsequent causal evolution.}
\label{fig:prompt-level1}
\end{figure}

\begin{figure}[t]
\begin{promptblock}

\promptheading{Trigger event:}

The small red ball rolls along the seesaw lever, tipping the lever and moving the blue ball.

\medskip

\textit{All remaining fields are identical to the Level~1 prompt.}

\end{promptblock}

\caption{\textbf{Additional trigger information provided in the representative Level~2 prompt.} Level~2 adds the initiating event while otherwise using the same structured physical-state specification as Level~1.}
\label{fig:prompt-level2}
\end{figure}

\begin{figure}[t]
\begin{promptblock}

In a room with a tan floor, a brown seesaw-like plank on a black triangular fulcrum sits on the left, holding a small red ball near the left side and a blue ball farther to the right; in the middle stands a short row of red domino blocks beside a black platform carrying several upright green bars, while a large orange ball waits motionless on the far right.

The plank slowly rotates so its right end drops, and both balls roll rightward along the board, with the blue ball leading. The blue ball reaches the lowered end first and hits the first red domino, starting a rightward domino cascade that tips the red blocks over in sequence and leaves them collapsed in a heap against the left side of the central block assembly.

The green bars stay standing and the large orange ball is never touched. After the blue ball stops near the fallen red pieces, the red ball continues sliding down the tilted plank toward the same end, catches up, and gently nudges the blue ball slightly to the right before both settle.

At the end, the plank remains slanted down to the right, the red dominoes are down, the green bars are still upright, the orange ball is unchanged, and the two small balls rest together near the right tip of the plank beside the block assembly. Keep the camera fixed.

\end{promptblock}

\caption{\textbf{Representative Level~3 prompt.} Unlike Levels~1--2, Level~3 provides a standalone natural-language specification of the complete event sequence, including the major interactions and terminal state.}
\label{fig:prompt-level3}
\end{figure}

\section{Counterfactual Variant Generation and Screening}
\label{app:edit_generation}
We construct controlled object-centric variants for all 50 base scenes using four edit types: \emph{delete}, \emph{move}, \emph{scale}, and \emph{mass}.
Candidate objects are identified from scene metadata, the event graph, and the simulator-derived reference rollout, with preference given to objects that participate in the physical process.
Each candidate modifies one object and one scene factor at a time, while all unrelated initial conditions are kept unchanged. Edits are represented as delta JSON files that can be applied to the original scene metadata and Blender file to reconstruct the edited scenes. Below, we describe the construction of each edit family, followed by the validation and screening procedure used to obtain the final controlled-variant dataset and RQ2 evaluation subset.

\subsection{Delete Edits}

Delete interventions remove a single causally relevant object from the initial scene. 

For standard scenes, candidate objects are ACTIVE rigid bodies with identifiable roles in the physical process, including balls, dominoes, movable blocks, supports, and other rigid bodies participating in the cascade.

In particular, Galton-board scenes require additional treatment because fixed pegs can also influence the downstream trajectory.
For each such scene, we select PASSIVE pegs that are contacted by, traversed by, or immediately adjacent to the observed reference trajectory.
For annotation and reconstruction, each peg is identified by its visual row and column within the array.

This procedure produces 450 delete candidates in total.

\subsection{Move Edits}

Move interventions change only the initial position of a selected object.
Rather than applying fixed offsets, we define candidate displacements relative to local scene geometry, object scale, and interaction structure.
Eligible objects are primarily ACTIVE rigid bodies participating in the event graph.
For long repeated structures, such as domino chains, we sample representative members near the beginning, approximately one-third and two-thirds along the structure, and near the end, rather than perturbing every member.

For isolated active objects, candidate displacements include:
(i) two movements toward the nearest causally interacting object, with magnitudes determined by object scale and current clearance;
(ii) movements away from that object by approximately $1.25\times$ and $2.0\times$ the characteristic object scale;
(iii) displacement along the local interaction normal toward the camera-facing side by approximately $1.0\times$ the object scale;
(iv) displacement toward the opposite side by approximately $1.5\times$ the object scale; 
(v) upward displacement by approximately $0.75\times$ the object height.

For members of chains or other repeated structures, the local tangent is estimated from neighboring members.
Candidates include lateral displacements of approximately $1.0\times$ and $1.75\times$ the transverse object width in opposite directions, as well as displacement away from a neighboring member by approximately $0.65\times$ the local inter-object spacing. 

Galton-board scenes additionally contribute 90 trajectory-aware PASSIVE peg moves.
For each scene, five pegs lying on the observed descent trajectory are shifted laterally out of the corresponding path by approximately $0.55\times$ the peg-column spacing. 

We generate 1,155 move candidates in total.

\subsection{Scale Edits}

Scale interventions modify the geometry of one causally relevant rigid body while preserving its mass and location. To avoid tying edit semantics to global Blender axes, we order the object's local mesh extents as \emph{shortest}, \emph{middle}, and \emph{longest}. Scale candidates are then defined according to the geometry class of the object.

For spheres, we use uniform factors $0.60$, $0.80$, $1.30$, and $1.70$.
For dominoes, candidates modify thickness by $0.55/1.65\times$, width by $0.65/1.50\times$, both short axes jointly by $1.45\times$, or height by $0.80/1.20\times$.
For cylinders, cans, and pegs, we modify the cross section by $0.65/1.45\times$, the principal axis by $0.75/1.25\times$, or apply uniform $0.75/1.35\times$ scaling.
For planks, seesaws, platforms, and rails, we vary length by $0.65/1.45\times$, width by $0.65/1.40\times$, thickness by $0.65/1.50\times$, or jointly enlarge the two longer axes by $1.25\times$.

For supported objects, scaling about the object center can otherwise displace the lower surface from its original support. We therefore compensate the object position after scaling so that its lowest world-coordinate point remains fixed. Objects that are initially airborne, falling, or intentionally centered about a pivot retain their original center position.

For Galton-board scenes, five trajectory-relevant passive pegs are selected per scene from the reference rollout. Each receives uniform $0.55\times$ and $1.55\times$ scale variants.

In total, we generate 2,091 scale candidates.

\subsection{Mass Edits}

Mass interventions modify only the mass of an ACTIVE rigid body while leaving its geometry, pose, and other rigid-body parameters unchanged.
For an object with original mass $m$, we define
\begin{equation}
    m' = r m,
    \qquad
    r = 10^z.
\end{equation}

For each of 408 eligible objects, we sample ten multiplicative factors: five that decrease the mass and five that increase it,
\begin{align}
    z_i^- &\sim U(-2,-\epsilon),
    && i=1,\ldots,5,\\
    z_i^+ &\sim U(\epsilon,2),
    && i=1,\ldots,5,
\end{align}
with $\epsilon=0.2$.

Sampling uniformly in log space provides broad coverage of relative mass changes while avoiding candidates arbitrarily close to the original value.
A fixed random seed is used for reproducibility.
Since no visible scene property is modified, the simulator-rendered initial RGB frame remains identical to that of the corresponding base scene.

This procedure produces 4,080 mass candidates.

\subsection{Automatic Validation and Manual Screening}

\paragraph{Automatic geometric validation.}
Before rendering, move and scale candidates undergo edit-specific geometric checks to remove invalid initial configurations.

For move edits, the edited object center must remain within the fixed camera frustum, the edit must not introduce a new initial intersection with another ACTIVE object, and Galton pegs must not overlap other pegs in the same array.
These checks reject 128 of the 1,155 move candidates, leaving 1,027 candidates for manual screening.

For scale edits, we require valid object extents, sufficient visibility within the fixed camera, no new initial ACTIVE/PASSIVE overlap, and, where support-preserving compensation is applied, a displacement of the lowest world-coordinate point below $10^{-3}$.
These checks reject 153 of the 2,091 scale candidates, leaving 1,938 candidates for manual screening.

Delete and mass edits do not require an analogous geometric filtering stage.

\begin{figure}[t]
    \centering
    \includegraphics[width=\linewidth]{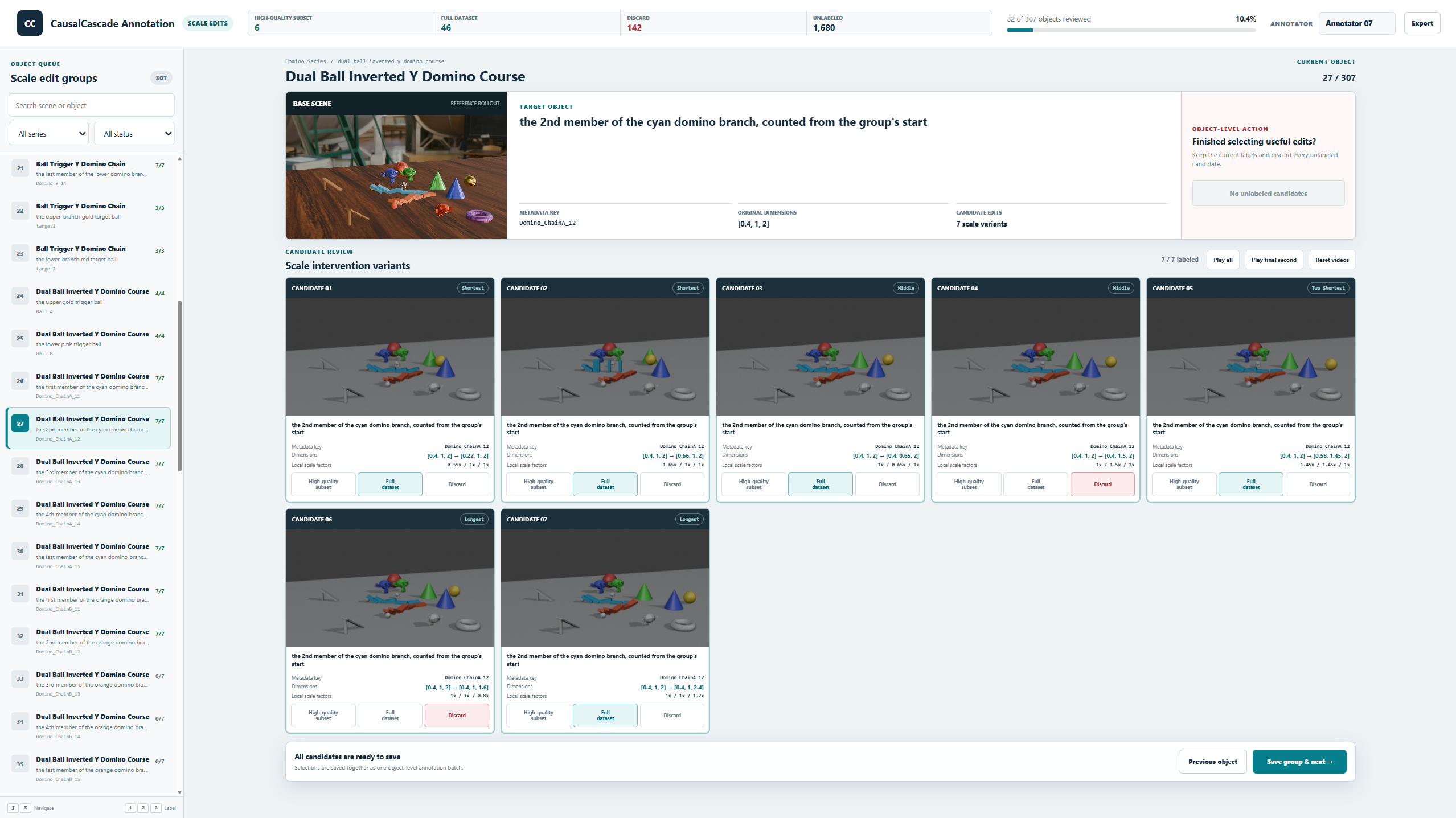}
    \caption{\textbf{Interface used for manual screening of edited rollouts.}
    The base and edited rollouts are shown together for comparison, and synchronized playback is supported.
    Annotators assign each candidate to \emph{High quality}, \emph{Full dataset}, or \emph{Discard}.
    }
    \label{fig:manual_screening}
\end{figure}

\paragraph{Manual screening.}
\label{app:manual_screening}
Candidates that pass automatic validation are re-simulated and rendered as low-resolution Workbench rollouts for manual inspection through a custom annotation interface illustrated in Figure~\ref{fig:manual_screening}.
The interface presents each edited rollout together with its corresponding base rollout and records one of three operational labels: \emph{High quality}, \emph{Full dataset}, or \emph{Discard}.
A candidate is labeled \emph{High quality} when the intervention produces a clear and visually verifiable downstream consequence while the rollout remains physically valid.
These candidates form an eligible pool for RQ2 rather than being included exhaustively.
From this larger pool, we select a compact evaluation subset of 50 delete, 50 move, 50 scale, and 75 mass pairs.
These quotas are imposed to keep the evaluation subset small and carefully curated, rather than reflecting a shortage of qualifying candidates.
All screening and subset selection are model-blind: labels and subset membership are determined solely from the simulator rollouts, without inspecting any model generations.
Candidates labeled \emph{Full dataset} remain valid counterfactual variants but are not included in the designated RQ2 subset, whereas \emph{Discard} candidates are removed.

For mass edits, candidates are grouped by edited object so that the base rollout and the ten corresponding mass variants can be inspected together with synchronized playback.
The other edit types are inspected through paired base-edited rollout comparisons.

The resulting complete variant dataset contains 326 delete, 1,469 mass, 747 move, and 290 scale variants, for a total of 2,832 retained edits.
The RQ2 subset contains 225 instances: 50 delete, 75 mass, 50 move, and 50 scale.

\subsection{Edit Representation and Reconstruction}

For each retained edit, we store a semantic edit description, the corresponding delta JSON file \texttt{diff.json}, and the edited initial RGB frame.

Together with the original scene metadata and Blender file, these records allow each controlled variant to be reconstructed and re-simulated without storing a separate edited \texttt{.blend} file.

\section{Evaluation Details}
\label{app:evaluation_details}

This section provides additional details of the motion-based evaluation and temporal-alignment diagnostics used in our experiments. We build on the motion-map-based evaluation framework of Physics-IQ~\citep{physics-iq}. Spatial IoU and Weighted Spatial IoU retain the corresponding motion-map formulations, while our spatiotemporal evaluation excludes jointly inactive frame pairs so that long periods with no motion in either video do not artificially increase the score. We refer to the resulting frame-aligned metric as \emph{Strict ST-IoU}. Global affine alignment and dynamic time warping (DTW) are used only as post-hoc temporal diagnostics within RQ1 and do not affect the primary benchmark ranking.

\subsection{Video Preprocessing and Strict Temporal Protocol}
\label{app:strict_protocol}

Let
\begin{equation}
    V^*=\{I_i^*\}_{i=0}^{N^*-1},
    \qquad
    \hat V=\{\hat I_j\}_{j=0}^{\hat N-1}
\end{equation}
denote the simulator-derived reference rollout and the generated rollout, respectively. Here, $N^*$ and $\hat N$ denote their native frame counts, while $f^*$ and $\hat f$ denote their native frame rates.

All decoded frames are converted to RGB, represented as floating-point values in $[0,1]$, and resized to $256\times256$ using area interpolation. No spatial cropping or padding is applied.

Because the evaluated models produce videos at different native frame rates and durations, strict evaluation uses a common physical-time grid defined by the reference rollout. Let
\begin{equation}
    D^*=\frac{N^*}{f^*}
\end{equation}
denote the reference duration and let $f_{\mathrm{eval}}$ denote the evaluation frame rate. In our experiments, $f_{\mathrm{eval}}=f^*$. The number of evaluation samples is
\begin{equation}
    K=
    \operatorname{round}
    \left(
        D^* f_{\mathrm{eval}}
    \right),
\end{equation}
with timestamps
\begin{equation}
    \tau_k=\frac{k}{f_{\mathrm{eval}}},
    \qquad
    k=0,\ldots,K-1.
\end{equation}

For a source sequence with frame rate $f_{\mathrm{src}}$, the source frame associated with $\tau_k$ is selected as
\begin{equation}
    j_k=
    \operatorname{round}
    \left(
        \tau_k f_{\mathrm{src}}
    \right).
\end{equation}
Thus, differences in native frame rate are normalized through nearest-frame sampling rather than frame interpolation or temporal warping. Importantly, the generated rollout is not stretched to match the duration of the reference: corresponding samples represent the same physical timestamps.

The evaluation horizon is the duration of the reference rollout. Generated videos extending beyond this duration are therefore ignored after the reference terminates. If a generated sequence terminates earlier, the last available sample is held for the remaining evaluation timestamps under our default short-video policy. The same temporal sampling procedure is applied to RGB frames and motion masks. No post-hoc temporal shift, scaling, or warping is applied in the strict protocol.

\subsection{Motion Masks and Motion Maps}
\label{app:motion_maps}

Following the motion-map formulation of Physics-IQ~\citep{physics-iq}, we represent observable physical activity using binary motion masks. Motion masks are extracted independently from the native decoded sequence of each video before strict temporal resampling.

Given an RGB frame $I_i$, we first convert it to grayscale:
\begin{equation}
    G_i
    =
    0.299 I_i^R
    +
    0.587 I_i^G
    +
    0.114 I_i^B.
\end{equation}
Since all benchmark scenes use fixed cameras, motion is estimated relative to an adaptive background initialized from the first frame,
\begin{equation}
    B_0=G_0.
\end{equation}
For each native frame $i$, we compute
\begin{equation}
    D_i=|G_i-B_i|.
\end{equation}
The difference map is smoothed using a $5\times5$ Gaussian kernel and thresholded at $\tau_{\mathrm{m}}=0.06$. Denoting the smoothed difference map by $\widetilde D_i$, the initial binary motion mask is
\begin{equation}
    M_i^{\mathrm{nat}}
    =
    \mathbb{I}
    \left[
        \widetilde D_i \geq \tau_{\mathrm{m}}
    \right].
\end{equation}
We then apply morphological opening followed by closing using a $3\times3$ kernel. The adaptive background is updated after each frame:
\begin{equation}
    B_{i+1}
    =
    (1-\alpha)B_i+\alpha G_i,
    \qquad
    \alpha=0.03.
\end{equation}
Since the first frame initializes the background, its motion mask is empty by construction.

Under the strict protocol, the native motion-mask sequences are sampled on the common physical-time grid defined above. We denote the resulting reference and generated masks by
\begin{equation}
    \mathcal{M}^*
    =
    \{M_k^*\}_{k=0}^{K-1},
    \qquad
    \hat{\mathcal{M}}
    =
    \{\hat M_k\}_{k=0}^{K-1}.
\end{equation}

\subsection{Motion-based Metrics}
\label{app:motion_metrics}

For two binary masks $A$ and $B$ with a nonempty union, we define
\begin{equation}
    \operatorname{IoU}(A,B)
    =
    \frac{|A\cap B|}
         {|A\cup B|}.
\end{equation}

For frame-wise spatiotemporal evaluation, jointly inactive frame pairs are not treated as evidence of successful motion prediction. We therefore define the set of motion-active strict frame pairs as
\begin{equation}
    \mathcal{A}_{\mathrm{strict}}
    =
    \left\{
        k:
        |M_k^*\cup\hat M_k|>0
    \right\}.
\end{equation}
A frame remains in the evaluation whenever either the reference or generated video contains motion. Only pairs for which both motion masks are empty are excluded.

\paragraph{Spatial IoU.}

Spatial IoU measures whether physical activity occurs in the appropriate image regions without requiring exact temporal correspondence. Following Physics-IQ, we collapse each mask sequence over time using a logical OR:
\begin{equation}
    M_{\mathrm{sp}}^*
    =
    \bigvee_{k=0}^{K-1}M_k^*,
    \qquad
    \hat M_{\mathrm{sp}}
    =
    \bigvee_{k=0}^{K-1}\hat M_k.
\end{equation}
The resulting binary spatial motion maps are compared as
\begin{equation}
    \operatorname{SpatialIoU}
    =
    \frac{
        |M_{\mathrm{sp}}^*
        \cap
        \hat M_{\mathrm{sp}}|
    }{
        |M_{\mathrm{sp}}^*
        \cup
        \hat M_{\mathrm{sp}}|
    }.
\end{equation}
If both spatial motion maps are empty, the score is defined as $1$.

\paragraph{Strict ST-IoU.}

Spatiotemporal IoU additionally requires motion to occur at the corresponding physical time. Under our strict protocol, we compute
\begin{equation}
    \operatorname{StrictSTIoU}
    =
    \frac{1}
         {|\mathcal{A}_{\mathrm{strict}}|}
    \sum_{k\in\mathcal{A}_{\mathrm{strict}}}
    \operatorname{IoU}
    \left(
        M_k^*,
        \hat M_k
    \right).
\end{equation}

Thus, a frame in which motion occurs only in the reference or only in the generated rollout remains in the average and contributes zero overlap, whereas a jointly inactive frame is excluded. This prevents long motionless intervals, including shared stable tails after a cascade has terminated, from artificially increasing spatiotemporal agreement. If no motion-active frame pair exists over the entire evaluation interval, the score is defined as $0$.

We refer to this metric as \emph{Strict ST-IoU} because temporal correspondence is fixed by physical time: no temporal offset, global rescaling, or local warping is permitted.

\paragraph{Weighted Spatial IoU.}

Weighted Spatial IoU additionally captures how frequently motion occurs at each spatial location. We construct weighted spatial motion maps
\begin{equation}
    M_{\mathrm{w}}^*(x)
    =
    \frac{1}{K}
    \sum_{k=0}^{K-1}
    M_k^*(x),
    \qquad
    \hat M_{\mathrm{w}}(x)
    =
    \frac{1}{K}
    \sum_{k=0}^{K-1}
    \hat M_k(x),
\end{equation}
where $x$ indexes spatial locations. The score is
\begin{equation}
    \operatorname{WeightedSpatialIoU}
    =
    \frac{
        \sum_x
        \min
        \left(
            M_{\mathrm{w}}^*(x),
            \hat M_{\mathrm{w}}(x)
        \right)
    }{
        \sum_x
        \max
        \left(
            M_{\mathrm{w}}^*(x),
            \hat M_{\mathrm{w}}(x)
        \right)
    }.
\end{equation}
If both weighted spatial motion maps are identically zero, the score is defined as $1$. Spatial IoU and Weighted Spatial IoU otherwise retain their original temporal aggregation and are not modified by the joint-empty-frame exclusion used for ST-IoU.

\subsection{Appearance Reconstruction Metrics}
\label{app:appearance_metrics}

We complement the motion-based overlap metrics with appearance-level reconstruction measures. We report motion-masked mean squared error(mMSE) and structural similarity (SSIM).

For strict evaluation frame $k$, let
\begin{equation}
    \Omega_k
    =
    \left\{
        x:
        M_k^*(x)\lor\hat M_k(x)
    \right\}
\end{equation}
denote the union of the reference and generated motion regions. For $|\Omega_k|>0$, motion-masked MSE is
\begin{equation}
    \operatorname{mMSE}_k
    =
    \frac{1}{3|\Omega_k|}
    \sum_{x\in\Omega_k}
    \sum_{c\in\{R,G,B\}}
    \left(
        I_{k,c}^*(x)
        -
        \hat I_{k,c}(x)
    \right)^2.
\end{equation}
Frames with an empty motion union are omitted from the video-level mMSE.
Defining
\begin{equation}
    \mathcal{K}_{\mathrm{m}}
    =
    \left\{
        k:
        |\Omega_k|>0
    \right\},
\end{equation}
we compute
\begin{equation}
    \operatorname{mMSE}
    =
    \frac{1}
         {|\mathcal{K}_{\mathrm{m}}|}
    \sum_{k\in\mathcal{K}_{\mathrm{m}}}
    \operatorname{mMSE}_k.
\end{equation}
If no frame contains a nonempty motion union, mMSE is left undefined and is omitted from aggregation.

SSIM is computed between corresponding RGB frames under the strict physical-time protocol. We compute SSIM separately for each RGB channel using an $11\times11$ Gaussian window with standard deviation $1.5$ and constants $C_1=0.01^2$ and $C_2=0.03^2$. Scores are averaged spatially, across the three channels, and finally across evaluation frames.

\subsection{Temporal Alignment Diagnostics}
\label{app:temporal_alignment}

Strict evaluation deliberately preserves physical-time correspondence and therefore directly penalizes timing discrepancies. Within RQ1, we additionally examine how much of this disagreement can be reduced when temporal correspondence is relaxed. We consider two post-hoc diagnostics: global affine temporal alignment and constrained dynamic time warping.

These diagnostics operate on the complete native motion-mask sequences after common spatial preprocessing and before strict physical-time resampling. Let
\begin{equation}
    \{M_i^{*,\mathrm{nat}}\}_{i=0}^{N^*-1},
    \qquad
    \{\hat M_j^{\mathrm{nat}}\}_{j=0}^{\hat N-1}
\end{equation}
denote the native reference and generated motion masks.

For each possible frame pair $(i,j)$, we define
\begin{equation}
    A_{ij}
    =
    \mathbb{I}
    \left[
        |M_i^{*,\mathrm{nat}}
        \cup
        \hat M_j^{\mathrm{nat}}|>0
    \right]
\end{equation}
and construct a pairwise motion-similarity matrix
\begin{equation}
    S_{ij}
    =
    \begin{cases}
    \displaystyle
    \frac{
        |M_i^{*,\mathrm{nat}}
        \cap
        \hat M_j^{\mathrm{nat}}|
    }{
        |M_i^{*,\mathrm{nat}}
        \cup
        \hat M_j^{\mathrm{nat}}|
    },
    & A_{ij}=1,\\[10pt]
    0,
    & A_{ij}=0.
    \end{cases}
\end{equation}
Jointly inactive pairs are therefore assigned zero similarity during alignment search. This prevents either alignment procedure from obtaining a high score simply by matching large numbers of static frames.

\paragraph{Global affine temporal alignment.}

We first consider an affine temporal correspondence
\begin{equation}
    j(i)
    =
    \operatorname{round}(ai+b),
\end{equation}
where $a$ represents relative temporal scale and $b$ represents temporal offset in native frame-index space.

The nominal scale implied by the native sequence lengths is
\begin{equation}
    a_0
    =
    \frac{\hat N-1}{N^*-1}.
\end{equation}
We evaluate 41 uniformly spaced scale values over
\begin{equation}
    a\in[0.5a_0,1.5a_0]
\end{equation}
and 61 uniformly spaced offsets over
\begin{equation}
    b\in[-B,B],
    \qquad
    B=
    \operatorname{round}
    \left(
        0.25\max(N^*,\hat N)
    \right).
\end{equation}

The search objective evaluates correspondence in both directions. For the reference-to-generated direction,
\begin{equation}
    S_{\rightarrow}
    =
    \frac{1}{N^*}
    \sum_{i=0}^{N^*-1}
    \phi
    \left(
        i,
        j(i)
    \right),
\end{equation}
where
\begin{equation}
    \phi(i,j)
    =
    \begin{cases}
        S_{ij}, & 0\leq j<\hat N,\\
        0, & \text{otherwise}.
    \end{cases}
\end{equation}
Using the inverse correspondence
\begin{equation}
    i(j)
    =
    \operatorname{round}
    \left(
        \frac{j-b}{a}
    \right),
\end{equation}
we analogously compute the generated-to-reference score $S_{\leftarrow}$. The affine search objective is
\begin{equation}
    S_{\mathrm{search}}
    =
    \frac{1}{2}
    \left(
        S_{\rightarrow}
        +
        S_{\leftarrow}
    \right).
\end{equation}
Because jointly inactive frame pairs have $S_{ij}=0$, they cannot improve this search objective. Out-of-range correspondences likewise contribute zero.

To prevent the optimizer from selecting only a short favorable part of the generated video, the generated indices selected by the forward mapping are required to span at least $95\%$ of the generated sequence. Among candidates satisfying this constraint, we choose the mapping with the highest bidirectional objective. If no candidate satisfies the coverage requirement, the implementation falls back to the highest-scoring candidate and records the coverage violation.

After selecting the affine mapping, the reported \emph{Tempo-aligned ST-IoU} excludes valid mapped pairs for which both motion masks are empty. In contrast, mappings falling outside the other sequence remain zero-valued penalties. The reported score is computed in both mapping directions and averaged:
\begin{equation}
    \operatorname{TempoAlignedSTIoU}
    =
    \frac{1}{2}
    \left(
        R_{\rightarrow}
        +
        R_{\leftarrow}
    \right),
\end{equation}
where each $R$ averages IoU over motion-active mapped pairs together with zero penalties for out-of-range mappings, while excluding jointly inactive valid pairs from its denominator.

\paragraph{Dynamic time warping.}

Global affine alignment can correct an overall difference in temporal scale or offset but cannot accommodate locally varying timing. We therefore also apply constrained DTW using
\begin{equation}
    C_{ij}=1-S_{ij}
\end{equation}
as the local alignment cost. Since jointly inactive frame pairs have $S_{ij}=0$, they incur unit cost during path optimization and cannot serve as high-reward correspondences.

The monotonic path is constrained to a normalized Sakoe-Chiba band,
\begin{equation}
    \left|
        \frac{i}{N^*-1}
        -
        \frac{j}{\hat N-1}
    \right|
    \leq0.15.
\end{equation}
The path must connect $(0,0)$ to $(N^*-1,\hat N-1)$. Diagonal, horizontal, and vertical transitions are allowed. A diagonal transition resets the non-diagonal run length, and at most eight consecutive horizontal or vertical transitions are allowed without a diagonal transition. These constraints discourage degenerate paths that repeatedly align many frames to a small temporal region.

Let $\mathcal{P}$ denote the minimum-cost DTW path. To apply the same joint-inactivity convention used by Strict ST-IoU, we define the motion-active subset of the selected path as
\begin{equation}
    \mathcal{A}_{\mathrm{DTW}}
    =
    \left\{
        (i,j)\in\mathcal{P}:
        A_{ij}=1
    \right\}.
\end{equation}
The reported DTW-aligned score is
\begin{equation}
    \operatorname{DTWAlignedSTIoU}
    =
    \frac{1}
         {|\mathcal{A}_{\mathrm{DTW}}|}
    \sum_{(i,j)\in\mathcal{A}_{\mathrm{DTW}}}
    S_{ij}.
\end{equation}
If the selected path contains no motion-active pair, the score is defined as $0$.

Both temporal-alignment metrics are diagnostic rather than primary benchmark scores. Comparing Strict ST-IoU with Tempo-aligned and DTW-aligned ST-IoU reveals how much frame-aligned motion disagreement can be reduced when increasingly flexible temporal correspondence is allowed.
The aligned scores should therefore be interpreted as temporal diagnostics rather than as measures that isolate purely non-temporal physical error.

\section{Counterfactual Evaluation Details}
\label{app:cf_metrics}

RQ2 evaluates whether a generated future responds selectively to a controlled change in the physical world. For each factual-counterfactual pair, the intervention should induce the corresponding changes in motion while preserving motion that remains shared between the two reference futures. We operationalize these complementary requirements with \emph{Effect} and \emph{Preserve} metrics.

The counterfactual metrics build on the spatial and weighted spatial motion representations defined in Appendix~\ref{app:motion_metrics}.
For RQ2, motion masks are
extracted from each video's native decoded frames and independently aggregated
over its full duration: binary maps take the temporal union, while weighted
maps record the per-pixel fraction of active frames. Effect and Preserve
compare these whole-video maps without common physical-time resampling or
framewise alignment.

\subsection{Motion Decomposition}
\label{app:cf_decomposition}

Let $V^{f*}$ and $V^{cf*}$ denote the simulator-rendered factual and counterfactual reference rollouts, and let $\hat V^{f}$ and $\hat V^{cf}$ denote the corresponding generated rollouts.

For the binary spatial representation defined in Appendix~\ref{app:motion_metrics}, we denote the resulting maps by $M_{\mathrm{sp}}^{f*}$, $M_{\mathrm{sp}}^{cf*}$, $\hat M_{\mathrm{sp}}^{f}$, and $\hat M_{\mathrm{sp}}^{cf}$.
For the weighted spatial representation, we analogously use $M_{\mathrm{w}}^{f*}$, $M_{\mathrm{w}}^{cf*}$, $\hat M_{\mathrm{w}}^{f}$, and $\hat M_{\mathrm{w}}^{cf}$.

For either representation $r\in\{\mathrm{sp},\mathrm{w}\}$, we decompose the reference factual-counterfactual pair into three components:
\begin{align}
    P_r^*
    &=
    \min\left(
        M_r^{f*},
        M_r^{cf*}
    \right),
    \\
    S_r^*
    &=
    \left[
        M_r^{f*}
        -
        M_r^{cf*}
    \right]_+,
    \\
    A_r^*
    &=
    \left[
        M_r^{cf*}
        -
        M_r^{f*}
    \right]_+,
\end{align}
where $[z]_+=\max(z,0)$ and all operations are applied element-wise.
$P_r^*$ represents motion preserved across the intervention, $S_r^*$ represents motion suppressed by the intervention, and $A_r^*$ represents motion induced by the intervention.

We apply the same decomposition to the generated factual-counterfactual pair:
\begin{align}
    \hat P_r
    &=
    \min\left(
        \hat M_r^{f},
        \hat M_r^{cf}
    \right),
    \\
    \hat S_r
    &=
    \left[
        \hat M_r^{f}
        -
        \hat M_r^{cf}
    \right]_+,
    \\
    \hat A_r
    &=
    \left[
        \hat M_r^{cf}
        -
        \hat M_r^{f}
    \right]_+.
\end{align}

For the binary spatial representation, these operations reduce to set intersection and set difference.
For the weighted representation, they retain the magnitude of motion activity at each spatial location.

This decomposition separates the two requirements evaluated in RQ2.
The Effect metrics compare $(S_r^*,A_r^*)$ with $(\hat S_r,\hat A_r)$ and measure whether the intervention produces the appropriate suppressed and induced motion.
The Preserve metrics compare $P_r^*$ with $\hat P_r$ and measure whether motion shared across the two reference futures remains shared across the generated pair.

\subsection{Effect Metrics}
\label{app:cf_effect_metrics}

The Effect metrics evaluate whether the generated pair reproduces the motion changes caused by the intervention.
Suppressed and induced motion are treated as separate channels so that generating induced motion where the reference requires suppression, or vice versa, does not receive overlap credit.

\paragraph{CF Spatial Effect IoU.}

For binary spatial motion maps, we jointly compare the spatial support of the suppressed and induced components:
\begin{equation}
    \operatorname{CFSpatialEffectIoU}
    =
    \frac{
        |S_{\mathrm{sp}}^*\cap\hat S_{\mathrm{sp}}|
        +
        |A_{\mathrm{sp}}^*\cap\hat A_{\mathrm{sp}}|
    }{
        |S_{\mathrm{sp}}^*\cup\hat S_{\mathrm{sp}}|
        +
        |A_{\mathrm{sp}}^*\cup\hat A_{\mathrm{sp}}|
    }.
\end{equation}

This metric is high when the generated factual-counterfactual pair suppresses and induces motion in the same spatial regions as the reference pair.
Here, suppression and induction describe the \emph{type of motion change} caused by the intervention rather than the physical direction of object motion.

\paragraph{CF Weighted Effect IoU.}

For weighted spatial motion maps, we retain the motion-activity magnitude of the suppressed and induced components and compute a joint fuzzy IoU:
\begin{equation}
    \operatorname{CFWeightedEffectIoU}
    =
    \frac{
        \sum_x
        \min\left(
            S_{\mathrm{w}}^*(x),
            \hat S_{\mathrm{w}}(x)
        \right)
        +
        \sum_x
        \min\left(
            A_{\mathrm{w}}^*(x),
            \hat A_{\mathrm{w}}(x)
        \right)
    }{
        \sum_x
        \max\left(
            S_{\mathrm{w}}^*(x),
            \hat S_{\mathrm{w}}(x)
        \right)
        +
        \sum_x
        \max\left(
            A_{\mathrm{w}}^*(x),
            \hat A_{\mathrm{w}}(x)
        \right)
    }.
\end{equation}

This metric evaluates both where intervention effects occur and their relative amount of motion activity.

\subsection{Preservation Metrics}
\label{app:cf_preserve_metrics}

The Preserve metrics evaluate the complementary requirement that dynamics unaffected by the intervention remain stable across the factual and counterfactual generations.

\paragraph{CF Preserve Spatial IoU.}

For binary spatial motion maps, we compare the preserved reference and generated components:
\begin{equation}
    \operatorname{CFPreserveSpatialIoU}
    =
    \frac{
        |P_{\mathrm{sp}}^*
        \cap
        \hat P_{\mathrm{sp}}|
    }{
        |P_{\mathrm{sp}}^*
        \cup
        \hat P_{\mathrm{sp}}|
    }.
\end{equation}

This metric measures whether spatial regions that remain active in both reference futures are also preserved across the generated factual-counterfactual pair.
Generated shared motion outside the reference preserved region contributes to the union and therefore lowers the score.

\paragraph{CF Preserve Weighted IoU.}

For weighted spatial motion maps, we compare preserved motion activity using fuzzy IoU:
\begin{equation}
    \operatorname{CFPreserveWeightedIoU}
    =
    \frac{
        \sum_x
        \min\left(
            P_{\mathrm{w}}^*(x),
            \hat P_{\mathrm{w}}(x)
        \right)
    }{
        \sum_x
        \max\left(
            P_{\mathrm{w}}^*(x),
            \hat P_{\mathrm{w}}(x)
        \right)
    }.
\end{equation}

This metric measures whether motion shared across the two reference futures is preserved both spatially and in its relative amount of motion activity.

Together, Effect and Preserve capture the two complementary requirements of selective counterfactual response: changing what should change while preserving what should remain unchanged.

\subsection{Aggregation and Reporting}
\label{app:cf_aggregation}

RQ2 is evaluated on the model-blind subset of 225 matched factual-counterfactual pairs described in Appendix~\ref{app:edit_generation}.
The subset contains 50 delete, 50 move, 50 scale, and 75 mass interventions.
Subset selection uses only simulator-derived reference rollouts and does not inspect generations from any evaluated model.

Because the number of selected interventions may differ across base scenes, we use scene-macro aggregation.
For model $m$, intervention type $k$, metric $q$, and base scene $b$, let $\mathcal{E}_{b,k}$ denote the evaluated interventions of type $k$ associated with scene $b$, and let $s_{m,e}^{(q)}$ denote the per-pair score for intervention $e$.
We first average all evaluated interventions of the same type within each base scene:
\begin{equation}
    \bar{s}_{m,b,k}^{(q)}
    =
    \frac{1}{|\mathcal{E}_{b,k}|}
    \sum_{e\in\mathcal{E}_{b,k}}
    s_{m,e}^{(q)}.
\end{equation}

We then macro-average these scene-level scores across the base scenes represented for intervention type $k$:
\begin{equation}
    S_{m,k}^{(q)}
    =
    \frac{1}{|\mathcal{B}_k|}
    \sum_{b\in\mathcal{B}_k}
    \bar{s}_{m,b,k}^{(q)},
\end{equation}
where $\mathcal{B}_k$ denotes the set of base scenes represented for intervention type $k$.

This aggregation prevents scenes with more selected interventions from receiving greater weight.
Because the intervention types apply to different subsets of base scenes, we report delete, move, scale, and mass separately rather than pooling them into a single score.
For each intervention type, we report CF Spatial Effect IoU, CF Weighted Effect IoU, CF Preserve Spatial IoU, and CF Preserve Weighted IoU.

\section{Implementation Details}   
\label{app:implementation_details}

\paragraph{Inference protocol.}
We evaluate all models using their released checkpoints and model-specific inference pipelines. Because the evaluated models differ in their native video configurations and sampling procedures, we do not force a common generation resolution, frame count, frame rate, or sampler. Instead, each model is generated using a configuration compatible with its released implementation, and the resulting videos are normalized
only during evaluation. Table~\ref{tab:model_inference_settings} summarizes the exact generation settings used in our experiments. Prompt rewriting or extension is disabled wherever the released implementation provides such an option, such that the benchmark prompt is passed directly to the model together with the corresponding first frame.

\paragraph{Randomness and reproducibility.}
All experiments use a global base seed of 42 with deterministic case-specific seeding. For factual cases, the seed key is the case identifier; for each counterfactual edit, it is the identifier of the corresponding base scene. Thus, paired factual and counterfactual generations receive the same assigned seed. We compute the SHA-256 hash of the UTF-8 string formed by the base seed, a colon, and this seed key, convert its first eight hexadecimal digits to an integer, and reduce the result modulo $10^9$. This makes the assigned seed independent of dataset ordering and reproducible across repeated runs.

\paragraph{Compute environment.}
All experiments were conducted on a single node with four NVIDIA A100 80GB GPUs using Python~3.10, PyTorch~2.6, and CUDA~12.4. We use bfloat16 inference where supported by the corresponding model implementation.

\paragraph{Distributed inference.}
For models that fit on a single GPU, inference is parallelized across independent benchmark cases, with different cases assigned to different GPUs. Larger models use the distributed inference strategies provided by their released implementations, including FSDP and Ulysses/USP sequence parallelism where applicable. These choices are used only for memory efficiency and throughput and do not alter the benchmark inputs or evaluation protocol.

\begin{table}[H]
\centering
\small
\setlength{\tabcolsep}{4.0pt}
\renewcommand{\arraystretch}{1.12}
\caption{
\textbf{Inference configurations of the models under test in our experiments.} We retain each model's released generation configuration rather than imposing a common resolution, frame rate, or sampling schedule. Guidance parameters follow their model-specific definitions and are therefore not necessarily directly comparable across architectures.
}
\label{tab:model_inference_settings}

\resizebox{\textwidth}{!}{%
\begin{tabular}{llccccl}
\toprule
\textbf{Model}
& \textbf{Checkpoint}
& \textbf{Resolution}
& \textbf{Frames}
& \textbf{FPS}
& \textbf{Steps}
& \textbf{Guidance / schedule} \\
\midrule

CogVideoX-Fun-V1.5
& CogVideoX-Fun-V1.5-5b-InP
& $1360{\times}768$
& 85 & 8 & 50
& guidance 6.0 \\

Cosmos3-Nano
& Cosmos3-Nano
& $832{\times}480$
& 189 & 24 & 35
& shift 10.0; guidance 6.0 \\

HunyuanVideo-I2V
& HunyuanVideo-I2V
& $960{\times}544$
& 129 & 24 & 50
& shift 7.0; embedded CFG 6.0 \\

LTX-Video-13B
& LTX-Video-13B
& $832{\times}480$
& 121 & 16 & $30,30^\dagger$
& scheduled CFG/STG$^\dagger$ \\

MAGI-1-4.5B
& MAGI-1-4.5B-base
& $832{\times}480$
& 120 & 24 & 64
& multi-stage text/prev.-chunk guidance \\

SkyReels-V2-I2V
& SkyReels-V2-I2V-14B-720P
& $1280{\times}720$
& 121 & 24 & 50
& shift 3.0; guidance 5.0 \\

Wan2.1-I2V-14B
& Wan2.1-I2V-14B-480P
& $832{\times}480$
& 121 & 16 & 40
& shift 3.0; guidance 5.0 \\

Wan2.2-TI2V-5B
& Wan2.2-TI2V-5B
& $832{\times}480$
& 121 & 24 & 50
& shift 5.0; guidance 5.0 \\

Wan2.2-I2V-A14B
& Wan2.2-I2V-A14B
& $832{\times}480$
& 121 & 16 & 40
& shift 5.0; guidance 3.5, 3.5; boundary 0.9 \\

\bottomrule
\end{tabular}%
}

\vspace{0.25em}
\raggedright
\footnotesize
$^\dagger$LTX-Video-13B follows the released two-pass multi-scale configuration. Both passes are defined over a 30-step schedule; the first pass skips the final three scheduled steps, while the second skips the initial 17 scheduled steps. The released configuration additionally uses time-dependent CFG and STG schedules.
\end{table}

\section{Detailed Quantitative Results}
\label{app:detailed_results}

This section provides the complete quantitative results underlying the compact analyses in the main paper. We first report the full Level~1 comparison and the per-model effects of future-event specification for RQ1. We then report the complete counterfactual results for RQ2, including both spatial and weighted-spatial Effect and Preserve metrics across the four intervention types.


\subsection{Complete Level~1 Model Comparison}

Table~\ref{tab:rq1_full_results} extends Figure~\ref{fig:main_results_summary} (left) with Weighted Spatial IoU, motion-masked MSE, and SSIM. The additional metrics support the same cross-model conclusion as the compact presentation: Wan2.2-I2V-A14B is strongest across the motion-overlap measures and mMSE, whereas Cosmos3-Nano obtains the highest SSIM and VQA Overall. Thus, no single model dominates both reference-level physical realization and event-level causal recovery.
\begin{table}[H]
\centering
\small
\setlength{\tabcolsep}{4.5pt}
\renewcommand{\arraystretch}{1.08}
\caption{
\textbf{Complete performance of current I2V models under the Level~1 prompt and strict evaluation protocol.}
Best results are shown in bold and second-best results are underlined.
}
\label{tab:rq1_full_results}
\resizebox{\textwidth}{!}{%
\begin{tabular}{lcccccc}
\toprule
Model
& Spatial IoU $\uparrow$
& Strict ST-IoU $\uparrow$
& Weighted Spatial IoU $\uparrow$
& mMSE $\downarrow$
& SSIM $\uparrow$
& VQA Overall $\uparrow$ \\
\midrule
CogVideoX-Fun-V1.5
& 0.147 & 0.037 & 0.067 & \underline{0.0539} & 0.771 & 0.095 \\

Cosmos3-Nano
& \underline{0.252} & 0.043 & \underline{0.137} & 0.0579 & \textbf{0.903} & \textbf{0.357} \\

HunyuanVideo-I2V
& 0.178 & 0.027 & 0.085 & 0.0645 & 0.608 & 0.174 \\

LTX-Video-13B
& 0.182 & 0.037 & 0.087 & 0.0583 & 0.727 & 0.189 \\

MAGI-1-4.5B
& 0.128 & 0.031 & 0.072 & 0.0673 & 0.697 & 0.153 \\

SkyReels-V2-I2V
& 0.193 & 0.047 & 0.105 & 0.0597 & 0.738 & 0.177 \\

Wan2.1-I2V-14B
& 0.204 & \underline{0.055} & 0.111 & 0.0590 & \underline{0.886} & 0.196 \\

Wan2.2-TI2V-5B
& 0.149 & 0.027 & 0.069 & 0.0557 & 0.662 & 0.141 \\

Wan2.2-I2V-A14B
& \textbf{0.312} & \textbf{0.080} & \textbf{0.157} & \textbf{0.0513} & 0.881 & \underline{0.252} \\
\bottomrule
\end{tabular}%
}
\end{table}

\subsection{Per-Model Effects of Future-Event Specification}
\label{app:future_event_specification}

Table~\ref{tab:rq1_full_prompt_levels} provides the per-model values summarized by the averages in Figure~\ref{fig:main_results_summary}(right).
Strict ST-IoU is higher at Level~3 than at Level~1 for every evaluated model, while VQA Overall improves for eight of the nine models.
The progression is nevertheless model-dependent: HunyuanVideo-I2V reaches its strongest motion overlap at Level~2 and shows a slight decrease in VQA Overall as future-event specification becomes more explicit.
These variations motivate retaining the full per-model breakdown while using the model-average trend to communicate the principal effect in the main paper.

\begin{table}[H]
\centering
\scriptsize
\setlength{\tabcolsep}{3pt}
\renewcommand{\arraystretch}{1.02}
\caption{
\textbf{Complete effect of future-event specification on causal cascade generation.}
We report Spatial IoU, Strict ST-IoU, and VQA Overall under the three specification levels.
The best level for each model and metric is shown in bold.
}
\label{tab:rq1_full_prompt_levels}
\begin{tabular}{
@{}l
ccc
@{\hspace{7pt}}
ccc
@{\hspace{7pt}}
ccc
@{}
}
\toprule
& \multicolumn{3}{c}{Spatial IoU $\uparrow$}
& \multicolumn{3}{c}{Strict ST-IoU $\uparrow$}
& \multicolumn{3}{c}{VQA Overall $\uparrow$}
\\
\cmidrule(lr){2-4}
\cmidrule(lr){5-7}
\cmidrule(lr){8-10}
Model
& L1 & L2 & L3
& L1 & L2 & L3
& L1 & L2 & L3 \\
\midrule
CogVideoX-Fun-V1.5
& 0.147 & 0.174 & \textbf{0.191}
& 0.037 & 0.042 & \textbf{0.062}
& 0.095 & 0.128 & \textbf{0.186} \\

Cosmos3-Nano
& 0.252 & 0.273 & \textbf{0.387}
& 0.043 & 0.061 & \textbf{0.080}
& 0.357 & 0.358 & \textbf{0.568} \\

HunyuanVideo-I2V
& 0.178 & \textbf{0.202} & 0.172
& 0.027 & \textbf{0.035} & 0.034
& \textbf{0.174} & 0.171 & 0.167 \\

LTX-Video-13B
& 0.182 & 0.180 & \textbf{0.217}
& 0.037 & 0.042 & \textbf{0.057}
& 0.189 & 0.213 & \textbf{0.227} \\

MAGI-1-4.5B
& 0.128 & 0.142 & \textbf{0.154}
& 0.031 & 0.038 & \textbf{0.045}
& 0.153 & 0.171 & \textbf{0.229} \\

SkyReels-V2-I2V
& 0.193 & 0.249 & \textbf{0.265}
& 0.047 & 0.060 & \textbf{0.069}
& 0.177 & 0.262 & \textbf{0.367} \\

Wan2.1-I2V-14B
& 0.204 & 0.219 & \textbf{0.324}
& 0.055 & 0.061 & \textbf{0.083}
& 0.196 & 0.272 & \textbf{0.385} \\

Wan2.2-TI2V-5B
& 0.149 & 0.157 & \textbf{0.170}
& 0.027 & 0.035 & \textbf{0.044}
& 0.141 & 0.189 & \textbf{0.265} \\

Wan2.2-I2V-A14B
& 0.312 & 0.351 & \textbf{0.440}
& 0.080 & 0.099 & \textbf{0.131}
& 0.252 & 0.416 & \textbf{0.532} \\
\midrule
Model Avg.
& 0.194 & 0.216 & \textbf{0.258}
& 0.043 & 0.053 & \textbf{0.067}
& 0.193 & 0.242 & \textbf{0.325} \\
\bottomrule
\end{tabular}
\end{table}

\textbf{Dimension-level VQA results.}
Table~\ref{tab:vqa_dimension_levels} reports per-model VQA performance across the five diagnostic dimensions under each future-event specification level.

\begin{table*}[t]
\centering
\scriptsize
\setlength{\tabcolsep}{7pt}
\renewcommand{\arraystretch}{1.02}
\caption{
\textbf{Dimension-level VQA performance under the three future-event specification levels.}
We report VQA Overall and scores for event occurrence, causal pathway, temporal ordering, outcome or termination, and scene consistency.
Best results within each panel are shown in bold and second-best results are underlined.
}
\label{tab:vqa_dimension_levels}

\textbf{(a) Level~1}\\[2pt]
\begin{tabular}{lcccccc}
\toprule
Model
& Overall
& \makecell{Event\\occurrence}
& \makecell{Causal\\pathway}
& \makecell{Temporal\\ordering}
& \makecell{Outcome /\\termination}
& \makecell{Scene\\consistency} \\
\midrule
CogVideoX-Fun-V1.5
& 0.095 & 0.172 & 0.036 & 0.027 & 0.022 & 0.220 \\

Cosmos3-Nano
& \textbf{0.357} & \textbf{0.473} & \textbf{0.223} & \textbf{0.253} & \textbf{0.250} & \underline{0.585} \\

HunyuanVideo-I2V
& 0.174 & 0.240 & 0.070 & 0.089 & 0.060 & 0.410 \\

LTX-Video-13B
& 0.189 & 0.209 & 0.063 & 0.073 & 0.052 & 0.550 \\

MAGI-1-4.5B
& 0.153 & 0.108 & 0.032 & 0.032 & \underline{0.103} & 0.490 \\

SkyReels-V2-I2V
& 0.177 & 0.232 & 0.070 & 0.054 & 0.085 & 0.445 \\

Wan2.1-I2V-14B
& 0.196 & 0.196 & 0.052 & 0.055 & 0.045 & \textbf{0.630} \\

Wan2.2-TI2V-5B
& 0.141 & 0.065 & 0.014 & 0.020 & 0.042 & 0.565 \\

Wan2.2-I2V-A14B
& \underline{0.252} & \underline{0.371} & \underline{0.140} & \underline{0.148} & 0.080 & 0.520 \\
\bottomrule
\end{tabular}

\vspace{0.8em}

\textbf{(b) Level~2}\\[2pt]
\begin{tabular}{lcccccc}
\toprule
Model
& Overall
& \makecell{Event\\occurrence}
& \makecell{Causal\\pathway}
& \makecell{Temporal\\ordering}
& \makecell{Outcome /\\termination}
& \makecell{Scene\\consistency} \\
\midrule
CogVideoX-Fun-V1.5
& 0.128 & 0.219 & 0.065 & 0.077 & 0.038 & 0.240 \\

Cosmos3-Nano
& \underline{0.358} & \underline{0.521} & \underline{0.256} & \underline{0.240} & \textbf{0.215} & 0.555 \\

HunyuanVideo-I2V
& 0.171 & 0.278 & 0.067 & 0.084 & 0.088 & 0.340 \\

LTX-Video-13B
& 0.213 & 0.260 & 0.106 & 0.120 & 0.077 & 0.505 \\

MAGI-1-4.5B
& 0.171 & 0.213 & 0.037 & 0.037 & 0.060 & 0.505 \\

SkyReels-V2-I2V
& 0.262 & 0.385 & 0.131 & 0.150 & 0.087 & 0.555 \\

Wan2.1-I2V-14B
& 0.272 & 0.395 & 0.105 & 0.116 & 0.067 & \textbf{0.675} \\

Wan2.2-TI2V-5B
& 0.189 & 0.234 & 0.073 & 0.085 & 0.070 & 0.485 \\

Wan2.2-I2V-A14B
& \textbf{0.416} & \textbf{0.643} & \textbf{0.287} & \textbf{0.352} & \underline{0.170} & \underline{0.630} \\
\bottomrule
\end{tabular}

\vspace{0.8em}

\textbf{(c) Level~3}\\[2pt]
\begin{tabular}{lcccccc}
\toprule
Model
& Overall
& \makecell{Event\\occurrence}
& \makecell{Causal\\pathway}
& \makecell{Temporal\\ordering}
& \makecell{Outcome /\\termination}
& \makecell{Scene\\consistency} \\
\midrule
CogVideoX-Fun-V1.5
& 0.186 & 0.383 & 0.125 & 0.161 & 0.093 & 0.170 \\

Cosmos3-Nano
& \textbf{0.568} & \underline{0.793} & \underline{0.427} & \underline{0.512} & \textbf{0.555} & \underline{0.555} \\

HunyuanVideo-I2V
& 0.167 & 0.281 & 0.113 & 0.089 & 0.108 & 0.245 \\

LTX-Video-13B
& 0.227 & 0.289 & 0.080 & 0.091 & 0.110 & \textbf{0.565} \\

MAGI-1-4.5B
& 0.229 & 0.405 & 0.133 & 0.164 & 0.170 & 0.275 \\

SkyReels-V2-I2V
& 0.367 & 0.585 & 0.245 & 0.303 & 0.255 & 0.445 \\

Wan2.1-I2V-14B
& 0.385 & 0.667 & 0.278 & 0.282 & 0.235 & 0.465 \\

Wan2.2-TI2V-5B
& 0.265 & 0.411 & 0.182 & 0.257 & 0.257 & 0.220 \\

Wan2.2-I2V-A14B
& \underline{0.532} & \textbf{0.865} & \textbf{0.474} & \textbf{0.522} & \underline{0.422} & 0.380 \\
\bottomrule
\end{tabular}

\end{table*}

\subsection{Full Counterfactual Results}
\label{app:cf_full_results}

Tables~\ref{tab:cf_delete_results}--\ref{tab:cf_mass_results} report the complete RQ2 results on the 225 matched factual-counterfactual pairs.
All results use the scene-macro aggregation defined in Appendix~\ref{app:cf_aggregation}, with the four intervention types reported separately.

\begin{table}[H]
\centering
\small
\setlength{\tabcolsep}{4.5pt}
\renewcommand{\arraystretch}{1.08}
\caption{
\textbf{Complete counterfactual results for delete interventions.}
All values are scene-macro averages.
Higher is better for all metrics.
Best results are shown in bold and second-best results are underlined.
}
\label{tab:cf_delete_results}
\resizebox{\textwidth}{!}{%
\begin{tabular}{lcccc}
\toprule
Model
& CF Spatial Effect IoU $\uparrow$
& CF Weighted Effect IoU $\uparrow$
& CF Preserve Spatial IoU $\uparrow$
& CF Preserve Weighted IoU $\uparrow$ \\
\midrule
CogVideoX-Fun-V1.5
& 0.0261 & 0.0461 & 0.1190 & 0.0507 \\

Cosmos3-Nano
& \underline{0.0410} & \underline{0.0488} & \underline{0.1983} & \underline{0.1223} \\

HunyuanVideo-I2V
& 0.0365 & 0.0389 & 0.1577 & 0.0895 \\

LTX-Video-13B
& 0.0324 & 0.0362 & 0.1471 & 0.0800 \\

MAGI-1-4.5B
& 0.0353 & 0.0372 & 0.1067 & 0.0829 \\

SkyReels-V2-I2V
& 0.0321 & 0.0468 & 0.1687 & 0.0946 \\

Wan2.1-I2V-14B
& 0.0298 & 0.0384 & 0.1730 & 0.0971 \\

Wan2.2-TI2V-5B
& 0.0268 & 0.0408 & 0.1288 & 0.0746 \\

Wan2.2-I2V-A14B
& \textbf{0.0454} & \textbf{0.0647} & \textbf{0.2505} & \textbf{0.1348} \\
\bottomrule
\end{tabular}%
}
\end{table}

\begin{table}[H]
\centering
\small
\setlength{\tabcolsep}{4.5pt}
\renewcommand{\arraystretch}{1.08}
\caption{
\textbf{Complete counterfactual results for move interventions.}
All values are scene-macro averages.
Higher is better for all metrics.
Best results are shown in bold and second-best results are underlined.
}
\label{tab:cf_move_results}
\resizebox{\textwidth}{!}{%
\begin{tabular}{lcccc}
\toprule
Model
& CF Spatial Effect IoU $\uparrow$
& CF Weighted Effect IoU $\uparrow$
& CF Preserve Spatial IoU $\uparrow$
& CF Preserve Weighted IoU $\uparrow$ \\
\midrule
CogVideoX-Fun-V1.5
& 0.0278 & 0.0349 & 0.1235 & 0.0537 \\

Cosmos3-Nano
& 0.0451 & \underline{0.0576} & \underline{0.1879} & \underline{0.1360} \\

HunyuanVideo-I2V
& 0.0367 & 0.0385 & 0.1720 & 0.0979 \\

LTX-Video-13B
& \underline{0.0457} & 0.0400 & 0.1274 & 0.0704 \\

MAGI-1-4.5B
& 0.0231 & 0.0283 & 0.1096 & 0.0665 \\

SkyReels-V2-I2V
& 0.0243 & 0.0389 & 0.1729 & 0.0935 \\

Wan2.1-I2V-14B
& 0.0397 & 0.0412 & 0.1546 & 0.0971 \\

Wan2.2-TI2V-5B
& 0.0275 & 0.0373 & 0.1143 & 0.0596 \\

Wan2.2-I2V-A14B
& \textbf{0.0534} & \textbf{0.0639} & \textbf{0.2883} & \textbf{0.1475} \\
\bottomrule
\end{tabular}%
}
\end{table}

\begin{table}[H]
\centering
\small
\setlength{\tabcolsep}{4.5pt}
\renewcommand{\arraystretch}{1.08}
\caption{
\textbf{Complete counterfactual results for scale interventions.}
All values are scene-macro averages.
Higher is better for all metrics.
Best results are shown in bold and second-best results are underlined.
}
\label{tab:cf_scale_results}
\resizebox{\textwidth}{!}{%
\begin{tabular}{lcccc}
\toprule
Model
& CF Spatial Effect IoU $\uparrow$
& CF Weighted Effect IoU $\uparrow$
& CF Preserve Spatial IoU $\uparrow$
& CF Preserve Weighted IoU $\uparrow$ \\
\midrule
CogVideoX-Fun-V1.5
& 0.0180 & 0.0278 & 0.1136 & 0.0547 \\

Cosmos3-Nano
& \underline{0.0482} & \underline{0.0525} & 0.1879 & 0.1215 \\

HunyuanVideo-I2V
& 0.0256 & 0.0264 & 0.1535 & 0.0749 \\

LTX-Video-13B
& 0.0333 & 0.0367 & 0.1654 & 0.0860 \\

MAGI-1-4.5B
& 0.0193 & 0.0263 & 0.1024 & 0.0566 \\

SkyReels-V2-I2V
& 0.0308 & 0.0357 & 0.1661 & 0.0856 \\

Wan2.1-I2V-14B
& 0.0413 & 0.0471 & \underline{0.1960} & \underline{0.1281} \\

Wan2.2-TI2V-5B
& 0.0290 & 0.0329 & 0.1229 & 0.0685 \\

Wan2.2-I2V-A14B
& \textbf{0.0550} & \textbf{0.0710} & \textbf{0.3037} & \textbf{0.1738} \\
\bottomrule
\end{tabular}%
}
\end{table}

\begin{table}[H]
\centering
\small
\setlength{\tabcolsep}{4.5pt}
\renewcommand{\arraystretch}{1.08}
\caption{
\textbf{Complete counterfactual results for mass interventions.}
All values are scene-macro averages.
Higher is better for all metrics.
Best results are shown in bold and second-best results are underlined.
}
\label{tab:cf_mass_results}
\resizebox{\textwidth}{!}{%
\begin{tabular}{lcccc}
\toprule
Model
& CF Spatial Effect IoU $\uparrow$
& CF Weighted Effect IoU $\uparrow$
& CF Preserve Spatial IoU $\uparrow$
& CF Preserve Weighted IoU $\uparrow$ \\
\midrule
CogVideoX-Fun-V1.5
& 0.0229 & 0.0344 & 0.1182 & 0.0494 \\

Cosmos3-Nano
& 0.0341 & 0.0397 & \underline{0.1847} & \underline{0.1099} \\

HunyuanVideo-I2V
& \textbf{0.0360} & 0.0387 & 0.1596 & 0.0950 \\

LTX-Video-13B
& 0.0108 & 0.0222 & 0.1502 & 0.0855 \\

MAGI-1-4.5B
& 0.0250 & 0.0331 & 0.1020 & 0.0636 \\

SkyReels-V2-I2V
& 0.0278 & \underline{0.0425} & 0.1745 & 0.0922 \\

Wan2.1-I2V-14B
& 0.0346 & 0.0371 & 0.1788 & 0.1053 \\

Wan2.2-TI2V-5B
& 0.0243 & 0.0368 & 0.1290 & 0.0702 \\

Wan2.2-I2V-A14B
& \underline{0.0359} & \textbf{0.0538} & \textbf{0.2680} & \textbf{0.1341} \\
\bottomrule
\end{tabular}%
}
\end{table}

\section{Qualitative Analysis of Counterfactual Responses}
\label{app:cf_cases}
To complement the aggregate counterfactual results in Section~\ref{sec:rq2}, we inspect six representative factual-counterfactual pairs spanning mass, move, scale, and delete interventions.
Each visualization compares the factual and counterfactual reference rollouts with the corresponding generated rollouts at matched normalized times.
The examples are selected to illustrate distinct response patterns rather than to estimate their prevalence.
We focus on two complementary questions aligned with the counterfactual evaluation: whether the generated pair realizes the consequence induced by the intervention, and whether dynamics that should remain shared are preserved.

\subsection{Intervention Neglect and Incorrect Consequences}
\label{app:cf_cases_effect}
\begin{figure*}[t]
    \centering
    \includegraphics[width=\textwidth]{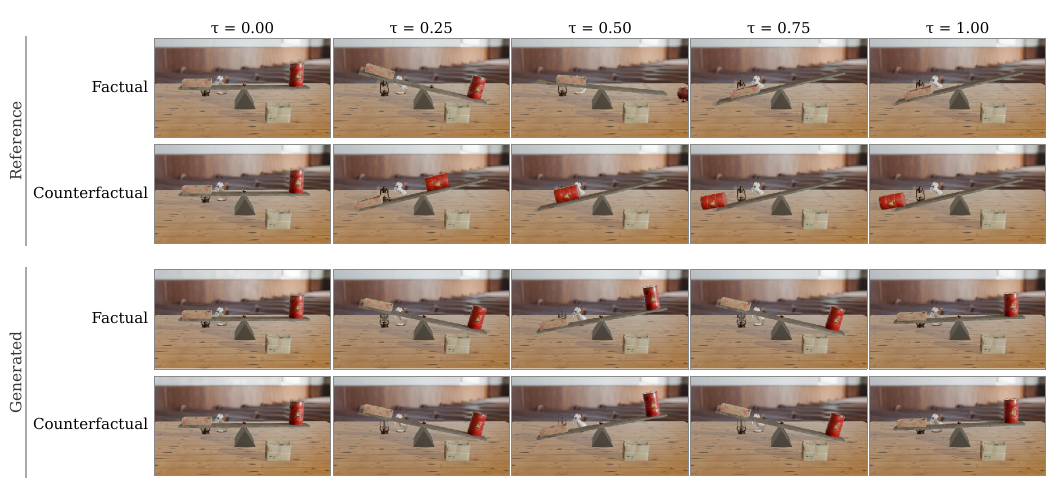}

    \medskip

    \includegraphics[width=\textwidth]{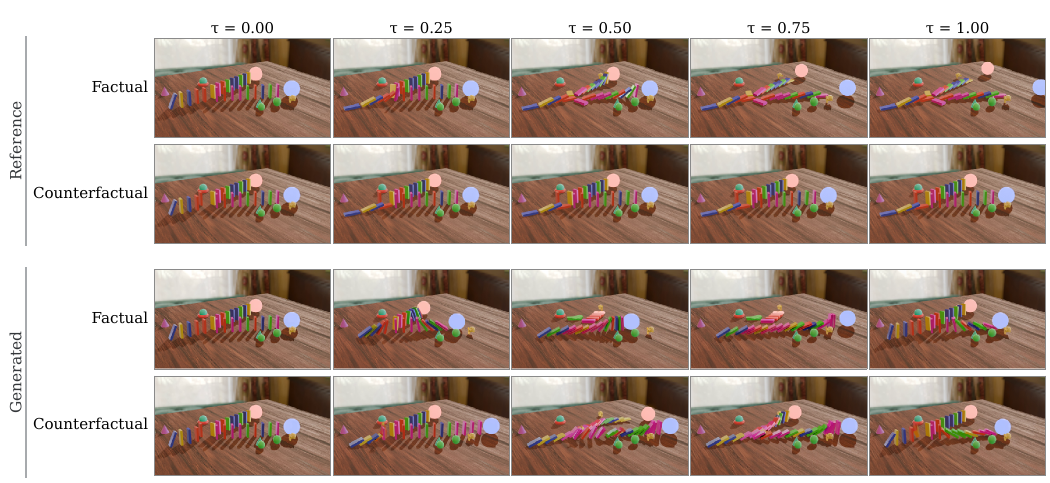}

    \caption{
    \textbf{Representative cases of intervention neglect despite active dynamics.}
    \textbf{Top:} For Wan2.2-I2V-A14B, changing the barrel mass substantially changes the reference trajectory, while the generated factual and counterfactual rollouts remain dynamic but evolve similarly.
    \textbf{Bottom:} For Wan2.2-I2V-A14B, moving a critical domino interrupts the reference counterfactual cascade before the fork, but the generated cascade continues through downstream branches in both rollouts.
    Reference and generated factual-counterfactual rollouts are shown at matched normalized times.
    }
    \label{fig:cf_intervention_neglect}
\end{figure*}

\textbf{Intervention neglect despite active dynamics.}
Figure~\ref{fig:cf_intervention_neglect} shows two cases in which the model generates substantial physical activity but fails to update the causal process according to the intervention.
In the mass-intervention case, changing the barrel mass substantially alters its reference trajectory and final settling side.
The generated factual and counterfactual rollouts remain dynamic, yet follow nearly the same repeated rocking evolution, indicating that the low Effect cannot be explained by a generic absence of motion.
Instead, the generated dynamics remain largely insensitive to the changed mass.

The move-intervention case exhibits a related failure under a visually explicit edit.
Displacing a critical domino interrupts the reference counterfactual cascade before the fork, whereas the factual cascade continues through both downstream branches.
The generated pair nevertheless activates most of the downstream chain in both rollouts.
Thus, although the model recognizes that the scene should produce a domino cascade, the edited contact geometry fails to gate downstream propagation.

\begin{figure*}[t]
    \centering
    \includegraphics[width=\textwidth]{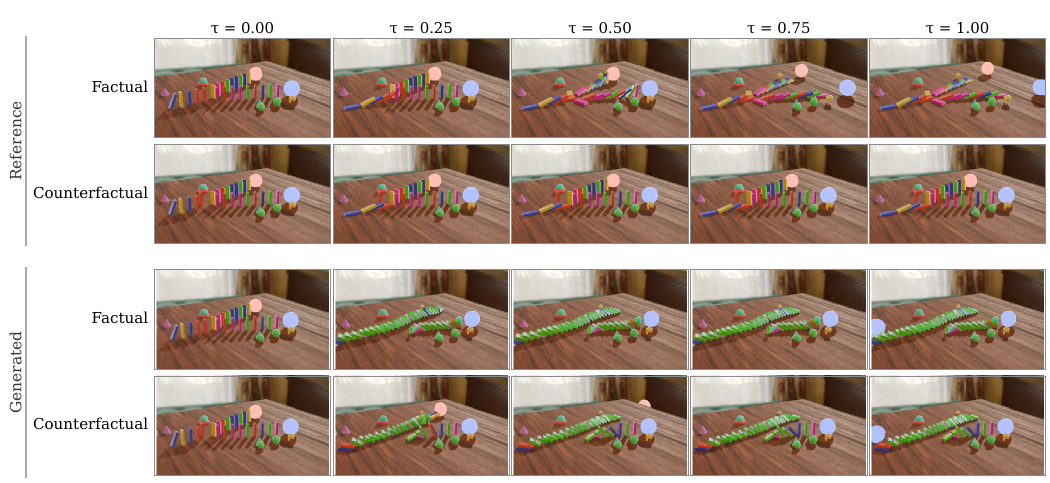}

    \medskip

    \includegraphics[width=\textwidth]{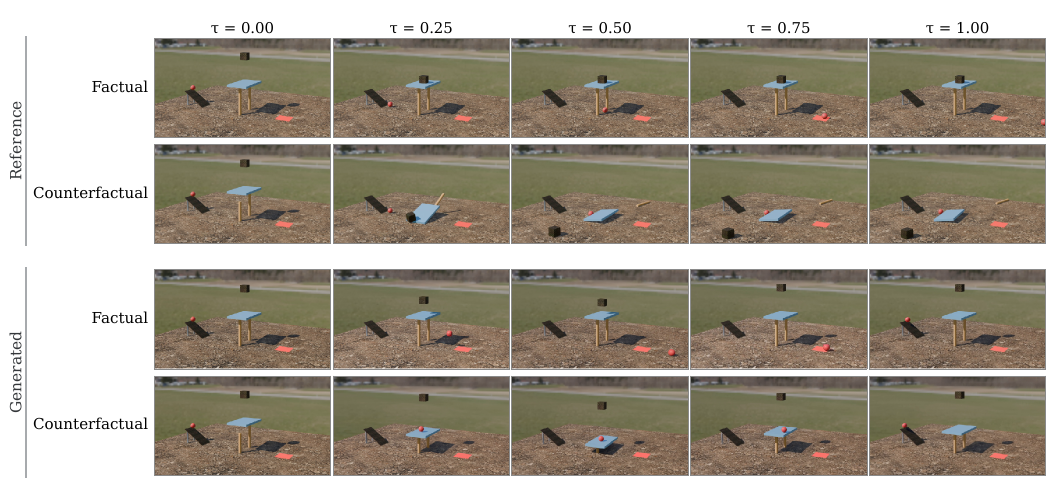}

    \caption{
    \textbf{Generated differences that do not realize the reference intervention consequence.}
    \textbf{Top:} For Cosmos3-Nano, the factual and counterfactual generations differ, but their difference is dominated by identity and geometry drift rather than the required interruption of the domino pathway.
    \textbf{Bottom:} For Wan2.2-I2V-A14B, the scale intervention produces a localized generated response, yet the counterfactual rollout does not reproduce the reference support-loss mechanism or collapsed terminal state.
    These examples show that pairwise change alone does not establish a correct causal response.
    }
    \label{fig:cf_incorrect_consequence}
\end{figure*}

\textbf{Pairwise change does not guarantee the correct consequence.}
The cases in Figure~\ref{fig:cf_incorrect_consequence} illustrate a complementary failure mode in which the generated factual and counterfactual rollouts differ, but not according to the reference physical consequence.
In the move-intervention case, the generated difference is dominated by changes in domino identity and geometry rather than by the required interruption of the edited contact pathway.
The pair therefore changes visibly, but the change does not realize the causal consequence of the intervention.

The scale-intervention case provides a less catastrophic but particularly informative example.
Shortening the support causes the reference counterfactual platform to lose support, collapse, and remain down, while the factual platform remains supported.
The generated pair exhibits a localized difference, but the counterfactual platform does not undergo the reference support-loss process or reach the collapsed terminal state.
Instead, much of the difference comes from rerouted ball motion and later replay-like dynamics.
This case shows that a localized response can be detected without the correct physical mechanism or outcome being realized.

\subsection{Failure to Preserve Unaffected Dynamics}
\label{app:cf_cases_selectivity}

\begin{figure*}[t]
    \centering
    \includegraphics[width=\textwidth]{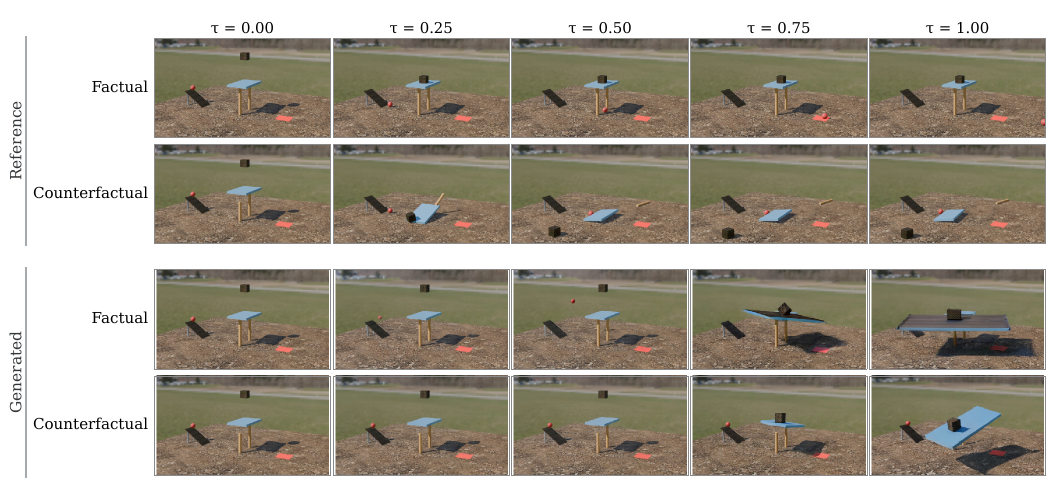}

    \medskip

    \includegraphics[width=\textwidth]{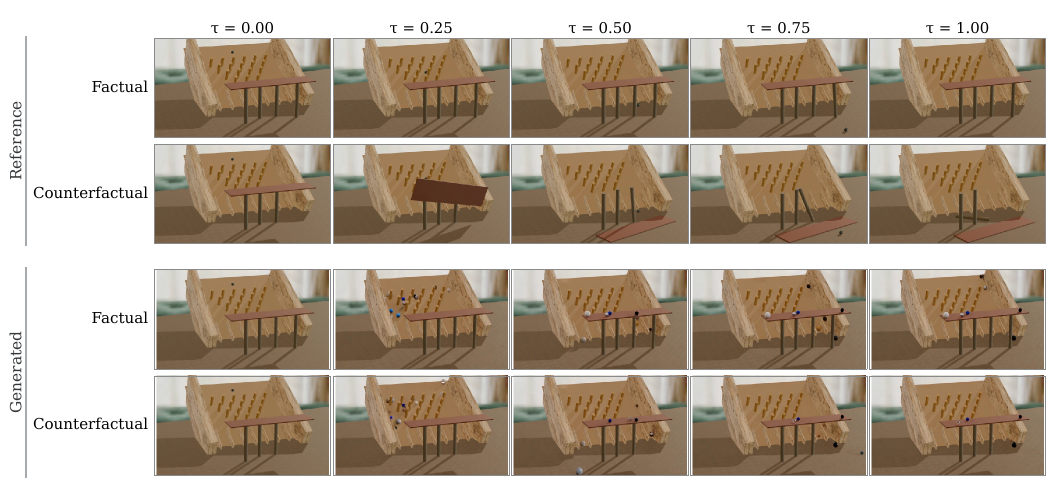}

    \caption{
    \textbf{Representative failures to preserve intervention-unaffected dynamics.}
    \textbf{Top:} Shortening one support should locally alter the platform support relation, but the generated pair exhibits large changes in platform scale and geometry that extend beyond the intervention-affected dynamics.
    \textbf{Bottom:} Deleting one support should primarily alter the platform response, but the generated pair introduces numerous additional balls whose motion is unrelated to the intervention and disrupts dynamics that should remain shared.
    Together, the cases show that differences between generated factual and counterfactual rollouts are not reliably confined to the consequences of the controlled intervention.
    }
    \label{fig:cf_preservation_failures}
\end{figure*}

Limited Preserve performance indicates that generated factual-counterfactual pairs often fail to retain dynamics and scene structure that remain shared in the reference pair.
Figure~\ref{fig:cf_preservation_failures} shows two distinct manifestations of this failure.

\textbf{Geometry corruption beyond the edited support.}
Shortening one support pillar causes a localized support loss in the reference counterfactual while much of the surrounding apparatus remains unchanged.
The generated factual and counterfactual rollouts instead exhibit large changes in platform scale, geometry, and orientation that extend well beyond the edited support.
The generated pair therefore fails to preserve scene structure and dynamics that should remain shared.

\textbf{Spurious motion from object multiplication.}
Deleting one support changes the platform dynamics in the reference counterfactual while leaving much of the remaining scene shared with the factual rollout.
Cosmos3-Nano instead introduces numerous additional balls whose motion is unrelated to the deleted support.
These spurious objects create large pairwise differences and contaminate motion that should remain common to the factual and counterfactual generations.

\begin{figure*}[t]
    \centering

    \begin{minipage}[t]{0.48\textwidth}
        \centering
        \includegraphics[width=\linewidth]{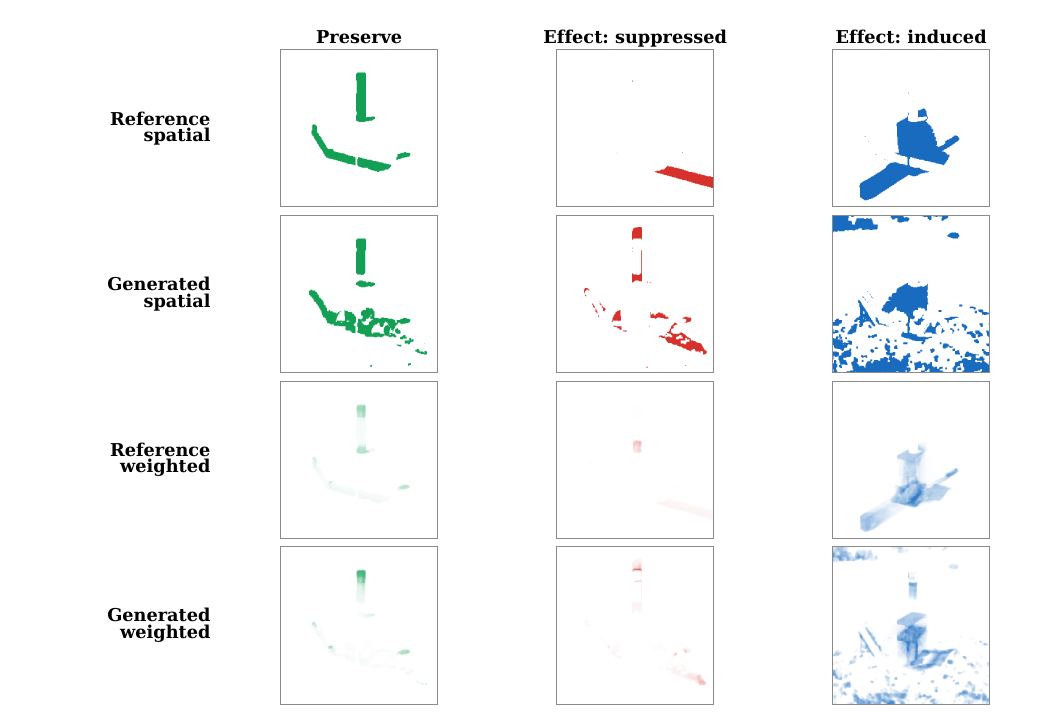}
    \end{minipage}
    \hfill
    \begin{minipage}[t]{0.48\textwidth}
        \centering
        \includegraphics[width=\linewidth]{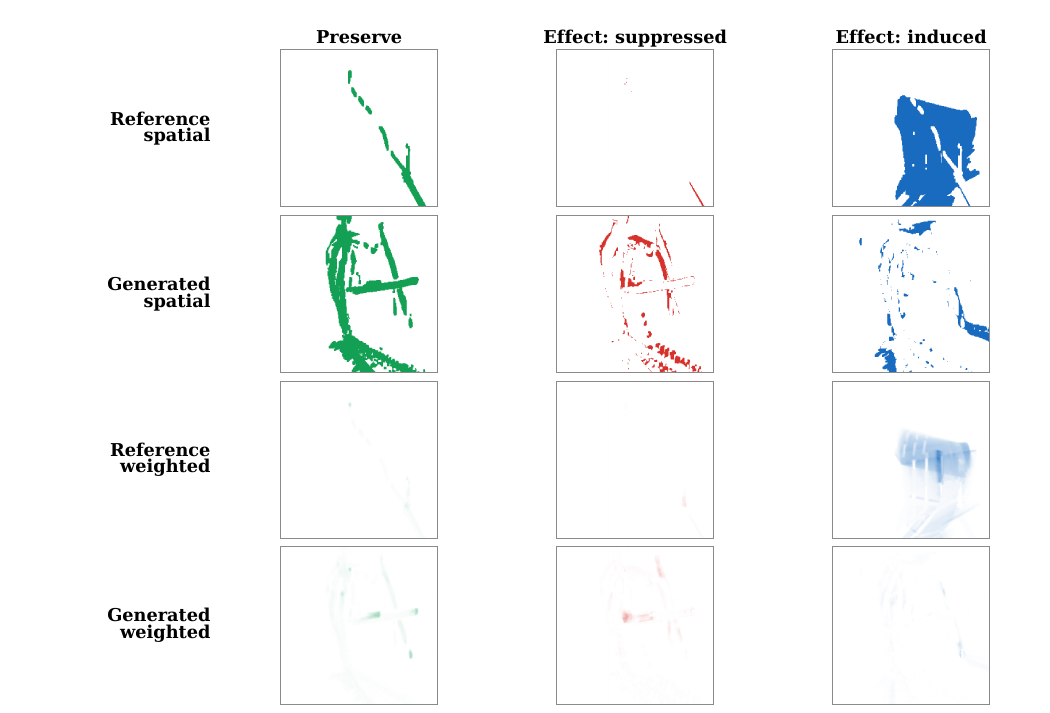}
    \end{minipage}

    \caption{
    \textbf{Motion-map diagnostics for representative counterfactual responses.}
    \textbf{Left:} For the scale intervention, the generated maps retain partial spatial correspondence with the reference response but introduce substantially more diffuse preserved, suppressed, and induced motion; the correct support-loss process is not recovered.
    \textbf{Right:} For the delete intervention, the generated pair introduces broad preserved and suppressed activity while recovering little of the concentrated induced-motion pattern present in the reference pair.
    Columns show preserved, intervention-suppressed, and intervention-induced motion, while rows compare spatial and weighted maps for the reference and generated factual-counterfactual pairs.
    }
    \label{fig:cf_motion_diagnostics}
\end{figure*}

\textbf{Relation to the motion-based metrics.}
Figure~\ref{fig:cf_motion_diagnostics} connects the rollout-level observations above with the spatial and weighted Effect/Preserve metrics.

For the scale-intervention case, the generated maps retain partial correspondence with the reference motion regions, but the response is substantially more diffuse.
In particular, generated induced motion extends well beyond the compact reference region, while suppressed and preserved motion are fragmented across additional areas.
The weighted maps attenuate many of these transient activations but still do not recover the reference support-loss pattern.
This explains why nontrivial spatial overlap can coexist with an incorrect physical mechanism and terminal state.

For the delete-intervention case, the discrepancy is more pronounced.
The reference pair contains a concentrated induced-motion region and relatively limited suppressed motion, whereas the generated pair shows broad preserved and suppressed activity in unrelated regions while recovering little of the reference induced pattern.
The weighted maps further reduce this correspondence, indicating that even where motion occurs in similar locations, its temporal occupancy is not reproduced.
Together, these cases show that the motion-based metrics capture where intervention-induced and preserved activity is reproduced, while rollout inspection remains necessary to determine whether the underlying physical mechanism and causal progression are correct.

\clearpage
\onecolumn
\section{Additional Qualitative Results}
\label{app:additional_qualitative_results}
\newcounter{qualcase}
\renewcommand{\thequalcase}{E.\arabic{qualcase}}

\paragraph{Scope and case selection.}
To support balanced cross-model and prompt-level comparisons, we construct the qualitative subset at the scene level. Four Level~1 scenes---a seesaw-triggered domino chain, a dual-ball inverted-Y domino course, a horizontal Galton board with object targets, and an inclined Galton board with domino targets---provide 36 rollouts for cross-model comparison. These scenes are selected to cover diverse causal structures, including chain propagation, branching interactions, and multi-stage physical transitions. The first three scenes are additionally examined under Levels~2 and~3, contributing 54 rollouts and forming 27 matched model-scene trajectories spanning all three prompt levels. Together, these yield 90 rollouts with uniform model coverage, enabling direct comparisons without selecting cases based on the success or failure of any individual model. Every judgment is based on reviewing the full generated videos and their corresponding ground-truth videos; the four-frame strips in the qualitative evidence atlas serve as compact visual summaries of the corresponding full-video analyses.

\subsection{Qualitative Analysis Protocol}
\label{app:qualitative_protocol}

Our goal is to determine whether a model generates, from the given initial state, a complete physical process that is consistent with the ground-truth causal cascade. We apply the same protocol to every rollout included in this qualitative study. We first reconstruct the reference process from the ground-truth video and event graph, including the participating entities, static scene structures, trigger and mediator events, direct causal edges, temporal stages, downstream outcomes, and terminal state. For scenes with parallel branches, direct dependencies are tracked separately so that temporal proximity is not mistaken for causation.

The qualitative analysis uses the same five diagnostic dimensions as the graph-grounded VQA evaluation: \emph{event occurrence}, \emph{causal pathway}, \emph{temporal ordering}, \emph{outcome and termination}, and \emph{scene consistency}. The deterministically generated, closed-ended VQA questions standardize what is assessed and how responses are scored across scenes and models, but their fixed templates and answer spaces cannot fully characterize scene-specific interactions or nuanced failure trajectories; human inspection therefore provides the interpretive flexibility needed for fine-grained, case-level analysis.

We compare each generated video with its reference chronologically and prioritize the earliest unambiguous deviation. This ordering prevents multiple later abnormalities from being counted as independent primary failures when they are consequences of the same upstream error. We first describe only what is directly observable, then assign the deviation to one or more diagnostic dimensions, identify the unique dimension containing the earliest primary failure, and trace its downstream effects through the event graph.

\paragraph{Observable manifestations.}
At this stage, we record visible abnormalities without inferring their internal causes. These include changes in object identity or geometry, duplication, disappearance, discontinuous displacement, corruption of fixed structures or spatial relations, unsupported motion, incorrect direction or physical response, reversed or recurrent motion, stalled evolution, failure to reach or maintain a terminal state, and camera or viewpoint drift. The manifestation labels are multilabel; the primary diagnostic dimension is assigned separately from this descriptive coding.

\paragraph{Position in the causal cascade and downstream effects.}
After describing an observable error, we identify the violated dimensions and trace its downstream effects through the causal process.

\begin{enumerate}
    \setlength{\itemsep}{0.4em}
    \setlength{\parskip}{0pt}
    \setlength{\parsep}{0pt}

    \item \textbf{Event occurrence} asks whether the reference trigger and mediator events occur. Human inspection additionally checks whether each event is realized by the correct participant and target, in the correct direction and interaction mode, with an appropriate state transition and degree of completion. Missing triggers or mediators, incorrectly realized reference events, and spurious events absent from the reference graph are coded here.

    \item \textbf{Causal pathway} asks whether an observed downstream event is produced through the direct physical dependency annotated in the reference graph. Qualitative failures include broken propagation, a causal shortcut that bypasses a required mediator, an incorrect intermediate pathway, reversed causal direction, and a spurious dependency or branch.

    \item \textbf{Temporal ordering} asks whether observed events occur in the correct relative order and whether parallel or successive stages remain coordinated. Human inspection extends the pairwise VQA questions to premature or delayed events, branch desynchronization, replay of an intermediate state, recurrent motion, and stalled progression.

    \item \textbf{Outcome and termination} asks whether the principal downstream outcomes occur, whether the reference terminal state is reached, and whether it remains stable. Human inspection further distinguishes an incomplete or incorrect outcome, premature termination, failure to terminate, continued dynamics after apparent completion, and post-terminal regression that revokes an already completed cascade.

    \item \textbf{Scene consistency} asks whether the generated process preserves stable referents and a coherent physical setting. It includes entity persistence, trajectory continuity, identity and geometry integrity, stable object count and spatial layout, fixed supports and structures, and camera stability.
\end{enumerate}

Together, the five dimensions connect the physical substrate of a generated scene to the events, relations, and final state of its causal cascade. Scene consistency supports traceable participants; event occurrence supplies the event nodes; causal pathway and temporal ordering provide complementary relational tests of why and when those events occur, and outcome and termination evaluates completion of the resulting cascade. By locating the earliest breakdown and tracing its downstream effects, our analysis provides a unified account of where a generated rollout first departs from the reference cascade, how much of the cascade it successfully realizes, and how the initial deviation shapes subsequent events and outcomes.

\paragraph{Cautious interpretation of causes and consequences.}
Only after the factual description and causal localization are complete do we offer an interpretation of potential causes. A failure may be consistent with incorrect inference of scene structure or support relations; inaccurate realization of gravity, collision, torque, or rigid-body dynamics; failure to preserve direct dependencies between events; failure to maintain object identity, world state, or completed history; or unstable grounding between the text condition, initial image, and spatial layout. These evidence-grounded interpretations relate observable failures to the model capabilities required for causal-cascade generation.

\subsection{Cross-Model Comparison under Level~1}
\label{app:level1_cross_model}

We first ask how the nine models behave when they receive only the initial physical state, without an additional event description. The analysis covers four Level~1 scenes and nine models, for 36 model-level cases (Cases~\ref{case:b01-m01}--\ref{case:b04-m09}).

For CogVideoX-Fun-V1.5, the most consistent pattern is that entities and scene structure break down before a causal cascade can be established. The seesaw, ramp, or central assembly undergoes severe early deformation in the two domino scenes, so subsequent motion can no longer be assigned to stable participants (Cases~\ref{case:b01-m01} and~\ref{case:b02-m01}). Again, in the horizontal Galton board, structural blur occurs (Case~\ref{case:b03-m01}). Compared with models that preserve the scene but connect causal events incorrectly, CogVideoX usually deviates first in scene consistency, after which event occurrence, causal pathway, and temporal ordering are often no longer reliably defined.

Cosmos3-Nano generally preserves the approximate appearance of the scene, and some generated objects exhibit locally plausible motion, but it struggles to maintain the uniqueness and continuity of causal participants. In the seesaw scene, the blue ball bypasses part of the mechanical route and is then duplicated (Case~\ref{case:b01-m02}); in the dual-ball scene, several balls change position without continuous trajectories (Case~\ref{case:b02-m02}); and in both Galton-board scenes, a single ball is replaced by multiple instances with different appearances and trajectories (Cases~\ref{case:b03-m02} and~\ref{case:b04-m02}). These failures rewrite a single-particle reference path as multiple spurious parallel branches.  

HunyuanVideo-I2V often preserves scene appearance better, but repeatedly generates motion without the required upstream cause. In both domino scenes, downstream structures begin to change before the ball or an intermediate mediator has completed the necessary contact, producing a clear causal shortcut (Cases~\ref{case:b01-m03} and~\ref{case:b02-m03}). In the inclined Galton board, pegs that should be fixed obstacles become moving participants (Case~\ref{case:b04-m03}). Therefore, HunyuanVideo-I2V's most consistent failure is spontaneous, causally unsupported motion: these motions introduce spurious events and causal branches even when the overall scene appearance remains relatively stable.

LTX-Video-13B is best characterized by an unstable commitment to the evolving physical state. In the seesaw scene, the model initiates the blue ball's motion but later returns it to a previously occupied region (Case~\ref{case:b01-m04}). In the dual-ball scene, the left ramp and nearby ball progressively lose their rigid structure and recognizable identities (Case~\ref{case:b02-m04}). In the horizontal Galton board, a fixed upper board becomes an active moving structure, followed by an unsupported upward return of the ball (Case~\ref{case:b03-m04}); in the inclined board, the single moving ball is duplicated into several divergent trajectories (Case~\ref{case:b04-m04}).  The model can initiate locally plausible transitions, but does not reliably treat their consequences as persistent constraints on subsequent events.

MAGI-1-4.5B exhibits difficulties primarily in entity persistence and the grounding of a fixed observation condition. In the seesaw scene, both balls change size(Case~\ref{case:b01-m05}). In the dual-ball scene, persistent camera drift is accompanied by incomplete realization of the initial gravity trigger (Case~\ref{case:b02-m05}). In the horizontal Galton board, the moving ball disappears before completing its descent, again with viewpoint change (Case~\ref{case:b03-m05}). In conclusion, MAGI's primary limitation is closer to entity and scene grounding that does not persist over time.

SkyReels-V2-I2V produces substantial visible activity, but this motion frequently lacks the correct trigger, direction, or contact basis. In the seesaw scene, the blue ball first moves in the wrong direction without a cause and later repeats the motion (Case~\ref{case:b01-m06}). In the dual-ball scene, several central objects rise and move laterally without contact, replacing the reference cascade with spurious branches (Case~\ref{case:b02-m06}). In the horizontal Galton board, peg collisions after the initial descent send the ball in the wrong direction and then activate irrelevant objects (Case~\ref{case:b03-m06}); in the inclined board, the ball does not continue its gravity-driven descent and the camera position changes (Case~\ref{case:b04-m06}). The common problem is that the dynamics are not reliably constrained by the initial state and the contact-mediated reference path. 

Wan2.1-I2V-14B is comparatively more likely to identify and initiate a salient early motion in the reference process, but it usually loses consistency in local dynamics, entity identity, temporal ordering, or terminal-state maintenance. In the seesaw scene, the blue ball initially moves in the expected direction, but the board produces the opposite torque response and the ball later returns to an earlier position (Case~\ref{case:b01-m07}). In the dual-ball scene, the model recognizes that a ball should descend along the ramp but changes the identity and subsequent path of the moving ball (Case~\ref{case:b02-m07}). In both Galton-board scenes, the ball departs from the correct gravity-driven trajectory but exhibits reversal or oscillation (Cases~\ref{case:b03-m07} and~\ref{case:b04-m07}). Wan2.1's relative strength is thus its ability to preserve a recognizable causal prefix. Its first clear deviation often occurs later than in models whose scenes collapse immediately, but incorrect trajectories, torque, or completed-state preservation still interrupt downstream propagation.

Wan2.2-TI2V-5B frequently deviates at the trigger or the first stage of motion immediately following it. In the seesaw scene, the red ball moves spontaneously, the blue ball departs from its target path, and the seesaw itself deforms (Case~\ref{case:b01-m08}). In the dual-ball scene, the triggering ball does not descend along the ramp, while irrelevant central objects move and change identity (Case~\ref{case:b02-m08}). In the horizontal Galton board, the ball begins descending and then returns upward spontaneously (Case~\ref{case:b03-m08}); in the inclined board, the gravity trigger is missing while identity change is introduced (Case~\ref{case:b04-m08}). The model therefore does not exhibit a correct causal prefix that is stable across scenes; its limitations jointly involve event occurrence, scene consistency, and temporal ordering.

Wan2.2-I2V-A14B is qualitatively distinguished by the greater causal depth it can attain before failure. In the seesaw scene, it preserves the relevant participants, realizes the initial ball motion, and produces the mechanically coupled board tilt, thereby maintaining one of the longest correct causal prefixes (Case~\ref{case:b01-m09}). The advantage is nevertheless scene-dependent: in the dual-ball scene, the trigger ball changes identity and stops descending before the central dominoes topple without contact (Case~\ref{case:b02-m09}); in the horizontal Galton board, the initial descent occurs but follows an incorrect route and later reverses (Case~\ref{case:b03-m09}); and in the inclined board, the gravity-driven trigger is missing while downstream dominoes move prematurely (Case~\ref{case:b04-m09}). These cases show that greater causal depth is a capability rather than a guarantee: when the initial state and early interactions are correctly grounded, Wan2.2-I2V-A14B carries the reference process farther and shifts the first failure from cascade initiation toward downstream propagation or temporal-state maintenance.

\subsection{Effect of Prompt Information}
\label{app:prompt_effect_qualitative}

To determine whether additional prompt information genuinely improves generation, we compare the same model on the same scene from Level~1 through Levels~2 and~3. A meaningful improvement should lengthen the correct causal prefix, move the earliest deviation to a later stage, recover more trigger and mediator occurrences, preserve the reference causal pathway and temporal ordering, or reach and maintain the terminal state more reliably. The analysis covers three scenes with all three prompt levels and nine models, producing 27 matched model-scene trajectories. Because a model may benefit in only some scenes, we organize the results by the response pattern of each trajectory rather than assigning each model permanently to one category.

The first group exhibits relatively consistent improvement. On the dual-ball scene, Wan2.1-I2V-14B changes from early identity and trajectory errors to correctly initiating gravity-driven motion at Level~3, failing only at later domino collisions and state preservation (Cases~\ref{case:b02-m07}, \ref{case:b06-m07}, and~\ref{case:b09-m07}). On the horizontal Galton board, it similarly improves from an early incorrect trajectory and upward return to completing most of the reference route, with state replay only after the process has ended (Cases~\ref{case:b03-m07}, \ref{case:b07-m07}, and~\ref{case:b10-m07}). Wan2.2-I2V-A14B follows the same pattern on the horizontal Galton board: lateral deviation and motion recurrence at Level~1 give way to a largely correct descent at Levels~2 and~3, leaving only slight temporal misalignment or terminal-state recurrence (Cases~\ref{case:b03-m09}, \ref{case:b07-m09}, and~\ref{case:b10-m09}). Cosmos3-Nano also progressively recovers more of the early ball motion and plank response in the seesaw scene, although downstream contact propagation remains incorrect (Cases~\ref{case:b01-m02}, \ref{case:b05-m02}, and~\ref{case:b08-m02}). Richer prompts can therefore preserve a longer part of the reference process for some trajectories, but the first gains tend to be in event structure and causal-prefix length rather than a complete solution to continuous dynamics.

The second group shows scene-dependent partial improvement: the same model benefits in one scene without transferring the gain to other structures. At Level~2, SkyReels-V2-I2V largely recovers the correct descent and contact path in the horizontal Galton board, leaving only slight displacement of fixed pegs; at Level~3, however, ball duplication and irrelevant object motion reappear, while its wrong directions, structural instability, and spurious motion in the two domino scenes do not disappear in parallel (Cases~\ref{case:b03-m06}, \ref{case:b07-m06}, and~\ref{case:b10-m06}). MAGI-1-4.5B better recovers the early descent in the Level~3 horizontal Galton board but then changes the structural role of a fixed board, while camera drift and a missing trigger persist across prompt levels in the dual-ball scene (Cases~\ref{case:b03-m05}, \ref{case:b07-m05}, and~\ref{case:b10-m05}; Cases~\ref{case:b02-m05}, \ref{case:b06-m05}, and~\ref{case:b09-m05}). Prompt effects are therefore jointly constrained by scene geometry, contact-chain length, and persistence requirements; a local gain does not imply a generalizable causal-generation capability.

The third group changes its error type as the prompt level changes but does not exhibit a substantive causal improvement. Across the seesaw levels, HunyuanVideo-I2V changes from unsupported downstream motion, to incorrect torque and collision responses, to a failure of the initial dynamics at Level~3. Its failures in the dual-ball and Galton-board scenes likewise shift among causal shortcuts, incorrect trajectories, identity changes, and viewpoint drift (Cases~\ref{case:b01-m03}, \ref{case:b05-m03}, and~\ref{case:b08-m03}; Cases~\ref{case:b02-m03}, \ref{case:b06-m03}, and~\ref{case:b09-m03}; Cases~\ref{case:b03-m03}, \ref{case:b07-m03}, and~\ref{case:b10-m03}). LTX-Video-13B similarly moves among motion recurrence, structural misinterpretation, entity changes, and a missing trigger, without consistently recovering direct physical dependencies across the three scenes (Cases~\ref{case:b01-m04}, \ref{case:b05-m04}, and~\ref{case:b08-m04}; Cases~\ref{case:b02-m04}, \ref{case:b06-m04}, and~\ref{case:b09-m04}; Cases~\ref{case:b03-m04}, \ref{case:b07-m04}, and~\ref{case:b10-m04}). Such failure substitution indicates that a more detailed prompt may alter the motion pattern selected by a model without bringing the generated process closer to the reference causal graph.

The fourth group is largely insensitive to prompt enrichment because its primary bottleneck lies in scene consistency or initial event occurrence rather than missing event information. CogVideoX-Fun-V1.5 repeatedly corrupts entity identity, rigid structure, and global scene topology in all three matched scenes; even when Level~3 occasionally recovers a local ball motion, the structure and object coupling that should carry the event remain unstable (Cases~\ref{case:b01-m01}, \ref{case:b05-m01}, and~\ref{case:b08-m01}; Cases~\ref{case:b02-m01}, \ref{case:b06-m01}, and~\ref{case:b09-m01}; Cases~\ref{case:b03-m01}, \ref{case:b07-m01}, and~\ref{case:b10-m01}). Trigger, structural, and temporal-state failures in Wan2.2-TI2V-5B likewise do not decrease consistently with prompt level, and some Level~3 generations exhibit even more severe entity or scene collapse (Cases~\ref{case:b01-m08}, \ref{case:b05-m08}, and~\ref{case:b08-m08}; Cases~\ref{case:b02-m08}, \ref{case:b06-m08}, and~\ref{case:b09-m08}; Cases~\ref{case:b03-m08}, \ref{case:b07-m08}, and~\ref{case:b10-m08}). MAGI's persistent viewpoint drift and missing initial gravity motion in the dual-ball scene further show that a description of later events cannot directly recover the full cascade when the limiting factor is scene consistency, structural grounding, fixed observation conditions, or basic dynamics.

Overall, Levels~2 and~3 improve event-level adherence for some models and scenes, and in several trajectories move failure from the trigger or route-selection stage to contact propagation, temporal control, or termination. These gains are neither universal nor monotonically related to prompt level. Additional information more reliably reduces uncertainty about \emph{what should happen} than it resolves \emph{how those events should happen}; entity persistence, structural constraints, contact dynamics, force transmission, and temporal history remain the principal bottlenecks. Richer prompts should therefore be understood as support for specifying causal events, not as a substitute for the capacity to realize the complete physical process.

\subsection{Recurring Failure Patterns}
\label{app:recurring_failures}

We finally ask which failure patterns recur across model families and prompt levels, where they first emerge in the causal cascade, and how they reshape the subsequent process. To answer this question, we synthesize the complete qualitative descriptions of all 90 cases. We first apply multilabel coding to the observable errors in the descriptions and then examine how these errors combine across models. Because a frequent manifestation may either be an upstream error that diverts the reference cascade or a downstream symptom of an earlier failure, we additionally assign each case one earliest primary failure. Finally, we trace how the deviation changes necessary events, propagation relations, alternative branches, and established temporal states, successively addressing how failures manifest, how they are distributed, where they originate, and how they propagate.

Table~\ref{tab:full_description_failures} first reports the observable errors that recur in the complete descriptions. The two most prominent shared problems concern the control of motion initiation and physical response: spontaneous or unsupported motion occurs in 51 cases, and incorrect physical response occurs in 47. The dominant difficulty is therefore not limited to omitting a complex downstream event. More fundamentally, models often fail to constrain when an object should move and how it should respond after contact, collision, or loss of support. Moreover, 74 cases contain at least two error categories, with an average of 2.36 per case. This extensive co-occurrence shows that generation failures are seldom isolated local defects; instead, they form interconnected combinations across object identity, spatial structure, motion response, and temporal state.

\begin{table}[htbp]
\centering
\caption{Multilabel observable failures coded from the complete qualitative descriptions.}
\label{tab:full_description_failures}
\small
\setlength{\tabcolsep}{5pt}
\renewcommand{\arraystretch}{1.12}
\begin{tabular*}{\textwidth}{@{\extracolsep{\fill}}lllll@{}}
\toprule
Failure manifestation & All cases & Level~1 & Level~2 & Level~3 \\
\midrule
Spontaneous motion & 51/90 (56.7\%) & 20/36 (55.6\%) & 16/27 (59.3\%) & 15/27 (55.6\%) \\
Incorrect physical response & 47/90 (52.2\%) & 20/36 (55.6\%) & 17/27 (63.0\%) & 10/27 (37.0\%) \\
Identity or attribute corruption & 26/90 (28.9\%) & 10/36 (27.8\%) & 8/27 (29.6\%) & 8/27 (29.6\%) \\
Motion recurrence or state replay & 23/90 (25.6\%) & 11/36 (30.6\%) & 6/27 (22.2\%) & 6/27 (22.2\%) \\
Topology corruption & 21/90 (23.3\%) & 8/36 (22.2\%) & 5/27 (18.5\%) & 8/27 (29.6\%) \\
Structural-role misinterpretation & 15/90 (16.7\%) & 7/36 (19.4\%) & 4/27 (14.8\%) & 4/27 (14.8\%) \\
Camera or viewpoint drift & 11/90 (12.2\%) & 4/36 (11.1\%) & 4/27 (14.8\%) & 3/27 (11.1\%) \\
Object duplication & 10/90 (11.1\%) & 5/36 (13.9\%) & 2/27 (7.4\%) & 3/27 (11.1\%) \\
Position jump & 7/90 (7.8\%) & 6/36 (16.7\%) & 1/27 (3.7\%) & 0/27 (0.0\%) \\
Pure temporal misalignment & 1/90 (1.1\%) & 0/36 (0.0\%) & 1/27 (3.7\%) & 0/27 (0.0\%) \\
\bottomrule
\end{tabular*}
\end{table}

These aggregate frequencies establish which errors recur in the benchmark, but not whether all models share the same capability bottleneck. Table~\ref{tab:model_failure_profiles} therefore decomposes the observable errors into model-specific profiles and reports the three most frequent categories in each model's ten cases.

\begin{table}[htbp]
\centering
\caption{Model-specific failure profiles on each model's three most frequent error manifestations.}
\label{tab:model_failure_profiles}
\scriptsize
\setlength{\tabcolsep}{2.5pt}
\renewcommand{\arraystretch}{1.12}
\begin{tabular*}{\textwidth}{@{\extracolsep{\fill}}lcccccccc@{}}
\toprule
Model
& \makecell{Topology}
& \makecell{Unsupported\\motion}
& \makecell{Identity\\error}
& \makecell{Physical\\response}
& \makecell{Duplication}
& \makecell{Structural\\role}
& \makecell{Viewpoint\\drift}
& \makecell{State\\replay} \\
\midrule
CogVideoX-Fun-V1.5
& \cellcolor{heatblue!100}\textbf{10} & \cellcolor{heatblue!60}6 & \cellcolor{heatblue!30}3 & -- & -- & -- & -- & -- \\
Cosmos3-Nano
& -- & \cellcolor{heatblue!50}5 & -- & \cellcolor{heatblue!70}\textbf{7} & \cellcolor{heatblue!50}5 & -- & -- & -- \\
HunyuanVideo-I2V
& -- & \cellcolor{heatblue!90}\textbf{9} & \cellcolor{heatblue!40}4 & \cellcolor{heatblue!40}4 & -- & -- & -- & -- \\
LTX-Video-13B
& -- & \cellcolor{heatblue!30}3 & -- & \cellcolor{heatblue!80}\textbf{8} & -- & \cellcolor{heatblue!50}5 & -- & -- \\
MAGI-1-4.5B
& -- & -- & \cellcolor{heatblue!50}\textbf{5} & \cellcolor{heatblue!30}3 & -- & -- & \cellcolor{heatblue!50}\textbf{5} & -- \\
SkyReels-V2-I2V
& \cellcolor{heatblue!30}3 & \cellcolor{heatblue!90}\textbf{9} & -- & \cellcolor{heatblue!50}5 & -- & -- & -- & -- \\
Wan2.1-I2V-14B
& -- & \cellcolor{heatblue!60}6 & -- & \cellcolor{heatblue!70}\textbf{7} & -- & -- & -- & \cellcolor{heatblue!70}\textbf{7} \\
Wan2.2-TI2V-5B
& -- & \cellcolor{heatblue!70}\textbf{7} & \cellcolor{heatblue!50}5 & \cellcolor{heatblue!60}6 & -- & -- & -- & -- \\
Wan2.2-I2V-A14B
& -- & \cellcolor{heatblue!40}4 & -- & \cellcolor{heatblue!50}\textbf{5} & -- & -- & -- & \cellcolor{heatblue!50}\textbf{5} \\
\bottomrule
\end{tabular*}
\end{table}

Table~\ref{tab:model_failure_profiles} reveals a failure structure shaped jointly by shared bottlenecks and model-specific differences. Unsupported motion and incorrect physical response each appear among the three most frequent errors for eight of the nine models, confirming that correct motion initiation and contact response are pervasive cross-model difficulties. The models differ mainly in which secondary failure accompanies this common dynamics bottleneck. CogVideoX combines motion errors with pronounced topology and identity corruption; Cosmos3-Nano also frequently duplicates objects; HunyuanVideo more often exhibits identity changes; and LTX-Video is distinguished by structural-role misinterpretation. MAGI combines identity corruption with viewpoint drift, whereas Wan2.1 and Wan2.2-14B both exhibit recurrent state replay. The resulting profiles expose additional limitations in structural representation, entity persistence, observation stability, and temporal-state maintenance.

The temporal position of these errors in the complete case descriptions further shows that the underlying capability requirements are not equally deep in the causal cascade. CogVideoX's topology and rigid-structure failures typically interrupt the process before a stable physical world can support the cascade. By contrast, recurrent state replay in Wan2.1 and Wan2.2-14B requires the rollout to establish a temporal state before returning to or overwriting it. A later-emerging failure can therefore indicate that a model has satisfied more upstream requirements and may preserve greater capability for causal-cascade generation. Multilabel error frequencies alone cannot establish this distinction, however, because a late manifestation may still rest on an earlier unobserved deviation.

To obtain an evaluation that is more sensitive to causal progress, we next locate the earliest unambiguous deviation in every rollout. The correct causal prefix retained before that point indicates how far a model advances through the reference process, while the dimension containing the primary failure identifies the first capability that interrupts this progress. Table~\ref{tab:primary_failure} summarizes this location across all 90 cases, distinguishing rollouts that fail before the scene and event foundations have been established from those that expose pathway, temporal-ordering, or terminal-state errors only after satisfying more upstream conditions.

\begin{table}[htbp]
\centering
\caption{Location of the earliest primary failure. Each rollout receives exactly one label.}
\label{tab:primary_failure}
\small
\begin{tabular}{@{}p{0.64\textwidth}>{\centering\arraybackslash}p{0.12\textwidth}>{\centering\arraybackslash}p{0.16\textwidth}@{}}
\toprule
Earliest primary failure & Cases & Proportion \\
\midrule
Event occurrence & 43 & 47.8\% \\
Causal pathway & 8 & 8.9\% \\
Temporal ordering & 4 & 4.4\% \\
Outcome and termination & 2 & 2.2\% \\
Scene consistency & 33 & 36.7\% \\
\bottomrule
\end{tabular}
\end{table}

Table~\ref{tab:primary_failure} reveals a failure structure concentrated at the foundational front end of the causal cascade. Scene consistency is the earliest point of failure in more than one third of the cases, showing that many models cannot reliably preserve the entity identities, geometry, support relations, and spatial configuration required for a continuous physical process. Event occurrence accounts for nearly half of the earliest deviations, indicating that even when the scene remains sufficiently interpretable, models frequently fail to realize the trigger or an early mediator needed to initiate and sustain the reference cascade. Together, these two dimensions account for 84.4\% of the primary failures, locating the dominant bottleneck in establishing a persistent physical world and its event-level foundations rather than in completing the final stages of an otherwise correct causal chain.

This funnel-shaped distribution also reflects conditional dependencies among the five diagnostic dimensions. Scene consistency supplies traceable participants and stable geometry, and event occurrence supplies the nodes required to evaluate causal relations. Once these conditions are met, causal pathway and temporal ordering can test why and when the observed events occur, while outcome and termination can further test whether the process completes and remains complete. The smaller number of primary failures at the relational and terminal dimensions therefore does not imply that models have solved these capabilities. A rollout must first preserve a sufficiently long and interpretable preceding process before a deeper failure can become observable. Only 14 of the 90 cases place their earliest deviation in a relational or terminal dimension, and only two preserve the reference process until terminal-state maintenance becomes the primary failure.

Locating the first breakdown establishes where a rollout leaves the reference process, but not how that deviation reshapes the remainder of the cascade. Starting from this deviation, we therefore trace the observable errors in each complete case through the reference event graph and record their effects on the subsequent process using multilabel coding. We combine missing necessary events and broken propagation into a single consequence because the absence of a trigger, mediator, or downstream event is often the specific form through which propagation is truncated. We separately retain temporal-progression and terminal-state failures to capture state replay, temporal regression, and loss of a correctly reached outcome, which cannot be represented by changes to event-graph structure alone. Table~\ref{tab:causal_consequences} thus extends the single upstream location identified in Table~\ref{tab:primary_failure} into five cascade-level consequences that may co-occur within a case.

\begin{center}
\begin{minipage}{0.82\textwidth}
\captionsetup{type=table}
\centering
\caption{Effects of observable errors. Categories are multilabel.}
\label{tab:causal_consequences}
\small
\begin{tabular}{lrr}
\toprule
Causal consequence & Cases & Proportion \\
\midrule
Missing required event or cascade truncation & 53 & 58.9\% \\
Spurious branch or unsupported activation & 53 & 58.9\% \\
Causal shortcut & 12 & 13.3\% \\
Reference causal-pathway deviation & 52 & 57.8\% \\
Temporal-progression or terminal-state failure & 26 & 28.9\% \\
\bottomrule
\end{tabular}
\end{minipage}
\end{center}

The 90 cases yield 196 consequence labels, or 2.18 per case on average. Every case receives at least one label, and 69 cases receive two or more. The rows in Table~\ref{tab:causal_consequences} therefore do not partition the 90 cases; each count instead reports the number of complete rollouts in which that consequence occurs.

The distribution reveals a dual bottleneck between weakening the reference cascade and rewriting it. Missing events or truncation occur in 53 cases, showing that models frequently fail to establish or sustain a physical process through successive interactions. At the same time, 53 cases introduce spurious branches or unsupported activations, and 52 deviate from the reference causal pathway. Models therefore do not merely terminate the intended process prematurely; they often replace it with alternative activity that remains locally visible but is organized by the wrong causal structure.

We identify a causal shortcut in 12 cases. Although less frequent than unsupported motion in general, shortcuts are especially consequential for evaluation because event occurrence or final-state similarity alone can misclassify them as successful generations. Temporal-progression or terminal-state failure occurs in a further 26 cases, showing that previously generated event history does not reliably constrain later generation even after part of the reference cascade has been realized. State replay and motion recurrence further return the rollout to a stage that should already have been resolved. Causal fidelity therefore requires not only the correct event content, but also the correct generative pathway and irreversible preservation of the world states established along it.

Taken together, the four analyses show that current models can generate substantial local motion and recognizable events, yet still struggle to make stable entities, physical interactions, and established event history jointly constrain a continuous causal process. Co-occurring observable errors show that failures commonly span several capability dimensions; model-specific profiles reveal different combinations of these bottlenecks; the earliest primary failure indicates the length of the correct causal prefix a model can preserve; and the downstream consequences show how an upstream deviation truncates or rewrites the remainder of the process. When these constraints fail early, scene corruption or missing necessary events directly truncate the cascade. When a model advances further, its errors more often bypass necessary mediators, introduce spurious branches, or regenerate states that should already be complete. Improved capability should therefore be reflected not merely in fewer errors or the eventual appearance of a target event, but in a progressively longer correct causal prefix whose established dependencies and world states continue to constrain subsequent generation. A model that exposes a deeper temporal or terminal failure may consequently preserve more causal-generation capability than one that fails at the scene-foundation stage. The central limitation of current models is their inability to organize locally plausible changes into a complete causal cascade with valid dependencies, persistent states, and irreversible forward progression.

\subsection{Key Findings on Causal Video Generation}
\label{app:qualitative_key_findings}

The results first show that the five diagnostic dimensions form a dependency-aware system rather than a flat collection of scores. Scene consistency must preserve traceable participants and structures; event occurrence must supply the relevant trigger and mediator nodes; causal pathway and temporal ordering must preserve the physical and temporal relations among those nodes; and outcome and termination must complete the resulting process without revoking its endpoint. Model capability is therefore determined not only by how many expected events eventually appear, but also by how deeply the reference causal chain survives before the first deviation. We call this property \emph{causal depth}: the length of the correct causal prefix that a model can preserve continuously from the initial state. The first-error funnel in Section~\ref{app:recurring_failures} provides direct evidence for this view. A lower raw count in a relational or endpoint dimension does not make that dimension unimportant; it often means that the model has already satisfied more upstream conditions before exposing the failure.

Second, many current models behave more like systems that complete a salient event script than systems that simulate a continuous causal process. A model may generate a prompted downstream event or approximate terminal state while omitting a necessary trigger or mediator, reaching the result through an incorrect pathway, or making the target event occur without physical contact. Such a rollout can have high surface-level event coverage without recovering the causal edges in the ground truth. Success therefore cannot be determined solely by whether expected events appear or whether the final frame looks similar. Evaluation must also verify why the events occur, through which mediators their effects propagate, and whether the generated path agrees with the reference causal graph. This distinction is central to the value of CausalCascade over event-recognition or outcome-matching evaluation.

Third, richer prompts primarily lengthen the correct causal prefix for some models rather than generally repairing event occurrence and physical realization. Levels~2 and~3 reduce uncertainty about which subsequent events should occur and, in several matched trajectories, postpone the earliest deviation from the trigger or initial route choice to pathway propagation, temporal ordering, or termination. The benefit nevertheless depends on both model and scene, is not monotonic in prompt level, and sometimes only substitutes one failure for another. Prompt enrichment thus constrains high-level event structure more effectively than it supplies missing contact dynamics, scene consistency, or temporal-state modeling. Whether the correct causal prefix becomes longer characterizes the actual benefit of added prompt information more faithfully than final-outcome comparison alone.

Fourth, scene consistency is not a generic visual-quality concern but an underestimated foundational bottleneck for causal generation. Once an object is duplicated, changes identity, loses its topology, or follows a discontinuous trajectory, subsequent events no longer refer to the same stable participants and contact relations, and causal edges lose their meaning. A video may remain active and locally plausible without forming a traceable causal cascade. The frequency with which scene consistency is the first failure, together with the lack of a consistent reduction in identity and topology errors under richer prompts, shows that a generator must preserve more than frame-to-frame appearance: it needs a stable physical world capable of carrying events and interactions.

Finally, temporal ordering and outcome and termination jointly reveal that a complete causal cascade requires an irreversible commitment to state transitions that have already occurred. In multiple cases, a rollout initiates the correct motion or completes part of the propagation, but then returns an object to an earlier position, regenerates a completed motion, or restores a structure that had already changed. These behaviors are not merely local trajectory noise: they revoke previously established events and render the causal history self-contradictory. A model must therefore predict not only a plausible next motion, but also which events have already happened, which states have been consumed or changed, and which causal branches have terminated. We refer to this requirement as \emph{irreversible state commitment}: generation must continue to advance while respecting established history, rather than resampling a locally plausible world that conflicts with its own past. Causal video generation is consequently not reducible to physical realism alone; it requires scene consistency, correct event occurrence, valid causal pathways, coherent temporal ordering, and stable termination to be maintained together as one continuous and verifiable causal process.  

\clearpage
\subsection*{Qualitative Evidence Atlas}
\label{app:qualitative_atlas}
The following atlas presents all 90 model-level cases used in the preceding analyses. Each page places the ground-truth rollout above three model results and preserves the original frame annotations and case-level descriptions.\par\vspace{3pt}
\setlength{\parindent}{0pt}
\raggedbottom
\newcommand{\qstrip}[1]{%
  \noindent\includegraphics[width=\linewidth]{#1}\par%
}
\noindent
{\fontsize{10}{11.5}\selectfont\bfseries Level 1 · State-only physical prompt}\par
\vspace{0.5pt}
{\fontsize{8}{9}\selectfont\itshape Domino\_Series.ball\_trigger\_step\_domino\_chain}\par
\vspace{1pt}\hrule\vspace{3pt}
{\fontsize{8}{9}\selectfont\bfseries Ground Truth}\par
\vspace{1pt}
\qstrip{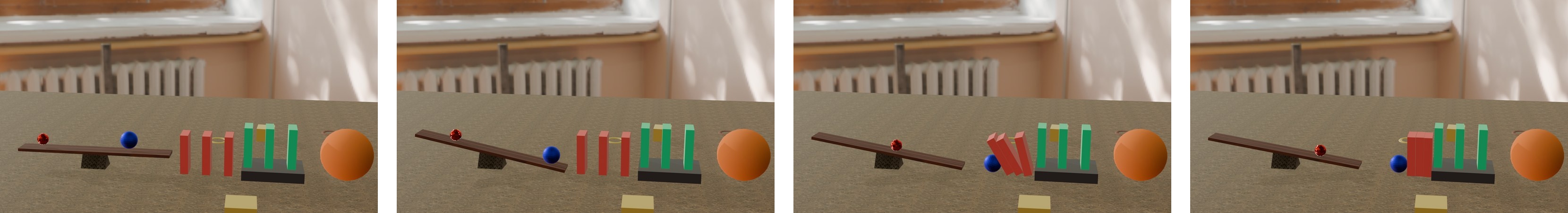}
\vspace{4pt}
\refstepcounter{qualcase}\label{case:b01-m01}
{\fontsize{8}{9}\selectfont\bfseries Case~\thequalcase\enspace\textperiodcentered\enspace CogVideoX-Fun-V1.5}\par
\vspace{1pt}
\qstrip{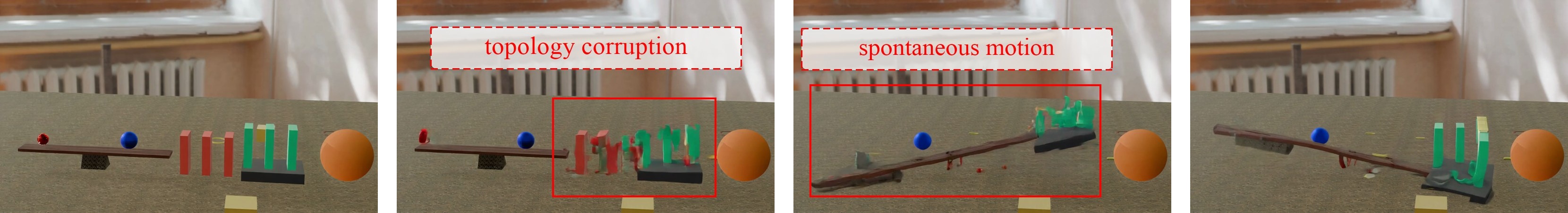}
\vspace{2pt}
{\fontsize{7.25}{8.5}\selectfont CogVideoX-Fun-V1.5. CogVideoX exhibits an early breakdown of entity and scene integrity before the intended causal cascade can be faithfully realized. The red and green structures rapidly undergo severe topology corruption, with their shapes and spatial organization becoming distorted rather than remaining persistent rigid objects. This corruption is followed by large unsupported motions of multiple scene components that are not preceded by the required physical interactions, constituting spontaneous dynamics. From the causal perspective, the failure first disrupts the persistent entity substrate on which the reference event graph is grounded, and subsequently prevents the intended ball-to-domino interaction and downstream propagation from being instantiated along the correct pathway. Instead, the corrupted scene introduces spurious events and unsupported branches of motion. These observations are most consistent with a failure to maintain stable object-centric and world-state representations, making it difficult to isolate a specific physical-law error once the underlying scene representation itself has already degraded.\par}

\vspace{3pt}
\hrule
\vspace{3pt}
\refstepcounter{qualcase}\label{case:b01-m02}
{\fontsize{8}{9}\selectfont\bfseries Case~\thequalcase\enspace\textperiodcentered\enspace Cosmos3-Nano}\par
\vspace{1pt}
\qstrip{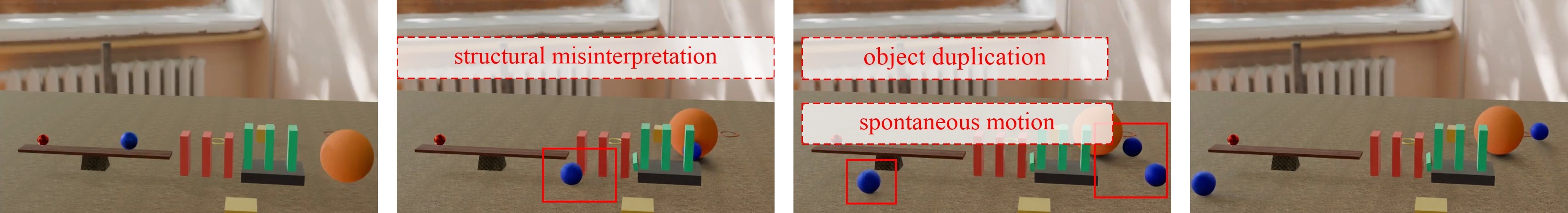}
\vspace{2pt}
{\fontsize{7.25}{8.5}\selectfont Cosmos3-Nano. Cosmos3-Nano largely preserves the visual appearance of the scene but exhibits errors in how the physical configuration is interpreted and maintained over time. The blue ball departs from its supported position and reaches the domino region without clearly following the mechanically required plank-mediated trajectory, suggesting that the support and contact structure of the scene may have been misinterpreted. The model subsequently duplicates the blue ball, after which multiple instances move independently through the scene, producing additional spontaneous dynamics. At the causal-graph level, the initially incorrect displacement bypasses part of the intended mechanical pathway, while object duplication breaks entity persistence and creates the possibility of spurious motion and collision events that are absent from the reference cascade. Thus, the observed behavior is consistent with a combination of structural or support-relation misinterpretation and unstable entity persistence.\par}

\vspace{3pt}
\hrule
\vspace{3pt}
\refstepcounter{qualcase}\label{case:b01-m03}
{\fontsize{8}{9}\selectfont\bfseries Case~\thequalcase\enspace\textperiodcentered\enspace HunyuanVideo-I2V}\par
\vspace{1pt}
\qstrip{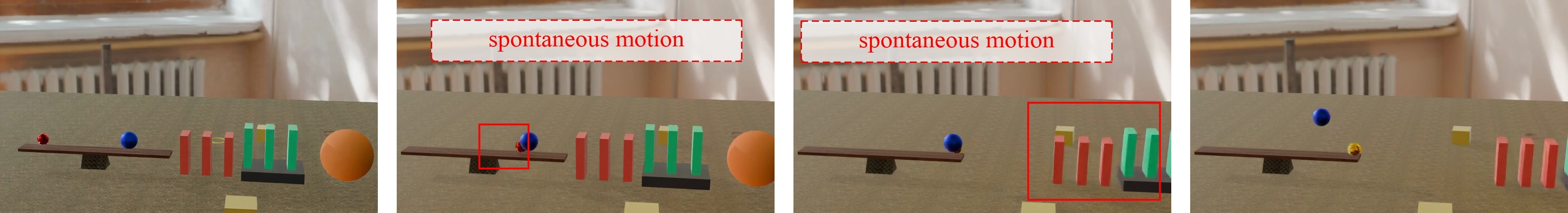}
\vspace{2pt}
{\fontsize{7.25}{8.5}\selectfont HunyuanVideo-I2V. HunyuanVideo primarily fails through unsupported dynamics. The red ball begins to move without a clearly established upstream mechanical trigger, and the domino-like structures subsequently change state without the intended contact-mediated interaction being faithfully realized. Although some downstream motion resembles events expected in the reference rollout, these events are not connected by the required physical dependency. In causal terms, this constitutes a causal shortcut: a downstream event is partially realized while the mediator that should physically produce it is missing or incorrectly instantiated. Additional abnormal object motion later in the rollout further introduces spurious events. This pattern is therefore more consistent with a failure to preserve causal dependencies between successive physical events than with a simple inability to generate the expected event types themselves.\par}

\clearpage
\noindent
{\fontsize{10}{11.5}\selectfont\bfseries Level 1 · State-only physical prompt}\par
\vspace{0.5pt}
{\fontsize{8}{9}\selectfont\itshape Domino\_Series.ball\_trigger\_step\_domino\_chain}\par
\vspace{1pt}\hrule\vspace{3pt}
{\fontsize{8}{9}\selectfont\bfseries Ground Truth}\par
\vspace{1pt}
\qstrip{appendix_e/assets/b01_m01_cogvideox_5b_i2v_gt_strip.jpg}
\vspace{4pt}
\refstepcounter{qualcase}\label{case:b01-m04}
{\fontsize{8}{9}\selectfont\bfseries Case~\thequalcase\enspace\textperiodcentered\enspace LTX-Video-13B}\par
\vspace{1pt}
\qstrip{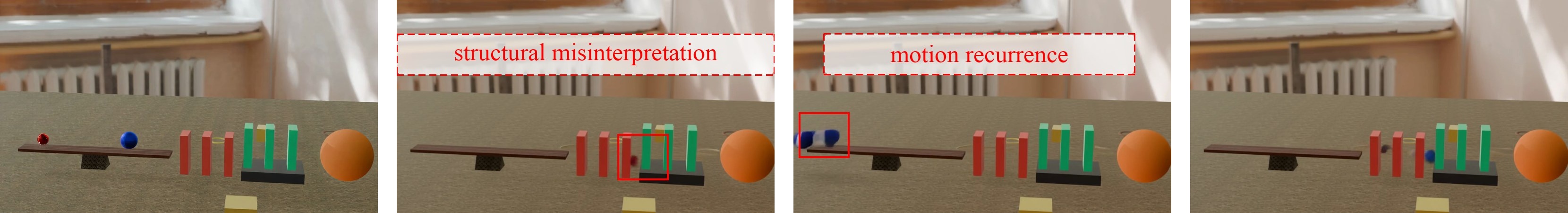}
\vspace{2pt}
{\fontsize{7.25}{8.5}\selectfont LTX-Video-13B. LTX-Video initially generates rightward motion of the blue ball, with the apparent duplicate-like appearance in the intermediate frame arising from motion blur or a temporal ghost rather than true object duplication. The more substantive failure occurs later, when the ball returns toward a previously occupied region after its initial rightward motion, producing a clear motion recurrence and partial replay of an earlier scene state. Unlike the other models, the primary error is therefore not entity corruption or an obviously incorrect local physical response, but a breakdown of causal-temporal progression: the generated world fails to advance monotonically toward the terminal state and instead revisits an earlier phase of the rollout. This behavior is most consistent with a temporal-state persistence failure, in which the model does not reliably preserve the fact that an earlier motion transition has already been completed.\par}

\vspace{3pt}
\hrule
\vspace{3pt}
\refstepcounter{qualcase}\label{case:b01-m05}
{\fontsize{8}{9}\selectfont\bfseries Case~\thequalcase\enspace\textperiodcentered\enspace MAGI-1-4.5B}\par
\vspace{1pt}
\qstrip{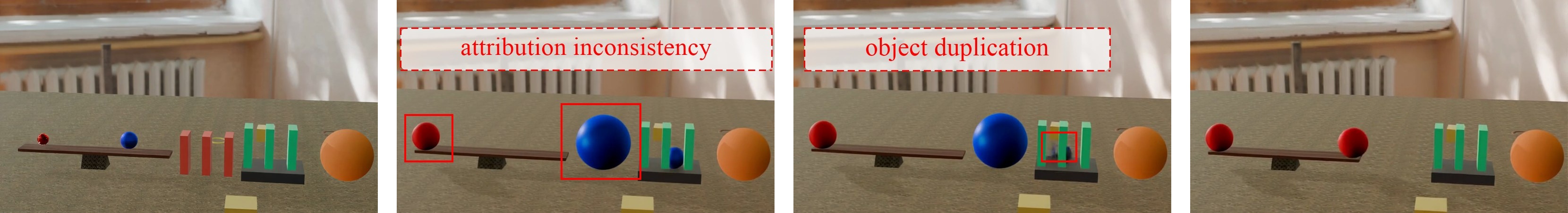}
\vspace{2pt}
{\fontsize{7.25}{8.5}\selectfont MAGI-1-4.5B. MAGI-1-4.5B exhibits a combination of entity-state corruption and incorrect mechanical realization. Early in the rollout, both the red and blue balls undergo substantial and unsupported changes in size, violating entity and visual integrity. Although the rightward motion of the blue ball is qualitatively consistent with the reference rollout, the seesaw itself remains approximately balanced rather than rotating downward on the right as the ball moves. From the causal perspective, this breaks the intended plank-mediated pathway, producing a causal shortcut in the early cascade. The rollout then further degrades through object duplication, with an additional blue-ball instance appearing near the downstream structures, thereby violating entity persistence and introducing a spurious participant that could support events absent from the reference graph. Taken together, the result potentially reflects a failure in torque-sensitive physical dynamics or in grounding the relevant mass and lever configuration to the visible scene.\par}

\vspace{3pt}
\hrule
\vspace{3pt}
\refstepcounter{qualcase}\label{case:b01-m06}
{\fontsize{8}{9}\selectfont\bfseries Case~\thequalcase\enspace\textperiodcentered\enspace SkyReels-V2-I2V}\par
\vspace{1pt}
\qstrip{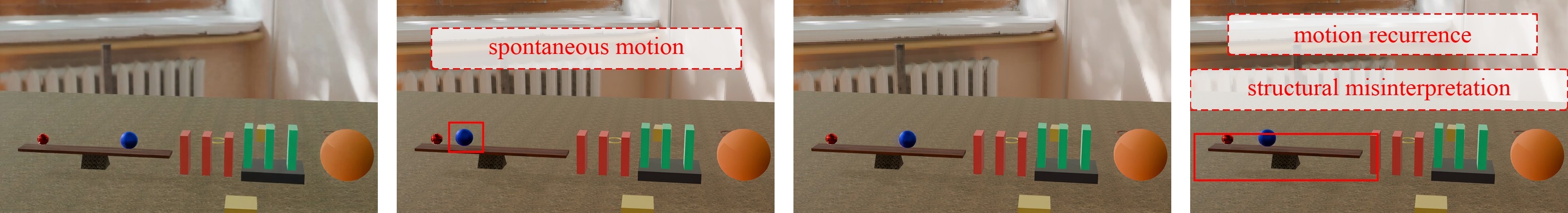}
\vspace{2pt}
{\fontsize{7.25}{8.5}\selectfont SkyReels-V2-I2V. SkyReels-V2-I2V primarily fails through unsupported and recurrent object motion. Early in the rollout, the blue ball moves spontaneously toward the left even though no corresponding trigger or mechanical change in the scene supports this displacement. From the causal perspective, this introduces a spurious motion event, thereby deviating from the reference cascade at its initiation. The ball subsequently reverses direction and moves back toward its initial position, after which the leftward motion is generated again. This repeated return to an earlier motion pattern constitutes motion recurrence and partial state replay, indicating that the generated physical process does not advance monotonically toward the downstream interaction and terminal state. Because the leftward displacement is produced without an evident change in support or lever configuration, the behavior may also reflect an incorrect interpretation of the scene’s mechanical structure; however, the repeated reversal and reappearance of the same motion pattern are more directly consistent with a temporal-state persistence failure, in which the model does not reliably preserve the fact that an earlier motion phase has already been completed.\par}

\clearpage
\noindent
{\fontsize{10}{11.5}\selectfont\bfseries Level 1 · State-only physical prompt}\par
\vspace{0.5pt}
{\fontsize{8}{9}\selectfont\itshape Domino\_Series.ball\_trigger\_step\_domino\_chain}\par
\vspace{1pt}\hrule\vspace{3pt}
{\fontsize{8}{9}\selectfont\bfseries Ground Truth}\par
\vspace{1pt}
\qstrip{appendix_e/assets/b01_m01_cogvideox_5b_i2v_gt_strip.jpg}
\vspace{4pt}
\refstepcounter{qualcase}\label{case:b01-m07}
{\fontsize{8}{9}\selectfont\bfseries Case~\thequalcase\enspace\textperiodcentered\enspace Wan2.1-I2V-14B}\par
\vspace{1pt}
\qstrip{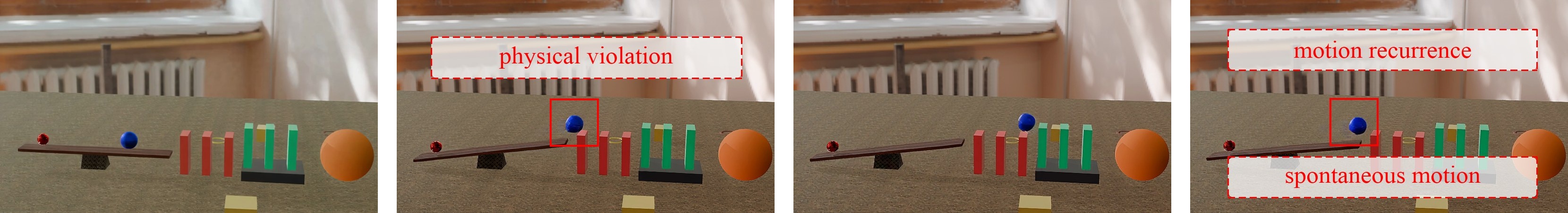}
\vspace{2pt}
{\fontsize{7.25}{8.5}\selectfont Wan2.1-I2V-14B. Wan2.1-I2V-14B initially produces the expected rightward motion of the blue ball, but realizes the accompanying seesaw dynamics incorrectly. Instead of rotating downward, the plank responds in the opposite manner and effectively lifts the blue ball, yielding a clear physical-law inconsistency in the lever response. From the causal perspective, the blue-ball motion event is therefore present, but the corresponding plank-mediated dynamics that should physically enable and guide this motion are incorrectly instantiated, breaking the intended early causal pathway before the ball reaches the domino sequence. Later in the rollout, after the blue ball has progressed toward the downstream red-domino region, it returns toward the earlier position near the right end of the seesaw, reproducing a previously completed portion of its trajectory. This constitutes motion recurrence and a partial temporal replay, introducing unsupported motion and preventing monotonic progression toward the reference terminal state. Overall, the sequence is consistent with an incorrect realization of torque-sensitive physical dynamics, followed by a temporal-state persistence failure that causes an already completed motion phase to be generated again.\par}

\vspace{3pt}
\hrule
\vspace{3pt}
\refstepcounter{qualcase}\label{case:b01-m08}
{\fontsize{8}{9}\selectfont\bfseries Case~\thequalcase\enspace\textperiodcentered\enspace Wan2.2-TI2V-5B}\par
\vspace{1pt}
\qstrip{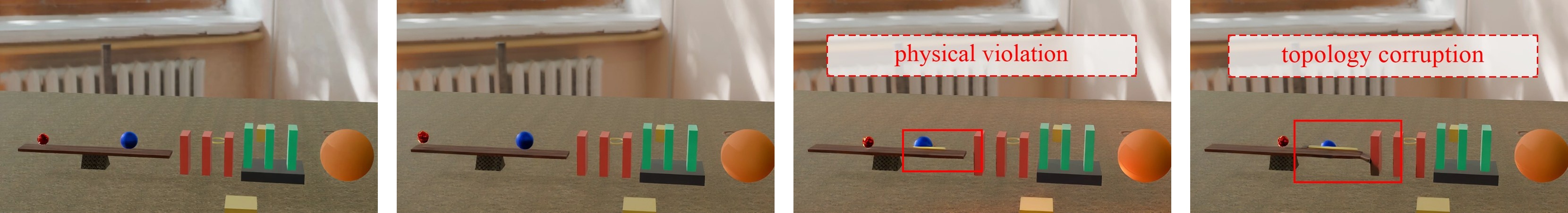}
\vspace{2pt}
{\fontsize{7.25}{8.5}\selectfont Wan2.2-TI2V-5B. Wan2.2-TI2V-5B exhibits both unsupported object motion and a breakdown of rigid-body integrity. Early in the rollout, the red ball spontaneously moves rightward despite the absence of a corresponding trigger in the reference process, introducing a spurious motion event before the intended cascade is realized. The blue ball also follows a mechanically inconsistent trajectory, moving away from the expected rightward path toward the domino sequence and instead deviating behind the seesaw in a manner incompatible with the visible support geometry and reference dynamics. From the causal perspective, these motions corrupt the intended early event sequence: the red-ball displacement introduces an unsupported branch, while the blue ball fails to realize the plank-mediated trajectory required to produce the subsequent ball–domino interaction, thereby breaking propagation toward the downstream domino cascade. The failure is compounded by topology corruption of the seesaw itself, whose shape becomes visibly distorted rather than remaining a persistent rigid lever.\par}

\vspace{3pt}
\hrule
\vspace{3pt}
\refstepcounter{qualcase}\label{case:b01-m09}
{\fontsize{8}{9}\selectfont\bfseries Case~\thequalcase\enspace\textperiodcentered\enspace Wan2.2-I2V-A14B}\par
\vspace{1pt}
\qstrip{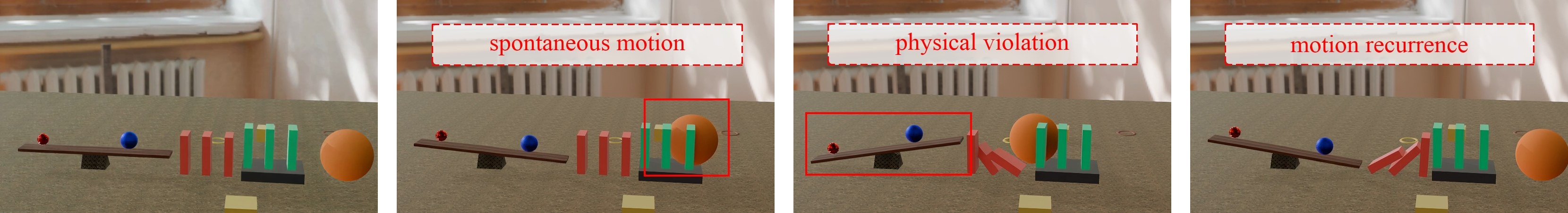}
\vspace{2pt}
{\fontsize{7.25}{8.5}\selectfont Wan2.2-I2V-A14B. Wan2.2-I2V-A14B initially captures the intended early dynamics reasonably well: the heavier blue ball moves rightward as the seesaw begins to tip in the corresponding direction. However, the rollout soon introduces an unsupported motion of the large orange ball on the right, which moves leftward despite no upstream event in the reference cascade that should trigger it. Although this motion subsequently produces a locally plausible interaction with the downstream structures, it constitutes a spurious event and creates an additional causal branch absent from the ground-truth process. The model then further departs from the reference dynamics as the seesaw reverses its rotation and later returns toward a previously generated orientation, producing a clear physical-law inconsistency together with motion recurrence. From the causal perspective, the rollout therefore combines a spurious branch with a breakdown of causal-temporal progression. These behaviors are consistent with both incorrect torque-sensitive physical dynamics and a temporal-state persistence failure, in which previously completed state transitions are not stably maintained.\par}

\clearpage
\noindent
{\fontsize{10}{11.5}\selectfont\bfseries Level 1 · State-only physical prompt}\par
\vspace{0.5pt}
{\fontsize{8}{9}\selectfont\itshape Domino\_Series.dual\_ball\_inverted\_y\_domino\_course}\par
\vspace{1pt}\hrule\vspace{3pt}
{\fontsize{8}{9}\selectfont\bfseries Ground Truth}\par
\vspace{1pt}
\qstrip{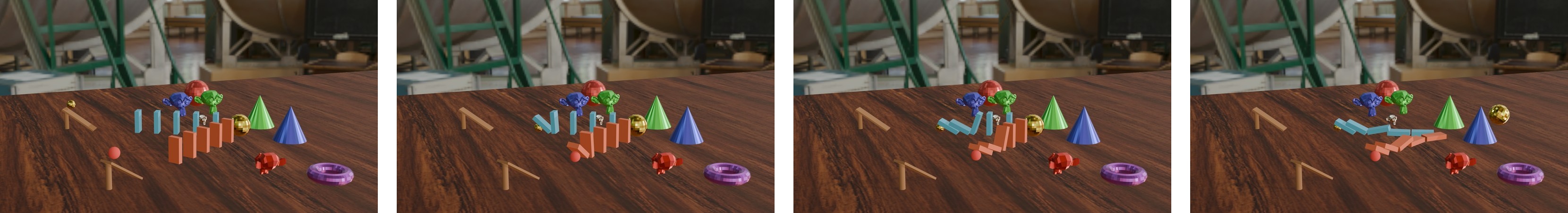}
\vspace{4pt}
\refstepcounter{qualcase}\label{case:b02-m01}
{\fontsize{8}{9}\selectfont\bfseries Case~\thequalcase\enspace\textperiodcentered\enspace CogVideoX-Fun-V1.5}\par
\vspace{1pt}
\qstrip{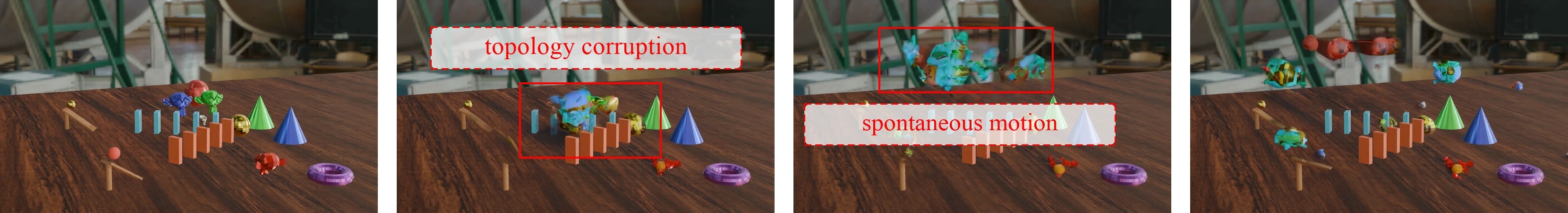}
\vspace{2pt}
{\fontsize{7.25}{8.5}\selectfont CogVideoX-Fun-V1.5. CogVideoX-Fun-V1.5 exhibits an immediate breakdown of entity integrity in the central interaction region, where several objects undergo severe topology corruption before the intended cascade can be realized. The corrupted structures subsequently move upward spontaneously and then disperse in multiple unsupported directions, producing dynamics that are incompatible with both the visible scene configuration and basic physical constraints. From the causal perspective, this failure first destroys the persistent object substrate required to ground the reference events, making the intended ball-to-domino and domino-to-domino interactions no longer well defined. The resulting upward and scattered motions then introduce multiple spurious events and causal branches while preventing the reference propagation toward the downstream gold ball from being faithfully realized. Overall, the sequence is most consistent with unstable persistent object/world-state representations, compounded by incorrect physical-dynamics realization once the scene structure has already become corrupted.\par}

\vspace{3pt}
\hrule
\vspace{3pt}
\refstepcounter{qualcase}\label{case:b02-m02}
{\fontsize{8}{9}\selectfont\bfseries Case~\thequalcase\enspace\textperiodcentered\enspace Cosmos3-Nano}\par
\vspace{1pt}
\qstrip{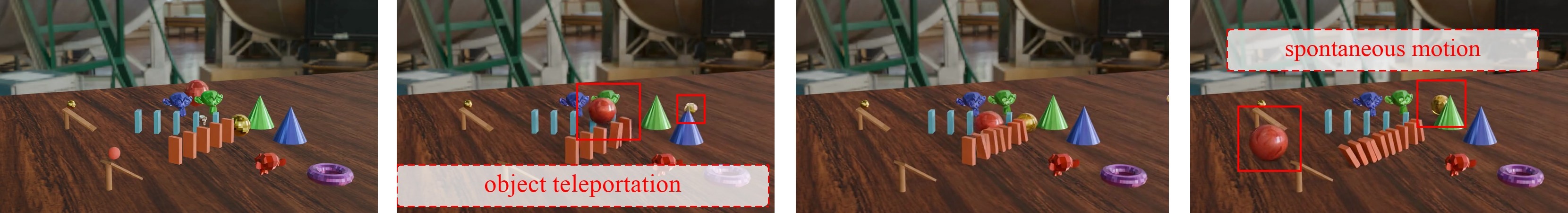}
\vspace{2pt}
{\fontsize{7.25}{8.5}\selectfont Cosmos3-Nano. Cosmos3-Nano primarily fails through severe spatiotemporal discontinuity and unsupported object motion. Early in the rollout, one of the central balls abruptly appears in a new position above the domino region, while the gold ball simultaneously jumps from its original position on the left side of the two cones to the opposite side, producing clear object-teleportation failures without continuous intermediate trajectories. Later, both balls move spontaneously away from one another and disturb the domino arrangement without the required upstream ramp-to-domino interactions. From the causal perspective, these discontinuous relocations create spurious motion events and bypass the intended sequence of gravity-driven ramp motion, ball–domino contact, and domino-mediated propagation. The later spontaneous motions further introduce spurious branches that perturb the domino cascade. Overall, the behavior is most consistent with failures in spatiotemporal state persistence and causal grounding, whereby object locations and their required physical dependencies are not stably maintained across the rollout.\par}

\vspace{3pt}
\hrule
\vspace{3pt}
\refstepcounter{qualcase}\label{case:b02-m03}
{\fontsize{8}{9}\selectfont\bfseries Case~\thequalcase\enspace\textperiodcentered\enspace HunyuanVideo-I2V}\par
\vspace{1pt}
\qstrip{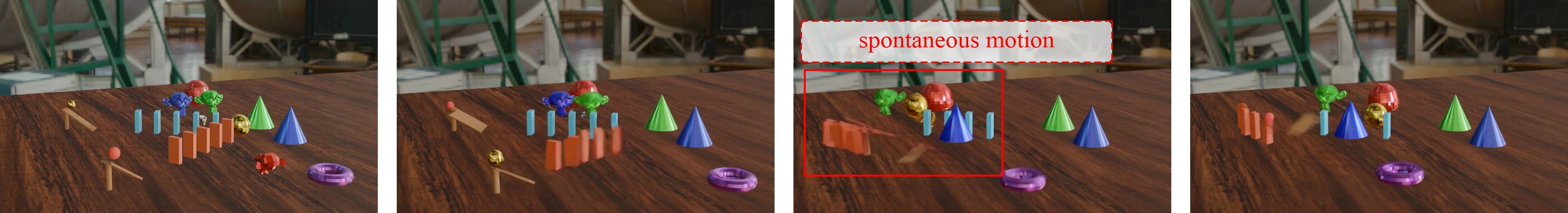}
\vspace{2pt}
{\fontsize{7.25}{8.5}\selectfont HunyuanVideo-I2V. HunyuanVideo-I2V primarily fails through unsupported downstream motion. In the generated rollout, the lower row of dominoes begins to move spontaneously even though the required upstream ramp-ball interaction has not yet propagated to this part of the scene. From the causal perspective, this introduces a spurious event within the domino cascade and effectively bypasses the physical mediator that should trigger the lower-row motion. The model therefore partially produces a causal shortcut rather than through the intended sequence of ball–domino contact and domino-to-domino propagation. This behavior is most consistent with a causal-dependency reasoning failure, in which the model captures that the dominoes are expected to move but does not preserve the specific upstream interaction required to cause that motion.\par}

\clearpage
\noindent
{\fontsize{10}{11.5}\selectfont\bfseries Level 1 · State-only physical prompt}\par
\vspace{0.5pt}
{\fontsize{8}{9}\selectfont\itshape Domino\_Series.dual\_ball\_inverted\_y\_domino\_course}\par
\vspace{1pt}\hrule\vspace{3pt}
{\fontsize{8}{9}\selectfont\bfseries Ground Truth}\par
\vspace{1pt}
\qstrip{appendix_e/assets/b02_m01_cogvideox_5b_i2v_gt_strip.jpg}
\vspace{4pt}
\refstepcounter{qualcase}\label{case:b02-m04}
{\fontsize{8}{9}\selectfont\bfseries Case~\thequalcase\enspace\textperiodcentered\enspace LTX-Video-13B}\par
\vspace{1pt}
\qstrip{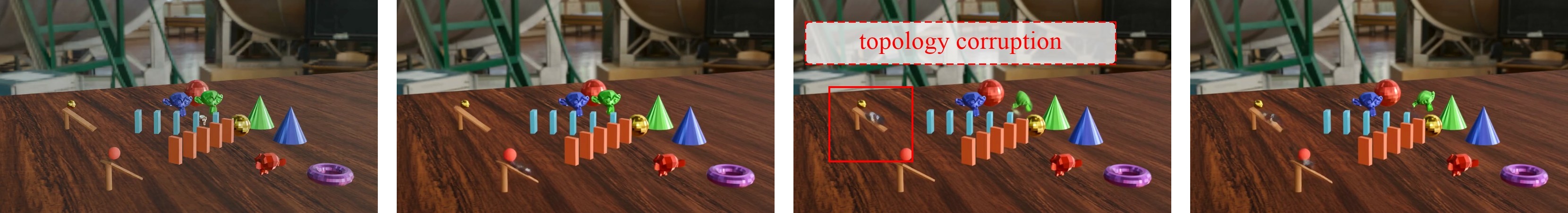}
\vspace{2pt}
{\fontsize{7.25}{8.5}\selectfont LTX-Video-13B. LTX-Video-13B primarily fails through a loss of entity and structural integrity in the left-side ramp region. As the rollout progresses, the ramp-like support and the nearby ball become visibly distorted and deformed rather than preserving their original rigid geometry and recognizable object identities. From the causal perspective, this topology corruption damages the persistent physical substrate required to realize the reference trigger events, because the gravity-driven descent of the ball can no longer be grounded to a stable ramp–ball configuration. As a result, the subsequent ball-to-domino interaction and downstream domino propagation become unreliable even before the main causal cascade is fully initiated. This behavior is most consistent with unstable persistent object and scene-state representations, rather than with a specific isolated error in the downstream causal dependencies themselves.\par}

\vspace{3pt}
\hrule
\vspace{3pt}
\refstepcounter{qualcase}\label{case:b02-m05}
{\fontsize{8}{9}\selectfont\bfseries Case~\thequalcase\enspace\textperiodcentered\enspace MAGI-1-4.5B}\par
\vspace{1pt}
\qstrip{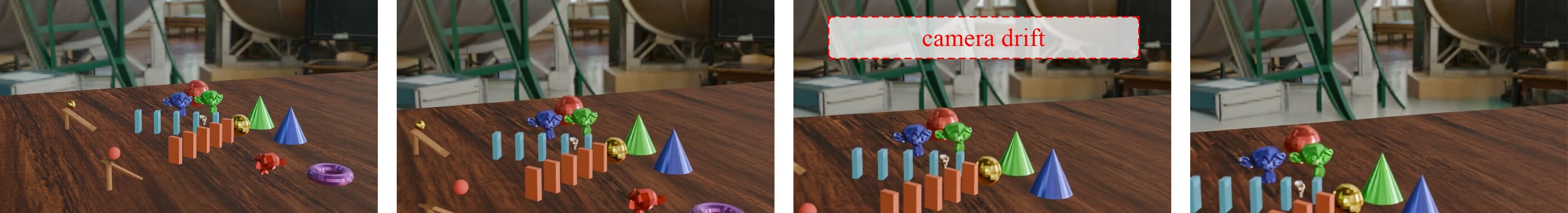}
\vspace{2pt}
{\fontsize{7.25}{8.5}\selectfont MAGI-1-4.5B. MAGI-1-4.5B primarily fails through viewpoint instability together with incomplete realization of the gravity-driven trigger. Across the rollout, the camera progressively drifts and moves closer to the central scene, violating the fixed-view condition and altering the apparent spatial relationships among the interacting objects. At the same time, the left-side ball does not clearly follow the expected downhill trajectory along the ramp, indicating an incorrect or incomplete physical realization of the initial gravity-driven motion. From the causal perspective, this weakens the very first stage of the reference cascade: the trigger event is not faithfully realized, and the subsequent ball–domino contact and downstream domino propagation therefore cannot be grounded in the intended physical pathway. The camera drift further complicates the persistence of the scene representation by changing the observation geometry during the rollout. Overall, the sequence is most consistent with a combination of conditioning or viewpoint-grounding failure and inaccurate realization of the initial gravity-driven dynamics.\par}

\vspace{3pt}
\hrule
\vspace{3pt}
\refstepcounter{qualcase}\label{case:b02-m06}
{\fontsize{8}{9}\selectfont\bfseries Case~\thequalcase\enspace\textperiodcentered\enspace SkyReels-V2-I2V}\par
\vspace{1pt}
\qstrip{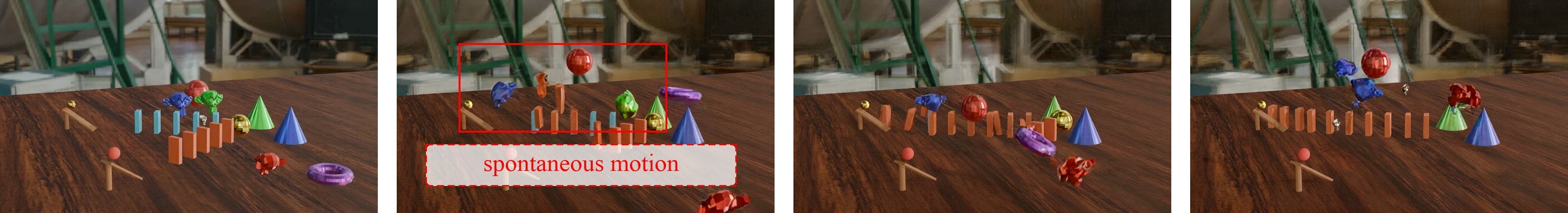}
\vspace{2pt}
{\fontsize{7.25}{8.5}\selectfont SkyReels-V2-I2V. SkyReels-V2-I2V exhibits severe unsupported dynamics throughout the rollout. Several central objects spontaneously rise into the air without any preceding contact, support loss, or other physical trigger, producing a clear combination of spontaneous motion and physical-law inconsistency. The dominoes then undergo additional unsupported horizontal displacement rather than being activated through the intended sequence of ramp-ball impacts and domino-to-domino propagation. From the causal perspective, these motions introduce multiple spurious events and bypass the required upstream mediators, replacing the reference cascade with unsupported causal branches that are not grounded in the initial conditions or prompt. As a result, the generated sequence no longer preserves the intended progression from gravity-driven ball motion to domino propagation and downstream activation of the gold ball. Overall, the behavior is most consistent with a failure to preserve causal dependencies together with inaccurate physical-dynamics realization, causing objects to move according to scene-level event associations rather than the required contact-mediated mechanism.\par}

\clearpage
\noindent
{\fontsize{10}{11.5}\selectfont\bfseries Level 1 · State-only physical prompt}\par
\vspace{0.5pt}
{\fontsize{8}{9}\selectfont\itshape Domino\_Series.dual\_ball\_inverted\_y\_domino\_course}\par
\vspace{1pt}\hrule\vspace{3pt}
{\fontsize{8}{9}\selectfont\bfseries Ground Truth}\par
\vspace{1pt}
\qstrip{appendix_e/assets/b02_m01_cogvideox_5b_i2v_gt_strip.jpg}
\vspace{4pt}
\refstepcounter{qualcase}\label{case:b02-m07}
{\fontsize{8}{9}\selectfont\bfseries Case~\thequalcase\enspace\textperiodcentered\enspace Wan2.1-I2V-14B}\par
\vspace{1pt}
\qstrip{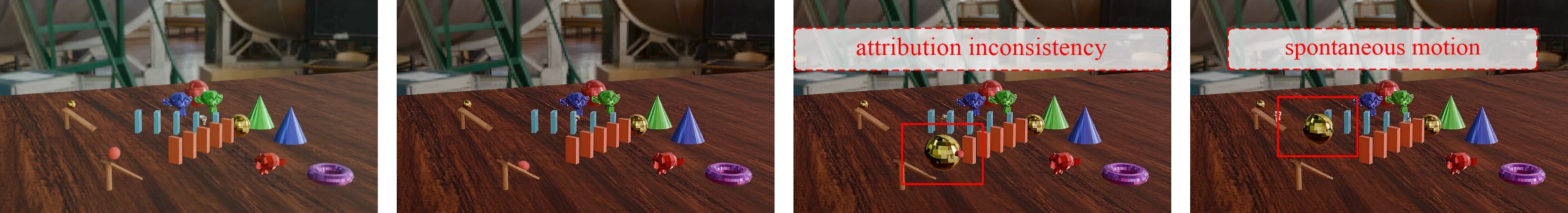}
\vspace{2pt}
{\fontsize{7.25}{8.5}\selectfont Wan2.1-I2V-14B. Wan2.1-I2V-14B partially captures the intended trigger by recognizing that the two left-side balls should roll downhill along their ramps, but it fails to preserve the identity and subsequent trajectory of the moving object. During the descent, the small ramp ball is transformed into a much larger gold ball, producing a clear identity-corruption failure and breaking the persistent entity grounding required by the reference event graph. After reaching the lower end of the ramp, the ball also deviates leftward rather than continuing along the intended path toward the dominoes, introducing unsupported motion inconsistent with both the visible geometry and the prescribed causal process. From the causal perspective, the initial gravity-driven motion is only partially realized, while the subsequent ball–domino contact event is missing, thereby breaking propagation before the domino cascade can be correctly initiated. Overall, the sequence is consistent with unstable entity-state persistence combined with incorrect trajectory grounding, causing the model to lose both the identity and the mechanically required downstream role of the trigger object.\par}

\vspace{3pt}
\hrule
\vspace{3pt}
\refstepcounter{qualcase}\label{case:b02-m08}
{\fontsize{8}{9}\selectfont\bfseries Case~\thequalcase\enspace\textperiodcentered\enspace Wan2.2-TI2V-5B}\par
\vspace{1pt}
\qstrip{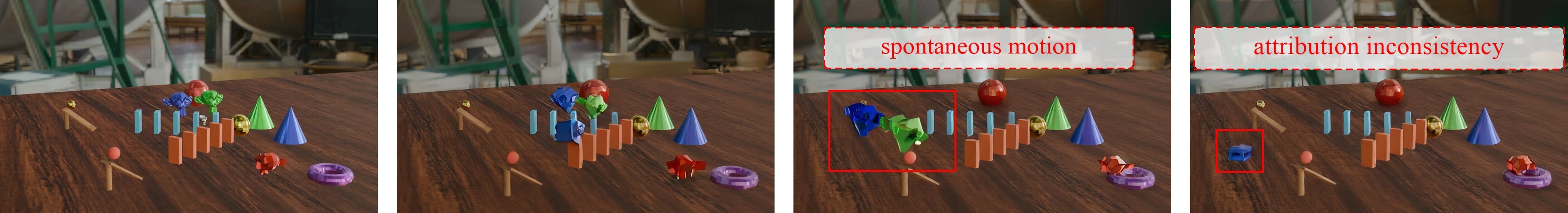}
\vspace{2pt}
{\fontsize{7.25}{8.5}\selectfont Wan2.2-TI2V-5B. Wan2.2-TI2V-5B primarily fails at the initiation of the reference causal cascade. The left-side trigger ball does not undergo the expected gravity-driven descent along the ramp, yielding a clear physical-law inconsistency and leaving the intended initiating event unrealized. From the causal perspective, this constitutes a missing trigger: because the initial ball motion is absent, the subsequent ball–domino contact and downstream domino propagation cannot be activated through the reference pathway. In parallel, several visually salient but causally irrelevant central objects move spontaneously toward the left while changing color, shape, and structural appearance, producing additional unsupported motion together with identity and topology corruption. These latter failures disrupt local scene integrity but are secondary to the more consequential omission of the trigger event. Overall, the rollout is most consistent with an incorrect realization of the initial gravity-driven dynamics, accompanied by unstable entity/world-state representations in unrelated parts of the scene.\par}

\vspace{3pt}
\hrule
\vspace{3pt}
\refstepcounter{qualcase}\label{case:b02-m09}
{\fontsize{8}{9}\selectfont\bfseries Case~\thequalcase\enspace\textperiodcentered\enspace Wan2.2-I2V-A14B}\par
\vspace{1pt}
\qstrip{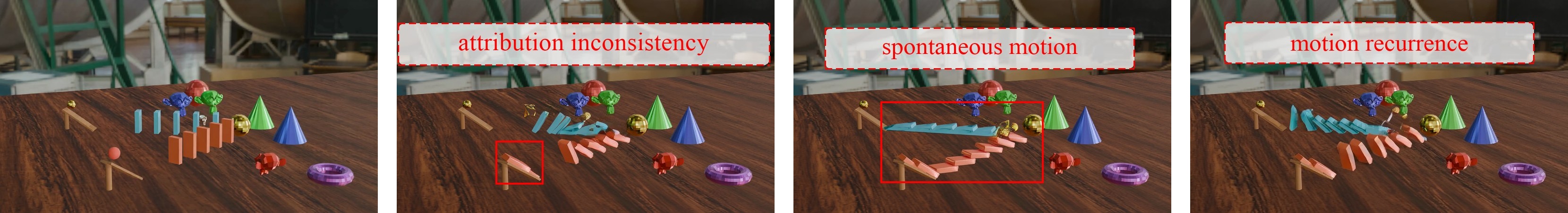}
\vspace{2pt}
{\fontsize{7.25}{8.5}\selectfont Wan2.2-I2V-A14B. Wan2.2-I2V-A14B exhibits both entity-identity corruption and a breakdown of causal-temporal progression. Early in the rollout, the small ball positioned at the top of the lower left ramp changes into a domino-like object, producing a clear identity inconsistency that disrupts the persistent representation of the intended trigger entity. This transformed object then remains on the ramp rather than undergoing the expected gravity-driven descent, leaving the reference trigger event unrealized. Despite the absence of this upstream trigger, the central domino chain begins to topple spontaneously, creating a spurious downstream event and a clear causal shortcut in which the model generates an expected part of the cascade without the required ball–domino interaction. The rollout subsequently regresses further when the fallen dominoes reverse their motion and return toward an upright configuration, constituting state replay and motion recurrence rather than forward progression toward the terminal state. Overall, the sequence is consistent with unstable entity-state persistence, failure to realize the initial gravity-driven dynamics, and weak preservation of causal dependencies and temporal history across the cascade.\par}

\clearpage
\noindent
{\fontsize{10}{11.5}\selectfont\bfseries Level 1 · State-only physical prompt}\par
\vspace{0.5pt}
{\fontsize{8}{9}\selectfont\itshape Galton\_Series.horizontal\_galton\_board\_object\_targets}\par
\vspace{1pt}\hrule\vspace{3pt}
{\fontsize{8}{9}\selectfont\bfseries Ground Truth}\par
\vspace{1pt}
\qstrip{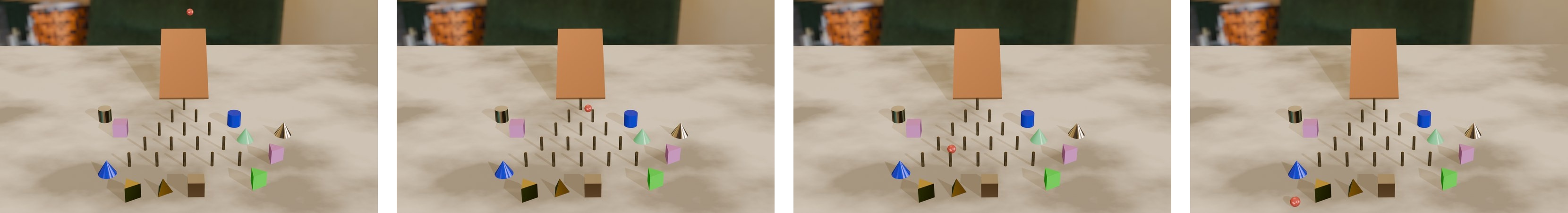}
\vspace{4pt}
\refstepcounter{qualcase}\label{case:b03-m01}
{\fontsize{8}{9}\selectfont\bfseries Case~\thequalcase\enspace\textperiodcentered\enspace CogVideoX-Fun-V1.5}\par
\vspace{1pt}
\qstrip{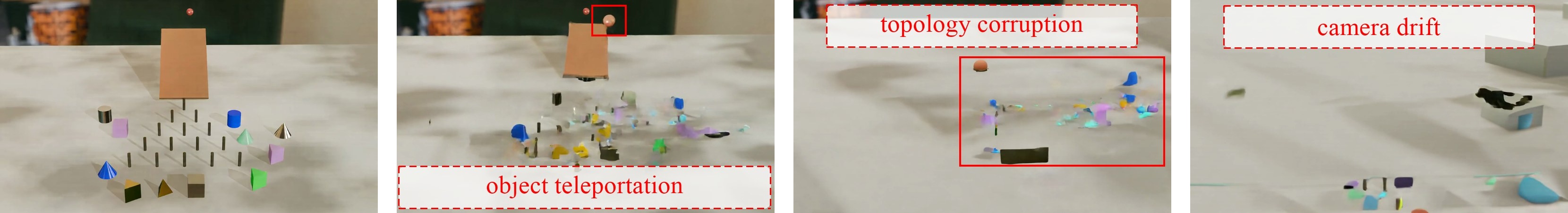}
\vspace{2pt}
{\fontsize{7.25}{8.5}\selectfont CogVideoX-Fun-V1.5. CogVideoX-Fun-V1.5 exhibits a severe breakdown of scene and entity integrity before the intended Galton-board trajectory can be faithfully realized. The lower board and surrounding structures rapidly become blurred and heavily deformed, producing clear topology and identity corruption and destroying the stable physical substrate required to track the ball–peg interactions over time. In addition, an extra ball abruptly appears in the upper region without a continuous preceding trajectory, constituting an object-teleportation failure and introducing a spurious event absent from the reference rollout. From the causal perspective, these errors prevent the generated sequence from preserving the intended contact-mediated chain of gravitational descent, successive peg collisions, and trajectory redirection toward the lower-left exit; instead, the corrupted scene and newly introduced object create unsupported states and potential spurious causal branches. The final frames further exhibit substantial camera drift, violating the fixed-view conditioning and making the spatial grounding of the causal process increasingly unstable.\par}

\vspace{3pt}
\hrule
\vspace{3pt}
\refstepcounter{qualcase}\label{case:b03-m02}
{\fontsize{8}{9}\selectfont\bfseries Case~\thequalcase\enspace\textperiodcentered\enspace Cosmos3-Nano}\par
\vspace{1pt}
\qstrip{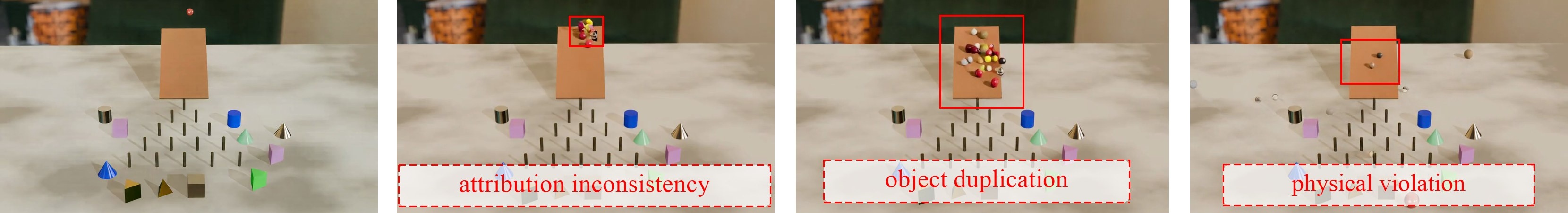}
\vspace{2pt}
{\fontsize{7.25}{8.5}\selectfont Cosmos3-Nano. Cosmos3-Nano exhibits a pronounced breakdown of entity identity and persistence during the initial descent. Rather than maintaining a single consistent red ball, the model progressively transforms and duplicates it into numerous spheres of different colors and sizes, producing both identity corruption and object duplication. Most of these generated instances nevertheless continue to move downward along approximately plausible gravity-driven trajectories through the Galton board, but their presence introduces multiple spurious participants and additional ball–peg interactions that are absent from the reference causal process. From the causal perspective, the intended single-particle trajectory is therefore replaced by several unsupported parallel branches, making the subsequent contact-mediated pathway fundamentally different from the Ground Truth even when individual motions appear locally plausible. In addition, two of the generated balls remain unnaturally suspended on the upper board rather than continuing their gravity-driven descent, yielding a clear physical-law inconsistency and incomplete realization of the expected motion.\par}

\vspace{3pt}
\hrule
\vspace{3pt}
\refstepcounter{qualcase}\label{case:b03-m03}
{\fontsize{8}{9}\selectfont\bfseries Case~\thequalcase\enspace\textperiodcentered\enspace HunyuanVideo-I2V}\par
\vspace{1pt}
\qstrip{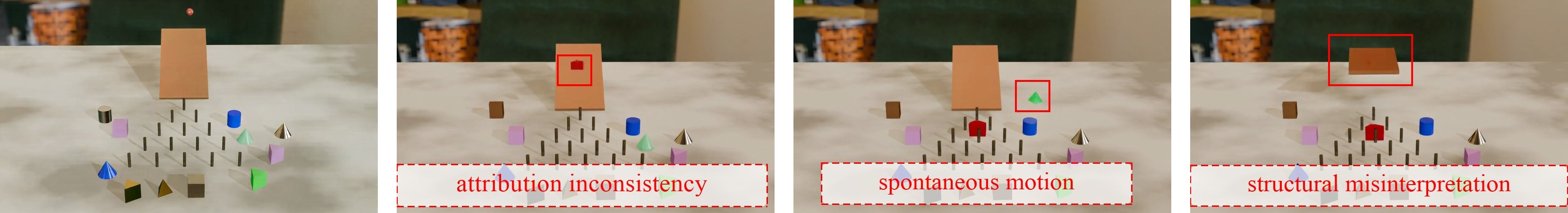}
\vspace{2pt}
{\fontsize{7.25}{8.5}\selectfont HunyuanVideo-I2V. HunyuanVideo-I2V initially captures the expected gravity-driven descent of the ball, but fails to preserve its identity during the subsequent motion: the falling object changes both shape and color, producing a clear identity-corruption failure and weakening the persistent grounding of the causal participant. The rollout then introduces an additional unsupported event on the right, where a conical object spontaneously moves upward without any preceding physical trigger and subsequently falls back toward the inclined board. This spurious motion produces an interaction absent from the reference Galton-board process and causes the board itself to tilt, further diverting the scene from the intended contact-mediated ball trajectory. From the causal perspective, the initial trigger is only partially realized, while the later cone motion creates a spurious causal branch unrelated to the reference sequence of ball–peg collisions and trajectory redirections. Overall, the behavior is consistent with unstable entity-state persistence together with a possible structural misinterpretation, in which the model incorrectly treats a causally irrelevant object and the inclined board as components of an active mechanical interaction.\par}

\clearpage
\noindent
{\fontsize{10}{11.5}\selectfont\bfseries Level 1 · State-only physical prompt}\par
\vspace{0.5pt}
{\fontsize{8}{9}\selectfont\itshape Galton\_Series.horizontal\_galton\_board\_object\_targets}\par
\vspace{1pt}\hrule\vspace{3pt}
{\fontsize{8}{9}\selectfont\bfseries Ground Truth}\par
\vspace{1pt}
\qstrip{appendix_e/assets/b03_m01_cogvideox_5b_i2v_gt_strip.jpg}
\vspace{4pt}
\refstepcounter{qualcase}\label{case:b03-m04}
{\fontsize{8}{9}\selectfont\bfseries Case~\thequalcase\enspace\textperiodcentered\enspace LTX-Video-13B}\par
\vspace{1pt}
\qstrip{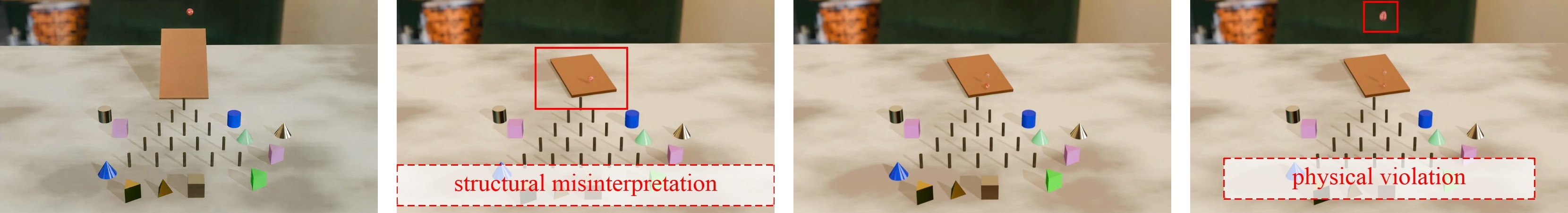}
\vspace{2pt}
{\fontsize{7.25}{8.5}\selectfont LTX-Video-13B. LTX-Video-13B first exhibits a structural misinterpretation of the upper inclined board, treating it as if it were unsupported and allowing it to collapse during the rollout even though the reference scene indicates that the structure should remain stable. This incorrect structural inference alters the physical configuration before the intended Galton-board trajectory is fully realized and introduces an interaction that is absent from the reference process. The ball then follows an additional physically inconsistent motion: after descending under gravity and reaching the board, it reverses its vertical motion and rises back into the air without any plausible impulse or supporting interaction. From the causal perspective, the model therefore replaces the intended sequence of gravitational descent and successive ball–peg interactions with a spurious structural event and an unsupported reversal of the ball trajectory, preventing faithful propagation through the reference pathway. The later upward return also constitutes motion recurrence and partial state replay, indicating a breakdown of causal-temporal progression. Overall, the sequence is consistent with an initial structural-inference failure, followed by incorrect physical-dynamics realization and weak temporal-state persistence.\par}

\vspace{3pt}
\hrule
\vspace{3pt}
\refstepcounter{qualcase}\label{case:b03-m05}
{\fontsize{8}{9}\selectfont\bfseries Case~\thequalcase\enspace\textperiodcentered\enspace MAGI-1-4.5B}\par
\vspace{1pt}
\qstrip{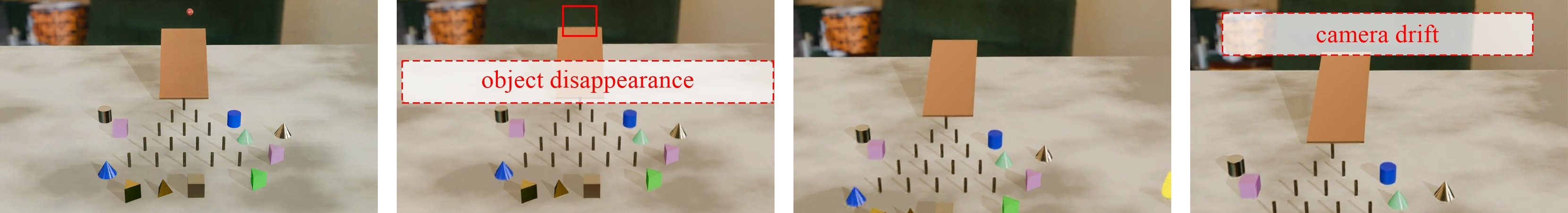}
\vspace{2pt}
{\fontsize{7.25}{8.5}\selectfont MAGI-1-4.5B. MAGI-1-4.5B primarily fails through object disappearance and viewpoint instability. During the rollout, the red ball that should descend from the upper release region vanishes before completing its gravity-driven trajectory through the Galton board, producing a clear entity-persistence failure and leaving the intended trigger and subsequent ball–peg interactions unrealized. From the causal perspective, the disappearance removes the primary participant from the reference process, thereby breaking the cascade at its initiation and preventing the expected sequence of contact-mediated trajectory redirections toward the lower-left exit. In addition, the camera progressively shifts from left to right across the scene, violating the fixed-view condition and altering the apparent spatial relationships among the board, pegs, and surrounding objects. This viewpoint drift further weakens the stable grounding of the generated physical process. Overall, the sequence is most consistent with unstable entity persistence together with a conditioning or viewpoint-grounding failure, rather than with an isolated error in the downstream collision dynamics.\par}

\vspace{3pt}
\hrule
\vspace{3pt}
\refstepcounter{qualcase}\label{case:b03-m06}
{\fontsize{8}{9}\selectfont\bfseries Case~\thequalcase\enspace\textperiodcentered\enspace SkyReels-V2-I2V}\par
\vspace{1pt}
\qstrip{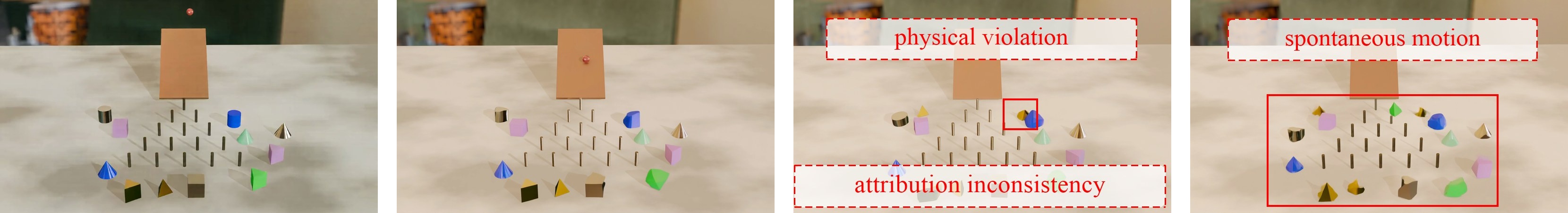}
\vspace{2pt}
{\fontsize{7.25}{8.5}\selectfont SkyReels-V2-I2V. SkyReels-V2-I2V first exhibits an incorrect realization of the ball–peg interaction. After the ball descends into the Galton board, the reference collision should redirect it toward the left, whereas the generated ball instead moves toward the right, producing a clear physical-law inconsistency in the contact-mediated trajectory. During this deviation, the ball also changes in color and shape, resulting in an identity-corruption failure that weakens the persistent grounding of the main causal participant. From the causal perspective, the gravity-driven trigger is realized, but the subsequent peg collision produces the wrong trajectory redirection, causing the rollout to leave the reference pathway at an intermediate mediator event. After the ball reaches and stalls near the blue object, several surrounding objects begin moving spontaneously despite having no corresponding upstream trigger, introducing additional spurious events and causal branches. Overall, the sequence is most consistent with incorrect collision-sensitive physical dynamics together with unstable entity-state persistence and weak preservation of causal dependencies for otherwise inactive scene objects.\par}

\clearpage
\noindent
{\fontsize{10}{11.5}\selectfont\bfseries Level 1 · State-only physical prompt}\par
\vspace{0.5pt}
{\fontsize{8}{9}\selectfont\itshape Galton\_Series.horizontal\_galton\_board\_object\_targets}\par
\vspace{1pt}\hrule\vspace{3pt}
{\fontsize{8}{9}\selectfont\bfseries Ground Truth}\par
\vspace{1pt}
\qstrip{appendix_e/assets/b03_m01_cogvideox_5b_i2v_gt_strip.jpg}
\vspace{4pt}
\refstepcounter{qualcase}\label{case:b03-m07}
{\fontsize{8}{9}\selectfont\bfseries Case~\thequalcase\enspace\textperiodcentered\enspace Wan2.1-I2V-14B}\par
\vspace{1pt}
\qstrip{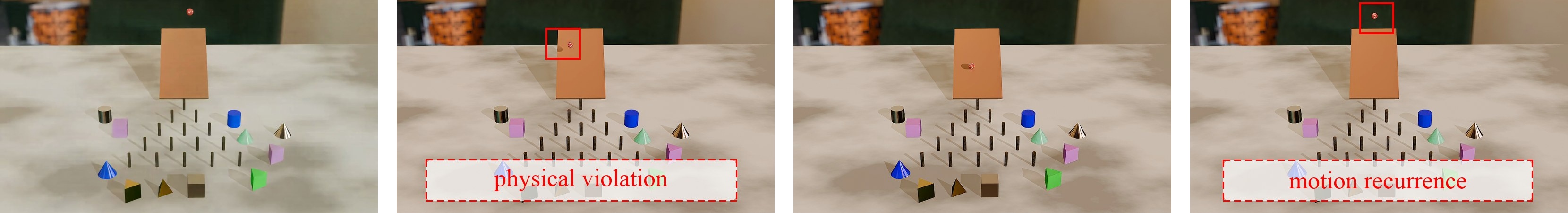}
\vspace{2pt}
{\fontsize{7.25}{8.5}\selectfont Wan2.1-I2V-14B. Wan2.1-I2V-14B first exhibits an incorrect realization of the ball’s gravity-driven descent. Instead of continuing approximately vertically downward into the Galton board as in the reference rollout, the ball deviates laterally toward the left, producing a physically inconsistent trajectory before the intended sequence of peg interactions can be faithfully established. From the causal perspective, the initial trigger is only partially realized: the ball begins to descend, but its incorrect trajectory changes the entry condition for the subsequent ball–peg contacts and therefore diverts the rollout from the reference interaction pathway at an early stage. The model then introduces a second failure when the ball reverses its motion and rises back toward its earlier position after descending, producing clear motion recurrence and partial state replay. This temporal regression prevents the physical process from progressing monotonically through the Galton board toward the lower-left exit. Overall, the sequence is most consistent with inaccurate gravity- and trajectory-sensitive physical dynamics, followed by a temporal-state persistence failure that causes a previously completed phase of motion to be generated again.\par}

\vspace{3pt}
\hrule
\vspace{3pt}
\refstepcounter{qualcase}\label{case:b03-m08}
{\fontsize{8}{9}\selectfont\bfseries Case~\thequalcase\enspace\textperiodcentered\enspace Wan2.2-TI2V-5B}\par
\vspace{1pt}
\qstrip{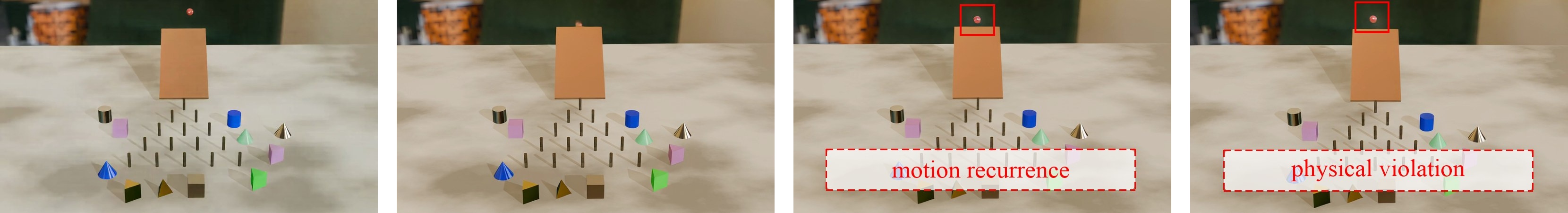}
\vspace{2pt}
{\fontsize{7.25}{8.5}\selectfont Wan2.2-TI2V-5B. Wan2.2-TI2V-5B primarily fails through motion recurrence combined with a physically inconsistent reversal of the ball trajectory. The ball initially begins the expected gravity-driven descent, but after moving downward it reverses direction and rises back toward its earlier position without any visible collision or impulse that could support such an upward motion. From the causal perspective, the initial trigger is only partially realized, while the subsequent reversal interrupts the intended sequence of entry into the Galton board and successive peg-mediated trajectory changes. Instead of progressing monotonically toward the lower-left exit, the rollout returns toward a previously occupied state, producing a clear causal-temporal regression and preventing faithful continuation of the reference pathway. Overall, the sequence is most consistent with incorrect gravity-sensitive physical dynamics together with a temporal-state persistence failure that causes an already completed phase of motion to recur.\par}

\vspace{3pt}
\hrule
\vspace{3pt}
\refstepcounter{qualcase}\label{case:b03-m09}
{\fontsize{8}{9}\selectfont\bfseries Case~\thequalcase\enspace\textperiodcentered\enspace Wan2.2-I2V-A14B}\par
\vspace{1pt}
\qstrip{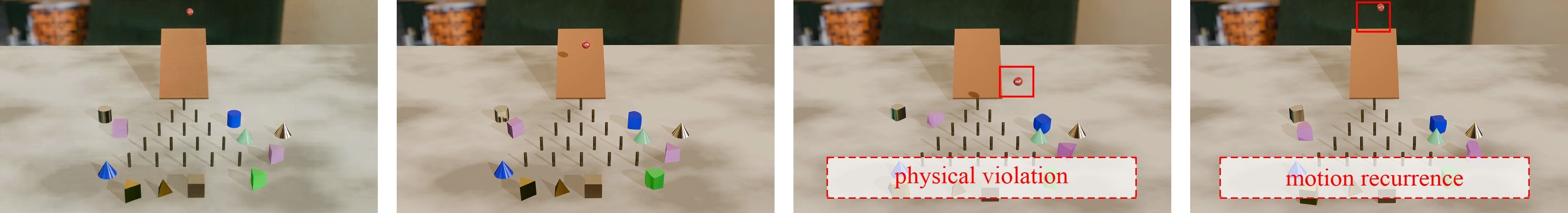}
\vspace{2pt}
{\fontsize{7.25}{8.5}\selectfont Wan2.2-I2V-A14B. Wan2.2-I2V-A14B first exhibits a physically inconsistent deviation in the ball trajectory. After beginning the expected gravity-driven descent, the ball drifts toward the right rather than following the reference path into the Galton board, thereby altering the spatial conditions for the subsequent peg interactions. From the causal perspective, the initial trigger is only partially realized, while the incorrect lateral motion causes the rollout to depart from the intended contact-mediated pathway before the required sequence of peg collisions and trajectory redirections can occur. The failure is then compounded when the ball reverses its motion and travels upward toward a previously occupied position, producing clear motion recurrence and partial state replay. This temporal regression prevents the physical process from advancing monotonically toward the lower-left exit and instead reintroduces an earlier stage of the rollout. Overall, the sequence is most consistent with inaccurate gravity- and trajectory-sensitive physical dynamics, followed by a temporal-state persistence failure that causes an already completed portion of the motion to recur.\par}

\clearpage
\noindent
{\fontsize{10}{11.5}\selectfont\bfseries Level 1 · State-only physical prompt}\par
\vspace{0.5pt}
{\fontsize{8}{9}\selectfont\itshape Galton\_Series.inclined\_galton\_board\_domino\_targets}\par
\vspace{1pt}\hrule\vspace{3pt}
{\fontsize{8}{9}\selectfont\bfseries Ground Truth}\par
\vspace{1pt}
\qstrip{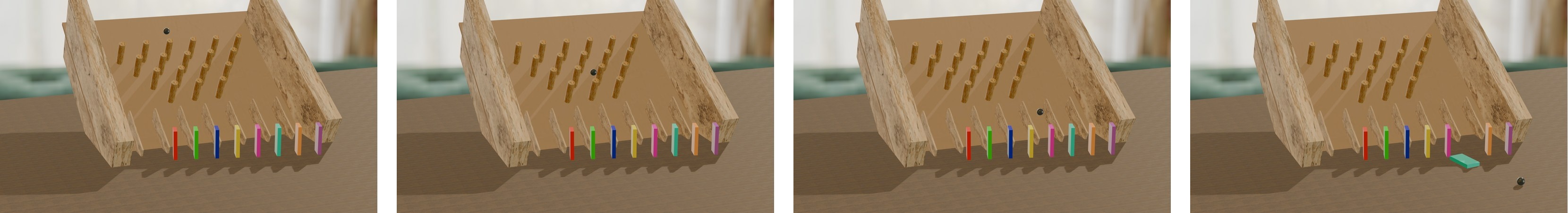}
\vspace{4pt}
\refstepcounter{qualcase}\label{case:b04-m01}
{\fontsize{8}{9}\selectfont\bfseries Case~\thequalcase\enspace\textperiodcentered\enspace CogVideoX-Fun-V1.5}\par
\vspace{1pt}
\qstrip{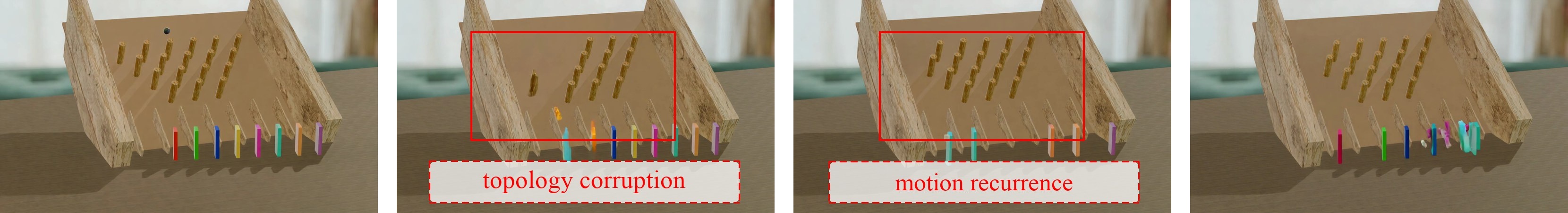}
\vspace{2pt}
{\fontsize{7.25}{8.5}\selectfont CogVideoX-Fun-V1.5. CogVideoX-Fun-V1.5 exhibits a pronounced failure of entity persistence and causal-temporal continuity. The falling ball disappears before completing its trajectory through the Galton board, while several pegs in the upper-left region also vanish from the scene. These missing structures subsequently reappear, producing a clear state replay in which an earlier scene configuration is partially restored rather than the physical process advancing monotonically toward the downstream domino interaction. From the causal perspective, the disappearance of the ball removes the primary participant required to realize the sequence of peg-mediated contacts, while the temporary loss and reappearance of the pegs makes the intermediate collision pathway itself unstable and prevents faithful grounding of the reference causal chain. In the final stage, several lower-right board elements further undergo visible deformation and inconsistent displacement, introducing topology corruption and additional scene-state instability near the intended exit region. Overall, the sequence is most consistent with unstable persistent representations of both dynamic and static entities, compounded by a temporal-state persistence failure that repeatedly reconstructs previously lost scene content.\par}

\vspace{3pt}
\hrule
\vspace{3pt}
\refstepcounter{qualcase}\label{case:b04-m02}
{\fontsize{8}{9}\selectfont\bfseries Case~\thequalcase\enspace\textperiodcentered\enspace Cosmos3-Nano}\par
\vspace{1pt}
\qstrip{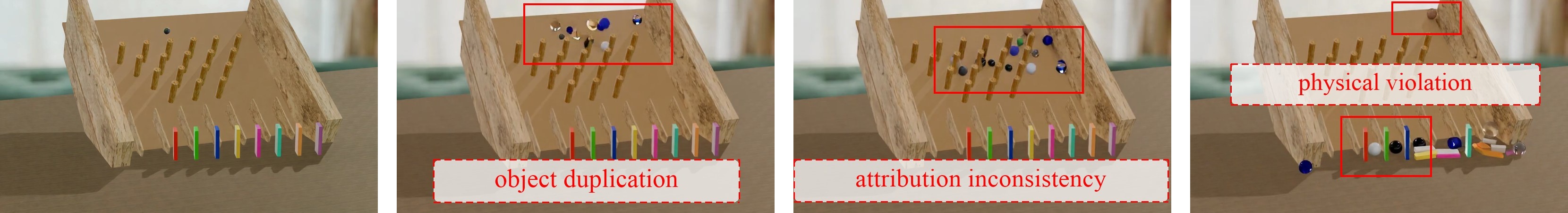}
\vspace{2pt}
{\fontsize{7.25}{8.5}\selectfont Cosmos3-Nano. Cosmos3-Nano exhibits a severe breakdown of entity persistence during the ball’s descent through the Galton board. Instead of maintaining a single causal participant, the original ball is duplicated into multiple spheres with different colors and sizes, and these instances continue to change in appearance as they move downward, producing both object-duplication and identity-corruption failures. From the causal perspective, the intended single-particle trajectory is therefore replaced by multiple spurious participants and parallel contact sequences, making the generated ball–peg interaction pathway fundamentally different from the reference cascade. The rollout also contains clear physical inconsistencies: several balls reach the lower region but then remain stationary instead of continuing along the expected exit trajectory, while others become suspended on the inclined board rather than continuing to descend under gravity. These failures interrupt the intended progression from peg-mediated redirection to the final domino interaction and scene exit. Overall, the sequence is most consistent with unstable entity-state persistence together with inaccurate gravity- and trajectory-sensitive physical dynamics.\par}

\vspace{3pt}
\hrule
\vspace{3pt}
\refstepcounter{qualcase}\label{case:b04-m03}
{\fontsize{8}{9}\selectfont\bfseries Case~\thequalcase\enspace\textperiodcentered\enspace HunyuanVideo-I2V}\par
\vspace{1pt}
\qstrip{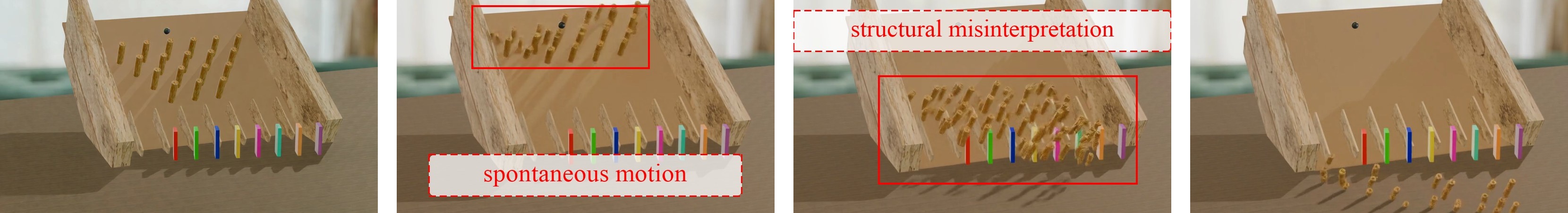}
\vspace{2pt}
{\fontsize{7.25}{8.5}\selectfont HunyuanVideo-I2V. HunyuanVideo-I2V primarily fails through a structural misinterpretation of the Galton-board pegs. Rather than treating the cylindrical pegs as fixed obstacles, the model causes several of them to move spontaneously upward and subsequently fall or roll downward through the board. The initial upward displacement is unsupported by any preceding interaction and therefore constitutes both spontaneous motion and a physical-law inconsistency, while the later downward motion is locally compatible with gravity but occurs only because the model has incorrectly treated the pegs as freely moving objects. From the causal perspective, this introduces spurious events and alters the interaction substrate itself: instead of the ball being redirected by a stable sequence of peg contacts, the obstacles become active participants and create an alternative, non-reference pathway through the scene. Overall, the sequence is most consistent with a structural-role inference failure, in which the model misclassifies fixed Galton-board pegs as dynamic objects, with the resulting unsupported motion further corrupting the intended contact-mediated causal process.\par}

\clearpage
\noindent
{\fontsize{10}{11.5}\selectfont\bfseries Level 1 · State-only physical prompt}\par
\vspace{0.5pt}
{\fontsize{8}{9}\selectfont\itshape Galton\_Series.inclined\_galton\_board\_domino\_targets}\par
\vspace{1pt}\hrule\vspace{3pt}
{\fontsize{8}{9}\selectfont\bfseries Ground Truth}\par
\vspace{1pt}
\qstrip{appendix_e/assets/b04_m01_cogvideox_5b_i2v_gt_strip.jpg}
\vspace{4pt}
\refstepcounter{qualcase}\label{case:b04-m04}
{\fontsize{8}{9}\selectfont\bfseries Case~\thequalcase\enspace\textperiodcentered\enspace LTX-Video-13B}\par
\vspace{1pt}
\qstrip{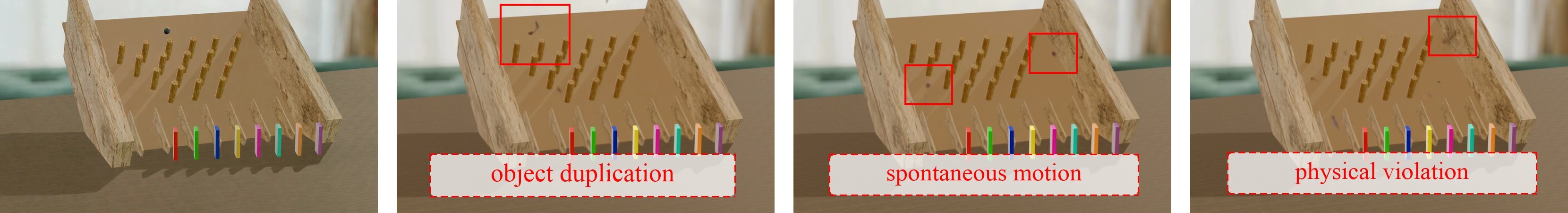}
\vspace{2pt}
{\fontsize{7.25}{8.5}\selectfont LTX-Video-13B. LTX-Video-13B first exhibits an unsupported lateral deviation of the falling ball, which moves leftward instead of continuing along the expected gravity-driven path through the Galton board. The failure is then compounded by object duplication, as the original ball splits into multiple instances that subsequently diverge toward different sides of the board. These duplicated trajectories introduce spurious participants and additional contact possibilities that are absent from the reference process. From the causal perspective, the intended single-ball pathway is therefore replaced by multiple unsupported branches, disrupting the sequence of peg-mediated trajectory redirections that should lead to the final domino interaction. Some of the duplicated balls also exhibit physically inconsistent motion; in particular, one remains suspended on the upper-right portion of the inclined board rather than continuing to descend under gravity. Overall, the sequence is most consistent with unstable entity-state persistence together with inaccurate gravity- and trajectory-sensitive physical dynamics, causing both duplication of the causal participant and failure to maintain a coherent descent through the board.\par}

\vspace{3pt}
\hrule
\vspace{3pt}
\refstepcounter{qualcase}\label{case:b04-m05}
{\fontsize{8}{9}\selectfont\bfseries Case~\thequalcase\enspace\textperiodcentered\enspace MAGI-1-4.5B}\par
\vspace{1pt}
\qstrip{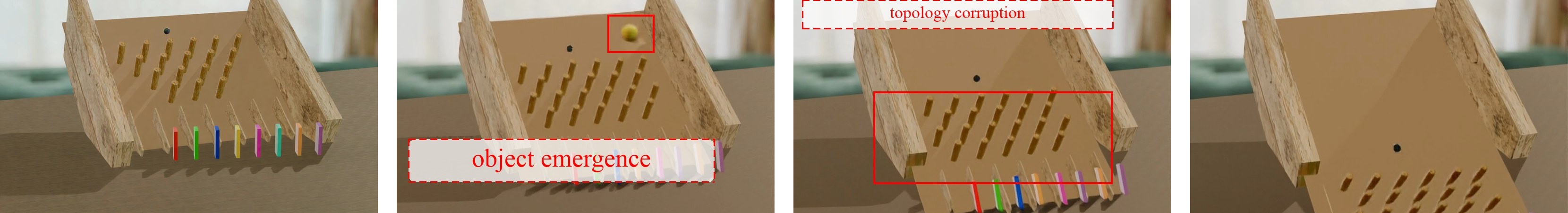}
\vspace{2pt}
{\fontsize{7.25}{8.5}\selectfont MAGI-1-4.5B. MAGI-1-4.5B exhibits both spurious object emergence and structural corruption of the Galton board. Early in the rollout, an additional ball suddenly appears on the right side of the scene despite having no corresponding source or preceding trajectory, introducing a new entity and a spurious event that are absent from the reference process. The board itself then undergoes a pronounced topology change, with the inclined surface extending downward and altering its geometry rather than remaining a fixed physical structure. From the causal perspective, this deformation corrupts the persistent substrate that should constrain the ball’s successive peg interactions, making the intended contact-mediated trajectory and final domino interaction no longer grounded in the reference scene. The newly emerged ball further introduces a potential spurious causal branch unrelated to the original single-ball cascade. Overall, the sequence is most consistent with unstable scene/entity persistence together with a structural-inference failure that causes the model to modify the geometry of what should remain a fixed Galton-board mechanism.\par}

\vspace{3pt}
\hrule
\vspace{3pt}
\refstepcounter{qualcase}\label{case:b04-m06}
{\fontsize{8}{9}\selectfont\bfseries Case~\thequalcase\enspace\textperiodcentered\enspace SkyReels-V2-I2V}\par
\vspace{1pt}
\qstrip{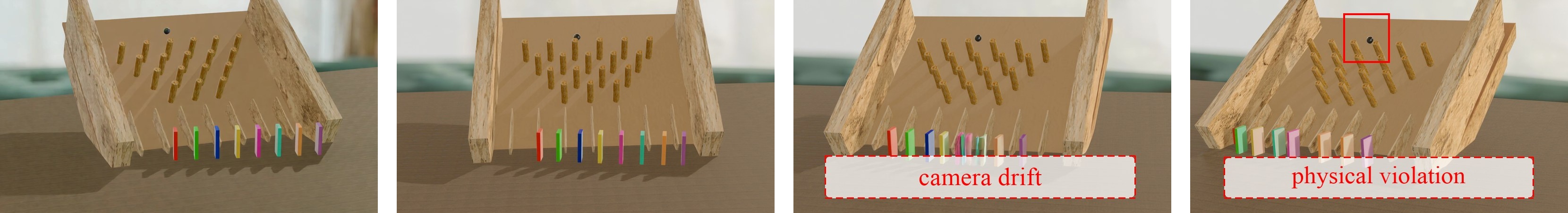}
\vspace{2pt}
{\fontsize{7.25}{8.5}\selectfont SkyReels-V2-I2V. SkyReels-V2-I2V primarily fails through viewpoint instability and incomplete realization of the gravity-driven trigger. Across the rollout, the camera progressively shifts relative to the Galton board, violating the fixed-view condition and altering the apparent spatial relationships among the board, pegs, and lower domino region. At the same time, the ball remains on the inclined surface instead of continuing to descend under gravity, yielding a clear physical-law inconsistency and preventing the intended trajectory through the peg field from being established. From the causal perspective, the cascade therefore breaks at its initiation: the required gravity-driven ball motion is not faithfully realized, so the subsequent peg-mediated redirections, final ball–domino collision, and scene exit cannot occur along the reference pathway. Overall, the sequence is most consistent with a combination of viewpoint-grounding failure and inaccurate realization of the initial gravity-sensitive dynamics.\par}

\clearpage
\noindent
{\fontsize{10}{11.5}\selectfont\bfseries Level 1 · State-only physical prompt}\par
\vspace{0.5pt}
{\fontsize{8}{9}\selectfont\itshape Galton\_Series.inclined\_galton\_board\_domino\_targets}\par
\vspace{1pt}\hrule\vspace{3pt}
{\fontsize{8}{9}\selectfont\bfseries Ground Truth}\par
\vspace{1pt}
\qstrip{appendix_e/assets/b04_m01_cogvideox_5b_i2v_gt_strip.jpg}
\vspace{4pt}
\refstepcounter{qualcase}\label{case:b04-m07}
{\fontsize{8}{9}\selectfont\bfseries Case~\thequalcase\enspace\textperiodcentered\enspace Wan2.1-I2V-14B}\par
\vspace{1pt}
\qstrip{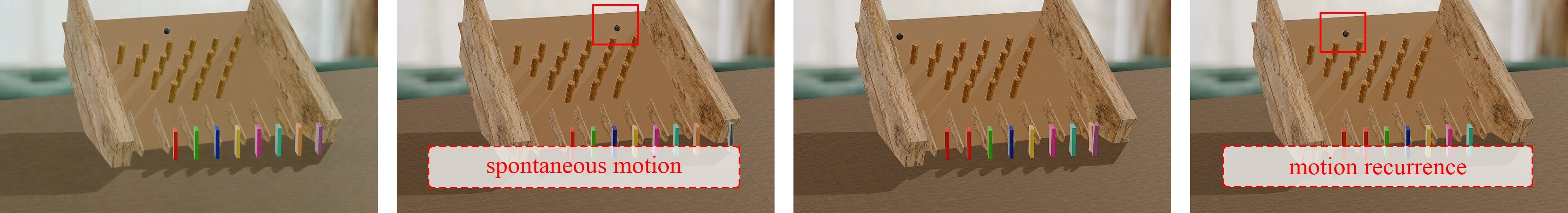}
\vspace{2pt}
{\fontsize{7.25}{8.5}\selectfont Wan2.1-I2V-14B. Wan2.1-I2V-14B primarily fails through unsupported lateral motion and repeated trajectory reversal. Rather than descending through the Galton board under gravity, the ball moves horizontally across the inclined surface, producing a clear physical-law inconsistency and preventing the intended gravity-driven trigger from being faithfully realized. From the causal perspective, the reference process therefore diverges at its earliest stage: because the ball does not enter the peg field along the expected downward trajectory, the subsequent sequence of ball–peg collisions, trajectory redirections, and final domino interaction cannot be established through the correct pathway. The failure is further compounded by motion recurrence, as the ball moves rightward, reverses toward the left, and then returns rightward again without any corresponding physical trigger. This oscillatory behavior introduces a causal-temporal regression rather than monotonic progression through the board. Overall, the sequence is most consistent with inaccurate gravity- and trajectory-sensitive physical dynamics together with a temporal-state persistence failure that repeatedly regenerates previously visited motion states.\par}

\vspace{3pt}
\hrule
\vspace{3pt}
\refstepcounter{qualcase}\label{case:b04-m08}
{\fontsize{8}{9}\selectfont\bfseries Case~\thequalcase\enspace\textperiodcentered\enspace Wan2.2-TI2V-5B}\par
\vspace{1pt}
\qstrip{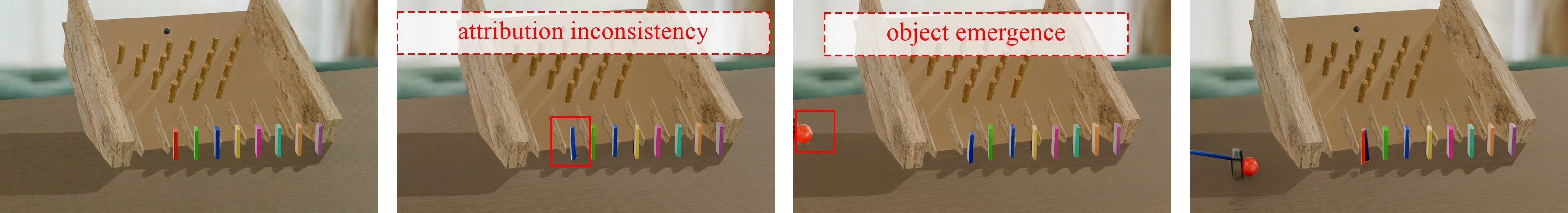}
\vspace{2pt}
{\fontsize{7.25}{8.5}\selectfont Wan2.2-TI2V-5B. Wan2.2-TI2V-5B exhibits a combination of identity inconsistency, spurious object emergence, and failure to realize the gravity-driven trigger. Early in the rollout, the first domino on the left changes color from its original appearance to blue, producing a localized identity-corruption failure in an otherwise static scene element. A more substantial error then occurs when an additional object suddenly appears on the far left without any preceding trajectory or physical source, introducing a spurious entity and an event absent from the reference rollout. At the same time, the ball itself fails to descend along the inclined Galton-board surface as required by gravity, so the intended trigger event is not faithfully realized. From the causal perspective, this prevents the subsequent sequence of peg-mediated trajectory redirections and the final ball–domino collision from being established along the reference pathway, while the newly emerged object creates an unrelated spurious branch in the generated scene. Overall, the sequence is most consistent with unstable entity/state persistence together with inaccurate realization of the initial gravity-sensitive dynamics.\par}

\vspace{3pt}
\hrule
\vspace{3pt}
\refstepcounter{qualcase}\label{case:b04-m09}
{\fontsize{8}{9}\selectfont\bfseries Case~\thequalcase\enspace\textperiodcentered\enspace Wan2.2-I2V-A14B}\par
\vspace{1pt}
\qstrip{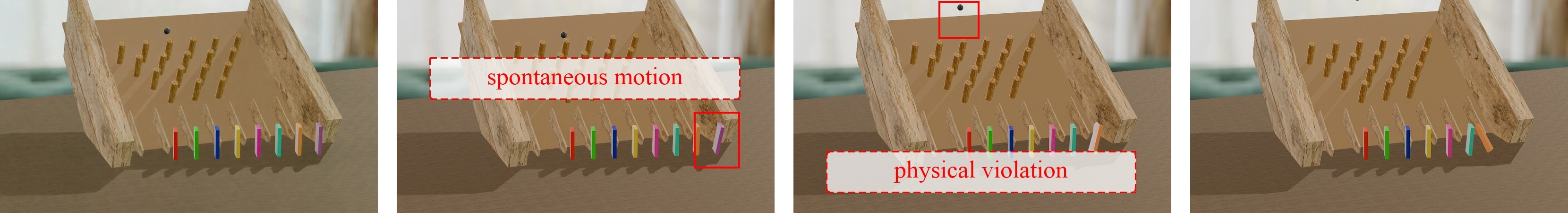}
\vspace{2pt}
{\fontsize{7.25}{8.5}\selectfont Wan2.2-I2V-A14B. Wan2.2-I2V-A14B exhibits both unsupported motion among the lower domino elements and a clear violation of the expected gravity-driven ball dynamics. Several objects near the bottom of the Galton board begin moving without any preceding ball contact, with the rightmost domino even toppling spontaneously despite having no physical trigger in the reference rollout. More critically, the ball itself fails to descend along the inclined board and instead rises upward into the air, producing a direct physical-law inconsistency and leaving the intended trigger event unrealized. From the causal perspective, the spontaneous domino motion introduces spurious downstream events and an unsupported causal branch, while the missing gravity-driven descent prevents the reference sequence of peg-mediated redirections and the final ball–domino interaction from being established. Overall, the sequence is most consistent with inaccurate gravity-sensitive physical dynamics together with weak preservation of causal dependencies and object-role grounding for scene elements that should remain inactive until contacted.\par}

\clearpage
\noindent
{\fontsize{10}{11.5}\selectfont\bfseries Level 2 · Trigger-aware prompt}\par
\vspace{0.5pt}
{\fontsize{8}{9}\selectfont\itshape Domino\_Series.ball\_trigger\_step\_domino\_chain}\par
\vspace{1pt}\hrule\vspace{3pt}
{\fontsize{8}{9}\selectfont\bfseries Ground Truth}\par
\vspace{1pt}
\qstrip{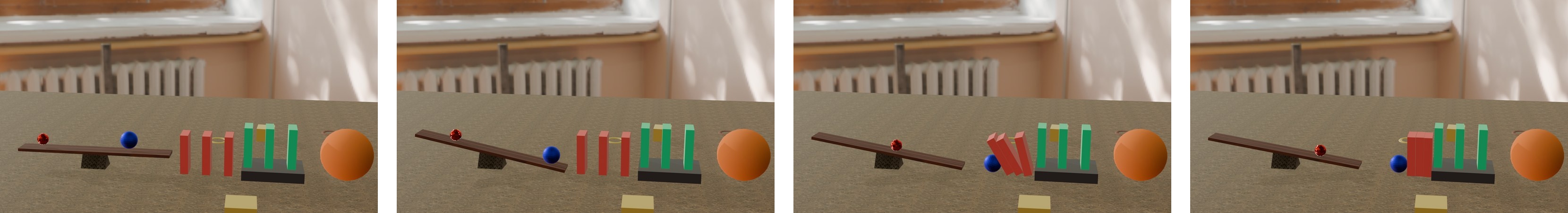}
\vspace{4pt}
\refstepcounter{qualcase}\label{case:b05-m01}
{\fontsize{8}{9}\selectfont\bfseries Case~\thequalcase\enspace\textperiodcentered\enspace CogVideoX-Fun-V1.5}\par
\vspace{1pt}
\qstrip{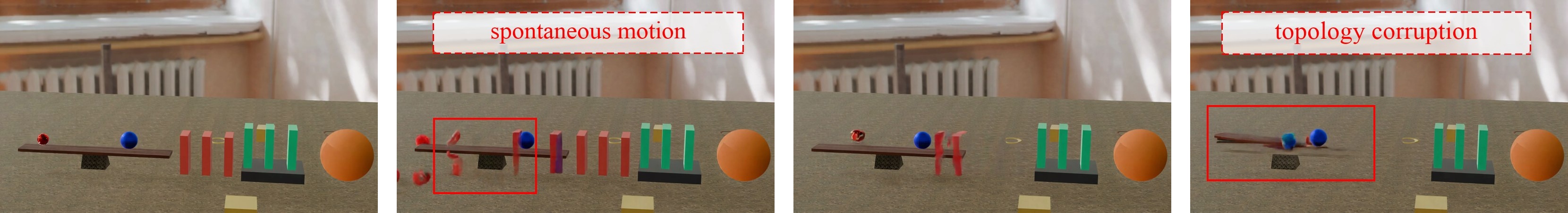}
\vspace{2pt}
{\fontsize{7.25}{8.5}\selectfont CogVideoX-Fun-V1.5. CogVideoX-Fun-V1.5 primarily fails through severe corruption of the seesaw structure, accompanied by object displacement and identity instability. As early as the second frame, the seesaw region develops a distorted, overlapping geometry rather than preserving a single rigid plank, indicating clear topology corruption and a loss of stable object identity in the central mechanical component. Although the moving ball does progress rightward in a manner that partially resembles the reference rollout, this motion is no longer grounded in a coherent lever mechanism because the supporting seesaw itself has already become structurally corrupted. From the causal perspective, the initial ball-motion event is therefore only partially realized, while the plank-mediated interaction that should physically support and propagate the downstream cascade is incorrectly instantiated. The later frames further preserve this instability through continued displacement and deformation of the affected objects, making the subsequent interactions with the dominoes unreliable. Overall, the sequence is most consistent with unstable entity/state persistence and a failure to maintain a coherent structural representation of the seesaw, which in turn undermines the causal grounding of the entire rollout.\par}

\vspace{3pt}
\hrule
\vspace{3pt}
\refstepcounter{qualcase}\label{case:b05-m02}
{\fontsize{8}{9}\selectfont\bfseries Case~\thequalcase\enspace\textperiodcentered\enspace Cosmos3-Nano}\par
\vspace{1pt}
\qstrip{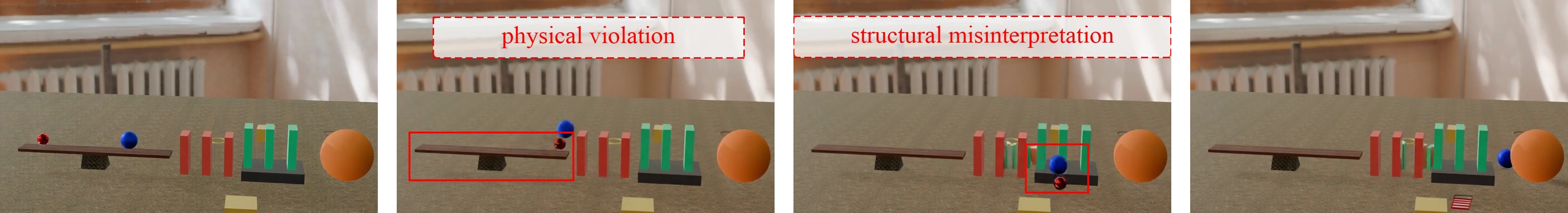}
\vspace{2pt}
{\fontsize{7.25}{8.5}\selectfont Cosmos3-Nano. Cosmos3-Nano partially captures the intended rightward motion of both the red and blue balls, but fails to realize the corresponding seesaw dynamics. As the balls move toward the right, the plank remains approximately horizontal instead of rotating into the expected left-high, right-low configuration, producing a clear physical-law inconsistency in the lever response. From the causal perspective, the ball-motion events are therefore present, but the mechanically coupled seesaw-rotation event that should accompany and enable the downstream interaction is incorrectly instantiated. A second failure occurs after the blue ball leaves the seesaw: rather than continuing along the spatially aligned path toward the domino row and producing the intended collision, the model places both balls in front of the domino structure. This indicates a structural and spatial misinterpretation of the scene geometry, causing the expected ball–domino contact event to be missed and breaking propagation toward the downstream domino cascade. Overall, the sequence is most consistent with inaccurate torque-sensitive physical dynamics together with incorrect spatial-structure inference of the ball trajectory and domino alignment.\par}

\vspace{3pt}
\hrule
\vspace{3pt}
\refstepcounter{qualcase}\label{case:b05-m03}
{\fontsize{8}{9}\selectfont\bfseries Case~\thequalcase\enspace\textperiodcentered\enspace HunyuanVideo-I2V}\par
\vspace{1pt}
\qstrip{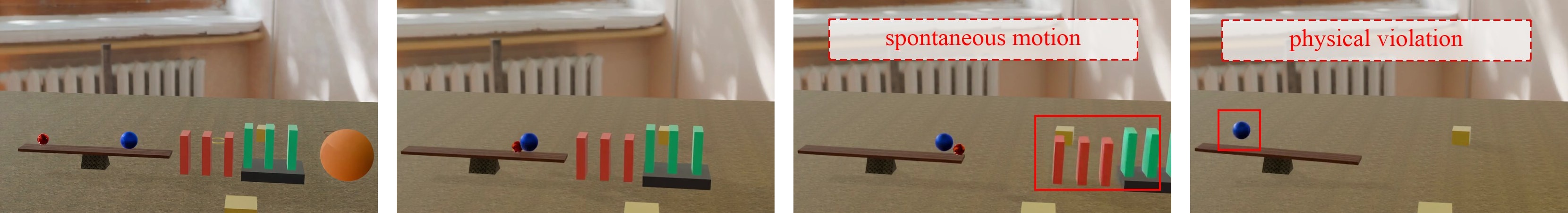}
\vspace{2pt}
{\fontsize{7.25}{8.5}\selectfont HunyuanVideo-I2V. HunyuanVideo-I2V partially captures the intended rightward motion of the red and blue balls, but fails to realize the corresponding mechanical response of the seesaw. As the balls move rightward, the plank does not rotate into the expected left-high, right-low configuration, yielding a clear physical-law inconsistency in the torque-driven dynamics. The rollout then introduces an additional downstream error: the domino row shifts rightward even though it has not been contacted by the ball, producing a spurious motion event and a causal shortcut that bypasses the required ball–domino interaction. The sequence departs further from the reference process when the blue ball, after reaching the right side of the seesaw, reverses direction and moves back leftward while also jumping upward, generating unsupported motion that again violates the expected physics. From the causal perspective, the model therefore preserves only part of the initial ball-motion pattern while breaking both the mechanically coupled seesaw response and the contact-mediated propagation to the dominoes. Overall, the behavior is most consistent with inaccurate torque- and collision-sensitive physical dynamics, together with weak preservation of causal dependencies and temporal-state consistency.\par}

\clearpage
\noindent
{\fontsize{10}{11.5}\selectfont\bfseries Level 2 · Trigger-aware prompt}\par
\vspace{0.5pt}
{\fontsize{8}{9}\selectfont\itshape Domino\_Series.ball\_trigger\_step\_domino\_chain}\par
\vspace{1pt}\hrule\vspace{3pt}
{\fontsize{8}{9}\selectfont\bfseries Ground Truth}\par
\vspace{1pt}
\qstrip{appendix_e/assets/b05_m01_cogvideox_5b_i2v_gt_strip.jpg}
\vspace{4pt}
\refstepcounter{qualcase}\label{case:b05-m04}
{\fontsize{8}{9}\selectfont\bfseries Case~\thequalcase\enspace\textperiodcentered\enspace LTX-Video-13B}\par
\vspace{1pt}
\qstrip{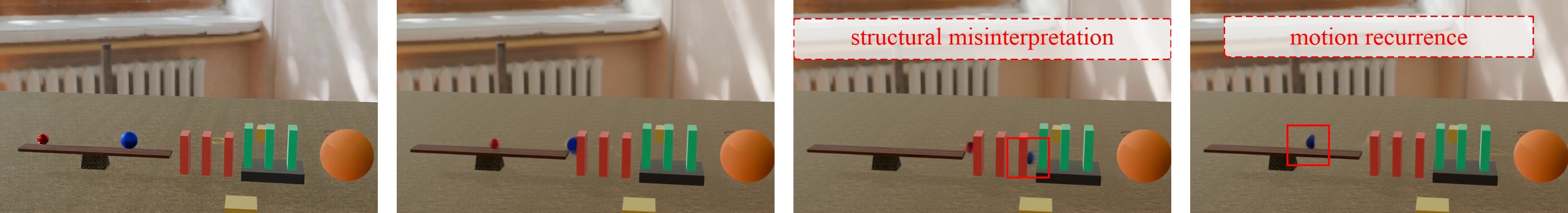}
\vspace{2pt}
{\fontsize{7.25}{8.5}\selectfont LTX-Video-13B. LTX-Video-13B exhibits a combination of spatial-structure misinterpretation, incorrect seesaw dynamics, and motion recurrence. After the blue ball moves rightward, it appears behind the domino row rather than following the spatially aligned path required to collide with the dominoes, indicating that the model has misinterpreted the relative depth and interaction geometry of the scene. From the causal perspective, the ball-motion event is present, but the expected ball–domino collision is missed, breaking propagation before the downstream domino cascade can be triggered. At the same time, the seesaw remains approximately horizontal instead of rotating into the expected left-high, right-low configuration. This may reflect either an incorrect torque-sensitive physical response or a structural misinterpretation in which the model treats the support beneath the plank as a stable platform rather than as a freely rotating fulcrum. The rollout then exhibits motion recurrence: after the blue ball has already moved rightward, it reappears on the left side and traverses the seesaw again, partially replaying an earlier phase of the trajectory. Overall, the sequence is most consistent with incorrect spatial and structural inference, compounded by weak temporal-state persistence.\par}

\vspace{3pt}
\hrule
\vspace{3pt}
\refstepcounter{qualcase}\label{case:b05-m05}
{\fontsize{8}{9}\selectfont\bfseries Case~\thequalcase\enspace\textperiodcentered\enspace MAGI-1-4.5B}\par
\vspace{1pt}
\qstrip{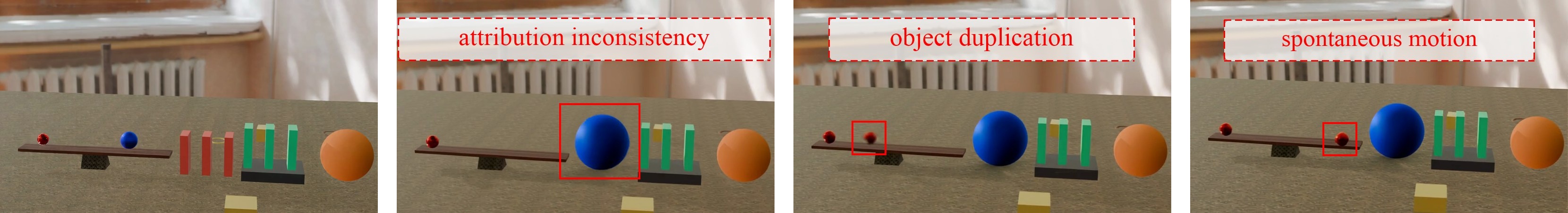}
\vspace{2pt}
{\fontsize{7.25}{8.5}\selectfont MAGI-1-4.5B. MAGI-1-4.5B exhibits a pronounced breakdown of entity identity and persistence during the rollout. After the blue ball moves rightward off the seesaw, it enlarges dramatically into a much bigger sphere, producing a clear identity-corruption failure that alters the size and role of the main causal participant. The sequence then further deteriorates through object duplication, as the red ball is replicated into an additional instance, while unsupported rightward motion is introduced without the required physical trigger. At the same time, the seesaw itself remains approximately balanced rather than rotating into the expected left-high, right-low configuration, indicating a clear physical-law inconsistency in the lever dynamics. From the causal perspective, these failures corrupt both the participants and the mechanism of the reference process: the main ball no longer preserves its identity, a spurious extra ball creates an additional unsupported branch, and the mechanically coupled seesaw response is not correctly realized. As a result, the downstream interaction with the dominoes is no longer grounded in the intended contact-mediated pathway. Overall, the sequence is most consistent with unstable entity-state persistence together with inaccurate torque-sensitive physical dynamics and weak preservation of causal dependencies.\par}

\vspace{3pt}
\hrule
\vspace{3pt}
\refstepcounter{qualcase}\label{case:b05-m06}
{\fontsize{8}{9}\selectfont\bfseries Case~\thequalcase\enspace\textperiodcentered\enspace SkyReels-V2-I2V}\par
\vspace{1pt}
\qstrip{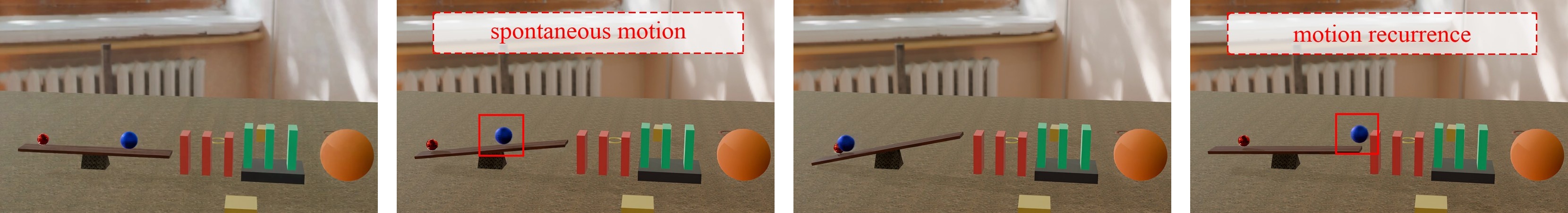}
\vspace{2pt}
{\fontsize{7.25}{8.5}\selectfont SkyReels-V2-I2V. SkyReels-V2-I2V fails to follow the intended initial evolution of the scene. Instead of producing the reference left-high, right-low seesaw configuration with both balls progressing rightward, the blue ball moves spontaneously toward the left, causing the plank to tilt in the opposite direction and sending the rollout along an incorrect early trajectory. From the causal perspective, the expected trigger dynamics are therefore not faithfully realized, and the subsequent ball motion begins from a pathway inconsistent with the reference cascade. The failure is then compounded by motion recurrence: after sliding toward the left side of the seesaw, the blue ball reverses direction and travels rightward again toward the opposite end, partially replaying the intended motion only after an unsupported preceding phase. Interestingly, the seesaw subsequently responds to this rightward motion in a locally plausible manner, tilting as the ball changes sides; however, this locally consistent response does not restore the correct causal history because it follows an earlier spurious trajectory. Overall, the sequence is most consistent with incorrect realization of the initial motion condition together with a temporal-state persistence failure that allows the ball to reverse and regenerate a previously expected motion pattern.\par}

\clearpage
\noindent
{\fontsize{10}{11.5}\selectfont\bfseries Level 2 · Trigger-aware prompt}\par
\vspace{0.5pt}
{\fontsize{8}{9}\selectfont\itshape Domino\_Series.ball\_trigger\_step\_domino\_chain}\par
\vspace{1pt}\hrule\vspace{3pt}
{\fontsize{8}{9}\selectfont\bfseries Ground Truth}\par
\vspace{1pt}
\qstrip{appendix_e/assets/b05_m01_cogvideox_5b_i2v_gt_strip.jpg}
\vspace{4pt}
\refstepcounter{qualcase}\label{case:b05-m07}
{\fontsize{8}{9}\selectfont\bfseries Case~\thequalcase\enspace\textperiodcentered\enspace Wan2.1-I2V-14B}\par
\vspace{1pt}
\qstrip{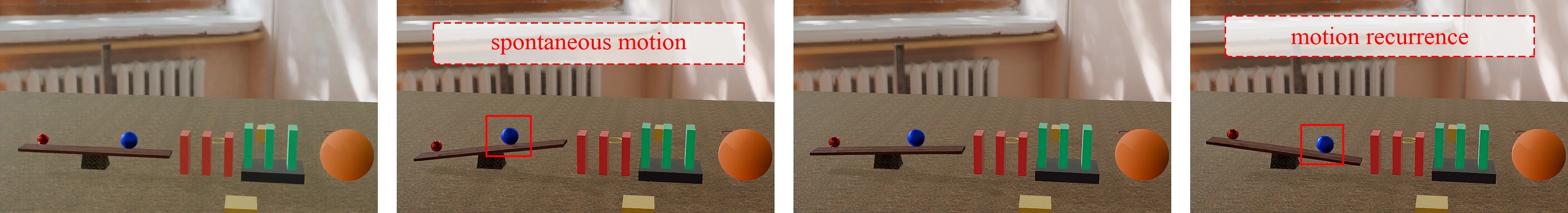}
\vspace{2pt}
{\fontsize{7.25}{8.5}\selectfont Wan2.1-I2V-14B. Wan2.1-I2V-14B exhibits the same early failure pattern as the preceding case, with the generated rollout departing from the intended initial seesaw dynamics. Instead of producing the reference left-high, right-low configuration with both balls moving rightward, the blue ball first moves spontaneously toward the left, causing the seesaw to tilt in the opposite left-low, right-high direction. From the causal perspective, this means that the initial motion event and its mechanically coupled lever response are both incorrectly realized, sending the rollout along the wrong early pathway before the intended downstream interaction can occur. The model then introduces motion recurrence: after this unsupported leftward phase, the blue and red balls move rightward again and the seesaw transitions into the expected left-high, right-low configuration. Although this later state is locally closer to the reference dynamics, it is reached only after an incorrect preceding evolution, producing a partial replay of the intended motion rather than faithful causal-temporal progression from the initial condition. Overall, the sequence is most consistent with incorrect realization of the initial seesaw dynamics together with a temporal-state persistence failure that allows the system to reverse and regenerate the expected motion pattern after an earlier spurious trajectory.\par}

\vspace{3pt}
\hrule
\vspace{3pt}
\refstepcounter{qualcase}\label{case:b05-m08}
{\fontsize{8}{9}\selectfont\bfseries Case~\thequalcase\enspace\textperiodcentered\enspace Wan2.2-TI2V-5B}\par
\vspace{1pt}
\qstrip{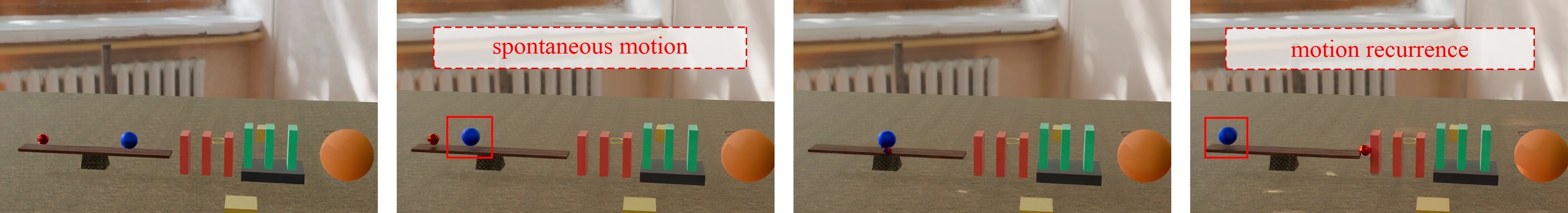}
\vspace{2pt}
{\fontsize{7.25}{8.5}\selectfont Wan2.2-TI2V-5B. Wan2.2-TI2V-5B primarily fails through unsupported lateral motion and repeated temporal regression. Instead of following the intended rightward evolution, the blue ball first moves spontaneously toward the left, violating the prescribed initial dynamics. More importantly, the seesaw remains nearly unchanged during this displacement rather than responding mechanically to the altered load, indicating either an incorrect realization of the lever dynamics or a structural misinterpretation of how the plank is supported and allowed to rotate. From the causal perspective, the initial ball-motion event is therefore inconsistent with the reference process, while the mechanically coupled seesaw response is effectively missing. The rollout then develops a clear motion loop: the blue ball moves back toward the center, repeats the same leftward displacement, and again returns to an earlier configuration rather than progressing toward the downstream domino interaction. This repeated reversal constitutes motion recurrence and causal-temporal regression. Overall, the sequence is most consistent with inaccurate seesaw-response modeling, potentially compounded by structural misinterpretation, together with a temporal-state persistence failure that causes the same unsupported motion pattern to recur.\par}

\vspace{3pt}
\hrule
\vspace{3pt}
\refstepcounter{qualcase}\label{case:b05-m09}
{\fontsize{8}{9}\selectfont\bfseries Case~\thequalcase\enspace\textperiodcentered\enspace Wan2.2-I2V-A14B}\par
\vspace{1pt}
\qstrip{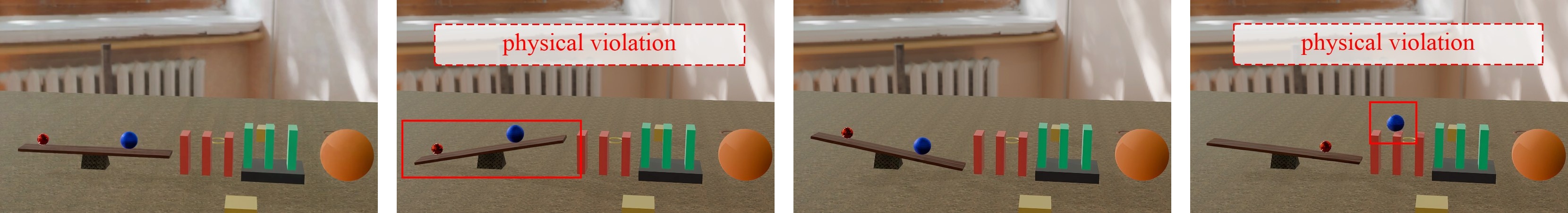}
\vspace{2pt}
{\fontsize{7.25}{8.5}\selectfont Wan2.2-I2V-A14B. Wan2.2-I2V-A14B exhibits multiple physically inconsistent responses in the seesaw dynamics. Early in the rollout, the plank rotates into a left-low, right-high configuration, opposite to the reference behavior expected from the relative loading of the two balls, indicating an incorrect realization of the torque- and mass-sensitive mechanics. The system then reverses into the opposite left-high, right-low orientation, producing a repeated change in the lever state rather than a stable progression toward the downstream interaction. Although the blue ball subsequently moves rightward in a direction that is locally consistent with the reference trajectory, it rises above the seesaw instead of remaining constrained by the plank surface, introducing a second clear physical-law violation in the ball–support interaction. From the causal perspective, the initial ball-motion and lever-response events are therefore incorrectly coupled, and the resulting trajectory is not grounded in the reference mechanical pathway required to reach the dominoes. Overall, the sequence is most consistent with inaccurate modeling of mass-sensitive torque and support dynamics, potentially compounded by an imperfect structural interpretation of the seesaw configuration.\par}

\clearpage
\noindent
{\fontsize{10}{11.5}\selectfont\bfseries Level 2 · Trigger-aware prompt}\par
\vspace{0.5pt}
{\fontsize{8}{9}\selectfont\itshape Domino\_Series.dual\_ball\_inverted\_y\_domino\_course}\par
\vspace{1pt}\hrule\vspace{3pt}
{\fontsize{8}{9}\selectfont\bfseries Ground Truth}\par
\vspace{1pt}
\qstrip{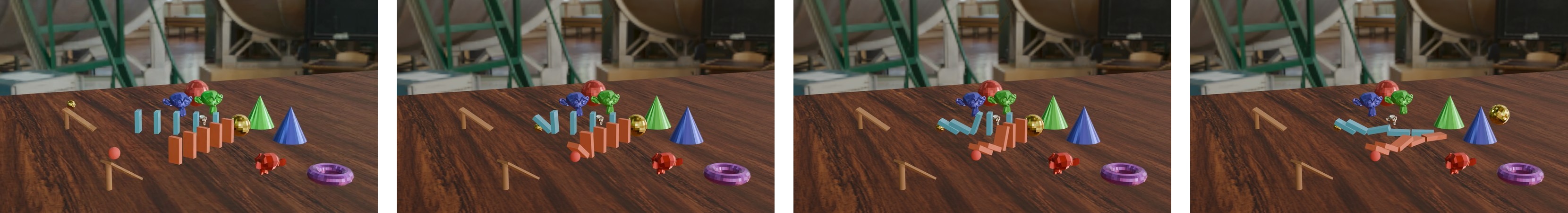}
\vspace{4pt}
\refstepcounter{qualcase}\label{case:b06-m01}
{\fontsize{8}{9}\selectfont\bfseries Case~\thequalcase\enspace\textperiodcentered\enspace CogVideoX-Fun-V1.5}\par
\vspace{1pt}
\qstrip{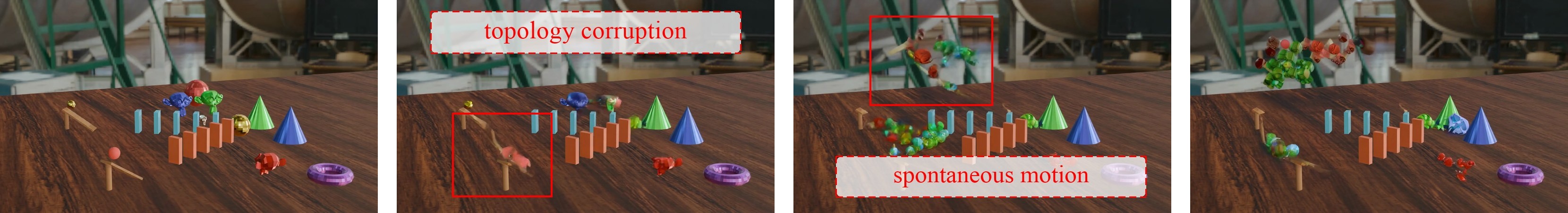}
\vspace{2pt}
{\fontsize{7.25}{8.5}\selectfont CogVideoX-Fun-V1.5. CogVideoX-Fun-V1.5 exhibits a severe collapse of scene integrity almost immediately after generation begins. The left-side ramp undergoes pronounced topology corruption, losing its original rigid geometry and becoming visibly distorted. This structural failure is accompanied by widespread spontaneous motion across multiple scene objects, which move in disordered and unsupported directions without the required upstream interactions. From the causal perspective, the corruption of the ramp and surrounding objects destroys the stable physical substrate needed to realize the reference sequence of gravity-driven ball motion, ball–domino contacts, and subsequent domino propagation. The numerous unsupported motions further introduce spurious events and causal branches, causing the generated rollout to depart almost entirely from the intended causal cascade. Overall, the sequence is most consistent with a severe failure to maintain persistent scene and entity representations, compounded by unstable physical-dynamics generation once the underlying scene structure has already collapsed.\par}

\vspace{3pt}
\hrule
\vspace{3pt}
\refstepcounter{qualcase}\label{case:b06-m02}
{\fontsize{8}{9}\selectfont\bfseries Case~\thequalcase\enspace\textperiodcentered\enspace Cosmos3-Nano}\par
\vspace{1pt}
\qstrip{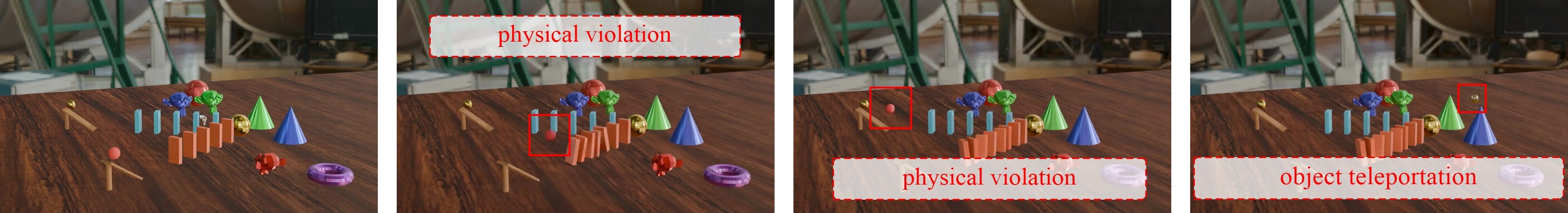}
\vspace{2pt}
{\fontsize{7.25}{8.5}\selectfont Cosmos3-Nano. Cosmos3-Nano exhibits a physically inconsistent trajectory immediately after the left-side ball leaves the lower ramp. Instead of continuing forward along the expected path toward the domino row, the ball moves upward and then further toward the upper-left region, producing a clear violation of the gravity- and trajectory-constrained dynamics. From the causal perspective, this incorrect motion prevents the intended ball–domino contact from being faithfully realized, yet the dominoes are nevertheless disturbed, suggesting a causal shortcut in which a downstream response occurs without the required upstream collision. The rollout then departs even further from the reference process through object teleportation: in the final frame, the ball abruptly appears near the far-right side of the scene without a continuous intermediate trajectory. This discontinuous relocation bypasses the required contact-mediated pathway and introduces an unsupported state that is absent from the Ground Truth. Overall, the sequence is most consistent with inaccurate trajectory-sensitive physical dynamics together with weak preservation of causal dependencies and spatiotemporal state continuity.\par}

\vspace{3pt}
\hrule
\vspace{3pt}
\refstepcounter{qualcase}\label{case:b06-m03}
{\fontsize{8}{9}\selectfont\bfseries Case~\thequalcase\enspace\textperiodcentered\enspace HunyuanVideo-I2V}\par
\vspace{1pt}
\qstrip{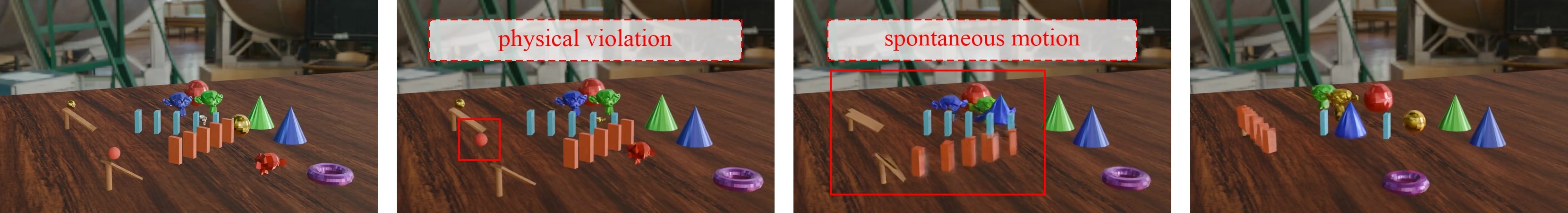}
\vspace{2pt}
{\fontsize{7.25}{8.5}\selectfont HunyuanVideo-I2V. HunyuanVideo-I2V exhibits a clear failure of gravity-consistent motion at the very beginning of the rollout. The ball in the lower-left region should descend along the ramp under gravity, but instead moves upward toward the left, producing a direct physical-law inconsistency and leaving the intended trigger trajectory unrealized. From the causal perspective, this incorrect initial motion prevents the ball from reaching and physically activating the domino sequence through the reference pathway. The rollout then introduces a second error when the dominoes undergo spontaneous horizontal displacement without the required preceding ball contact, creating spurious downstream events and a causal shortcut that bypasses the intended mediator. As a result, the generated process departs from the reference cascade both at its initiation and during downstream propagation. Overall, the sequence is most consistent with inaccurate gravity- and trajectory-sensitive physical dynamics together with weak preservation of the contact-mediated causal dependencies that should govern the domino motion.\par}

\clearpage
\noindent
{\fontsize{10}{11.5}\selectfont\bfseries Level 2 · Trigger-aware prompt}\par
\vspace{0.5pt}
{\fontsize{8}{9}\selectfont\itshape Domino\_Series.dual\_ball\_inverted\_y\_domino\_course}\par
\vspace{1pt}\hrule\vspace{3pt}
{\fontsize{8}{9}\selectfont\bfseries Ground Truth}\par
\vspace{1pt}
\qstrip{appendix_e/assets/b06_m01_cogvideox_5b_i2v_gt_strip.jpg}
\vspace{4pt}
\refstepcounter{qualcase}\label{case:b06-m04}
{\fontsize{8}{9}\selectfont\bfseries Case~\thequalcase\enspace\textperiodcentered\enspace LTX-Video-13B}\par
\vspace{1pt}
\qstrip{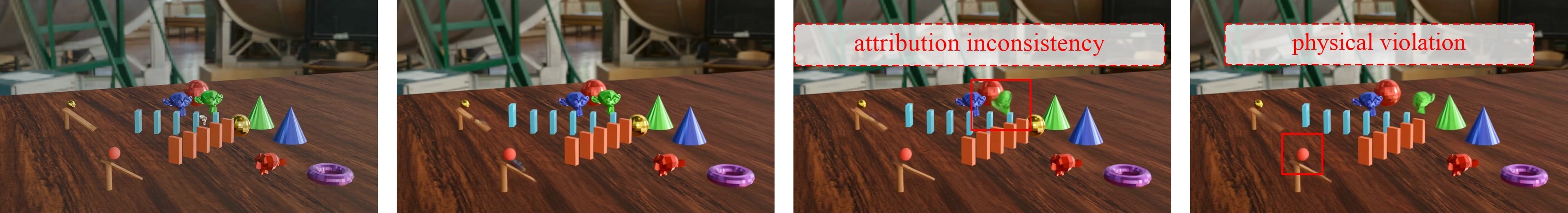}
\vspace{2pt}
{\fontsize{7.25}{8.5}\selectfont LTX-Video-13B. LTX-Video-13B exhibits both entity-identity corruption and failure to realize the initial gravity-driven motion. During the rollout, the rightmost object among the three central decorative structures changes its shape and appearance, producing a clear identity-corruption failure and weakening the persistence of the local scene representation. More importantly, the small ball on the lower-left ramp remains essentially stationary instead of descending under gravity, yielding a direct physical-law inconsistency and leaving the intended trigger event unrealized. From the causal perspective, this missing ball motion prevents the subsequent ball–domino interaction from occurring through the reference pathway, so the downstream domino cascade cannot be causally initiated as intended. The identity change of the central object is largely separate from the main causal chain, but further indicates instability in maintaining persistent scene entities. Overall, the sequence is most consistent with inaccurate realization of the initial gravity-sensitive dynamics, accompanied by localized entity-state persistence failure in causally secondary objects.\par}

\vspace{3pt}
\hrule
\vspace{3pt}
\refstepcounter{qualcase}\label{case:b06-m05}
{\fontsize{8}{9}\selectfont\bfseries Case~\thequalcase\enspace\textperiodcentered\enspace MAGI-1-4.5B}\par
\vspace{1pt}
\qstrip{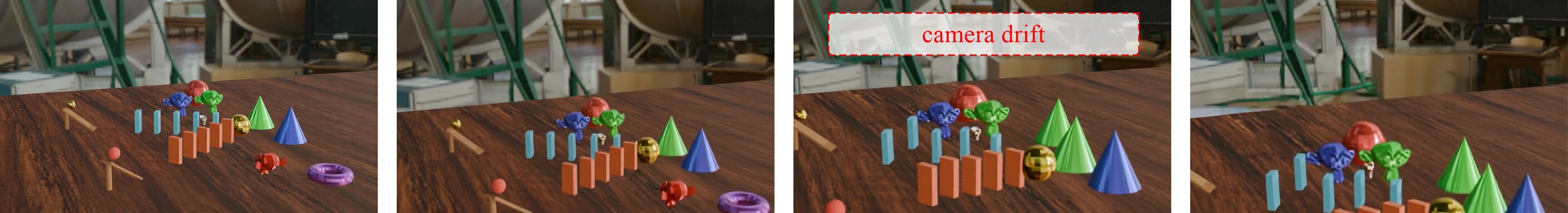}
\vspace{2pt}
{\fontsize{7.25}{8.5}\selectfont MAGI-1-4.5B. MAGI-1-4.5B primarily fails through viewpoint instability and omission of the expected gravity-driven trigger. Across the rollout, the camera progressively shifts relative to the scene, producing clear camera drift and altering the apparent spatial relationships among the ramps, dominoes, and downstream objects. At the same time, the ball on the left-side ramp does not undergo the expected downhill motion under gravity, yielding a physical-law inconsistency and leaving the initiating event unrealized. From the causal perspective, the reference cascade therefore breaks at its earliest stage: because the ball never correctly descends from the ramp, the subsequent ball–domino contact and downstream domino propagation cannot be activated through the intended pathway. The camera drift further weakens the stable spatial grounding of the generated process, making the evolving scene harder to align with the fixed reference configuration. Overall, the sequence is most consistent with a combination of viewpoint-grounding failure and inaccurate realization of the initial gravity-sensitive dynamics.\par}

\vspace{3pt}
\hrule
\vspace{3pt}
\refstepcounter{qualcase}\label{case:b06-m06}
{\fontsize{8}{9}\selectfont\bfseries Case~\thequalcase\enspace\textperiodcentered\enspace SkyReels-V2-I2V}\par
\vspace{1pt}
\qstrip{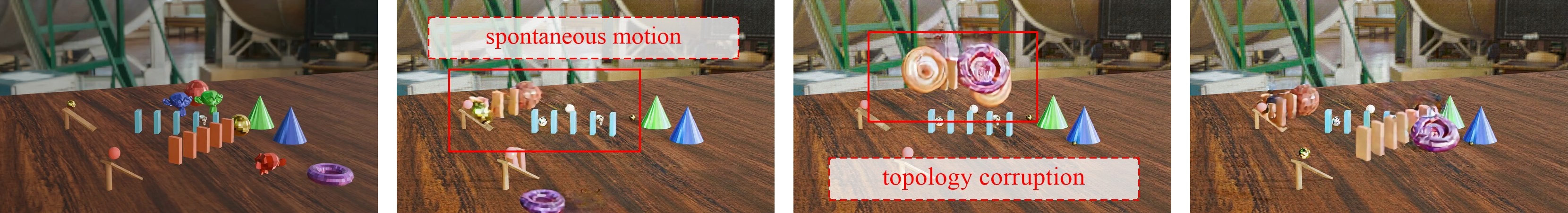}
\vspace{2pt}
{\fontsize{7.25}{8.5}\selectfont SkyReels-V2-I2V. SkyReels-V2-I2V exhibits a severe breakdown of both scene integrity and motion consistency. Several objects begin moving spontaneously without any corresponding upstream trigger, while their shapes, colors, and overall appearances change dramatically across the rollout, producing pronounced topology and identity corruption. These visual transformations make it difficult to maintain a stable correspondence between generated entities and the physical participants in the reference scene. From the causal perspective, the spontaneous motions introduce multiple spurious events and unsupported causal branches, while the loss of persistent object identity corrupts the substrate on which the intended sequence of ramp motion, ball–domino contact, and downstream propagation should be grounded. As the scene progressively departs from its original geometry and object configuration, the reference causal cascade can no longer be meaningfully realized. Overall, the sequence is most consistent with a severe failure of persistent entity/world-state representation, compounded by unsupported dynamics that cause otherwise stable scene elements to move without valid physical causes.\par}

\clearpage
\noindent
{\fontsize{10}{11.5}\selectfont\bfseries Level 2 · Trigger-aware prompt}\par
\vspace{0.5pt}
{\fontsize{8}{9}\selectfont\itshape Domino\_Series.dual\_ball\_inverted\_y\_domino\_course}\par
\vspace{1pt}\hrule\vspace{3pt}
{\fontsize{8}{9}\selectfont\bfseries Ground Truth}\par
\vspace{1pt}
\qstrip{appendix_e/assets/b06_m01_cogvideox_5b_i2v_gt_strip.jpg}
\vspace{4pt}
\refstepcounter{qualcase}\label{case:b06-m07}
{\fontsize{8}{9}\selectfont\bfseries Case~\thequalcase\enspace\textperiodcentered\enspace Wan2.1-I2V-14B}\par
\vspace{1pt}
\qstrip{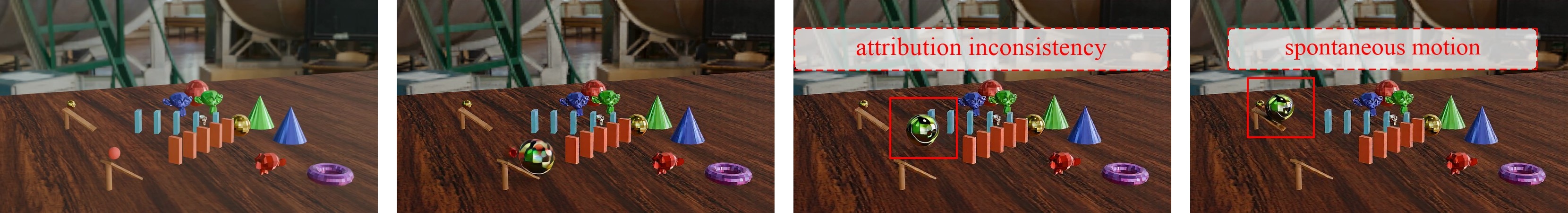}
\vspace{2pt}
{\fontsize{7.25}{8.5}\selectfont Wan2.1-I2V-14B. Wan2.1-I2V-14B exhibits a clear identity-corruption failure in the very first causal participant. The red ball descending from the left ramp does not preserve its appearance, but instead transforms into a much larger glossy sphere with a greenish appearance, indicating that the model fails to maintain stable object identity across time. The rollout also contains a strong physical-law inconsistency: after leaving the ramp, the ball should continue forward along the expected rightward path toward the domino row, but instead moves anomalously toward the upper-left direction. From the causal perspective, this both changes the identity of the trigger object and breaks the intended ball–domino contact event, preventing the downstream cascade from being realized through the reference pathway. Overall, the sequence is most consistent with unstable entity-state persistence together with inaccurate trajectory-sensitive physical dynamics.\par}

\vspace{3pt}
\hrule
\vspace{3pt}
\refstepcounter{qualcase}\label{case:b06-m08}
{\fontsize{8}{9}\selectfont\bfseries Case~\thequalcase\enspace\textperiodcentered\enspace Wan2.2-TI2V-5B}\par
\vspace{1pt}
\qstrip{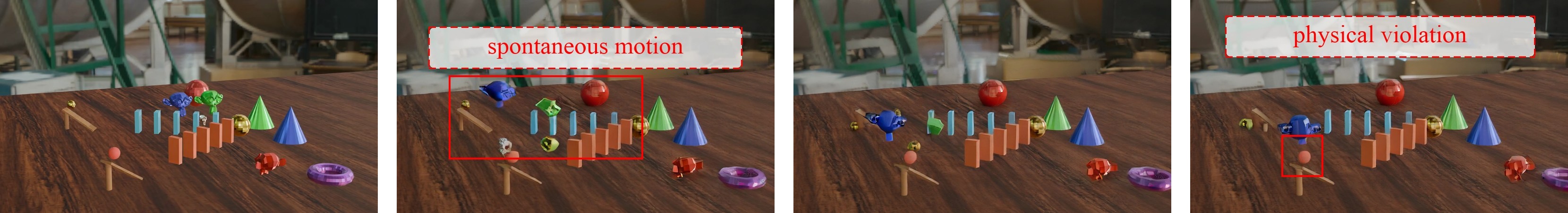}
\vspace{2pt}
{\fontsize{7.25}{8.5}\selectfont Wan2.2-TI2V-5B. Wan2.2-TI2V-5B exhibits a combined failure of entity persistence, unsupported dynamics, and basic physical plausibility. Several scene objects do not preserve stable identity, but instead change shape and appearance over time, indicating clear identity and topology corruption. At the same time, multiple objects undergo spontaneous motion in midair and disperse in unsupported directions, producing a series of spurious events that are not grounded in the prompt or in any preceding physical interaction. Most importantly, the red ball on the left ramp does not follow the expected gravity-driven descent, but fails to roll down along the ramp, yielding a clear physical-law inconsistency and leaving the intended trigger event unrealized. From the causal perspective, this means that the reference cascade is disrupted at its initiation, while the subsequent airborne motions introduce unrelated branches that further detach the rollout from the intended contact-mediated process. Overall, the sequence is most consistent with unstable entity-state persistence together with inaccurate gravity-sensitive dynamics and weak causal grounding of downstream motion.\par}

\vspace{3pt}
\hrule
\vspace{3pt}
\refstepcounter{qualcase}\label{case:b06-m09}
{\fontsize{8}{9}\selectfont\bfseries Case~\thequalcase\enspace\textperiodcentered\enspace Wan2.2-I2V-A14B}\par
\vspace{1pt}
\qstrip{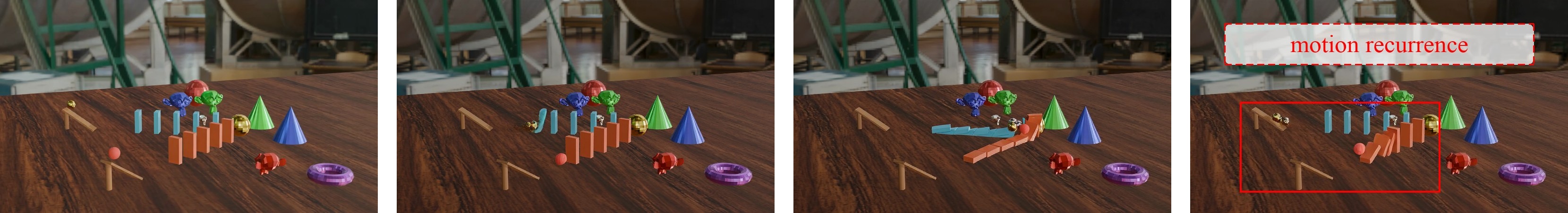}
\vspace{2pt}
{\fontsize{7.25}{8.5}\selectfont Wan2.2-I2V-A14B. Wan2.2-I2V-A14B performs substantially better than the other models in this scene and reproduces most of the intended causal pathway: the ball rolls down the ramp, follows the expected route, and successfully triggers the domino sequence through successive contacts. The main remaining causal error occurs near the downstream end of the cascade, where the falling dominoes fail to impart the required motion to the final ball, producing a broken propagation event and leaving the terminal outcome incomplete. The rollout then introduces an additional temporal failure: after the dominoes have already fallen, several of them reverse their motion and return toward an upright configuration. This constitutes motion recurrence and partial state replay, causing the physical process to regress rather than remain in its completed terminal state. From the causal perspective, the model therefore preserves most of the correct event sequence and dependencies but fails at the final transmission step and at maintaining the irreversibility of the completed cascade. Overall, the behavior is most consistent with relatively strong causal-pathway realization, limited by imperfect contact-mediated propagation at the terminal interaction and a residual temporal-state persistence failure.\par}

\clearpage
\noindent
{\fontsize{10}{11.5}\selectfont\bfseries Level 2 · Trigger-aware prompt}\par
\vspace{0.5pt}
{\fontsize{8}{9}\selectfont\itshape Galton\_Series.horizontal\_galton\_board\_object\_targets}\par
\vspace{1pt}\hrule\vspace{3pt}
{\fontsize{8}{9}\selectfont\bfseries Ground Truth}\par
\vspace{1pt}
\qstrip{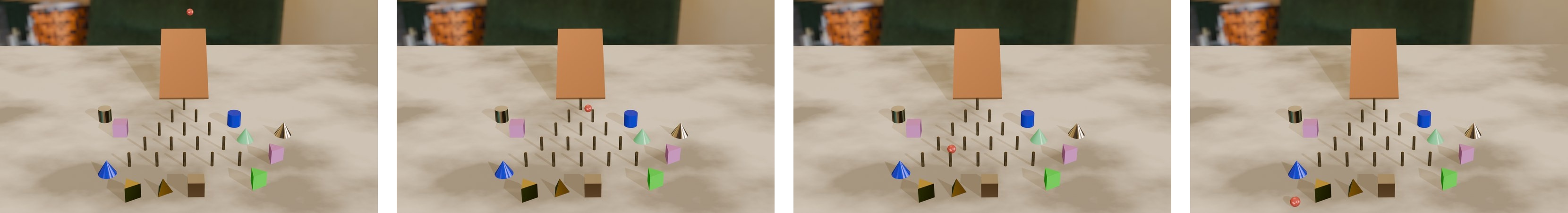}
\vspace{4pt}
\refstepcounter{qualcase}\label{case:b07-m01}
{\fontsize{8}{9}\selectfont\bfseries Case~\thequalcase\enspace\textperiodcentered\enspace CogVideoX-Fun-V1.5}\par
\vspace{1pt}
\qstrip{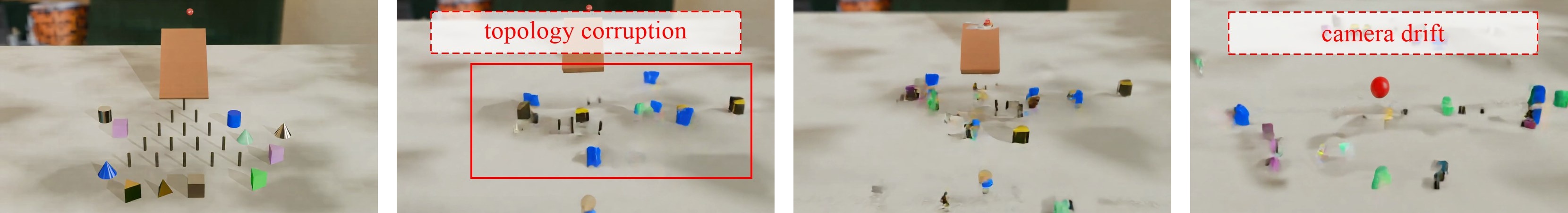}
\vspace{2pt}
{\fontsize{7.25}{8.5}\selectfont CogVideoX-Fun-V1.5. CogVideoX-Fun-V1.5 exhibits a severe breakdown of entity persistence, scene integrity, and viewpoint stability. Across the rollout, multiple objects fail to preserve their original shape and color, producing pronounced identity and topology corruption throughout the Galton-board scene. These distortions are accompanied by unsupported object motions, as several scene elements drift or move spontaneously in space without any valid upstream trigger, indicating that the generated dynamics are no longer grounded in the intended physical process. The failure further escalates through clear camera drift, which shifts the viewpoint and destabilizes the spatial frame needed to track the causal evolution consistently. From the causal perspective, the reference cascade is disrupted almost immediately: once the objects no longer maintain stable identities and the scene geometry collapses, the intended ball trajectory through the peg field cannot be faithfully realized, and the subsequent contact-mediated process is replaced by spurious motions and unsupported state changes. Overall, the sequence is most consistent with a severe failure of persistent entity/world-state representation, compounded by weak physical grounding and unstable viewpoint control.\par}

\vspace{3pt}
\hrule
\vspace{3pt}
\refstepcounter{qualcase}\label{case:b07-m02}
{\fontsize{8}{9}\selectfont\bfseries Case~\thequalcase\enspace\textperiodcentered\enspace Cosmos3-Nano}\par
\vspace{1pt}
\qstrip{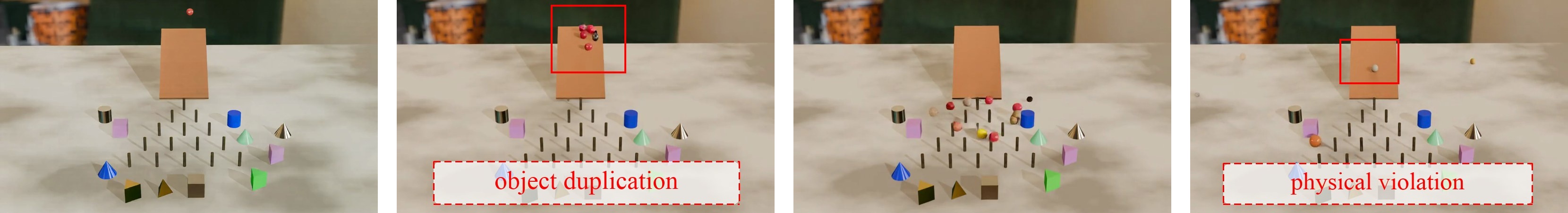}
\vspace{2pt}
{\fontsize{7.25}{8.5}\selectfont Cosmos3-Nano. Cosmos3-Nano exhibits a pronounced failure of entity persistence through uncontrolled object duplication. Instead of preserving a single ball, the model generates multiple additional spheres with varying appearances, creating numerous spurious participants throughout the rollout. Many of these duplicated balls nevertheless follow approximately gravity-consistent trajectories, descending through the Galton board and dispersing across different regions of the peg field. From the causal perspective, however, the intended single-particle pathway is replaced by multiple parallel and unsupported interaction branches, so the generated peg contacts no longer correspond to the reference causal process even when individual motions appear locally plausible. The rollout also contains a clear physical-law inconsistency: at least one ball remains suspended on the inclined upper board rather than continuing to descend under gravity. Overall, the sequence is most consistent with unstable entity-state persistence, combined with imperfect gravity-sensitive dynamics that allow some generated instances to follow plausible local motion while others fail to complete the expected descent.\par}

\vspace{3pt}
\hrule
\vspace{3pt}
\refstepcounter{qualcase}\label{case:b07-m03}
{\fontsize{8}{9}\selectfont\bfseries Case~\thequalcase\enspace\textperiodcentered\enspace HunyuanVideo-I2V}\par
\vspace{1pt}
\qstrip{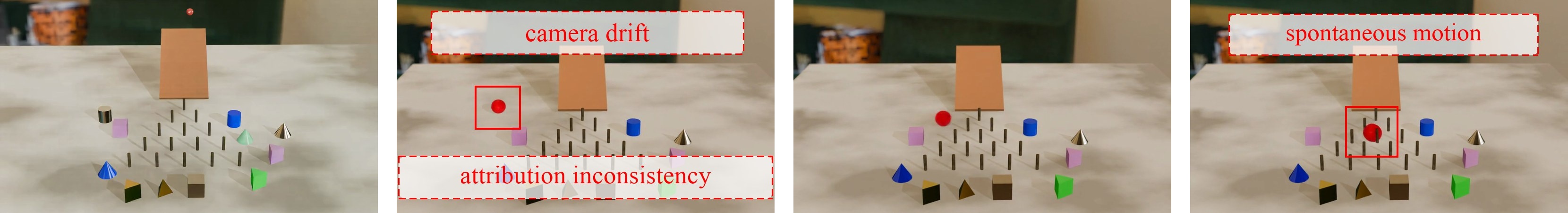}
\vspace{2pt}
{\fontsize{7.25}{8.5}\selectfont HunyuanVideo-I2V. HunyuanVideo-I2V exhibits a combination of viewpoint instability, entity-identity corruption, and physically inconsistent ball motion. Across the rollout, the camera progressively drifts relative to the scene, violating the fixed-view condition and changing the apparent spatial relationships within the Galton board. At the same time, the falling ball does not preserve a stable appearance but becomes noticeably larger, indicating a failure of entity-state persistence for the primary causal participant. Its trajectory is also incorrect: rather than continuing along the expected gravity-driven path through the peg field, the ball undergoes unsupported horizontal displacement while descending. From the causal perspective, the trigger event is only partially realized, because the ball enters the scene but no longer follows the contact-mediated trajectory required to reproduce the reference sequence of peg interactions. The identity change further weakens the grounding of subsequent events to a persistent object, while the camera drift destabilizes the spatial frame in which those interactions should be interpreted. Overall, the sequence is most consistent with unstable entity-state persistence, inaccurate trajectory-sensitive physical dynamics, and viewpoint-grounding failure.\par}

\clearpage
\noindent
{\fontsize{10}{11.5}\selectfont\bfseries Level 2 · Trigger-aware prompt}\par
\vspace{0.5pt}
{\fontsize{8}{9}\selectfont\itshape Galton\_Series.horizontal\_galton\_board\_object\_targets}\par
\vspace{1pt}\hrule\vspace{3pt}
{\fontsize{8}{9}\selectfont\bfseries Ground Truth}\par
\vspace{1pt}
\qstrip{appendix_e/assets/b07_m01_cogvideox_5b_i2v_gt_strip.jpg}
\vspace{4pt}
\refstepcounter{qualcase}\label{case:b07-m04}
{\fontsize{8}{9}\selectfont\bfseries Case~\thequalcase\enspace\textperiodcentered\enspace LTX-Video-13B}\par
\vspace{1pt}
\qstrip{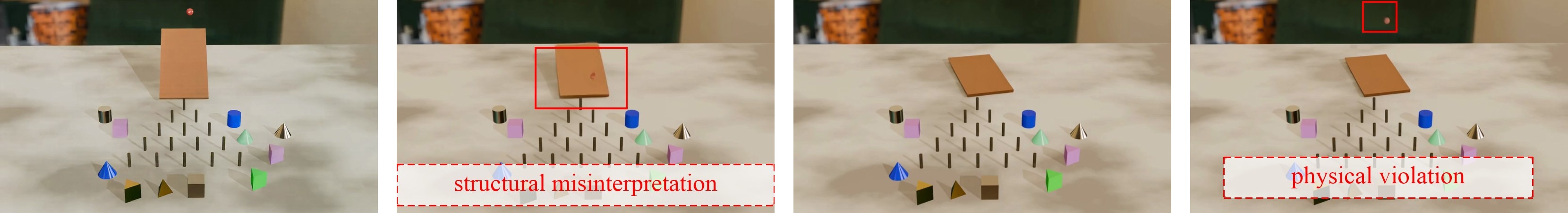}
\vspace{2pt}
{\fontsize{7.25}{8.5}\selectfont LTX-Video-13B. LTX-Video-13B first exhibits a structural misinterpretation of the upper board, treating it as if it were unsupported and allowing it to collapse even though the reference scene indicates that the structure should remain stable. This alters the physical configuration before the intended Galton-board trajectory can be faithfully realized. The ball then undergoes a second abnormal transition: after descending onto the board, it rebounds upward to nearly its original height and remains elevated rather than continuing to fall back under gravity. While a limited rebound could in principle be physically plausible, the magnitude and persistence of this upward motion are inconsistent with the visible interaction and therefore constitute a physical-law inconsistency. From the causal perspective, the incorrect board collapse introduces a spurious structural event and changes the interaction substrate, while the abnormal rebound prevents the ball from progressing through the intended sequence of peg-mediated contacts. Overall, the sequence is most consistent with an initial structural-inference failure, compounded by inaccurate gravity- and collision-sensitive physical dynamics.\par}

\vspace{3pt}
\hrule
\vspace{3pt}
\refstepcounter{qualcase}\label{case:b07-m05}
{\fontsize{8}{9}\selectfont\bfseries Case~\thequalcase\enspace\textperiodcentered\enspace MAGI-1-4.5B}\par
\vspace{1pt}
\qstrip{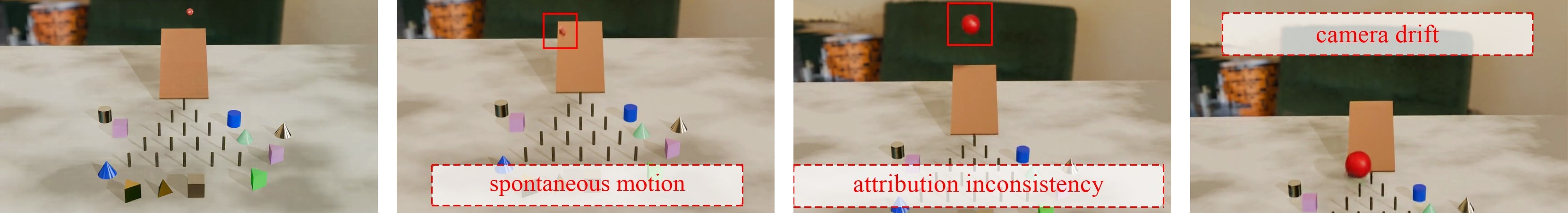}
\vspace{2pt}
{\fontsize{7.25}{8.5}\selectfont MAGI-1-4.5B. MAGI-1-4.5B exhibits a combination of incorrect ball dynamics, identity corruption, and viewpoint instability. Instead of following the expected gravity-driven descent through the Galton board, the ball first deviates laterally toward the left and then moves upward without any visible physical trigger, producing both a trajectory inconsistency and spontaneous motion. During this abnormal motion, the ball also changes substantially in appearance, becoming a much larger red sphere and therefore failing to preserve the identity of the original causal participant. From the causal perspective, the intended sequence of gravitational descent and successive peg-mediated redirections is disrupted at an early stage, so the generated trajectory no longer follows the reference interaction pathway. Although the enlarged ball later descends again, this occurs only after an unsupported upward excursion and therefore does not restore the correct causal history. The final part of the rollout additionally exhibits camera drift, further destabilizing the spatial grounding of the scene. Overall, the sequence is most consistent with inaccurate gravity- and trajectory-sensitive physical dynamics, unstable entity-state persistence, and viewpoint-grounding failure.\par}

\vspace{3pt}
\hrule
\vspace{3pt}
\refstepcounter{qualcase}\label{case:b07-m06}
{\fontsize{8}{9}\selectfont\bfseries Case~\thequalcase\enspace\textperiodcentered\enspace SkyReels-V2-I2V}\par
\vspace{1pt}
\qstrip{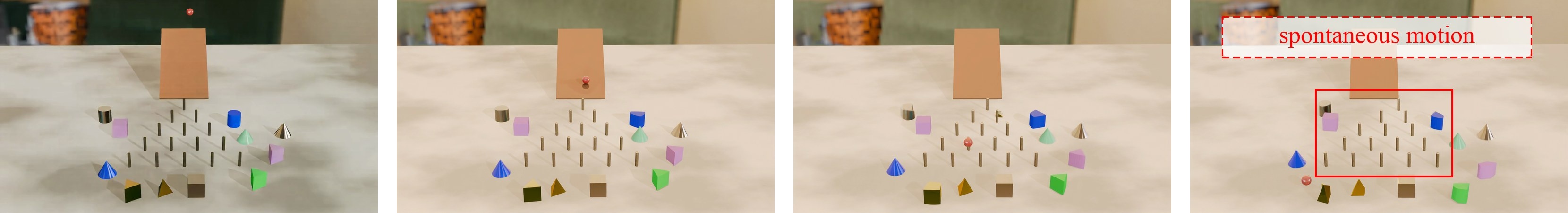}
\vspace{2pt}
{\fontsize{7.25}{8.5}\selectfont SkyReels-V2-I2V. SkyReels-V2-I2V performs relatively well in this Galton-board scene, successfully preserving the main gravity-driven trajectory of the ball as it descends through the peg field and is redirected toward the left in a manner broadly consistent with the Ground Truth. The primary remaining failure is a mild instability of the static scene structure: several pegs undergo small horizontal displacements even though they should remain fixed throughout the rollout. From the causal perspective, the main ball-motion pathway is largely preserved, but these peg movements perturb the physical substrate that is supposed to mediate the sequence of ball–peg interactions, introducing small spurious state changes that are absent from the reference process. Because the ball trajectory itself remains mostly coherent, this case represents a comparatively minor failure rather than a breakdown of the causal cascade. Overall, the sequence is most consistent with slight structural-state instability in otherwise largely correct contact-mediated dynamics.\par}

\clearpage
\noindent
{\fontsize{10}{11.5}\selectfont\bfseries Level 2 · Trigger-aware prompt}\par
\vspace{0.5pt}
{\fontsize{8}{9}\selectfont\itshape Galton\_Series.horizontal\_galton\_board\_object\_targets}\par
\vspace{1pt}\hrule\vspace{3pt}
{\fontsize{8}{9}\selectfont\bfseries Ground Truth}\par
\vspace{1pt}
\qstrip{appendix_e/assets/b07_m01_cogvideox_5b_i2v_gt_strip.jpg}
\vspace{4pt}
\refstepcounter{qualcase}\label{case:b07-m07}
{\fontsize{8}{9}\selectfont\bfseries Case~\thequalcase\enspace\textperiodcentered\enspace Wan2.1-I2V-14B}\par
\vspace{1pt}
\qstrip{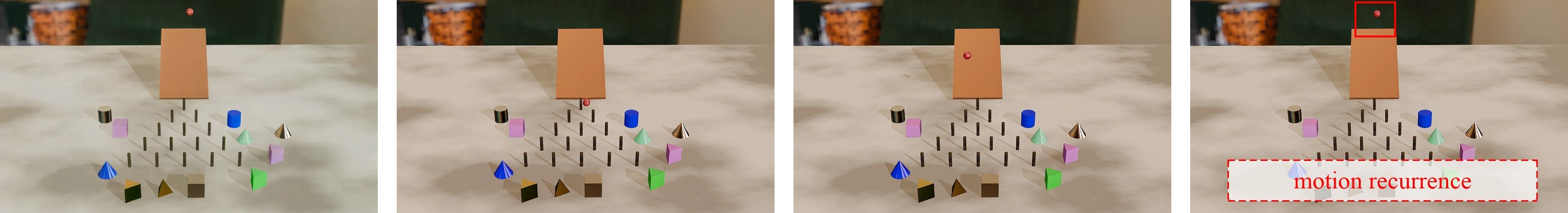}
\vspace{2pt}
{\fontsize{7.25}{8.5}\selectfont Wan2.1-I2V-14B. In this example, Wan2.1-I2V-14B partially realizes the initial event by letting the ball descend toward the board, but it fails to sustain the correct physical progression afterward. The main manifestation is motion recurrence accompanied by a physical-law violation: after falling downward, the ball rises back upward to a position close to its earlier state, creating an implausible replay-like trajectory rather than continuing through the Galton board as in the Ground Truth. From the causal-process perspective, the initial drop event is only partially realized, but the subsequent propagation through the board is not completed; instead, the sequence regresses to an earlier state, producing a breakdown in causal-temporal progression and preventing the expected downstream interactions with the peg field. This means the failure is not primarily one of object identity or scene corruption, but rather a temporal-dynamics error in which the model cannot maintain a coherent forward evolution of the event. Overall, this case is best characterized as a failure of physical dynamics and temporal state persistence, leading to an unphysical upward return and a repeated-motion pattern.\par}

\vspace{3pt}
\hrule
\vspace{3pt}
\refstepcounter{qualcase}\label{case:b07-m08}
{\fontsize{8}{9}\selectfont\bfseries Case~\thequalcase\enspace\textperiodcentered\enspace Wan2.2-TI2V-5B}\par
\vspace{1pt}
\qstrip{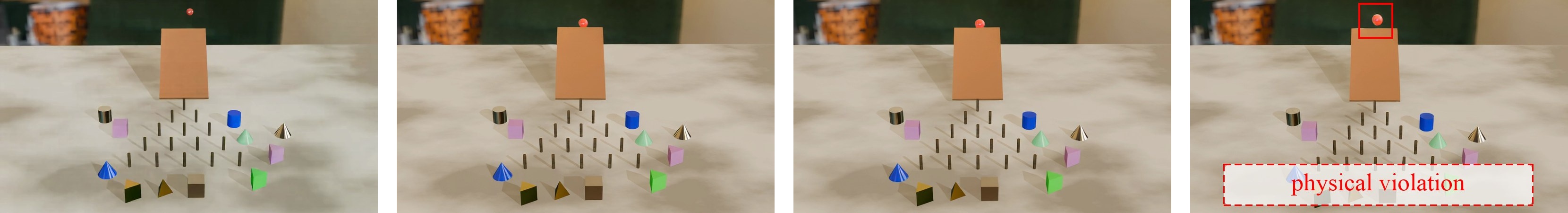}
\vspace{2pt}
{\fontsize{7.25}{8.5}\selectfont Wan2.2-TI2V-5B. Wan2.2-TI2V-5B primarily fails to realize the initial gravity-driven motion of the ball. Instead of descending from the upper release position and entering the Galton board, the ball remains suspended near the top of the scene throughout the rollout, producing a clear physical-law inconsistency in the absence of any visible support or upward force. From the causal perspective, this constitutes a missing trigger: because the initial descent never occurs, none of the subsequent ball–peg collisions or contact-mediated trajectory redirections can be realized along the reference pathway, and the ball therefore never progresses toward the lower-left exit. The generated sequence consequently fails at the very first stage of the causal cascade rather than during downstream propagation. Overall, the behavior is most consistent with an inaccurate realization of gravity-sensitive physical dynamics, which prevents the model from initiating the reference process at all.\par}

\vspace{3pt}
\hrule
\vspace{3pt}
\refstepcounter{qualcase}\label{case:b07-m09}
{\fontsize{8}{9}\selectfont\bfseries Case~\thequalcase\enspace\textperiodcentered\enspace Wan2.2-I2V-A14B}\par
\vspace{1pt}
\qstrip{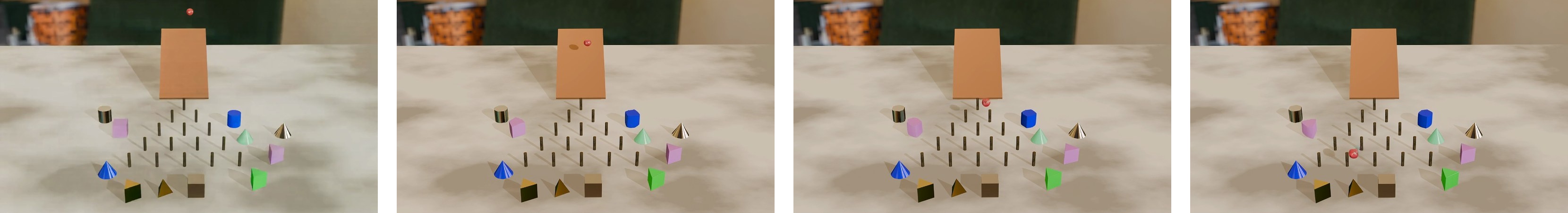}
\vspace{2pt}
{\fontsize{7.25}{8.5}\selectfont Wan2.2-I2V-A14B. Wan2.2-I2V-A14B performs very well in this example and largely reproduces the Ground Truth physical process. The ball descends through the Galton board along a qualitatively correct trajectory, and the sequence of interactions remains consistent with the intended contact-mediated pathway. The main remaining discrepancy is temporal rather than causal: the ball appears to descend more slowly than in the reference rollout, shifting the timing of the intermediate states while preserving the overall event sequence and spatial progression. From the causal perspective, the relevant events and dependencies are therefore largely maintained, with only a mild delay in causal-temporal progression rather than a failure of event realization or physical interaction. Overall, this case represents a near-correct rollout in which the primary error is a small temporal misalignment, likely reflecting imperfect control of motion speed rather than a substantive misunderstanding of the scene dynamics.\par}

\clearpage
\noindent
{\fontsize{10}{11.5}\selectfont\bfseries Level 3 · Full event-chain prompt}\par
\vspace{0.5pt}
{\fontsize{8}{9}\selectfont\itshape Domino\_Series.ball\_trigger\_step\_domino\_chain}\par
\vspace{1pt}\hrule\vspace{3pt}
{\fontsize{8}{9}\selectfont\bfseries Ground Truth}\par
\vspace{1pt}
\qstrip{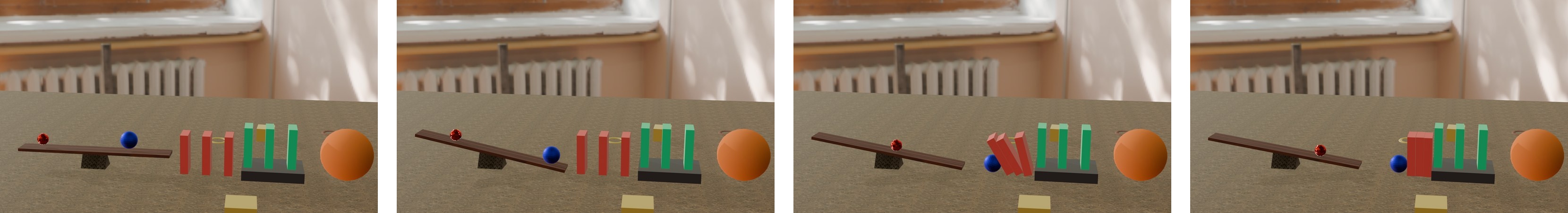}
\vspace{4pt}
\refstepcounter{qualcase}\label{case:b08-m01}
{\fontsize{8}{9}\selectfont\bfseries Case~\thequalcase\enspace\textperiodcentered\enspace CogVideoX-Fun-V1.5}\par
\vspace{1pt}
\qstrip{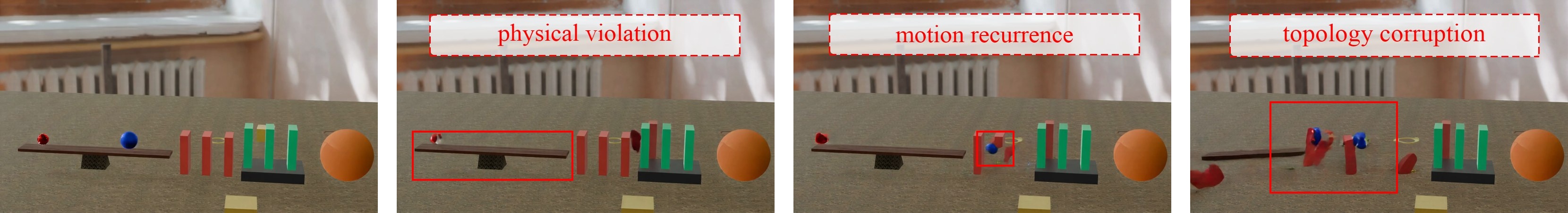}
\vspace{2pt}
{\fontsize{7.25}{8.5}\selectfont CogVideoX-Fun-V1.5. In this example, CogVideoX-Fun-V1.5 partially captures the intended rightward motion of the blue ball, but it fails to produce the correct physical response of the seesaw system. Although the blue ball moves to the right, the red ball remains near its original position and the seesaw unrealistically stays almost horizontal, rather than tilting into the expected left-high, right-low configuration. The rollout then develops a motion recurrence failure: after moving past the domino region, the blue ball reverses direction and bends back leftward, reappearing around the middle of the domino chain instead of continuing the causal progression. In the final frame, both the seesaw and the dominoes undergo clear topology corruption, with their shapes and geometric structure becoming distorted and inconsistent with the original objects. From the perspective of the causal chain, the model therefore breaks the scene at the first key interaction, because it fails to translate the ball’s displacement into the correct support and balance dynamics of the seesaw; this in turn destabilizes the downstream event sequence and leads to a spurious replay-like motion pattern. A plausible mechanistic origin is that the model can recognize a salient local motion cue—namely, the blue ball moving rightward—but fails to maintain a coherent representation of object coupling, rigid-body structure, and state transition consistency over time.\par}

\vspace{3pt}
\hrule
\vspace{3pt}
\refstepcounter{qualcase}\label{case:b08-m02}
{\fontsize{8}{9}\selectfont\bfseries Case~\thequalcase\enspace\textperiodcentered\enspace Cosmos3-Nano}\par
\vspace{1pt}
\qstrip{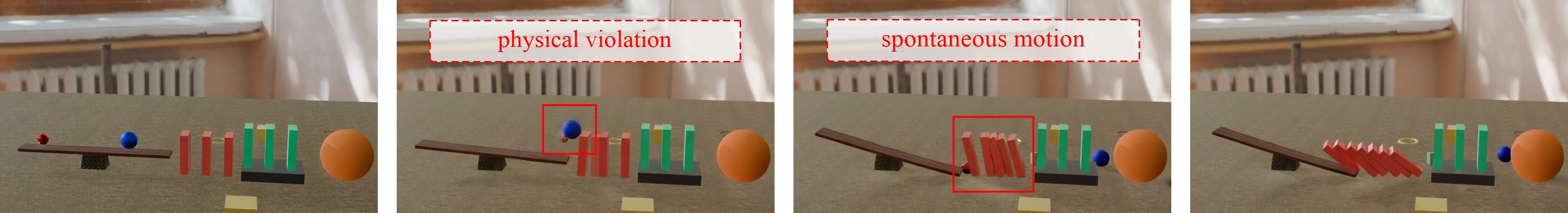}
\vspace{2pt}
{\fontsize{7.25}{8.5}\selectfont Cosmos3-Nano. In this example, the model initially captures the coarse early dynamics correctly: the red and blue balls both move rightward, and the seesaw correspondingly tilts into a left-high, right-low configuration. However, once the balls approach the domino region, the simulation breaks down. Instead of continuing through a plausible grounded collision, the balls abruptly jump upward, which is a clear physical violation. The dominoes then begin to topple leftward without a valid contact-based trigger, constituting an additional spontaneous motion error. These failures disrupt the intended causal chain: rather than the moving balls transferring force into the domino line through a physically consistent impact, the cascade is replaced by unsupported upward motion and an incorrectly induced domino fall. A plausible mechanistic origin is that the model captures the high-level event sequence—ball motion followed by domino activity—but fails to maintain local contact dynamics and force transmission, causing it to hallucinate the downstream cascade instead of simulating it faithfully.\par}

\vspace{3pt}
\hrule
\vspace{3pt}
\refstepcounter{qualcase}\label{case:b08-m03}
{\fontsize{8}{9}\selectfont\bfseries Case~\thequalcase\enspace\textperiodcentered\enspace HunyuanVideo-I2V}\par
\vspace{1pt}
\qstrip{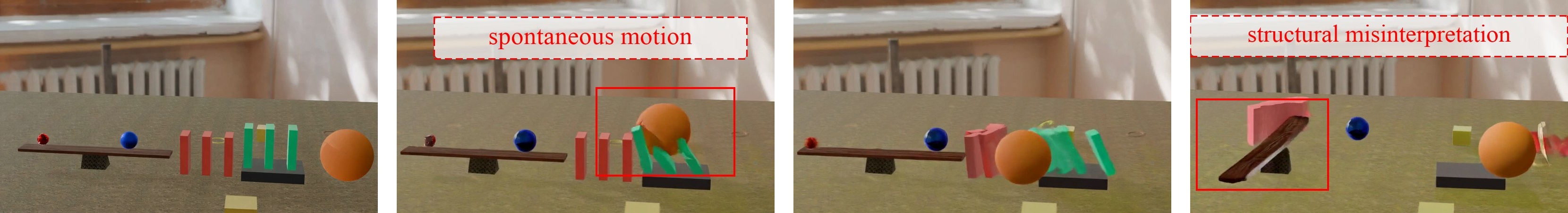}
\vspace{2pt}
{\fontsize{7.25}{8.5}\selectfont HunyuanVideo-I2V. HunyuanVideo-I2V exhibits failures in both the initiation of the intended dynamics and the interpretation of the seesaw structure. At the beginning of the rollout, neither the seesaw nor the two balls resting on it undergo the expected motion, leaving the reference trigger dynamics unrealized. Instead, the large orange ball on the far right moves spontaneously toward the left and directly disturbs the dominoes, producing a spurious downstream event and effectively bypassing the intended ball–seesaw–domino interaction pathway. From the causal perspective, the model therefore replaces the reference cascade with an unsupported causal branch initiated by an otherwise inactive object. A further structural error appears in the final frame, where the seesaw undergoes an excessively large rotation about its support, as if the support were an unconstrained hinge permitting free rotation rather than a fulcrum that constrains the plank to the expected seesaw-like tilting motion. Overall, the sequence is most consistent with a combination of missing initial physical dynamics, weak causal-dependency preservation, and structural misinterpretation of the seesaw’s support and rotational constraints.\par}

\clearpage
\noindent
{\fontsize{10}{11.5}\selectfont\bfseries Level 3 · Full event-chain prompt}\par
\vspace{0.5pt}
{\fontsize{8}{9}\selectfont\itshape Domino\_Series.ball\_trigger\_step\_domino\_chain}\par
\vspace{1pt}\hrule\vspace{3pt}
{\fontsize{8}{9}\selectfont\bfseries Ground Truth}\par
\vspace{1pt}
\qstrip{appendix_e/assets/b08_m01_cogvideox_5b_i2v_gt_strip.jpg}
\vspace{4pt}
\refstepcounter{qualcase}\label{case:b08-m04}
{\fontsize{8}{9}\selectfont\bfseries Case~\thequalcase\enspace\textperiodcentered\enspace LTX-Video-13B}\par
\vspace{1pt}
\qstrip{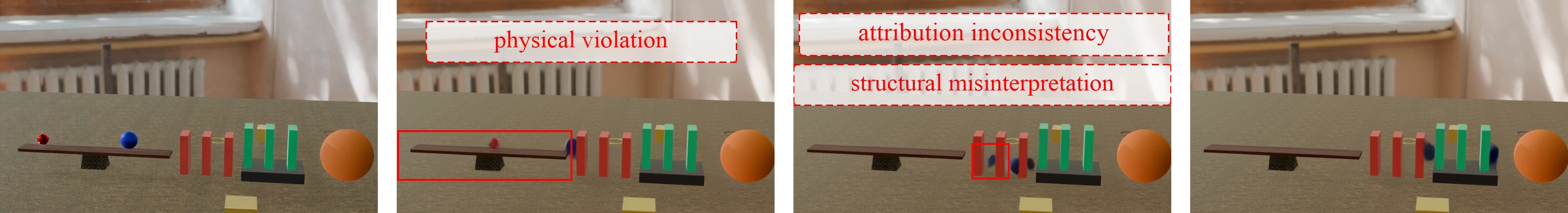}
\vspace{2pt}
{\fontsize{7.25}{8.5}\selectfont LTX-Video-13B. LTX-Video-13B partially captures the intended object motion, in that both balls move rightward as expected, but it fails to produce the corresponding physical response of the seesaw. In the ground-truth dynamics, the shifting mass should cause the plank to tilt, creating a clear imbalance over time. Instead, the generated seesaw remains essentially level, indicating a violation of the underlying mechanics. The sequence also shows attribution inconsistency: as the ball passes the domino region, the originally red ball appears to change into a blue one, so the object identity is not preserved across frames. In addition, the model exhibits a structural or spatial misinterpretation of the scene. Rather than colliding with and triggering the domino chain, the balls appear to pass behind the dominoes, as if the model had misread their spatial arrangement and failed to enforce the intended contact interaction. Overall, the main failure modes here are incorrect physical response, object identity inconsistency, and erroneous spatial understanding of the ball–domino interaction.\par}

\vspace{3pt}
\hrule
\vspace{3pt}
\refstepcounter{qualcase}\label{case:b08-m05}
{\fontsize{8}{9}\selectfont\bfseries Case~\thequalcase\enspace\textperiodcentered\enspace MAGI-1-4.5B}\par
\vspace{1pt}
\qstrip{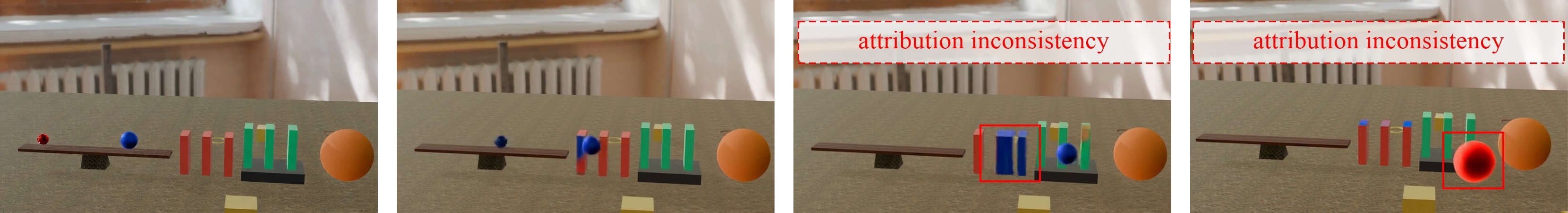}
\vspace{2pt}
{\fontsize{7.25}{8.5}\selectfont MAGI-1-4.5B. MAGI-1-4.5B exhibits pronounced identity corruption together with an incorrect interpretation of the ball–domino geometry. In the reference scene, the first domino row is red, whereas in the generated rollout these dominoes progressively change to blue, indicating that their visual identities are not preserved across time. The blue ball itself also undergoes a substantial identity change: after moving rightward past the domino region, it transforms into a much larger orange sphere. More importantly, the model fails to realize the intended spatial interaction between the ball and the dominoes. Rather than colliding with and toppling the red domino row, the ball appears to pass through it, leaving the expected contact event unrealized. From the causal perspective, the initial rightward ball motion is present, but the critical ball–domino mediator is missing, thereby breaking propagation toward the downstream domino cascade. Overall, the sequence is most consistent with unstable entity-state persistence combined with a structural/spatial misinterpretation of the relative depth and collision geometry of the ball and dominoes.\par}

\vspace{3pt}
\hrule
\vspace{3pt}
\refstepcounter{qualcase}\label{case:b08-m06}
{\fontsize{8}{9}\selectfont\bfseries Case~\thequalcase\enspace\textperiodcentered\enspace SkyReels-V2-I2V}\par
\vspace{1pt}
\qstrip{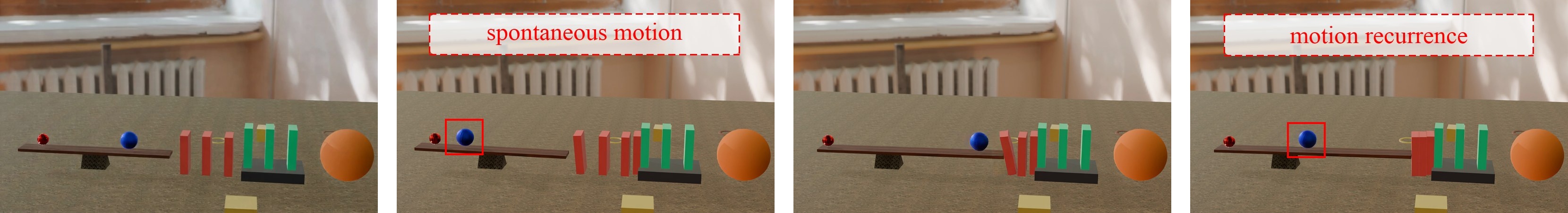}
\vspace{2pt}
{\fontsize{7.25}{8.5}\selectfont SkyReels-V2-I2V. SkyReels-V2-I2V exhibits a clear physical violation at the very beginning of the rollout: instead of continuing to roll rightward as required by the scene dynamics, the blue ball moves leftward along the seesaw. The sequence also shows motion recurrence, since the ball later reappears in a repeated or reversed state rather than following a single forward causal progression. In addition, the dominoes do not respond through the correct contact-driven interaction; instead, they undergo an incorrect displacement, suggesting spurious object motion rather than a physically grounded collision. The seesaw itself is also not stably preserved: its geometry appears to change and even become elongated, indicating topology corruption / structural inconsistency. Overall, the main failure modes here are incorrect motion direction, repeated motion, spurious domino movement, and deformation of the seesaw structure.\par}

\clearpage
\noindent
{\fontsize{10}{11.5}\selectfont\bfseries Level 3 · Full event-chain prompt}\par
\vspace{0.5pt}
{\fontsize{8}{9}\selectfont\itshape Domino\_Series.ball\_trigger\_step\_domino\_chain}\par
\vspace{1pt}\hrule\vspace{3pt}
{\fontsize{8}{9}\selectfont\bfseries Ground Truth}\par
\vspace{1pt}
\qstrip{appendix_e/assets/b08_m01_cogvideox_5b_i2v_gt_strip.jpg}
\vspace{4pt}
\refstepcounter{qualcase}\label{case:b08-m07}
{\fontsize{8}{9}\selectfont\bfseries Case~\thequalcase\enspace\textperiodcentered\enspace Wan2.1-I2V-14B}\par
\vspace{1pt}
\qstrip{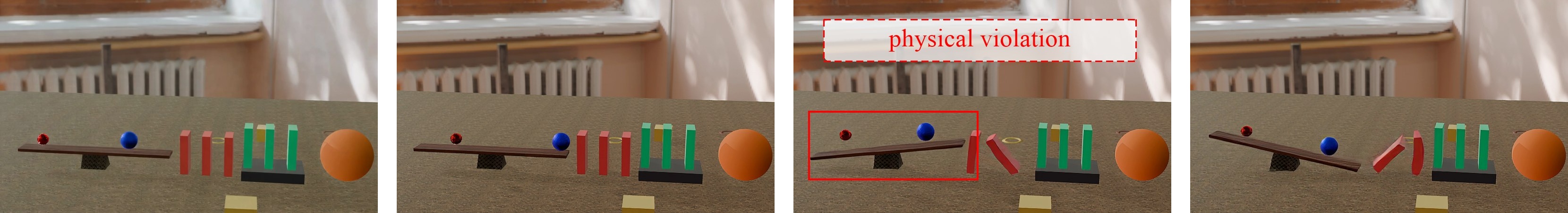}
\vspace{2pt}
{\fontsize{7.25}{8.5}\selectfont Wan2.1-I2V-14B. Wan2.1-I2V-14B exhibits a clear physical violation and a breakdown of the intended causal chain. Although in the second frame both balls move rightward along the designed trajectory, the seesaw does not respond correctly: it should become left-high and right-low, but instead remains nearly level. The subsequent motion is even less consistent with the scene physics: both balls reverse direction and move back to the left, one ball appears to jump upward, and the seesaw flips into a left-low, right-high configuration, all of which contradict the expected dynamics. In addition, the dominoes on the right topple without being properly struck, indicating spontaneous motion rather than contact-driven interaction. While the final frame shows a locally plausible rightward motion of the balls and a more reasonable seesaw tilt, the overall rollout is still physically inconsistent and does not preserve a coherent forward causal progression.\par}

\vspace{3pt}
\hrule
\vspace{3pt}
\refstepcounter{qualcase}\label{case:b08-m08}
{\fontsize{8}{9}\selectfont\bfseries Case~\thequalcase\enspace\textperiodcentered\enspace Wan2.2-TI2V-5B}\par
\vspace{1pt}
\qstrip{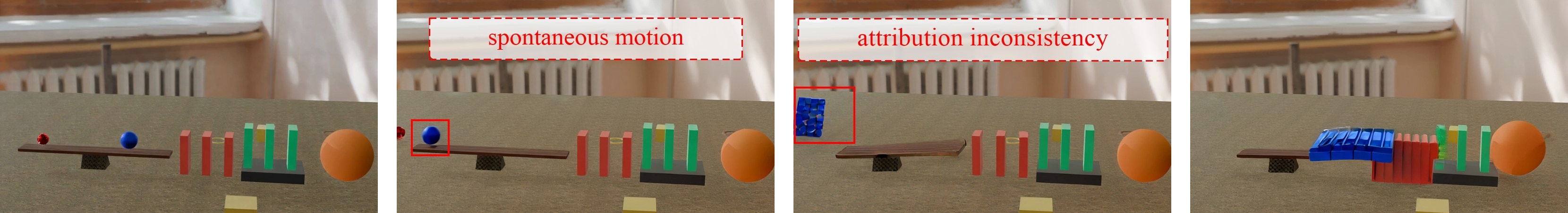}
\vspace{2pt}
{\fontsize{7.25}{8.5}\selectfont Wan2.2-TI2V-5B. Wan2.2-TI2V-5B shows severe failure in both physical reasoning and object identity consistency. Instead of following the intended causal chain, the two balls on the seesaw drift leftward on their own, which is inconsistent with the expected rightward progression and therefore constitutes clear spontaneous motion. At the same time, the blue ball does not preserve its identity: it first transforms into a blue cube-like object, and then further changes into something resembling a horizontally collapsed domino-like block sequence that continues moving to the right. This indicates strong attribution inconsistency, together with substantial deformation of the object’s shape and role in the scene. In addition, the seesaw itself also changes form, suggesting a breakdown in structural consistency. Overall, the rollout fails to maintain stable object identity, violates the intended physics, and does not preserve a coherent causal progression.\par}

\vspace{3pt}
\hrule
\vspace{3pt}
\refstepcounter{qualcase}\label{case:b08-m09}
{\fontsize{8}{9}\selectfont\bfseries Case~\thequalcase\enspace\textperiodcentered\enspace Wan2.2-I2V-A14B}\par
\vspace{1pt}
\qstrip{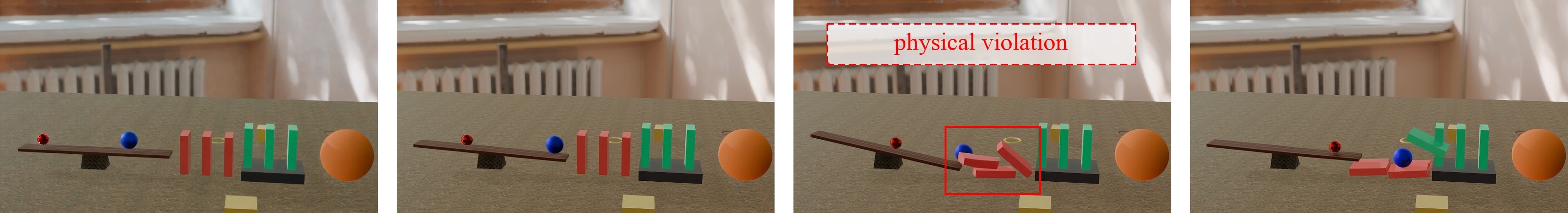}
\vspace{2pt}
{\fontsize{7.25}{8.5}\selectfont Wan2.2-I2V-A14B. Wan2.2-I2V-A14B performs relatively well in the early stage of the sequence: the seesaw correctly becomes left-high and right-low, and the two balls both roll rightward as expected. The main failure occurs at the domino stage, where the generated motion violates the intended physics. Instead of being knocked down to the right by the incoming causal interaction, the dominoes topple to the left, reversing the expected direction of the cascade. Thus, although the initial motion is largely plausible, the model ultimately exhibits a clear physical violation in the domino-toppling dynamics.\par}

\clearpage
\noindent
{\fontsize{10}{11.5}\selectfont\bfseries Level 3 · Full event-chain prompt}\par
\vspace{0.5pt}
{\fontsize{8}{9}\selectfont\itshape Domino\_Series.dual\_ball\_inverted\_y\_domino\_course}\par
\vspace{1pt}\hrule\vspace{3pt}
{\fontsize{8}{9}\selectfont\bfseries Ground Truth}\par
\vspace{1pt}
\qstrip{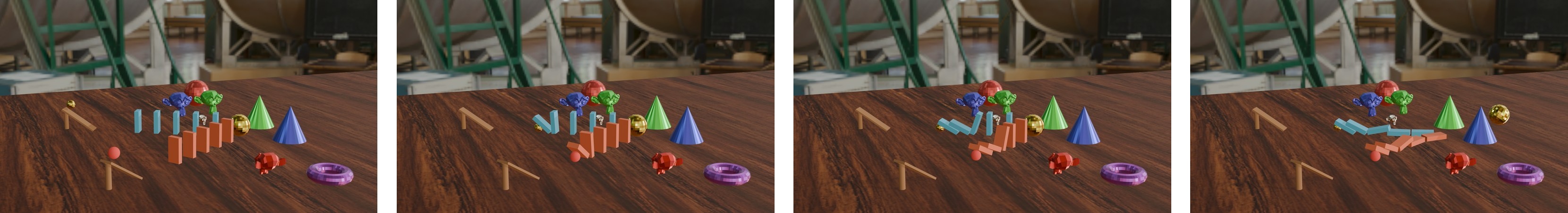}
\vspace{4pt}
\refstepcounter{qualcase}\label{case:b09-m01}
{\fontsize{8}{9}\selectfont\bfseries Case~\thequalcase\enspace\textperiodcentered\enspace CogVideoX-Fun-V1.5}\par
\vspace{1pt}
\qstrip{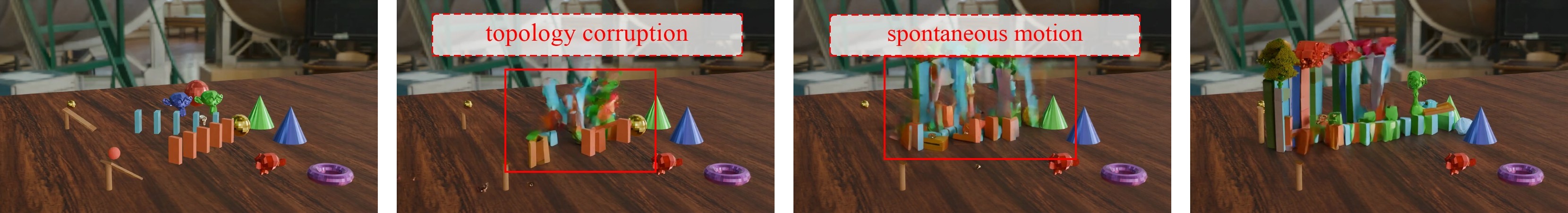}
\vspace{2pt}
{\fontsize{7.25}{8.5}\selectfont CogVideoX-Fun-V1.5. CogVideoX-Fun-V1.5 exhibits a severe collapse of entity and scene consistency almost immediately after generation begins. Multiple objects in the central region change their shapes, colors, and structural appearances, producing pronounced identity corruption together with extensive topology corruption. These deformed entities then undergo large spontaneous motions without any valid upstream trigger, causing the scene to evolve through unsupported and highly disordered dynamics rather than through the intended ramp–domino causal process. From the causal perspective, the failure first destroys the persistent physical substrate required to ground the reference events, after which numerous spurious motion events and causal branches emerge in place of the intended contact-mediated propagation. As the original objects can no longer be reliably tracked, the reference causal cascade effectively becomes unrecoverable. Overall, the sequence is most consistent with a severe failure of persistent entity/world-state representation, compounded by unstable dynamics once the underlying scene structure has already collapsed.\par}

\vspace{3pt}
\hrule
\vspace{3pt}
\refstepcounter{qualcase}\label{case:b09-m02}
{\fontsize{8}{9}\selectfont\bfseries Case~\thequalcase\enspace\textperiodcentered\enspace Cosmos3-Nano}\par
\vspace{1pt}
\qstrip{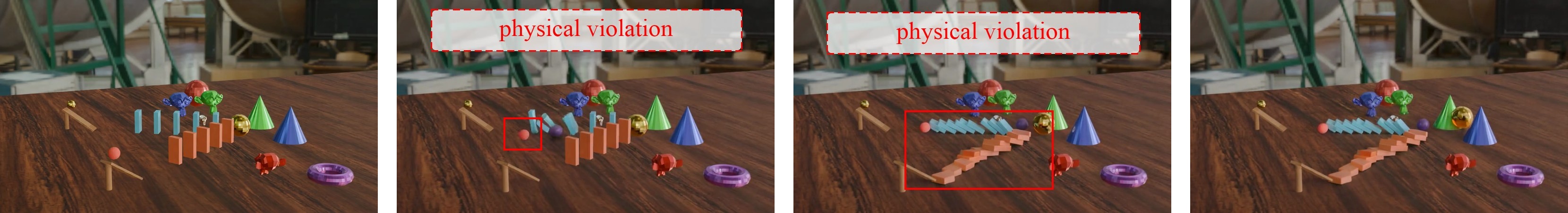}
\vspace{2pt}
{\fontsize{7.25}{8.5}\selectfont Cosmos3-Nano. Cosmos3-Nano exhibits a clear physical violation in the early stage of the sequence. The orange ball on the lower-left ramp should roll forward along the ramp and continue along the intended straight trajectory, but instead it moves upward and to the right, which is inconsistent with the scene geometry and the expected effect of gravity. A second physical failure occurs in the domino chain: the dominoes should be knocked over to the right only after being contacted by the incoming ball, yet in the generated sequence they instead tip leftward without the correct triggering interaction. Although the final collapsed configuration partially resembles the reference outcome, the intermediate causal process is incorrect. In other words, the model reaches a superficially similar end state through an invalid causal shortcut, replacing the intended contact-driven chain reaction with spontaneous motion and physically implausible dynamics.\par}

\vspace{3pt}
\hrule
\vspace{3pt}
\refstepcounter{qualcase}\label{case:b09-m03}
{\fontsize{8}{9}\selectfont\bfseries Case~\thequalcase\enspace\textperiodcentered\enspace HunyuanVideo-I2V}\par
\vspace{1pt}
\qstrip{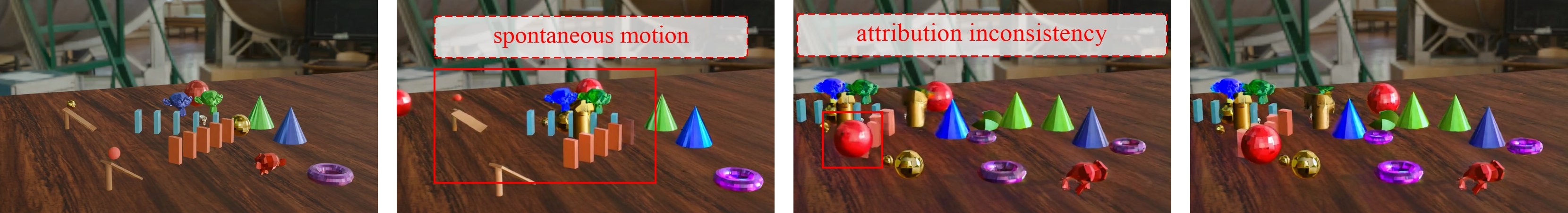}
\vspace{2pt}
{\fontsize{7.25}{8.5}\selectfont HunyuanVideo-I2V. HunyuanVideo-I2V shows two major failure modes in this sequence: spontaneous motion and attribution inconsistency. Instead of preserving the original causal chain, the red ball on the lower-left ramp and several other objects in the scene begin to move on their own, without being properly triggered by contact or gravity. At the same time, object identity is not maintained: multiple items change their appearance over time, including noticeable shifts in color, size, and shape. These changes affect not only the rolling ball but also other scene elements, making the generated dynamics highly unstable and inconsistent with the ground-truth setup. Overall, the model fails both to preserve object identity and to maintain a physically plausible, causally grounded evolution of the scene.\par}

\clearpage
\noindent
{\fontsize{10}{11.5}\selectfont\bfseries Level 3 · Full event-chain prompt}\par
\vspace{0.5pt}
{\fontsize{8}{9}\selectfont\itshape Domino\_Series.dual\_ball\_inverted\_y\_domino\_course}\par
\vspace{1pt}\hrule\vspace{3pt}
{\fontsize{8}{9}\selectfont\bfseries Ground Truth}\par
\vspace{1pt}
\qstrip{appendix_e/assets/b09_m01_cogvideox_5b_i2v_gt_strip.jpg}
\vspace{4pt}
\refstepcounter{qualcase}\label{case:b09-m04}
{\fontsize{8}{9}\selectfont\bfseries Case~\thequalcase\enspace\textperiodcentered\enspace LTX-Video-13B}\par
\vspace{1pt}
\qstrip{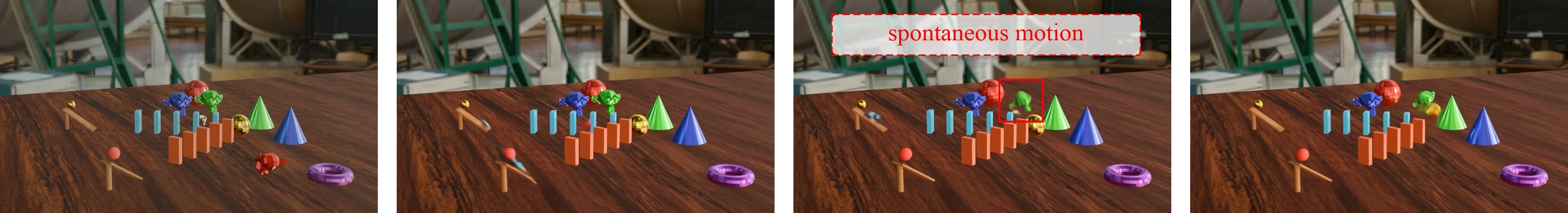}
\vspace{2pt}
{\fontsize{7.25}{8.5}\selectfont LTX-Video-13B. LTX-Video-13B exhibits two main failure modes in this example: spontaneous motion and physical violation. Among the three objects in the center, the rightmost one undergoes an unprompted motion even though it is not causally triggered by any preceding interaction. In addition, the generated dynamics violate basic physical expectations: both balls remain on their respective ramps instead of rolling downward under gravity, so the intended causal process never properly begins. As a result, the scene fails both in physical plausibility and in maintaining a correct cause-and-effect progression.\par}

\vspace{3pt}
\hrule
\vspace{3pt}
\refstepcounter{qualcase}\label{case:b09-m05}
{\fontsize{8}{9}\selectfont\bfseries Case~\thequalcase\enspace\textperiodcentered\enspace MAGI-1-4.5B}\par
\vspace{1pt}
\qstrip{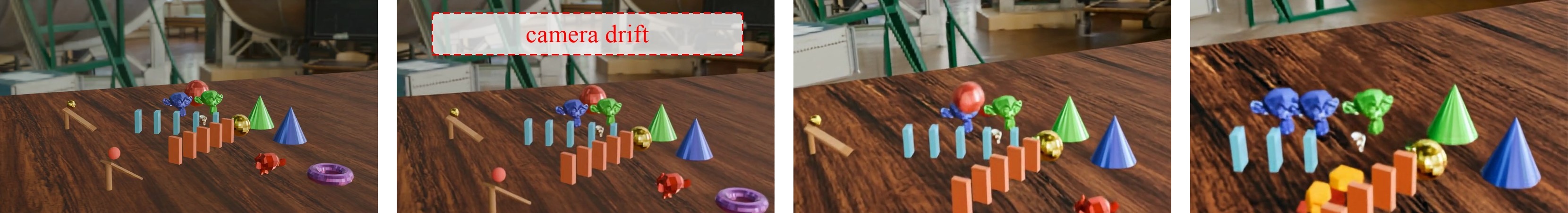}
\vspace{2pt}
{\fontsize{7.25}{8.5}\selectfont MAGI-1-4.5B. MAGI-1-4.5B primarily fails through viewpoint instability and omission of the initial gravity-driven trigger. Across the rollout, the camera progressively drifts and moves closer to the scene, altering the apparent spatial relationships among the ramps, dominoes, and downstream objects. More importantly, neither of the two balls on the left-side ramps undergoes the expected downhill motion under gravity, leaving the initiating events of the reference process unrealized. From the causal perspective, the cascade therefore fails before it can properly begin: because the balls never descend from the ramps, the subsequent ball–domino contacts and downstream domino propagation cannot be activated through the intended pathway. The camera drift further destabilizes the spatial grounding of the scene, but the dominant causal failure is the missing trigger itself. Overall, the sequence is most consistent with inaccurate realization of the initial gravity-sensitive dynamics, accompanied by a viewpoint-grounding failure.\par}

\vspace{3pt}
\hrule
\vspace{3pt}
\refstepcounter{qualcase}\label{case:b09-m06}
{\fontsize{8}{9}\selectfont\bfseries Case~\thequalcase\enspace\textperiodcentered\enspace SkyReels-V2-I2V}\par
\vspace{1pt}
\qstrip{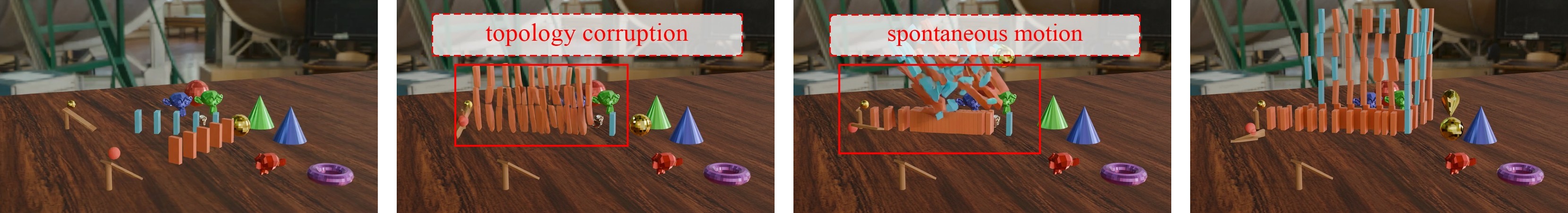}
\vspace{2pt}
{\fontsize{7.25}{8.5}\selectfont SkyReels-V2-I2V. SkyReels-V2-I2V exhibits a severe breakdown of entity persistence and scene integrity, with the central domino-like structures and blue elements rapidly changing in number, geometry, and spatial organization rather than remaining stable physical objects. These structures proliferate, deform, and expand vertically, producing pronounced topology corruption together with large-scale spontaneous motion that is unsupported by any preceding ramp-ball interaction. From the causal perspective, these failures destroy the persistent physical substrate required to ground the intended ball–domino interactions and replace the reference cascade with multiple spurious events and unsupported causal branches, so the original contact-mediated pathway can no longer be meaningfully recovered. Overall, the sequence is most consistent with a severe failure of persistent entity/world-state representation, compounded by unstable causal-temporal dynamics that repeatedly regenerate and reorganize scene structures without valid physical causes.\par}

\clearpage
\noindent
{\fontsize{10}{11.5}\selectfont\bfseries Level 3 · Full event-chain prompt}\par
\vspace{0.5pt}
{\fontsize{8}{9}\selectfont\itshape Domino\_Series.dual\_ball\_inverted\_y\_domino\_course}\par
\vspace{1pt}\hrule\vspace{3pt}
{\fontsize{8}{9}\selectfont\bfseries Ground Truth}\par
\vspace{1pt}
\qstrip{appendix_e/assets/b09_m01_cogvideox_5b_i2v_gt_strip.jpg}
\vspace{4pt}
\refstepcounter{qualcase}\label{case:b09-m07}
{\fontsize{8}{9}\selectfont\bfseries Case~\thequalcase\enspace\textperiodcentered\enspace Wan2.1-I2V-14B}\par
\vspace{1pt}
\qstrip{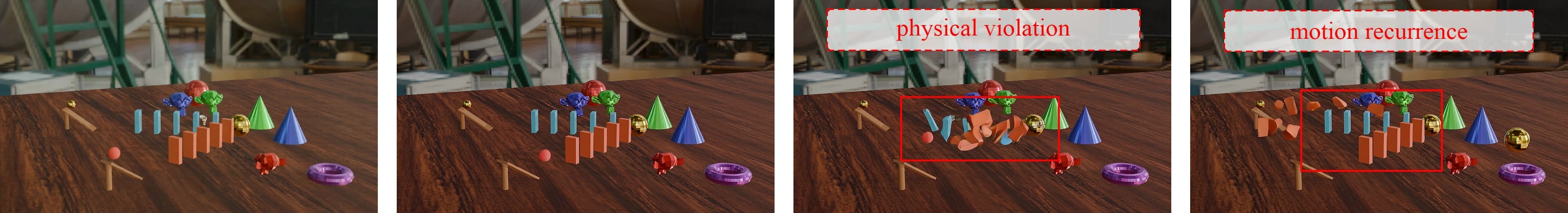}
\vspace{2pt}
{\fontsize{7.25}{8.5}\selectfont Wan2.1-I2V-14B. Wan2.1-I2V-14B initially captures the gravity-driven motion of the ball along the ramp reasonably well, but the subsequent domino interaction is physically incorrect. After the ball reaches the domino chain, the dominoes do not topple sequentially through contact-mediated propagation as in the reference rollout; instead, they collapse and scatter into a disordered configuration, indicating an incorrect realization of the collision and downstream force transmission. From the causal perspective, the initiating ball motion and approximate contact event are present, but the mediator sequence within the domino chain is corrupted, so the cascade no longer propagates through the intended domino-to-domino dependencies. The rollout then exhibits a second failure in causal-temporal progression: after the dominoes have already fallen, they rise again and reorganize into an ordered upright configuration, producing clear motion recurrence and partial state replay. Thus, even though the process begins along a plausible trajectory, the model fails both to realize the local collision dynamics correctly and to preserve the irreversibility of the completed cascade. Overall, the sequence is most consistent with inaccurate contact- and propagation-sensitive physical dynamics, compounded by a temporal-state persistence failure that reconstructs an earlier scene state after the domino collapse.\par}

\vspace{3pt}
\hrule
\vspace{3pt}
\refstepcounter{qualcase}\label{case:b09-m08}
{\fontsize{8}{9}\selectfont\bfseries Case~\thequalcase\enspace\textperiodcentered\enspace Wan2.2-TI2V-5B}\par
\vspace{1pt}
\qstrip{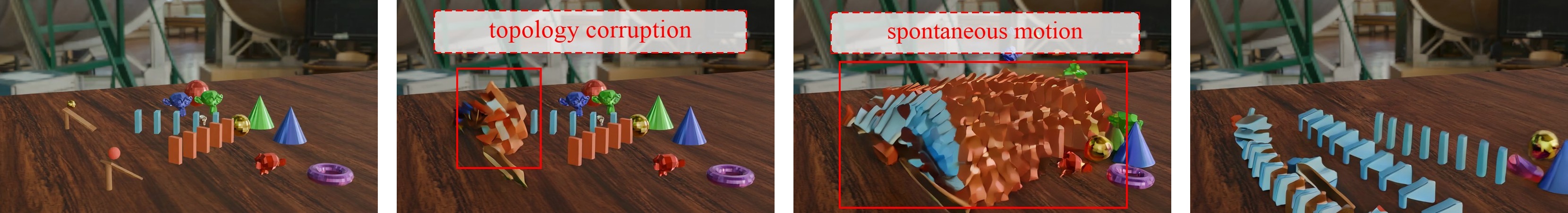}
\vspace{2pt}
{\fontsize{7.25}{8.5}\selectfont Wan2.2-TI2V-5B. Wan2.2-TI2V-5B exhibits a severe collapse of entity and structural integrity almost immediately after the rollout begins. The left-side ramp, the ball associated with it, and several nearby scene elements lose their original shapes and identities, transforming into irregular block-like structures that no longer resemble the corresponding objects in the reference scene. These corrupted structures then expand and move spontaneously through space without any valid upstream trigger, producing large-scale unsupported dynamics rather than the intended gravity-driven ramp motion and contact-mediated cascade. From the causal perspective, the topology and identity corruption first destroy the persistent physical substrate required to ground the reference events, after which the generated sequence is dominated by spurious motions and unsupported causal branches that effectively replace the intended ball–domino pathway. Overall, the sequence is most consistent with a severe failure of persistent entity/world-state representation, compounded by unstable dynamics once the underlying scene geometry and object identities have already collapsed.\par}

\vspace{3pt}
\hrule
\vspace{3pt}
\refstepcounter{qualcase}\label{case:b09-m09}
{\fontsize{8}{9}\selectfont\bfseries Case~\thequalcase\enspace\textperiodcentered\enspace Wan2.2-I2V-A14B}\par
\vspace{1pt}
\qstrip{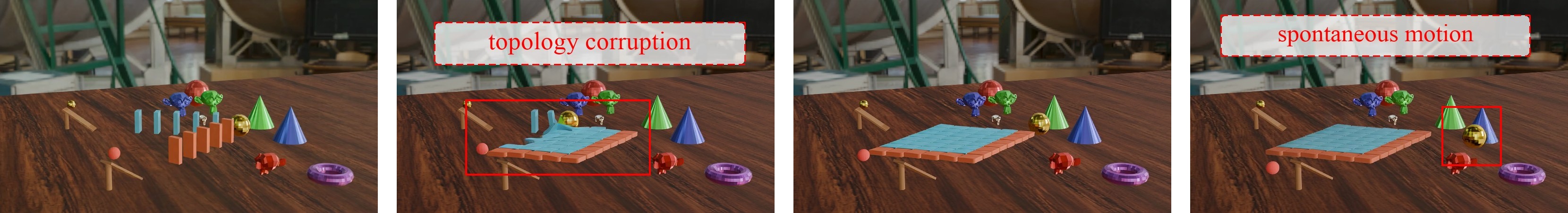}
\vspace{2pt}
{\fontsize{7.25}{8.5}\selectfont Wan2.2-I2V-A14B. Wan2.2-I2V-A14B exhibits a clear failure of object persistence and scene topology. The central domino chain, which should remain upright in the middle of the scene and then be knocked down sequentially by the rolling ball, instead collapses into a flat, spread-out structure on the tabletop, indicating severe topology corruption and loss of identity consistency. At the same time, the red ball at the top of the ramp fails to follow the expected gravity-driven trajectory, since it does not roll down the ramp as it should. Additional unsupported motion also appears elsewhere in the scene: objects on the tabletop, including the ball at the right side, undergo spontaneous displacement without a valid causal trigger. Overall, the sequence is dominated by topology corruption, failure to preserve object identity, and spontaneous motion that breaks the intended physical and causal structure of the event.\par}

\clearpage
\noindent
{\fontsize{10}{11.5}\selectfont\bfseries Level 3 · Full event-chain prompt}\par
\vspace{0.5pt}
{\fontsize{8}{9}\selectfont\itshape Galton\_Series.horizontal\_galton\_board\_object\_targets}\par
\vspace{1pt}\hrule\vspace{3pt}
{\fontsize{8}{9}\selectfont\bfseries Ground Truth}\par
\vspace{1pt}
\qstrip{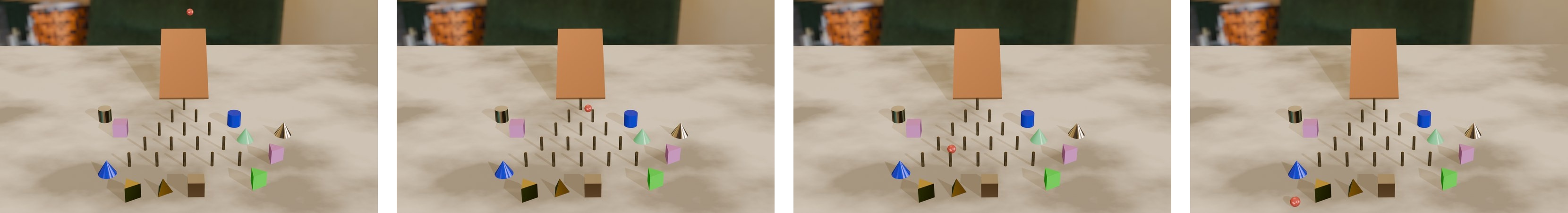}
\vspace{4pt}
\refstepcounter{qualcase}\label{case:b10-m01}
{\fontsize{8}{9}\selectfont\bfseries Case~\thequalcase\enspace\textperiodcentered\enspace CogVideoX-Fun-V1.5}\par
\vspace{1pt}
\qstrip{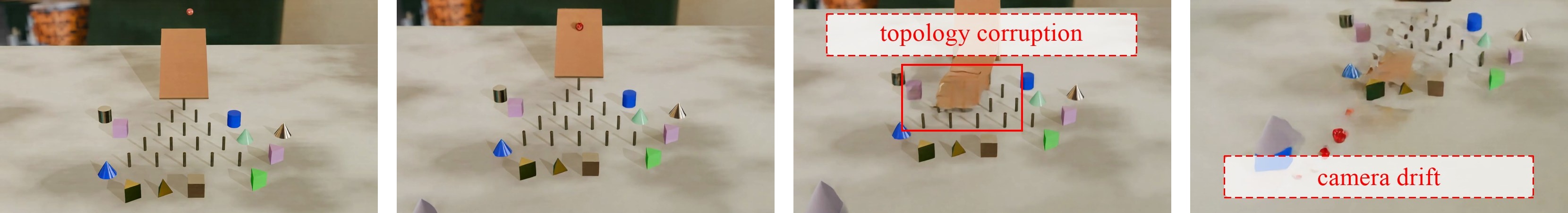}
\vspace{2pt}
{\fontsize{7.25}{8.5}\selectfont CogVideoX-Fun-V1.5. CogVideoX-Fun-V1.5 primarily fails through structural corruption of the inclined board together with viewpoint instability. As the rollout progresses, the board no longer preserves its original rigid geometry but instead becomes elongated, distorted, and visibly deformed, producing a clear topology-corruption failure in a structure that should remain fixed throughout the sequence. From the causal perspective, this deformation alters the physical substrate that is supposed to constrain the ball’s gravity-driven descent and subsequent peg-mediated interactions, making the reference trajectory and contact geometry increasingly unreliable. The failure is further compounded by substantial camera drift, which shifts the viewpoint and changes the apparent spatial relationships among the board, pegs, and surrounding objects. Overall, the sequence is most consistent with unstable structural scene representation combined with a viewpoint-grounding failure, both of which undermine the stable spatial and mechanical configuration required for faithful realization of the reference causal process.\par}

\vspace{3pt}
\hrule
\vspace{3pt}
\refstepcounter{qualcase}\label{case:b10-m02}
{\fontsize{8}{9}\selectfont\bfseries Case~\thequalcase\enspace\textperiodcentered\enspace Cosmos3-Nano}\par
\vspace{1pt}
\qstrip{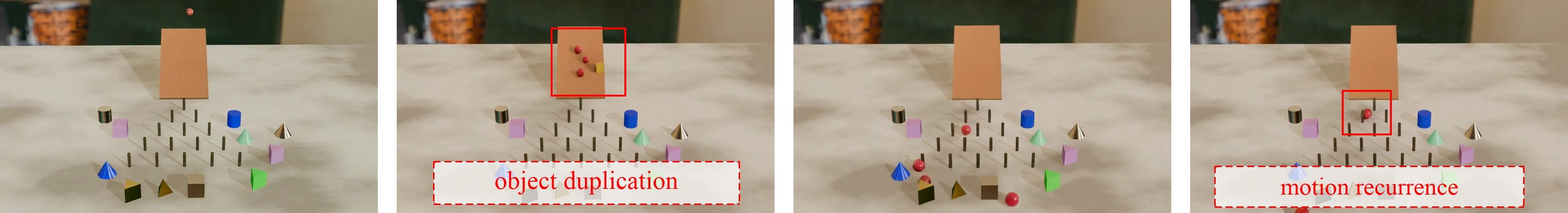}
\vspace{2pt}
{\fontsize{7.25}{8.5}\selectfont Cosmos3-Nano. Cosmos3-Nano mainly fails through object duplication combined with motion recurrence. Instead of preserving a single falling ball, the model replicates the original ball into several visually distinct balls while it is still on or near the upper board region, creating spurious entities that were never present in the initial condition. Some of these duplicated balls then continue to descend through the peg field in roughly gravity-consistent ways, so the local downward motion is partially plausible; however, the overall event structure is no longer correct because one initial object has been expanded into multiple independent trajectories. In addition, the sequence exhibits a clear recurrence failure: after the ball(s) have already descended and dispersed through the peg array, one ball reappears higher up near the pegs, effectively reversing part of the realized motion and replaying an earlier state. From a causal perspective, this introduces both spurious event generation and temporal inconsistency, since the model first invents extra objects and then partially resets the system to a previous stage of the rollout. Overall, the result suggests weak identity preservation together with unstable temporal-state tracking, leading to duplicated entities and a non-monotonic causal evolution.\par}

\vspace{3pt}
\hrule
\vspace{3pt}
\refstepcounter{qualcase}\label{case:b10-m03}
{\fontsize{8}{9}\selectfont\bfseries Case~\thequalcase\enspace\textperiodcentered\enspace HunyuanVideo-I2V}\par
\vspace{1pt}
\qstrip{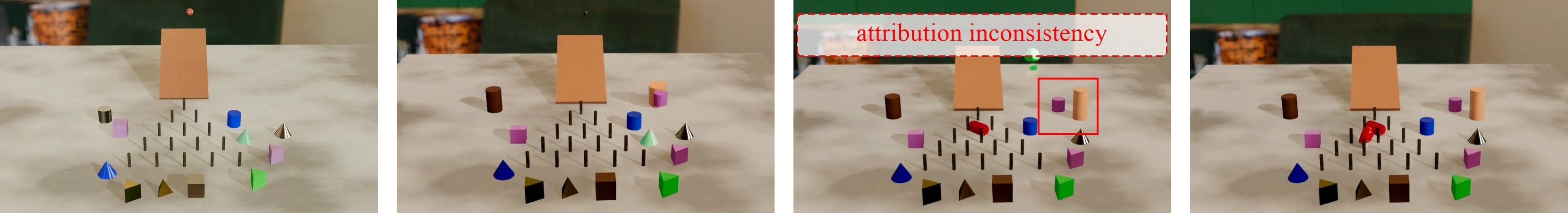}
\vspace{2pt}
{\fontsize{7.25}{8.5}\selectfont HunyuanVideo-I2V exhibits clear attribution inconsistency in this example. Several surrounding tabletop objects change their appearance over time, indicating that the model fails to preserve stable object identities across frames. The falling object itself also does not remain consistent: during its descent, it morphs from a sphere into a cylindrical shape, further revealing identity corruption. In addition, the ball does not complete the expected motion trajectory. Instead of continuing through the peg field and propagating downward as it should, it becomes stuck around the upper-middle region of the board. Overall, the result combines object-identity inconsistency with incorrect event realization, as the model neither preserves the object’s form nor allows the motion to unfold as expected.\par}

\clearpage
\noindent
{\fontsize{10}{11.5}\selectfont\bfseries Level 3 · Full event-chain prompt}\par
\vspace{0.5pt}
{\fontsize{8}{9}\selectfont\itshape Galton\_Series.horizontal\_galton\_board\_object\_targets}\par
\vspace{1pt}\hrule\vspace{3pt}
{\fontsize{8}{9}\selectfont\bfseries Ground Truth}\par
\vspace{1pt}
\qstrip{appendix_e/assets/b10_m01_cogvideox_5b_i2v_gt_strip.jpg}
\vspace{4pt}
\refstepcounter{qualcase}\label{case:b10-m04}
{\fontsize{8}{9}\selectfont\bfseries Case~\thequalcase\enspace\textperiodcentered\enspace LTX-Video-13B}\par
\vspace{1pt}
\qstrip{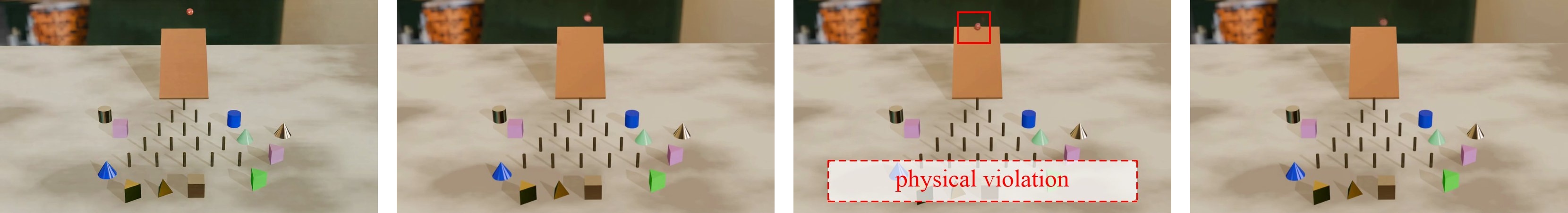}
\vspace{2pt}
{\fontsize{7.25}{8.5}\selectfont LTX-Video-13B. LTX-Video-13B primarily fails to realize the expected gravity-driven descent of the ball. Instead of falling from the upper region and entering the Galton board, the ball remains suspended above the inclined board while undergoing only slight horizontal displacement, producing a clear physical-law inconsistency in the absence of any visible support or balancing force. From the causal perspective, this constitutes a missing trigger: because the initial downward motion is never properly realized, the subsequent sequence of ball–peg contacts and trajectory redirections cannot begin along the reference pathway. The small lateral drift further indicates that the model generates unsupported motion while failing to execute the dominant gravity-driven transition required by the scene. Overall, the sequence is most consistent with inaccurate gravity-sensitive physical dynamics, which prevents the causal process from being initiated at all.\par}

\vspace{3pt}
\hrule
\vspace{3pt}
\refstepcounter{qualcase}\label{case:b10-m05}
{\fontsize{8}{9}\selectfont\bfseries Case~\thequalcase\enspace\textperiodcentered\enspace MAGI-1-4.5B}\par
\vspace{1pt}
\qstrip{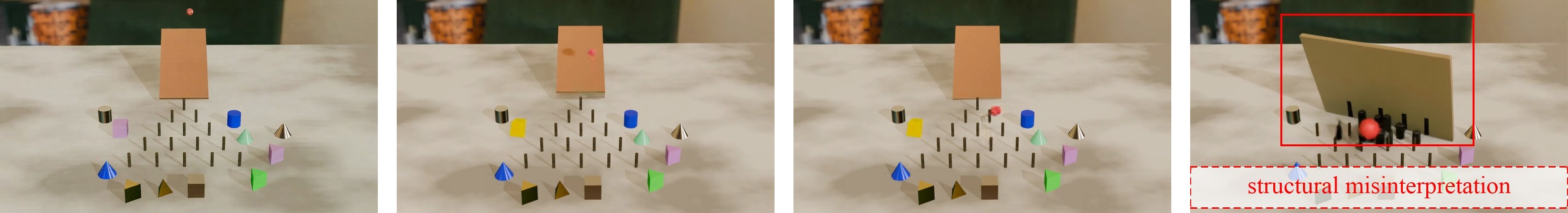}
\vspace{2pt}
{\fontsize{7.25}{8.5}\selectfont MAGI-1-4.5B. MAGI-1-4.5B initially reproduces the early dynamics reasonably well: the ball descends under gravity and enters the peg region along a trajectory that is broadly consistent with the Ground Truth. The rollout then diverges when the model misinterprets the spatial and structural role of the upper board, causing it to rotate or flip into an orientation that is not supported by the reference scene. This structural change alters the physical substrate through which the subsequent ball–peg interactions should be realized. At the same time, the falling ball no longer preserves a stable identity, but grows into a substantially larger sphere, producing clear identity corruption of the primary causal participant. From the causal perspective, the trigger and early descent events are therefore correctly initiated, but the later structural corruption and entity inconsistency prevent faithful continuation of the reference contact-mediated pathway. Overall, the sequence is most consistent with relatively accurate early gravity-sensitive dynamics followed by a structural-inference failure and unstable entity-state persistence.\par}

\vspace{3pt}
\hrule
\vspace{3pt}
\refstepcounter{qualcase}\label{case:b10-m06}
{\fontsize{8}{9}\selectfont\bfseries Case~\thequalcase\enspace\textperiodcentered\enspace SkyReels-V2-I2V}\par
\vspace{1pt}
\qstrip{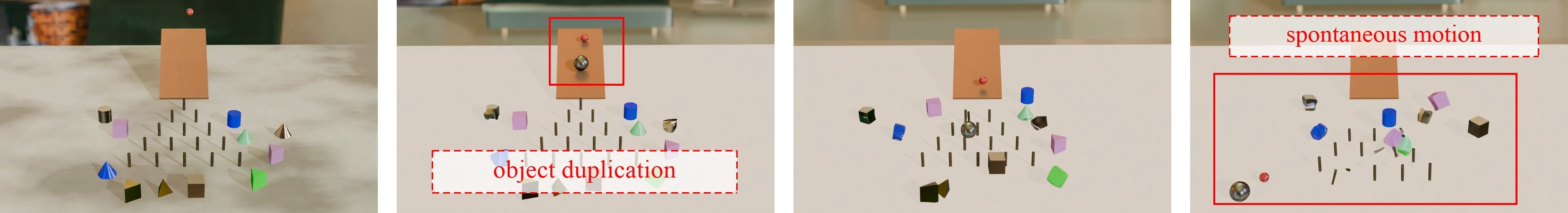}
\vspace{2pt}
{\fontsize{7.25}{8.5}\selectfont SkyReels-V2-I2V. SkyReels-V2-I2V initially captures the expected gravity-driven descent of the red ball reasonably well, but then fails to preserve the uniqueness of the causal participant. During the rollout, the original ball is duplicated into a second instance, so two balls proceed downward along the board instead of the single-ball trajectory defined by the Ground Truth. From the causal perspective, this duplication introduces a spurious participant and therefore creates additional ball–peg interaction branches that are absent from the reference process, even though the local downward motion of each ball is individually plausible. The sequence then develops a second failure after the balls move through the peg field: several pegs and surrounding tabletop objects begin to move spontaneously without any valid contact-based trigger, producing spurious downstream events and further corrupting the intended causal pathway. Overall, the rollout is most consistent with unstable entity-state persistence combined with weak preservation of causal dependencies, whereby a correctly initiated descent is progressively replaced by duplicated participants and unsupported motion of otherwise static scene elements.\par}

\clearpage
\noindent
{\fontsize{10}{11.5}\selectfont\bfseries Level 3 · Full event-chain prompt}\par
\vspace{0.5pt}
{\fontsize{8}{9}\selectfont\itshape Galton\_Series.horizontal\_galton\_board\_object\_targets}\par
\vspace{1pt}\hrule\vspace{3pt}
{\fontsize{8}{9}\selectfont\bfseries Ground Truth}\par
\vspace{1pt}
\qstrip{appendix_e/assets/b10_m01_cogvideox_5b_i2v_gt_strip.jpg}
\vspace{4pt}
\refstepcounter{qualcase}\label{case:b10-m07}
{\fontsize{8}{9}\selectfont\bfseries Case~\thequalcase\enspace\textperiodcentered\enspace Wan2.1-I2V-14B}\par
\vspace{1pt}
\qstrip{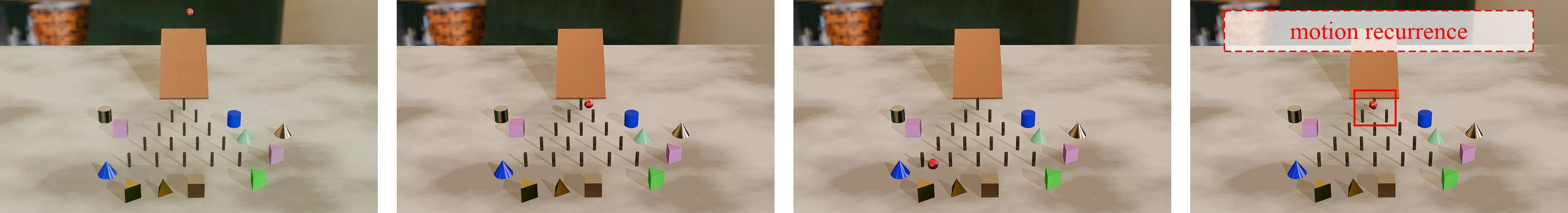}
\vspace{2pt}
{\fontsize{7.25}{8.5}\selectfont Wan2.1-I2V-14B. Wan2.1-I2V-14B performs very well in this Galton-board example and closely reproduces both the spatial trajectory and timing of the Ground Truth during most of the rollout. The ball descends under gravity, passes through the peg field along the expected route, and exits near the lower-left region with a causal progression that is largely consistent with the reference process. The main failure appears only after this otherwise correct sequence has been completed: instead of remaining in the terminal state after leaving the peg field, the ball reverses direction and returns upward toward an earlier position within the board. This constitutes clear motion recurrence and partial state replay, introducing a causal-temporal regression after the correct trajectory has already been realized. From the causal perspective, the event sequence and contact-mediated dependencies are therefore preserved for most of the rollout, but the model fails to maintain the completed terminal state and instead regenerates an earlier phase of motion. Overall, the sequence is most consistent with strong physical and causal realization accompanied by a residual temporal-state persistence failure.\par}

\vspace{3pt}
\hrule
\vspace{3pt}
\refstepcounter{qualcase}\label{case:b10-m08}
{\fontsize{8}{9}\selectfont\bfseries Case~\thequalcase\enspace\textperiodcentered\enspace Wan2.2-TI2V-5B}\par
\vspace{1pt}
\qstrip{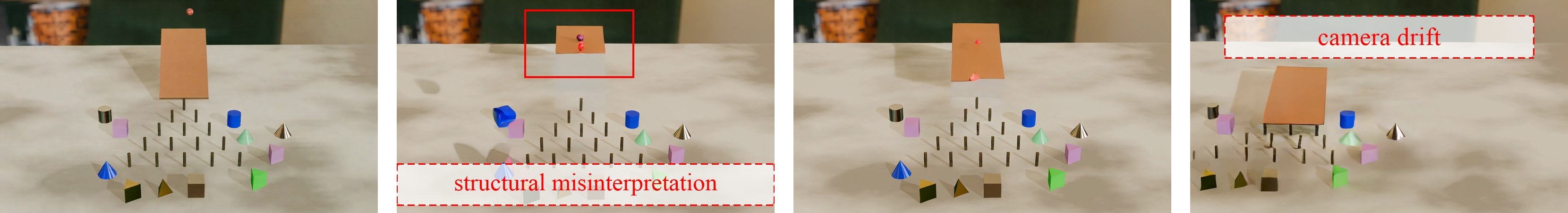}
\vspace{2pt}
{\fontsize{7.25}{8.5}\selectfont Wan2.2-TI2V-5B. Wan2.2-TI2V-5B primarily fails through structural misinterpretation and viewpoint instability. Rather than preserving the upper inclined structure as a fixed component of the scene, the model treats it as a freely moving board and causes it to rotate or flip during the rollout, thereby altering the physical substrate through which the ball should descend and interact with the peg field. This structural error is accompanied by spontaneous motion of other scene elements that move without any valid preceding contact or trigger, introducing spurious events absent from the reference process. The rollout also develops clear camera drift, further changing the apparent spatial relationships among the board, pegs, and surrounding objects. From the causal perspective, these failures corrupt both the scene geometry and the stable interaction framework required for the intended ball trajectory, preventing faithful realization of the reference cascade. Overall, the sequence is most consistent with structural-inference failure compounded by weak causal grounding and viewpoint control.\par}

\vspace{3pt}
\hrule
\vspace{3pt}
\refstepcounter{qualcase}\label{case:b10-m09}
{\fontsize{8}{9}\selectfont\bfseries Case~\thequalcase\enspace\textperiodcentered\enspace Wan2.2-I2V-A14B}\par
\vspace{1pt}
\qstrip{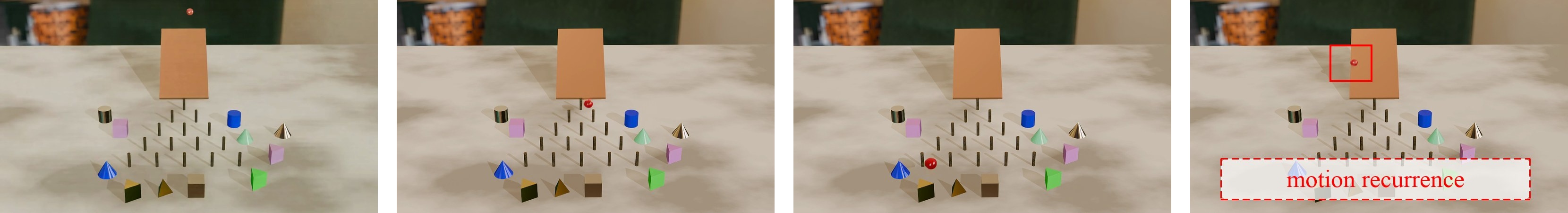}
\vspace{2pt}
{\fontsize{7.25}{8.5}\selectfont Wan2.2-I2V-A14B. Wan2.2-I2V-A14B performs very well in this example and closely reproduces both the spatial trajectory and timing of the Ground Truth during most of the rollout. The ball descends under gravity, passes through the peg field in the correct order, and exits near the lower-left region with a causal progression that is largely consistent with the reference process. The main failure appears only after this otherwise correct trajectory has been completed: instead of remaining in the terminal state after leaving the peg field, the ball reverses direction and moves upward toward the upper board, producing clear motion recurrence and partial state replay. From the causal perspective, the event sequence and contact-mediated dependencies are therefore preserved for most of the rollout, but the model fails to maintain the completed terminal state and instead regenerates an earlier phase of motion. Overall, the sequence is most consistent with strong physical and causal realization accompanied by a residual temporal-state persistence failure.\par}

\clearpage

\end{document}